%% file: main.tex
\documentclass[10pt,twocolumn,letterpaper]{article}

\usepackage[pagenumbers]{cvpr} 
\usepackage{multirow}
\usepackage{algorithm}
\usepackage{algpseudocode}
\usepackage{setspace}
\usepackage{array}
\usepackage{subcaption}
\usepackage{xfrac}
\usepackage{wrapfig}
\usepackage{tikz}

\input{preamble}
\definecolor{cvprblue}{rgb}{0.21,0.49,0.74}
\usepackage[pagebackref,breaklinks,colorlinks,allcolors=cvprblue]{hyperref}

\def\paperID{} 
\def\confName{CVPR}
\def\confYear{2026}

\title{Speeding Up the Learning of 3D Gaussians with Much Shorter Gaussian Lists}

\author{
Jiaqi Liu \quad \quad Zhizhong Han\\
Machine Perception Lab, 
Wayne State University, Detroit, USA\\
{\tt\small \{ig1017, h312h\}@wayne.edu}
}

\begin{document}

\twocolumn[{
\maketitle

\begin{center}
\vspace{-6mm}
    \centering
    \captionsetup{type=figure}
    \includegraphics[width=\textwidth]{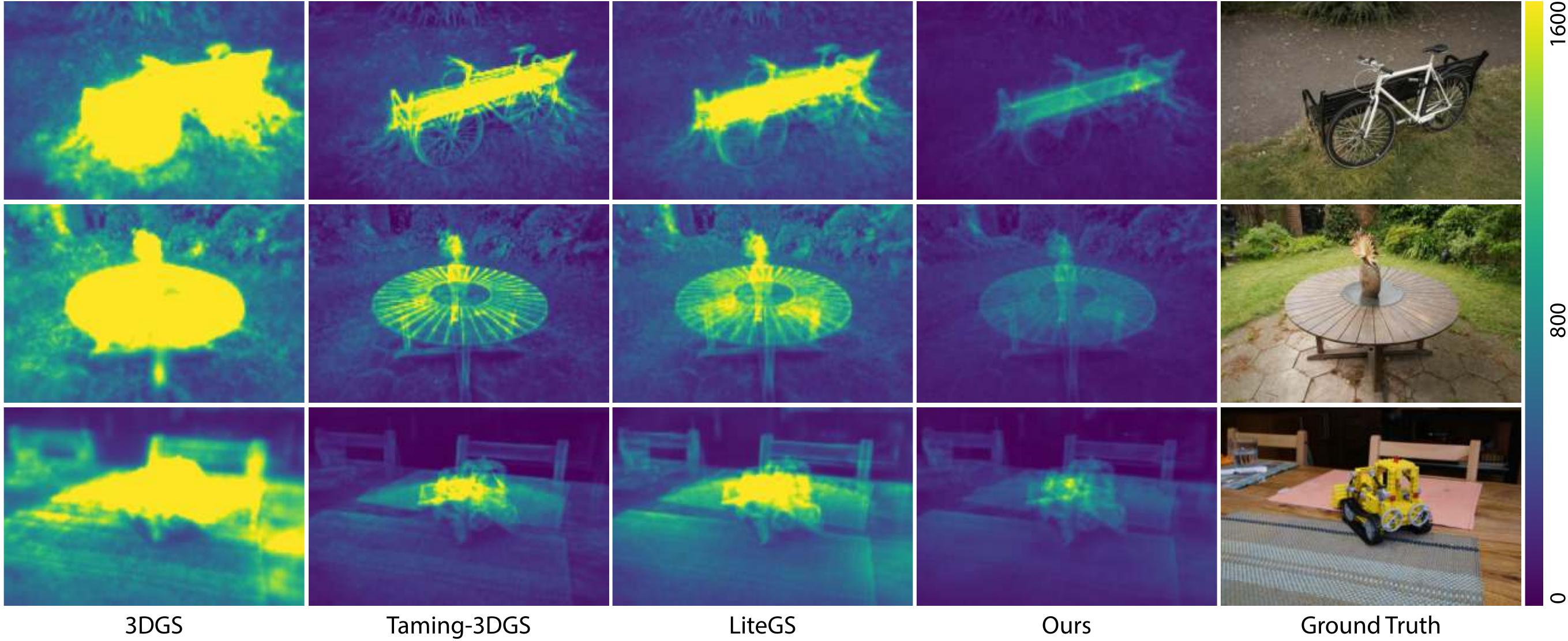}
    \vspace{-6mm}
    \captionof{figure}{
    Heatmaps of per-tile Gaussian list lengths measured during testing.
    Colors encode counts from low to high, with purple indicating fewer Gaussians and yellow indicating more. 
    All methods use identical tile sizes for fair comparison. 
    Our method consistently achieves the shortest lists across all scenes. 
    Training times (in seconds): 3DGS (919.51), Taming-3DGS (402.54), LiteGS (191.17), Ours (99.58).
    Additional results are provided in \cref{fig:supplementary_gaussian_count_viz}.}
    \label{fig:gaussian_count_viz}
\vspace{0mm}
\end{center}
}]



\begin{abstract}

3D Gaussian splatting (3DGS) has become a vital tool for learning a radiance field from multiple posed images. Although 3DGS shows great advantages over NeRF in terms of rendering quality and efficiency, it remains a research challenge to further improve the efficiency of learning 3D Gaussians. To overcome this challenge, we propose novel training strategies and losses to shorten each Gaussian list used to render a pixel, which speeds up the splatting by involving fewer Gaussians along a ray. Specifically, we shrink the size of each Gaussian by resetting their scales regularly, encouraging smaller Gaussians to cover fewer nearby pixels, which shortens the Gaussian lists of pixels. Additionally, we introduce an entropy constraint on the alpha blending procedure to sharpen the weight distribution of Gaussians along each ray, which drives dominant weights larger while making minor weights smaller. As a result, each Gaussian becomes more focused on the pixels where it is dominant, which reduces its impact on nearby pixels, leading to even shorter Gaussian lists. Eventually, we integrate our method into a rendering resolution scheduler which further improves efficiency through progressive resolution increase. We evaluate our method by comparing it with state-of-the-art methods on widely used benchmarks. Our results show significant advantages over others in efficiency without sacrificing rendering quality. Code is available at: \href{https://github.com/MachinePerceptionLab/ShorterSplatting}{https://github.com/MachinePerceptionLab/ShorterSplatting}.

\end{abstract}

\section{Introduction}



Novel view synthesis is an important task in computer vision, AR, VR, and robotics~\cite{irshad2024neural, zhang2025advances}. The widely used strategy is to first infer a radiance field from image observations, then render the radiance field into images from specific perspectives. Neural radiance field (NeRF)~\cite{mildenhall2020nerf} represents a radiance field using a neural network that can be queried at any point for color and opacity in volume rendering. Although NeRFs can model the geometry, appearance, and lighting of scenes well, both root finding with sampling and querying implicit neural networks are expensive.



Recently, 3D Gaussian splatting (3DGS)~\cite{kerbl3Dgaussians} has shown advantages over NeRF in terms of rendering efficiency and quality. 3DGS employs discrete 3D Gaussians with attributes like color, opacity, and shape to explicitly represent a radiance field which can be rendered into an image by a splatting operation. Although splatting runs faster, learning 3D Gaussians from image observations still requires a large number of splatting operations, which limits 3DGS in time-sensitive applications. Thus, more recent methods have adopted different strategies to speed up the learning of 3D Gaussians, such as more concise implementations~\cite{ye2025gsplat, liao2025litegshighperformanceframeworktrain}, more reasonable Gaussian density control~\cite{mallick2024taming, fang2024miniv2, kim2024color, chen2025dashgaussian}, and more advanced training strategies~\cite{hoellein_2025_3dgslm, lan20253dgs2, pehlivan2025second}. However, there is still room for improvement in training efficiency.

\begin{figure}[t]
  \centering
   \includegraphics[width=1.0\linewidth]{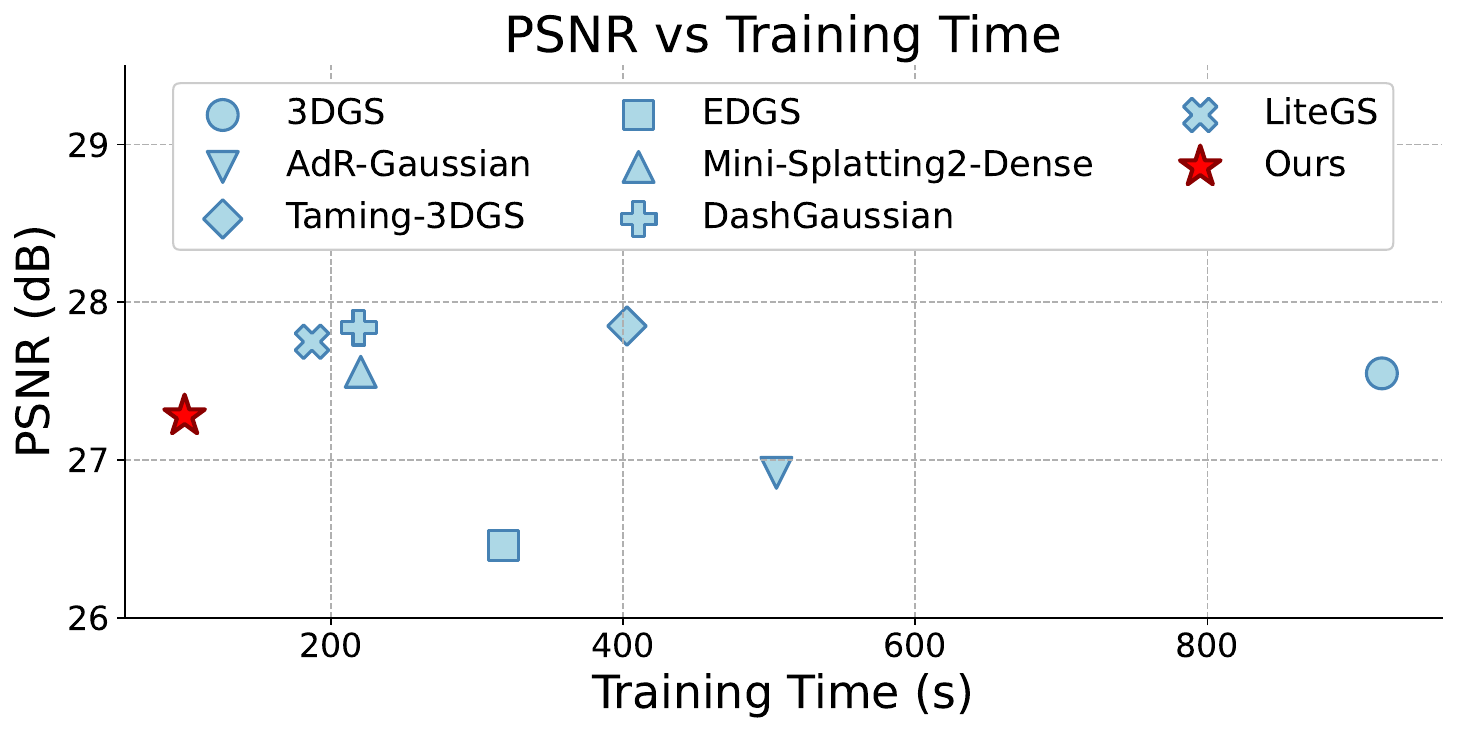}
   \vspace{-8mm}
   \caption{Our method achieves the fastest training time while maintaining comparable reconstruction quality.}
   \label{fig:psnr_all_methods_scatter}
   \vspace{-7.2mm}
\end{figure}

To achieve this goal, we propose to speed up the learning of 3D Gaussians by encouraging shorter Gaussian lists when rendering a color at each pixel, as shown by the Gaussian length heatmap in \cref{fig:gaussian_count_viz}. The efficiency averaged over all scenes in Mip-NeRF~360 highlights our improvement with comparable rendering quality in \cref{fig:psnr_all_methods_scatter}. Unlike the strategy of reducing the total number of 3D Gaussians~\cite{hanson2025speedy, fang2024miniv2}, our method mainly shortens the Gaussian lists by adjusting the Gaussian distributions at each pixel for more efficient modeling of radiance fields. Specifically, we propose scale reset which regularly decreases Gaussian scales by a ratio, encouraging smaller Gaussians covering fewer pixels, leading to shorter Gaussian lists at most pixels. Additionally, we introduce an entropy constraint on the alpha blending procedure to sharpen the weight distribution of Gaussians along each ray, which drives dominant weights larger while making minor weights smaller. This constraint increases the modeling efficiency by adjusting the weight distribution. It makes each Gaussian focus more on pixels where it is dominant, which weakens the Gaussian's impact on nearby pixels, leading to even shorter Gaussian lists. 
Eventually, we integrate our scale reset and entropy constraint into a rendering resolution scheduler, which further improves efficiency by rendering images at progressively increasing resolutions during training. 
We evaluate our method by comparing it with state-of-the-art methods on widely used benchmarks. Our results show significant advantages over others in efficiency with comparable rendering quality. Our main contributions are listed below.
\begin{itemize}
\item We present a method for speeding up the learning of 3D Gaussians by rendering pixels with shorter Gaussian lists.
\item We introduce scale reset to encourage smaller 3D Gaussians and an entropy constraint on the alpha blending procedure to adjust the weight distribution along a ray.
\item We achieve state-of-the-art training efficiency without degrading rendering quality.
\end{itemize}


\input{sec/related_works}
\input{sec/background}
\input{sec/method}
\input{sec/experiments}
\input{sec/ending}

{
    \small
    \bibliographystyle{ieeenat_fullname}
    \bibliography{main}
}

\input{sec/X_suppl}

\end{document}

%% file: sec/related_works.tex
\section{Related Work}

3DGS has been broadly explored in novel view synthesis~\cite{MaterialRefGS,MonoInstance,Binocular3DGS}, surface reconstruction~\cite{SelfConstrainedPriors3DGS,VGGS,GaussianUDF,GSPull}, and SLAM~\cite{SGADSLAM,VTGaussianSLAM,QQSLAM}. 
As our work focuses on efficiency, we mainly review prior studies on training acceleration.

\subsection{Reducing Training Load}
Recent work has substantially reduced the training iterations required by the original 3DGS pipeline from 30K~\cite{kerbl3Dgaussians}.
Fang and Wang~\cite{fang2024miniv2} achieve competitive quality within 8K iterations via aggressive densification, while Kotovenko \etal~\cite{kotovenko2025edgs} attain a 6 times reduction by directly triangulating dense 2D keypoints as initialization.
Fan \etal~\cite{fan2024instantsplat} leverage geometric foundation models MASt3R~\cite{mast3r_eccv24} for sparse-view reconstruction in just 1K iterations.
Liu \etal~\cite{liu2024mvsgaussian} propose MVSGaussian to eliminate per-scene training through generalization, allowing rapid scene-specific fine-tuning.


\subsection{Optimizer and CUDA Implementations}
Another line of work focuses on replacing Adam with second-order optimizers for faster convergence.
H{\"o}llein \etal~\cite{hoellein_2025_3dgslm} apply Levenberg–Marquardt optimization within 3DGS-LM, while Lan \etal~\cite{lan20253dgs2} propose 3DGS$^2$, decomposing the optimization at the kernel-attribute level to construct small Newton systems requiring 10 times fewer iterations. 
Pehlivan \etal~\cite{pehlivan2025second} combine Levenberg–Marquardt and Conjugate Gradient methods tailored for splatting.

Complementary efforts target low-level CUDA implementations for better GPU utilization.
Durvasula \etal~\cite{durvasula2023distwar} address atomic-operation bottlenecks via warp-level reduction, while Ye \etal~\cite{ye2025gsplat} release \texttt{gsplat} with optimized kernels.
Mallick \etal~\cite{mallick2024taming} restructure backpropagation from per-pixel to per-splat parallelization. 
Liao \etal~\cite{liao2025litegshighperformanceframeworktrain} present LiteGS with warp-based rasterization and a Cluster–Cull–Compact pipeline for significant acceleration.

Other works improve GPU efficiency through better parallelization strategies.
Wang \etal~\cite{wang2024faster} propose group-training to organize Gaussians into manageable cohorts, 
while Wang \etal~\cite{wang2024adr} introduce AdR-Gaussian with load-balancing to mitigate uneven per-pixel rendering workloads. 
Gui \etal~\cite{gui2024balanced} present Balanced 3DGS with Gaussian-wise parallelization for uniform thread workload. Complementary rendering works address popping artifacts via hierarchical sorting~\cite{radl2024stopthepop} and achieve 4 times speedup through stochastic rasterization~\cite{KheradmandVicini2025stochasticsplats}.


\subsection{Training Strategies}
Armagan \etal~\cite{armagan2025gsta} selectively freeze converged Gaussians via gradient thresholding. 
Hanson \etal~\cite{hanson2025speedy} achieve 90\% model reduction through dual pruning, 
Ren \etal~\cite{ren2025fastgs} accelerate training to 100 seconds via multi-view-consistent densification and pruning that adaptively regulates Gaussian counts, together with a compact bounding box for faster rasterization. 
Kim \etal~\cite{kim2024color} achieve 9 times compression via color-cued densification that combines spherical harmonics with positional gradients.
Progressive optimization strategies include coarse-to-fine resolution scheduling with quantization~\cite{girish2024eagles}, leading to 10-20 times memory reduction, progressive frequency-component fitting~\cite{chen2025dashgaussian}, and locality-aware near-real-time optimization during capture~\cite{xu2025gaussian}.

While not targeting training acceleration, Sun \etal~\cite{sun2025svraster} and Zoomers \etal~\cite{zoomers2025nvgs} improve rendering speed through sparse voxel rasterization and neural occlusion culling.

Unlike these approaches, we accelerate 3D Gaussian learning by shortening Gaussian lists at all pixels. 
Our method does not rely on data-driven priors, Gaussian reduction, or higher-order optimizers, while still achieving significantly improved learning efficiency.

%% file: sec/background.tex
\section{Background}
\label{se:background}
In 3DGS, each 3D Gaussian primitive $\mathcal{G}_i$ is parameterized by a mean $\mu_i \in \mathbb{R}^3$, scale $s_i \in \mathbb{R}^3$, rotation $r_i \in \mathbb{R}^4$, opacity $\sigma_i \in \mathbb{R}$, and color $c_i \in \mathbb{R}^{16\times3}$ represented in spherical harmonics.
The 3D Gaussian density function is defined as
\begin{align}
    G_i(x) = \exp\left(
    -\frac{1}{2} (x-\mu_i)^T \Sigma_i^{-1} (x-\mu_i)
    \right),
\end{align}
where the covariance matrix is $\Sigma_i = R_i S_i S_i^T R_i^T$.

For rendering from a given camera viewpoint, all 3D Gaussians are projected onto the 2D image plane using the EWA splatting approach~\cite{zwicker2001ewa}. In 2D image space, the projected Gaussian density becomes
\begin{align}
    g_i(x) = \exp\left(
    -\frac{1}{2} (x-\mu_i')^T (\Sigma_i')^{-1} (x-\mu_i')
    \right),
\end{align}
where $\mu_i'$ and $\Sigma_i'$ denote the projected mean and covariance.

The splatting process produces a Gaussian list at each pixel, containing all Gaussians that contribute to the volume rendering along the corresponding ray.
At pixel $p \in \mathbb{R}^2$, the alpha value of the $i$-th Gaussian is computed as
\begin{align}
    \alpha_i(p) = \sigma_i g_i(p),
    \label{eq:alpha_definition}
\end{align}
and the transmittance is recursively defined as
\begin{align}
    T_i(p) &= T_{i-1}(p) \cdot (1-\alpha_{i-1}(p))= \prod_{j=1}^{i-1} (1-\alpha_j(p)),
\end{align}
where $T_1(p)=1$. For a Gaussian list containing $N$ Gaussians, sorted by depth, the final blended color is
\begin{align}
    C(p) = \sum_{i=1}^N T_i(p) \alpha_i(p) c_i(p) = \sum_{i=1}^N w_i(p) c_i(p),
\end{align}
where we assume a black background.


\begin{figure*}[t]
    \centering
    \includegraphics[width=\textwidth]{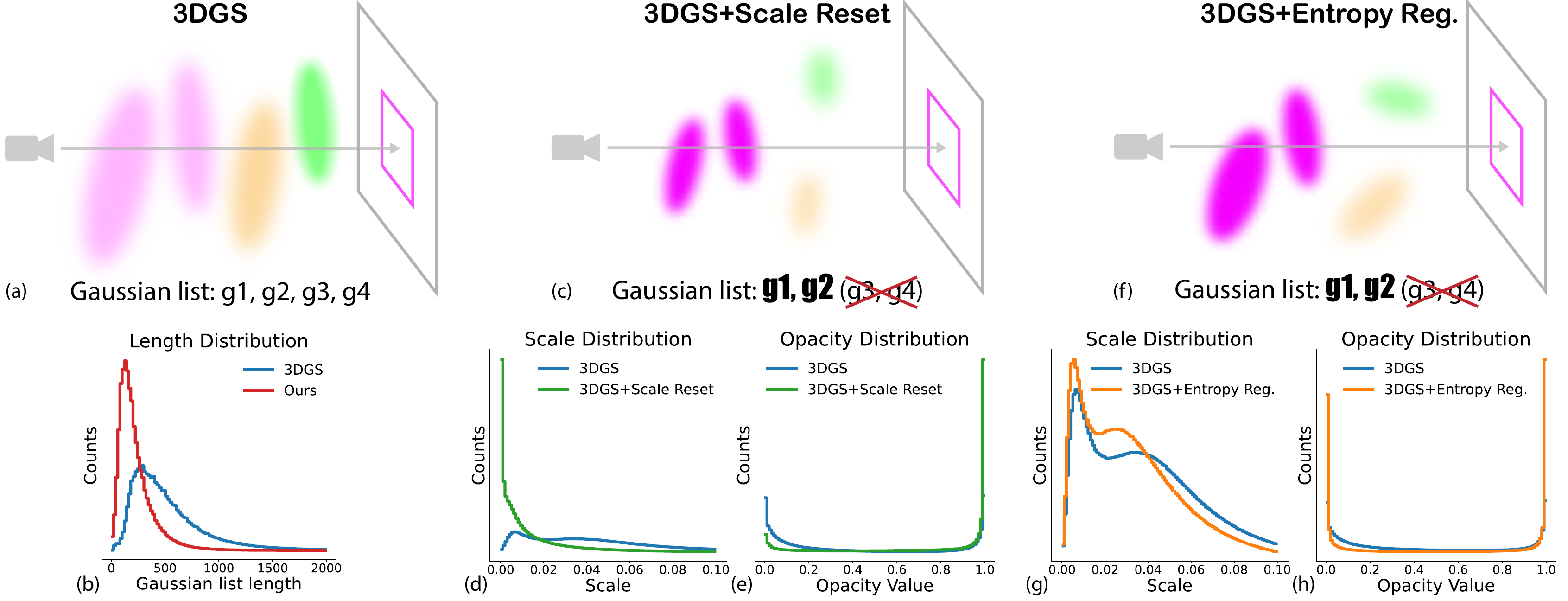}
    \vspace{-6mm}
    \caption{Overview of our method and effects. (a) The Gaussian list of 3DGS. (b) Distribution of Gaussian list length showing our method produces significantly shorter lists. (c) Gaussian list reduction after applying scale reset. (d, e) Scale and opacity distributions comparing 3DGS and 3DGS with scale reset, showing scale reset produces smaller Gaussians with higher opacities. (f) Gaussian list reduction after applying entropy regularization. (g, h) Scale and opacity distributions comparing 3DGS and 3DGS with entropy regularization, demonstrating entropy constraint produces smaller Gaussians and more polarized opacities. “3DGS” results are produced with LiteGS.
    }
    \label{fig:overview}
    \vspace{-4mm}
\end{figure*}


The training loss combines an $\mathcal{L}_1$ MAE term and a D-SSIM term with $\lambda$ denoting the weighting factor:
\begin{align}
\label{eq:base}
    \mathcal{L}_{\text{base}} = (1-\lambda) \mathcal{L}_1 + \lambda \mathcal{L}_{\text{D-SSIM}}.
\end{align}

%% file: sec/method.tex
\section{Method}
\label{se:method}

\subsection{Overview}
To render a pixel, rasterization-based volume rendering constructs a Gaussian list containing all 3D Gaussians that can contribute in alpha blending along the ray, as shown in~\cref{fig:overview}~(a). Longer lists contain more Gaussians, which directly increases both memory access and computational costs during forward rendering and backward gradient computation.
Our method accelerates Gaussian learning by producing significantly shorter Gaussian lists, as evidenced by the comparison of list length distributions in~\cref{fig:overview}~(b). All statistics reported below are computed as averages over all scenes in Mip-NeRF~360.

Prior work accelerates the learning of 3D Gaussians by reducing the total number of Gaussians~\cite{fang2024miniv2, hanson2025speedy} or by estimating the pixel coverage of each Gaussian more precisely~\cite{hanson2025speedy, wang2024adr, liao2025litegshighperformanceframeworktrain}, both of which result in shorter per-pixel Gaussian lists. However, reducing the total Gaussian count is impractical for complex or large-scale scenes with rich geometric details, while more precise coverage estimation only provides marginal speedups (e.g., 10\% in \cite{hanson2025speedy}). 

In contrast, our method operates effectively even with a large number of Gaussians, making it suitable for large-scale scenes. Our key idea is to encourage each Gaussian to concentrate its influence on a localized image region rather than dispersing its contribution across many pixels, thereby producing shorter Gaussian lists. This principle of spatial concentration motivates the following two approaches.

As shown in \cref{fig:overview}, we regularly reset the scale of each Gaussian by a ratio, encouraging smaller Gaussians to reduce the impact of each Gaussian on neighboring pixels, as shown by the illustration in \cref{fig:overview}~(c) and the comparison of scale distributions in \cref{fig:overview}~(d). The scale reset also encourages larger opacity, as shown by the opacity distribution in \cref{fig:overview}~(e). Moreover, we impose the entropy constraint on weights in the alpha blending procedure, which drives dominant weights larger while making minor weights smaller, reducing the impact of Gaussians with minor weights on neighboring pixels, as shown by the illustration in \cref{fig:overview}~(f) and the comparison of opacity distributions in \cref{fig:overview}~(h). The entropy constraint also encourages smaller scales, as shown by the scale distribution in \cref{fig:overview}~(g). 


\subsection{Scale Reset}

Since larger Gaussians cover more pixels and lengthen 
\begin{wrapfigure}{r}{0.52\linewidth}
  \centering
  \vspace{-2mm}
  \includegraphics[width=\linewidth]{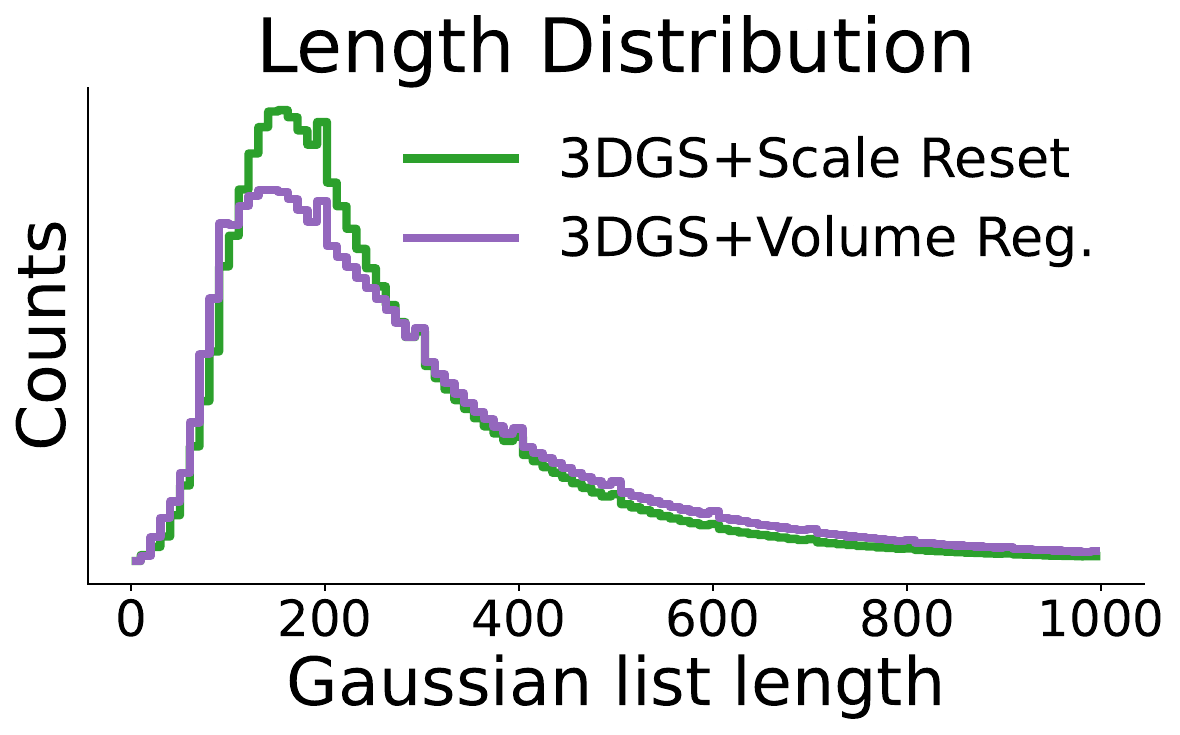}
  \vspace{-7mm}
  \caption{Scale reset achieves shorter Gaussian lists (limited x-range for clarity).}
  \label{fig:reset_vs_volume}
  \vspace{-6mm}
\end{wrapfigure}
Gaussian lists, encouraging smaller scales is a natural strategy. 
A straightforward approach is to add a volume penalty to the loss, but tuning its weight or threshold is difficult, as Gaussians can otherwise become too large or too small.

Our scale reset adopts a simpler strategy while yielding superior quality and speed. Specifically, we periodically reset the scale of all Gaussians with a shrinking factor $\zeta<1$,
\begin{align}
    s_i \leftarrow \zeta \cdot s_i, \quad \forall i.
\end{align}

The comparison of list length distributions in \cref{fig:reset_vs_volume} shows that scale reset produces shorter Gaussian lists under similar quality control.
One reason for its effectiveness is that scale reset takes effect much faster than volume regularization, providing immediate acceleration by instantly reducing scales and achieving shorter per-pixel Gaussian lists in all subsequent iterations.
Moreover, scale reset provides sufficient iterations before the next reset for adjusting other attributes to approximate the radiance field well, thereby ensuring rendering quality.
See \cref{sec:ablation_study} for the ablation studies and its superiority over volume regularization.


\subsection{Entropy Constraint}
\begin{figure}[t]
  \centering
  \includegraphics[width=1\linewidth]{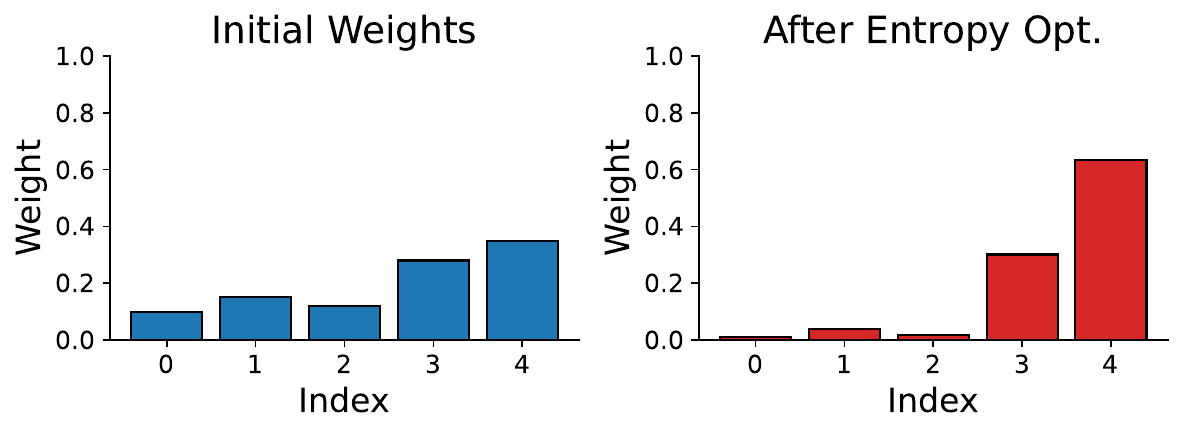}
  \vspace{-9mm}
  \caption{
  A toy example demonstrating how entropy regularization polarizes Gaussian weights along a ray.
  }
  \label{fig:entropy_effect}
  \vspace{-6mm}
\end{figure}

Entropy measures the disorder of a distribution. By minimizing entropy, we sharpen the distribution of Gaussian contributions in the alpha blending when rendering each pixel. With our entropy constraint, Gaussians with significant contribution become even more dominant while those with minor contribution become even weaker. This constraint encourages each Gaussian to concentrate on its designated region, while shrinking nondominant Gaussians, thereby shortening per-pixel Gaussian lists. \cref{fig:entropy_effect} demonstrates the effect of entropy loss on the weights along a ray.

Our entropy regularization can be applied to various Gaussian attributes, such as opacity. We impose the regularization on the Gaussian weights along each ray,
\begin{align}
    w_i = T_i \alpha_i.
\end{align}
For a Gaussian list of length $N$, these weights form a valid probability distribution, as the sum over all $N$ Gaussians plus the background contribution equals unity,
\begin{align}
    \sum_{i=1}^{N+1} w_i = 1,
\end{align}
where $w_{N+1} = T_{N+1}$ denotes the residual transmittance representing the background contribution after all Gaussians have been processed. A formal proof is provided in \cref{se:weights_sum_to_unity} of the supplementary material.

This normalization property eliminates the need for an explicit normalization pass during entropy computation and is the primary reason we apply entropy to weights. Without this property, normalization would require maintaining global statistics across all Gaussians and accessing them during backpropagation, thereby breaking the streaming computation model and incurring additional memory bandwidth costs that degrade cache performance.

The entropy of the Gaussian list at pixel $j$ is defined as
\begin{align}
    H_j &= -\sum_{i=1}^{N+1} w_{i,j} \log w_{i,j},
\end{align}
and the entropy loss is the average across all $M$ pixels,
\begin{align}
    \mathcal{L}_{\text{E}} &= \frac{1}{M} \sum_{j=1}^{M} H_j.
\end{align}
The overall loss combines losses in \cref{eq:base} and $\mathcal{L}_{E}$,
\begin{align}
    \mathcal{L} &= \mathcal{L}_{\text{base}} + \gamma \mathcal{L}_{\text{E}},
\end{align}
where $\gamma$ is a weighting coefficient.

Our proposed techniques, i.e., scale reset and entropy constraint, can be used together in a highly effective and complementary manner. Scale reset provides immediate geometric regularization by directly reducing Gaussian sizes, while entropy constraint continuously refines the contribution distribution during optimization, encouraging each Gaussian to focus on its designated region. Together, they achieve substantial speedup with much shorter Gaussian lists while maintaining rendering quality.


\subsection{Gradient Calculation}
Let $x_i$ denote any attribute (i.e., opacity, position, scale, rotation) of the $i$-th Gaussian. The gradient of total loss is
\begin{align}
    \frac{\partial \mathcal{L}}{\partial x_i}
    &= \frac{\partial \mathcal{L}_{\text{base}}}{\partial x_i} + \frac{\gamma}{M} \sum_{j=1}^{M} \frac{\partial H_j}{\partial \alpha_{i,j}} \frac{\partial \alpha_{i,j}}{\partial x_i}.
\end{align}
Note that $\partial \mathcal{L}_{\text{base}}/\partial x_i$ and $\partial \alpha_{i,j}/\partial x_i$ are standard in existing 3DGS frameworks. 

Introducing the intermediate variable,
\begin{align}
    R_{i,j} = \sum_{k=i}^{N+1} (\log w_{k,j} + 1) w_{k,j},
\end{align}
we obtain the gradient of entropy with respect to $\alpha_{i,j}$,
\begin{align}
    \frac{\partial H_j}{\partial \alpha_{i, j}} = (-\log w_{i, j} - 1) T_{i, j} + \frac{R_{i+1, j}}{1-\alpha_{i, j}}.
\end{align}
The complete derivation is provided in \cref{se:entropy_gradient} of the supplementary material. \cref{alg:H_alpha} presents an efficient $O(N)$ algorithm for computing these gradients.
\begin{algorithm}[H]
\caption{\small Compute $\partial H_j/\partial \alpha_i$ for all $i$ given pixel $j$}
\label{alg:H_alpha}
\begin{algorithmic}[1]
\small
\setstretch{1.3}
\State $R_j \gets (\log w_{N+1,j} + 1) \cdot w_{N+1,j}$ \Comment{Initialize $R_j$}
\For{$i = N$ \textbf{down to} $1$}
    \State $\frac{\partial H_j}{\partial \alpha_{i, j}} \gets (-\log w_{i,j} - 1) T_{i,j} + \frac{R_j}{1 - \alpha_{i,j}}$
    \State $R_j \gets (\log w_{i,j} + 1) \cdot w_{i,j} + R_j$ \Comment{Update $R_j$}
\EndFor
\end{algorithmic}
\end{algorithm}

%% file: sec/experiments.tex
\input{sec/stats_1}

\section{Experiments and Analysis}
\label{se:experiments}

\subsection{Datasets}

We follow standard protocols and evaluate our method on real-world scenes from Mip-NeRF~360~\cite{barron2022mipnerf360} (nine scenes: four indoor, five outdoor), Tanks \& Temples~\cite{Knapitsch2017} (train, truck), and Deep Blending~\cite{HPPFDB18} (drjohnson, playroom).


\subsection{Implementation Details}

\noindent\textbf{Baseline. }We implement our method based on LiteGS~\cite{liao2025litegshighperformanceframeworktrain} \texttt{stable} branch. 
Unlike the original 3DGS codebase that samples images randomly for a fixed 30K iterations, LiteGS organizes training into epochs where each epoch processes the entire dataset once. 
To ensure consistent training dynamics across scenes with varying numbers of input views, we standardize each epoch to process exactly 200 images through random sampling with replacement. For scenes with fewer than 200 images, some images are sampled multiple times per epoch; for scenes with more, we sample a subset each time. We train all scenes for 150 epochs (30K total iterations), following the original 3DGS.


\noindent\textbf{Resolution Scheduler. }We adopt the resolution scheduler from DashGaussian~\cite{chen2025dashgaussian} to further improve efficiency. The scheduler accelerates training through coarse-to-fine rendering with downsampling factor $r \ge 1$, progressively training from low resolution (large $r$) to full resolution ($r = 1$). However, excessively large $r$ can paradoxically increase training time in tile-based rasterization. At very low resolutions, each tile covers a large scene portion, creating excessive Gaussian overlaps and long per-tile lists that limit GPU parallelism. We empirically observe that efficiency degrades when per-tile Gaussian counts exceed approximately 150. Therefore, we adaptively determine $r_{\text{max}}$ to maintain counts below this threshold, capping $r_{\text{max}}$ at 4.

\noindent\textbf{Scale Reset and Entropy Constraint. }We apply scale reset every 20 epochs with shrinking factor $\zeta = 0.2$, and apply entropy constraint every other epoch during each scale-reset period with weight $\gamma=0.015$. When integrating with the resolution scheduler, we adopt a stage-adaptive strategy which applies weaker regularization during early coarse-resolution stages to preserve scene structures, then increases both $\zeta$ and $\gamma$ in later full-resolution epochs.


\subsection{Results}

\begin{figure*}[t]
    \centering
    
    \begin{subfigure}[t]{0.195\textwidth}
        \centering
        \includegraphics[width=\linewidth]{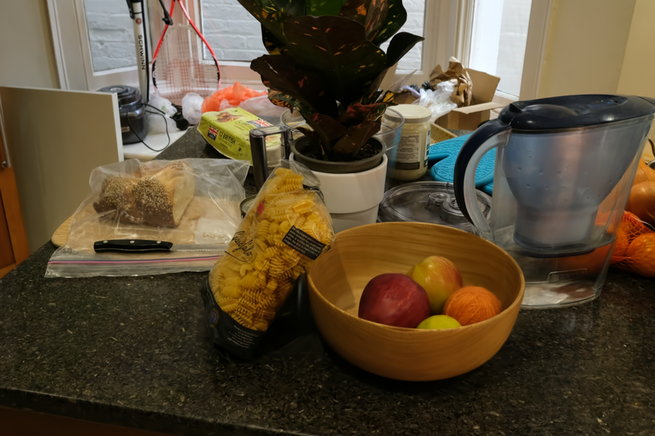}
        \caption{\scriptsize 3DGS (30.91dB / 821s)}
    \end{subfigure}
    \begin{subfigure}[t]{0.195\textwidth}
        \centering
        \includegraphics[width=\linewidth]{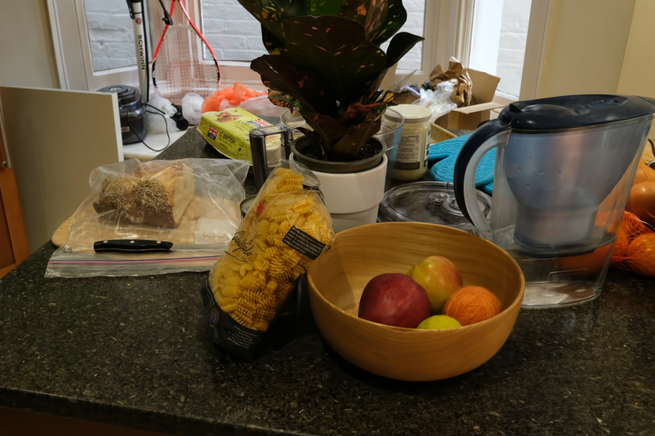}
        \caption{\scriptsize Taming (31.10dB / 298s)}
    \end{subfigure}
    \begin{subfigure}[t]{0.195\textwidth}
        \centering
        \includegraphics[width=\linewidth]{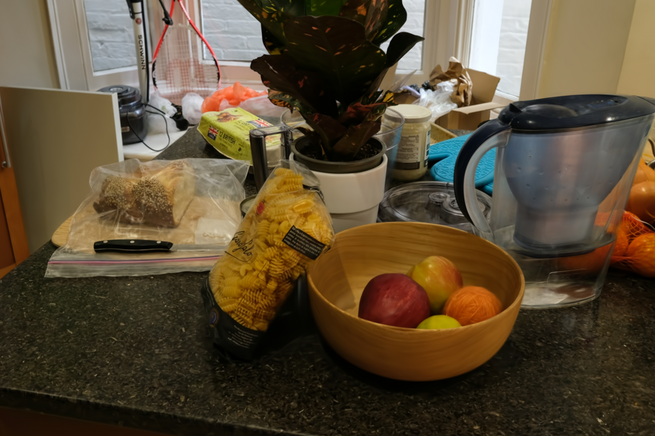}
        \caption{\scriptsize DashGaussian(30.69dB/158s)}
    \end{subfigure}
    \begin{subfigure}[t]{0.195\textwidth}
        \centering
        \includegraphics[width=\linewidth]{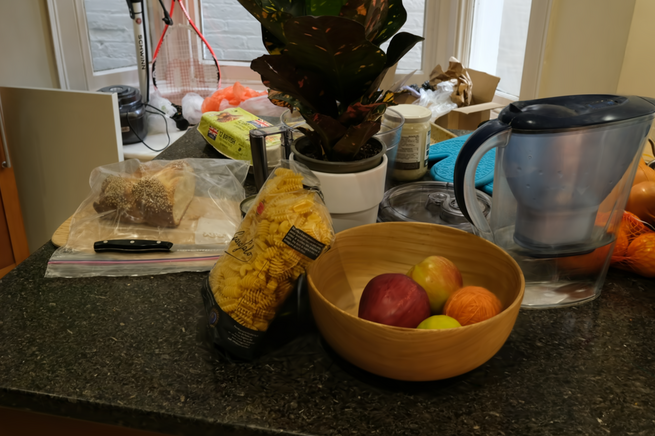}
        \caption{\scriptsize LiteGS (30.72dB / 179s)}
    \end{subfigure}
    \begin{subfigure}[t]{0.195\textwidth}
        \centering
        \includegraphics[width=\linewidth]{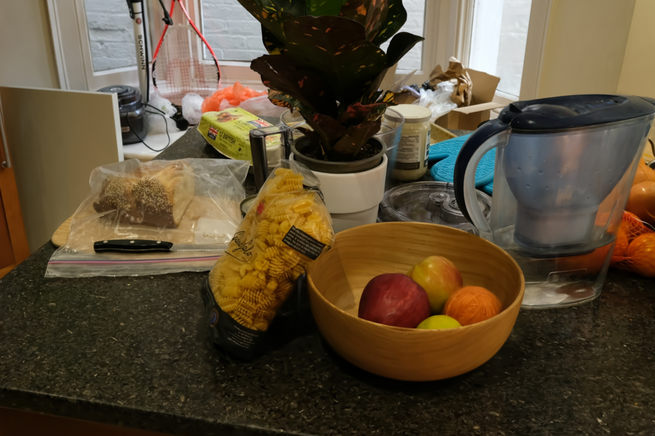}
        \caption{\scriptsize Ours (30.38dB / 86s)}
    \end{subfigure}\\
    
    \begin{subfigure}[t]{0.195\textwidth}
        \centering
        \includegraphics[width=\linewidth]{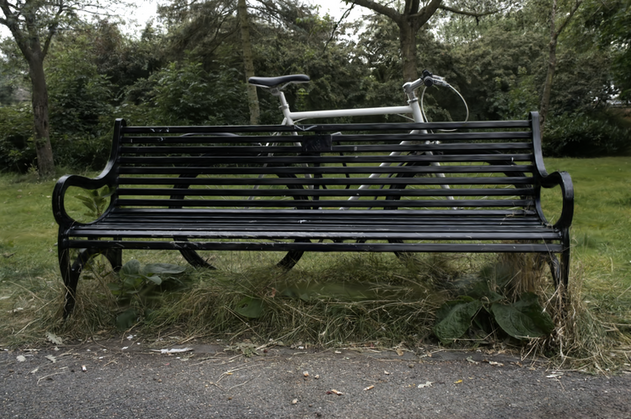}
        \caption{\scriptsize 3DGS (21.16dB / 1182s)}
    \end{subfigure}
    \begin{subfigure}[t]{0.195\textwidth}
        \centering
        \includegraphics[width=\linewidth]{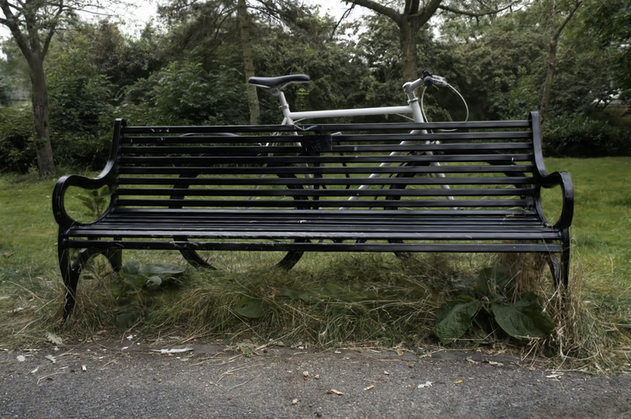}
        \caption{\scriptsize Taming (21.64dB / 598s)}
    \end{subfigure}
    \begin{subfigure}[t]{0.195\textwidth}
        \centering
        \includegraphics[width=\linewidth]{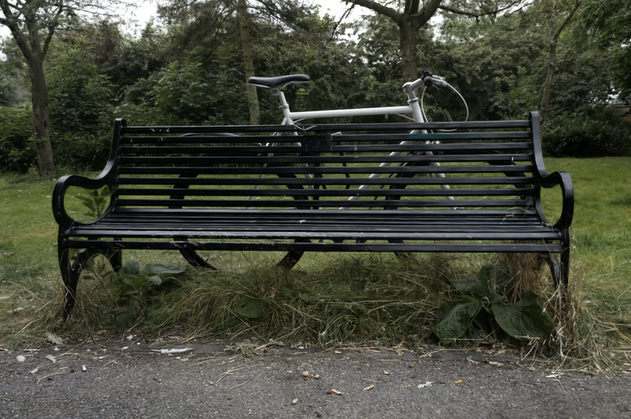}
        \caption{\scriptsize DashGaussian(21.16dB/293s)}
    \end{subfigure}
    \begin{subfigure}[t]{0.195\textwidth}
        \centering
        \includegraphics[width=\linewidth]{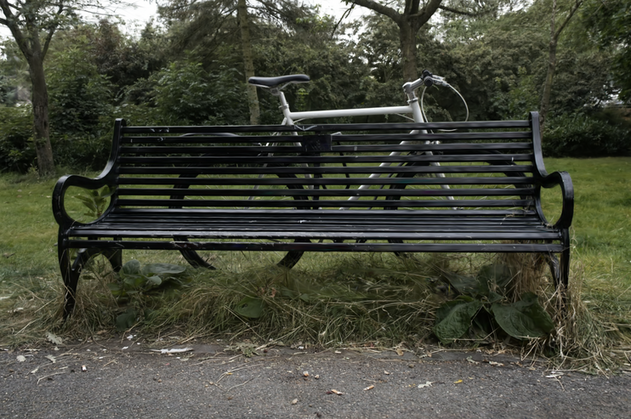}
        \caption{\scriptsize LiteGS (21.27dB / 238s)}
    \end{subfigure}
    \begin{subfigure}[t]{0.195\textwidth}
        \centering
        \includegraphics[width=\linewidth]{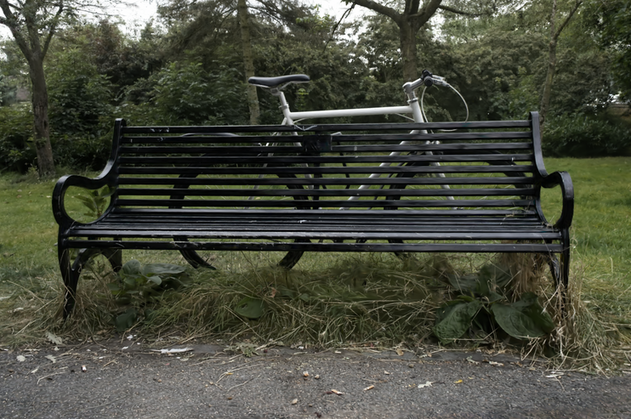}
        \caption{\scriptsize Ours (21.29dB / 115s)}
    \end{subfigure}
    
    \vspace{-1mm}
    \caption{Qualitative comparison of rendered results. Each row shows results from 3DGS, Taming 3DGS, DashGaussian, LiteGS, and ours with PSNR and training time. More comparisons are detailed in \cref{fig:rendered_comparison1}, \cref{fig:rendered_comparison2}, \cref{fig:rendered_comparison3} and \cref{fig:rendered_comparison4} of the supplementary material.}
    \label{fig:rendered_comparisons_overview}
    \vspace{-2mm}
\end{figure*}

\begin{figure*}[t]
    \centering
    \includegraphics[width=\textwidth]{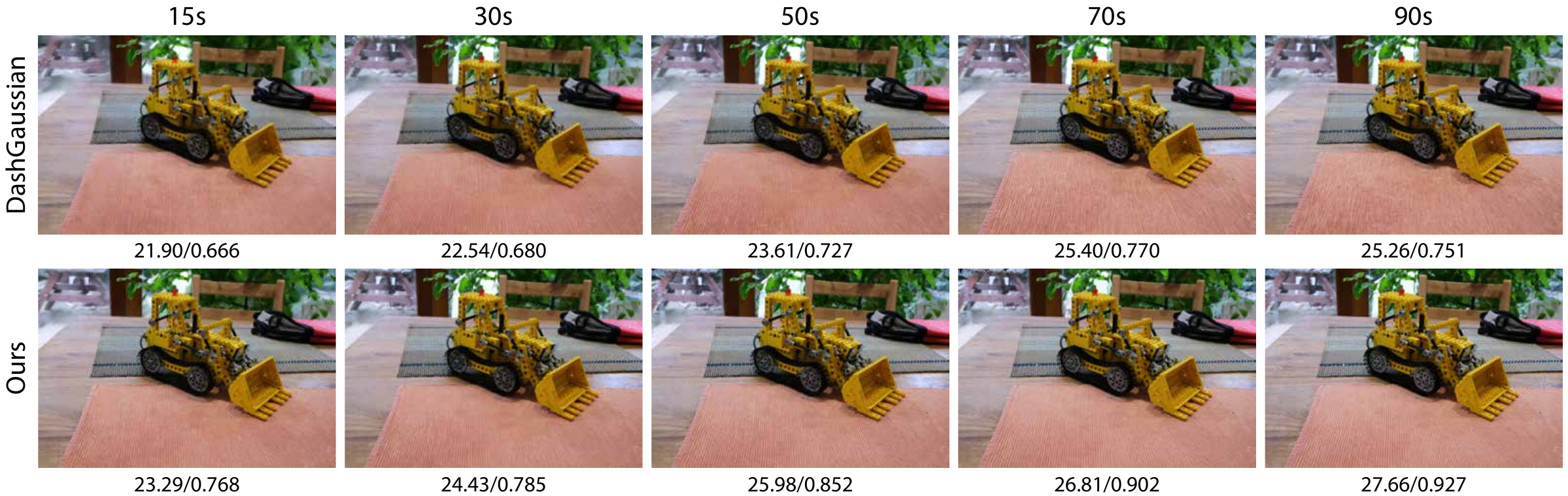}
    \vspace{-5mm}
    \caption{Qualitative comparison of rendering results, comparing our method with DashGaussian. Each row shows results at certain training times with PSNR and SSIM metrics displayed. More comparisons are detailed in \cref{fig:render_process_p2} of the supplementary material.}
    \label{fig:render_process_p1}
    \vspace{-4mm}
\end{figure*}

We present quantitative results in \cref{tab:main_quantitative_comparison_33M}.
Following the evaluation protocol established by Taming-3DGS~\cite{mallick2024taming} and LiteGS~\cite{liao2025litegshighperformanceframeworktrain}, we fix the Gaussian count for each dataset to ensure fair comparison.
We report the visual comparisons in \cref{fig:rendered_comparisons_overview} and \cref{fig:render_process_p1}.
All experiments were conducted on a single NVIDIA GeForce RTX 5090 D GPU. 

\begin{figure}[t]
  \centering
   \includegraphics[width=1.0\linewidth]{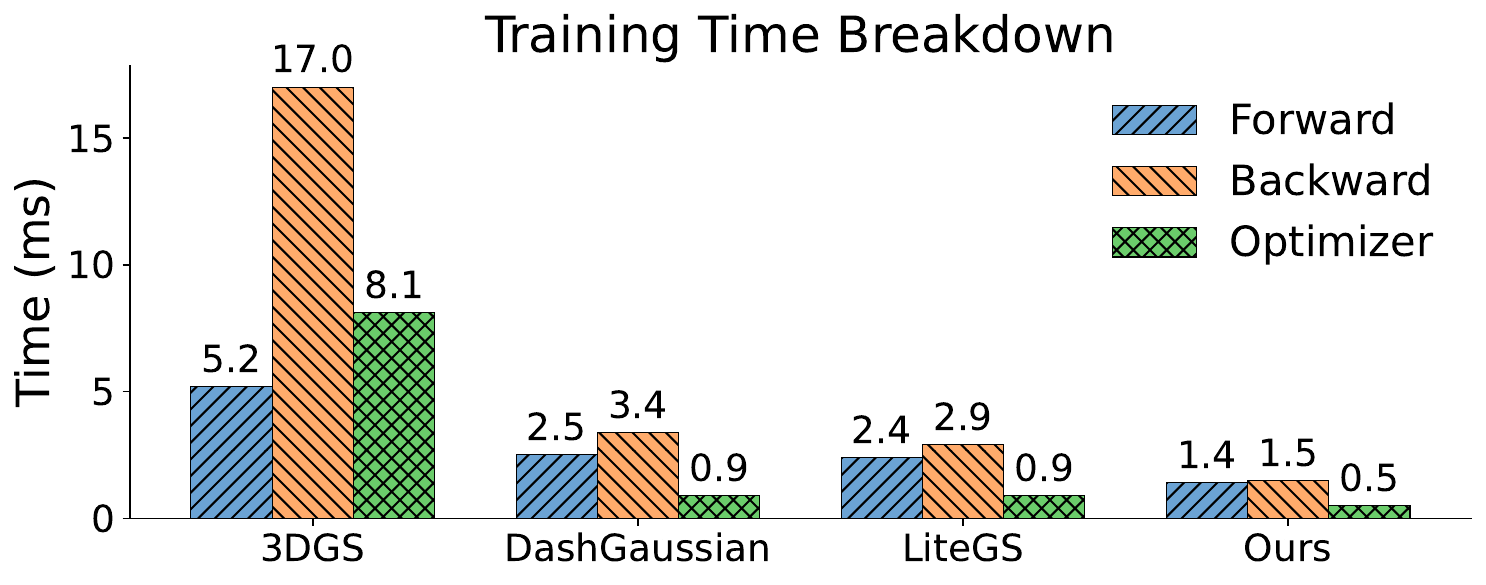}
   \vspace{-8mm}
   \caption{Per-iteration training time breakdown.}
   \label{fig:timing_breakdown}
   \vspace{-6mm}
\end{figure}


Our method substantially accelerates training across all datasets, achieving speedups of 9.2 times on Mip-NeRF~360 (99.58s vs. 919.51s), 11.9 times on Deep Blending (80.68s vs. 963.66s), and 5.3 times on Tanks and Temples (106.06s vs. 560.52s) compared to 3DGS. Relative to the recent LiteGS baseline, our method delivers nearly 50\% speedup.

The speedup comes with modest quality trade-offs. On Mip-NeRF~360, our PSNR of 27.28 dB is comparable to 3DGS (27.55 dB) and LiteGS (27.75 dB). Similar patterns hold for Deep Blending and Tanks and Temples, where perceptual metrics show minimal degradation. The more modest acceleration on Tanks and Temples is expected, as this dataset features smaller images and fewer Gaussians.

\begin{figure}[t]
  \centering
  \vspace{-2mm}
  \includegraphics[width=1.0\linewidth]{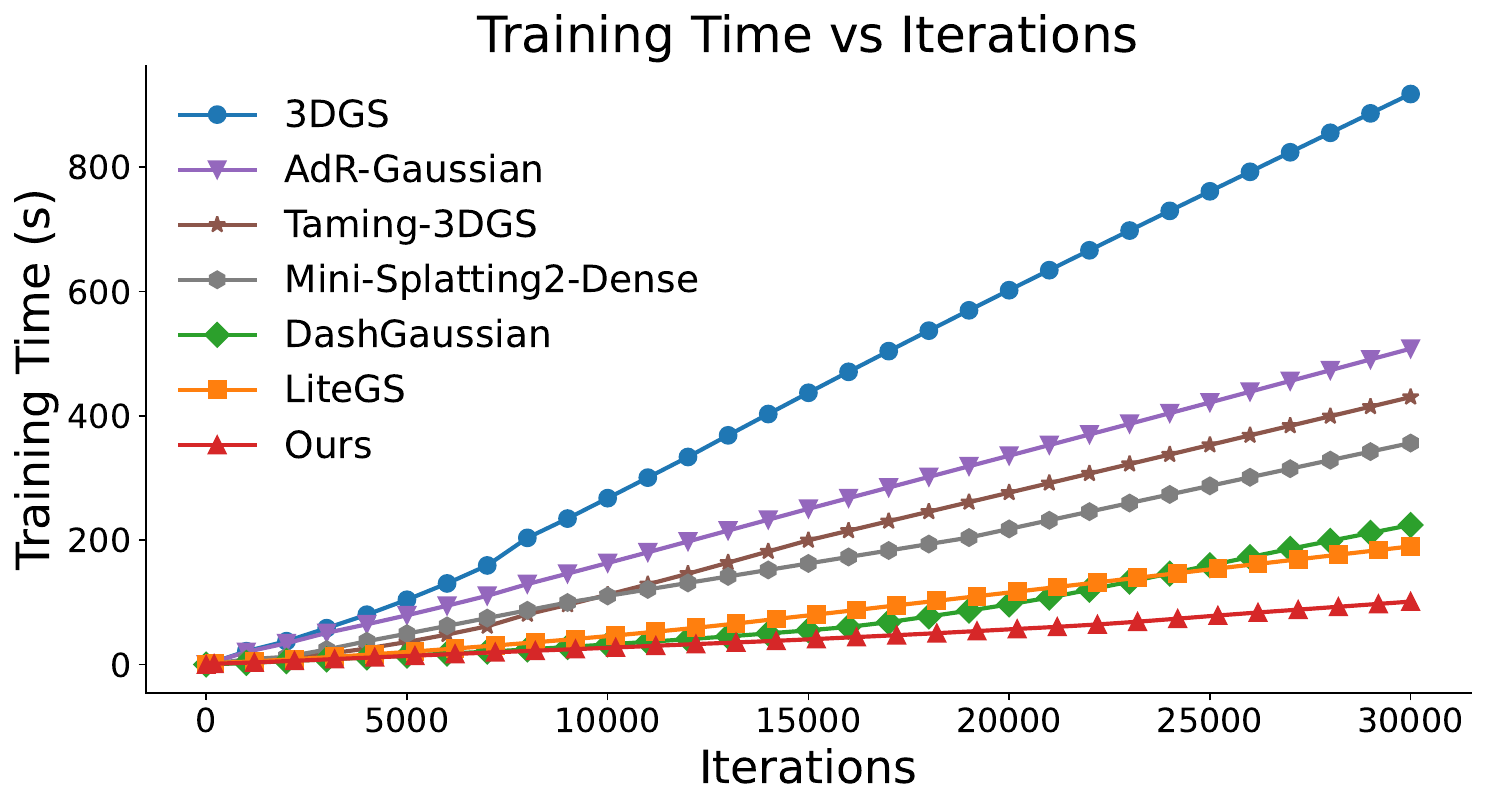}
  \vspace{-8mm}
  \caption{Training time versus iterations. Plot for compact models are shown in Supplementary \cref{fig:supplementary_time_vs_iter}.}
  \label{fig:time_vs_iter}
  \vspace{-4mm}
\end{figure}

\begin{figure}[t]
  \centering
  \includegraphics[width=1.0\linewidth]{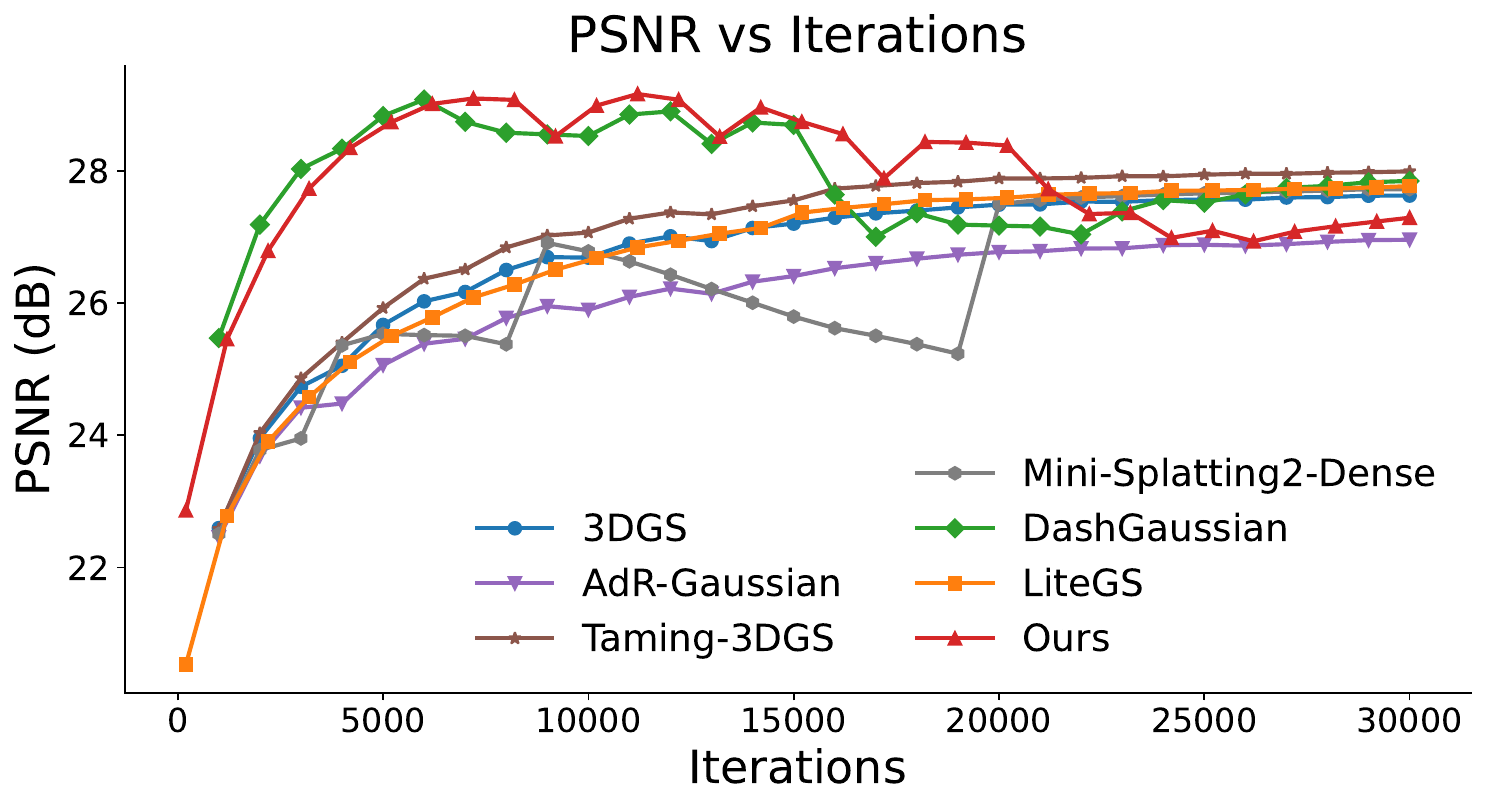}
  \vspace{-8mm}
  \caption{PSNR convergence across iterations. Plot for compact models is shown in Supplementary \cref{fig:supplementary_psnr_vs_iter}.}
  \label{fig:psnr_vs_iter}
  \vspace{-4mm}
\end{figure}

We show efficiency gains from shorter Gaussian lists through timing breakdowns in \cref{fig:timing_breakdown}. 
Training efficiency across iterations in \cref{fig:time_vs_iter} shows our method converges fastest among all compared approaches. 
For PSNR progression in \cref{fig:psnr_vs_iter}, both DashGaussian and ours use resolution scheduling, resulting in higher PSNR at lower resolutions than other methods. 
We visualize list lengths in individual views in \cref{fig:gaussian_count_viz}, demonstrating our shorter Gaussian lists.

\begin{table}[t]
\centering
\caption{
Quantitative comparison with FastGS on Mip-NeRF~360 under FastGS's default setting (0.4M Gaussians).
}
\vspace{-2mm}
\label{tab:comparison_fastgs}
\scriptsize
\begin{tabular}{l|c|c|ccc|c}
\toprule
\textbf{Method} & Iters & $N_{\text{G}}$ & PSNR$\uparrow$ & SSIM$\uparrow$ & LPIPS$\downarrow$ & Time(s)$\downarrow$ \\
\midrule
FastGS & 30K & 0.4M & \textbf{27.43} & \textbf{0.797} & \textbf{0.261} & 88.55 \\
\midrule
\textbf{Ours} & 30K & 0.4M & 26.50 & 0.754 & 0.314 & \textbf{66.32} \\
\bottomrule
\end{tabular}
\vspace{-6mm}
\end{table}

We additionally evaluate our method in a resource-constrained regime following the setup of Mini-Splatting2~\cite{fang2024miniv2} (18K iterations, much fewer Gaussians) in \cref{tab:main_quantitative_comparison_06M}.
Our method achieves up to 2.1 times speedup over Mini-Splatting2. While quality is lower than our 30K results in \cref{tab:main_quantitative_comparison_33M}, this gap reflects LiteGS backbone limitations under constraints rather than our acceleration strategy, demonstrating robustness across regimes.
We further compare with FastGS~\cite{ren2025fastgs} under its default setting (0.4M Gaussians) in \cref{tab:comparison_fastgs}, where our method achieves 1.4 times faster training.


\subsection{Ablation Study and Analysis}
\label{sec:ablation_study}

All ablation studies are conducted on Mip-NeRF~360 with $N_G = 3.3$M. All results are averaged across the nine scenes in the dataset. We use the following notations: L: LiteGS, D: DashGaussian, R($\zeta$): scale reset, E($\gamma$): entropy constraint, V($\eta$): volume regularization, and O($\xi$): opacity regularization. Our default parameters are $\zeta=0.2, \gamma=0.015$.

\noindent\textbf{Effect of Modules. }We report results with either one or both modules in \cref{tab:ablation_individual_mipnerf360}. Both the scale reset and entropy constraint accelerate 3D Gaussian learning, with or without DashGaussian's resolution scheduler. Each module provides incremental improvements when combined. Detailed curves are shown in \cref{fig:supplementary_psnr_R_E_ablation} of the supplementary material, confirming the effectiveness of both modules in each scene.


\begin{table}[t]
\centering
\caption{
Ablation study analyzing effects of individual modules.
}
\vspace{-2mm}
\label{tab:ablation_individual_mipnerf360}
\scriptsize
\begin{tabular}{l|cccc}
\toprule
\multirow{2}{*}{\textbf{Method}} & \multicolumn{4}{c}{\textbf{Mip-NeRF~360}} \\
\cmidrule(lr){2-5}
& PSNR$\uparrow$ & SSIM$\uparrow$ & LPIPS$\downarrow$ & Time(s)$\downarrow$ \\
\midrule
L & 27.75 & 0.822 & 0.208 & 191.17 \\
~~L+R & 27.33 & 0.815 & 0.212 & 147.33 \\
~~L+E & 27.35 & 0.815 & 0.218 & 162.53 \\
~~L+R+E & 27.14 & 0.812 & 0.215 & 141.28 \\
L+D & 27.85 & 0.822 & 0.213 & 134.99 \\
~~L+D+R & 27.52 & 0.815 & 0.219 & 108.62 \\
~~L+D+E & 27.38 & 0.813 & 0.223 & 112.13 \\
~~\textbf{L+D+R+E} & 27.28 & 0.810 & 0.224 & \textbf{99.58} \\
\bottomrule
\end{tabular}
\vspace{-2mm}
\end{table}


\begin{table}[t]
\centering
\caption{
Ablation study on parameter selection.
}\vspace{-2mm}
\label{tab:ablation_mipnerf360}
\scriptsize
\begin{tabular}{l|cccc}
\toprule
\multirow{2}{*}{\textbf{Method}} & \multicolumn{4}{c}{\textbf{Mip-NeRF~360}} \\
\cmidrule(lr){2-5}
& PSNR$\uparrow$ & SSIM$\uparrow$ & LPIPS$\downarrow$ & Time(s)$\downarrow$ \\
\midrule
L+D+R(0.2)+E(0.01) & 27.35 & 0.811 & 0.222 & 102.86 \\
L+D+R(0.2)+E(0.02) & 27.14 & 0.808 & 0.226 & 99.10 \\
L+D+R(0.3)+E(0.015) & 27.31 & 0.811 & 0.223 & 104.39 \\
L+D+R(0.1)+E(0.015) & 27.14 & 0.807 & 0.227 & 98.34 \\
\textbf{L+D+R(0.2)+E(0.015)} & 27.28 & 0.810 & 0.224 & \textbf{99.58} \\
\bottomrule
\end{tabular}
\vspace{-4mm}
\end{table}


\noindent\textbf{Effect of Parameters. }Both techniques exhibit a speed-quality trade-off. More aggressive regularization (lower $\zeta$ or higher $\gamma$) accelerates training by reducing Gaussian overlap and per-pixel list lengths, but can degrade quality if over-applied. As shown in \cref{tab:ablation_mipnerf360}, $\zeta = 0.2$ and $\gamma = 0.015$ achieve favorable balances, providing substantial speedup while maintaining quality. 
DashGaussian provides further acceleration and combines effectively with both techniques.


\noindent\textbf{Superiority over Alternatives. }We compare scale reset with volume regularization in \cref{tab:ablation_mipnerf360_volume}. Scale reset achieves both better rendering quality and substantially faster training. This validates our design choice, as periodic scale reset provides immediate geometric regularization that accelerates training while improving visual quality through direct manipulation rather than gradient-based constraints.


Dai and Xu~\etal~\cite{Dai2024GaussianSurfels} proposed an opacity loss that encourages Gaussian opacity toward extreme values (0 or 1), $\mathcal{L}_{\text{O}} = \exp\left(-(\sigma-0.5)^2 / 0.05\right)$.
As shown in \cref{tab:ablation_mipnerf360_opacity}, our entropy constraint achieves greater acceleration while maintaining better quality.
This is because entropy operates on blending weights that depend on multiple Gaussian attributes (opacity, scale, position, and rotation). In contrast, constraining only opacity with extreme values is overly restrictive, resulting in inferior speedup.


\begin{table}[t]
\centering
\caption{
Comparison of volume regularization versus scale reset.
}\vspace{-2mm}
\label{tab:ablation_mipnerf360_volume}
\scriptsize
\begin{tabular}{l|cccc}
\toprule
\multirow{2}{*}{\textbf{Method}} & \multicolumn{4}{c}{\textbf{Mip-NeRF~360}} \\
\cmidrule(lr){2-5}
& PSNR$\uparrow$ & SSIM$\uparrow$ & LPIPS$\downarrow$ & Time(s)$\downarrow$ \\
\midrule
L+D+V(1.0)+E(0.015) & 27.17 & 0.808 & 0.230 & 107.91 \\
\textbf{L+D+R(0.2)+E(0.015)} & 27.28 & 0.810 & 0.224 & \textbf{99.58} \\
\bottomrule
\end{tabular}
\vspace{-2mm}
\end{table}

\begin{table}[t]
\centering
\caption{
Comparison of opacity versus entropy constraint.
}\vspace{-2mm}
\label{tab:ablation_mipnerf360_opacity}
\scriptsize
\begin{tabular}{l|cccc}
\toprule
\multirow{2}{*}{\textbf{Method}} & \multicolumn{4}{c}{\textbf{Mip-NeRF~360}} \\
\cmidrule(lr){2-5}
& PSNR$\uparrow$ & SSIM$\uparrow$ & LPIPS$\downarrow$ & Time(s)$\downarrow$ \\
\midrule
L+D+R(0.2)+O(3.0) & 27.29 & 0.807 & 0.231 & 106.16 \\
\textbf{L+D+R(0.2)+E(0.015)} & 27.28 & 0.810 & 0.224 & \textbf{99.58} \\
\bottomrule
\end{tabular}
\vspace{-2mm}
\end{table}

\begin{table}[t]
\centering
\caption{
Ablation study of tile size impact on time and quality.
}\vspace{-2mm}
\label{tab:ablation_mipnerf360_tile_size}
\scriptsize
\begin{tabular}{l|cccc}
\toprule
\multirow{2}{*}{\textbf{Method}} & \multicolumn{4}{c}{\textbf{Mip-NeRF~360}} \\
\cmidrule(lr){2-5}
& PSNR$\uparrow$ & SSIM$\uparrow$ & LPIPS$\downarrow$ & Time(s)$\downarrow$ \\
\midrule
3DGS (16$\times$16) & 27.55 & 0.819 & 0.209 & 919.51 \\
3DGS (8$\times$8) & 27.56 & 0.819 & 0.208 & 1004.30 \\
\bottomrule
\end{tabular}
\end{table}


\noindent\textbf{Effect of tile sizes.} Since 3DGS uses $16\times16$ tiles while LiteGS and our method use $8\times8$, we tested 3DGS with $8\times8$ tiles for fair comparison. As shown in \cref{tab:ablation_mipnerf360_tile_size}, the quality and training time remain similar. This is because smaller tiles produce shorter per-tile Gaussian lists but more tiles overall, yielding similar training times.

\begin{figure}[t]
  \centering
  \includegraphics[width=1.0\linewidth]{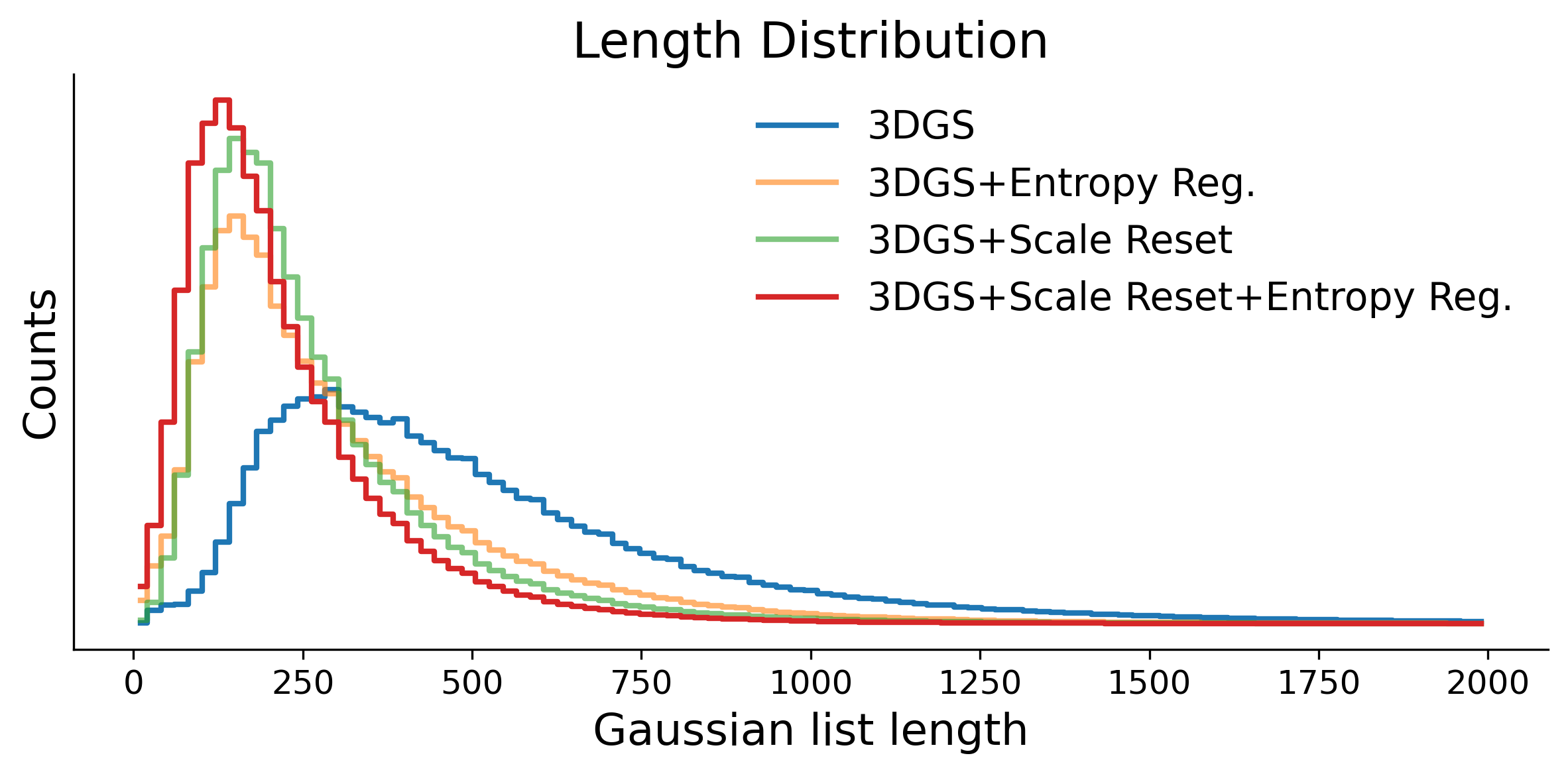}
  \vspace{-8mm}
  \caption{Comparison of Gaussian list length distributions across individual and combined modules.}
  \label{fig:ablation_length}
  \vspace{-4mm}
\end{figure}
\begin{figure}[t]
    \centering

    \begin{subfigure}[t]{0.115\textwidth}
        \centering
        \includegraphics[width=\linewidth]{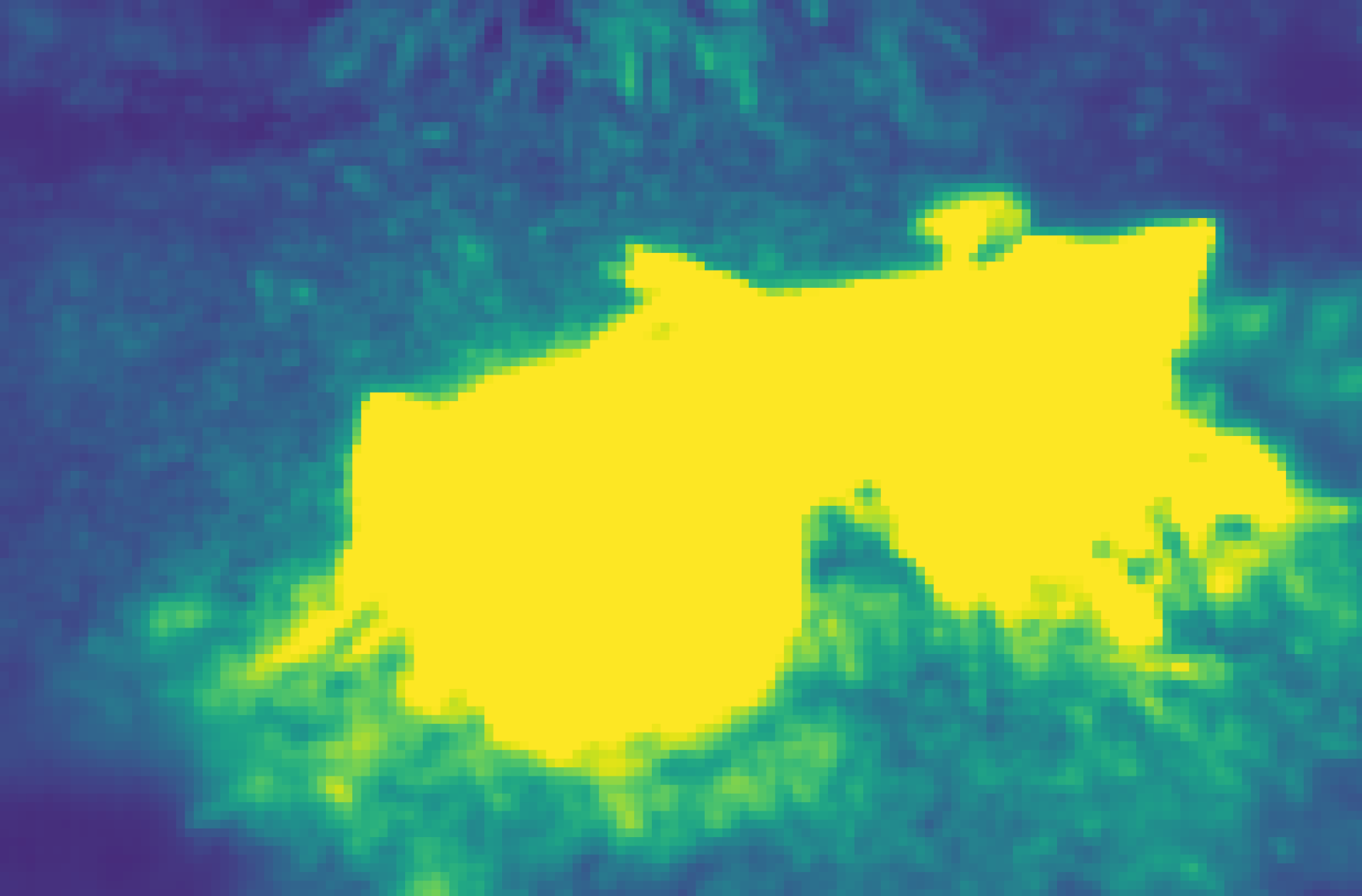}
    \end{subfigure}
    \begin{subfigure}[t]{0.115\textwidth}
        \centering
        \includegraphics[width=\linewidth]{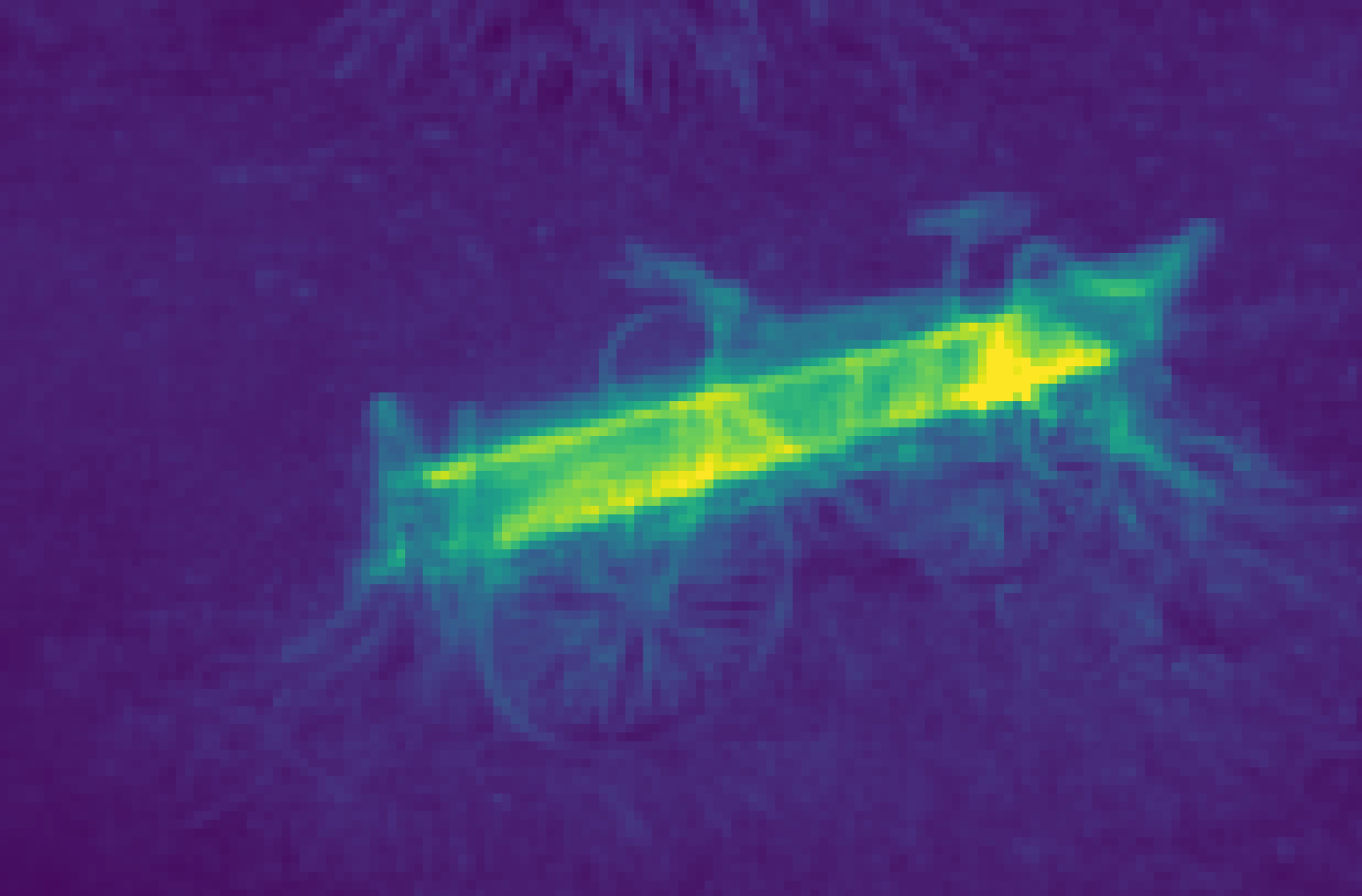}
    \end{subfigure}
    \begin{subfigure}[t]{0.115\textwidth}
        \centering
        \includegraphics[width=\linewidth]{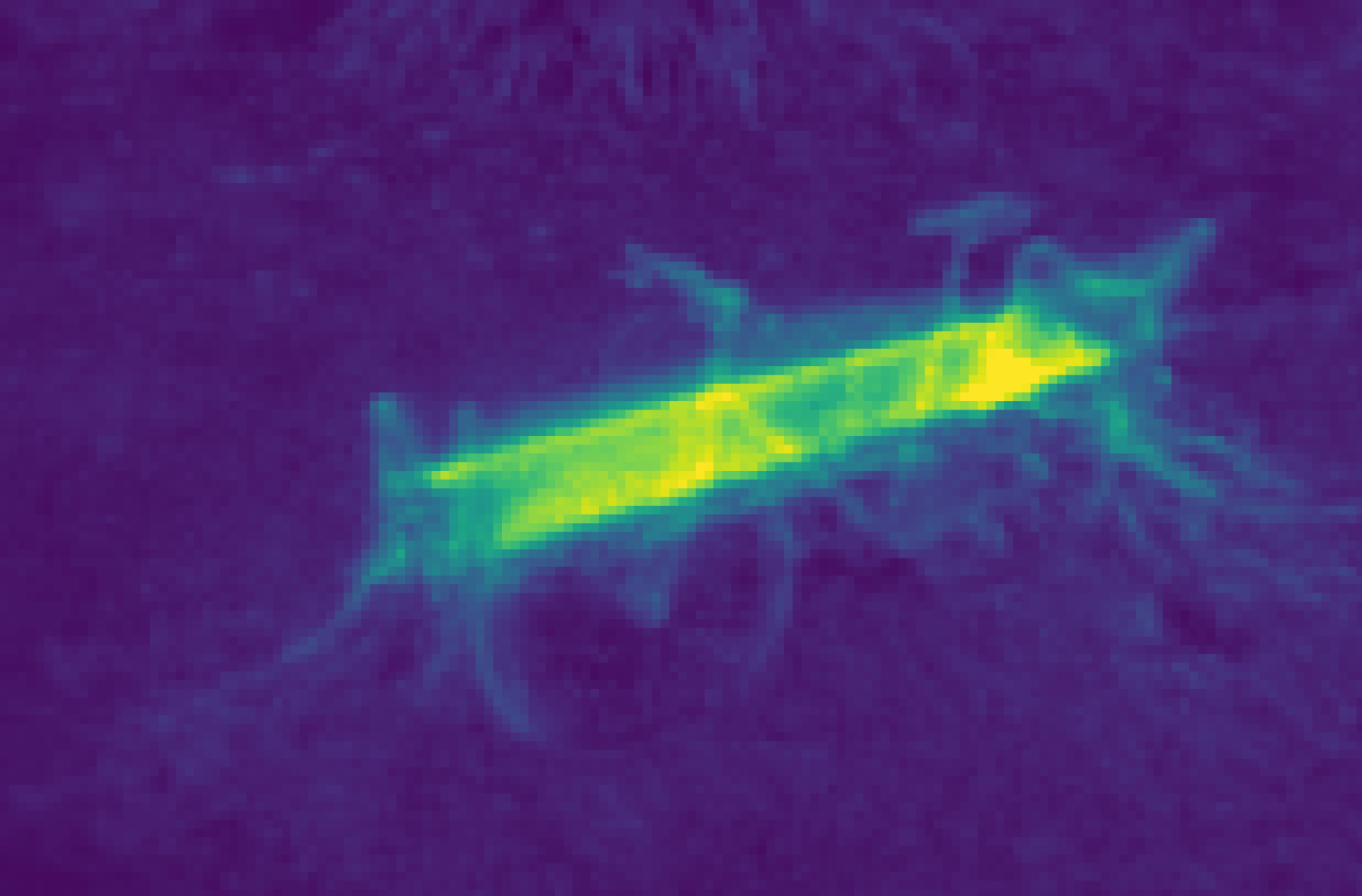}
    \end{subfigure}
    \begin{subfigure}[t]{0.115\textwidth}
        \centering
        \includegraphics[width=\linewidth]{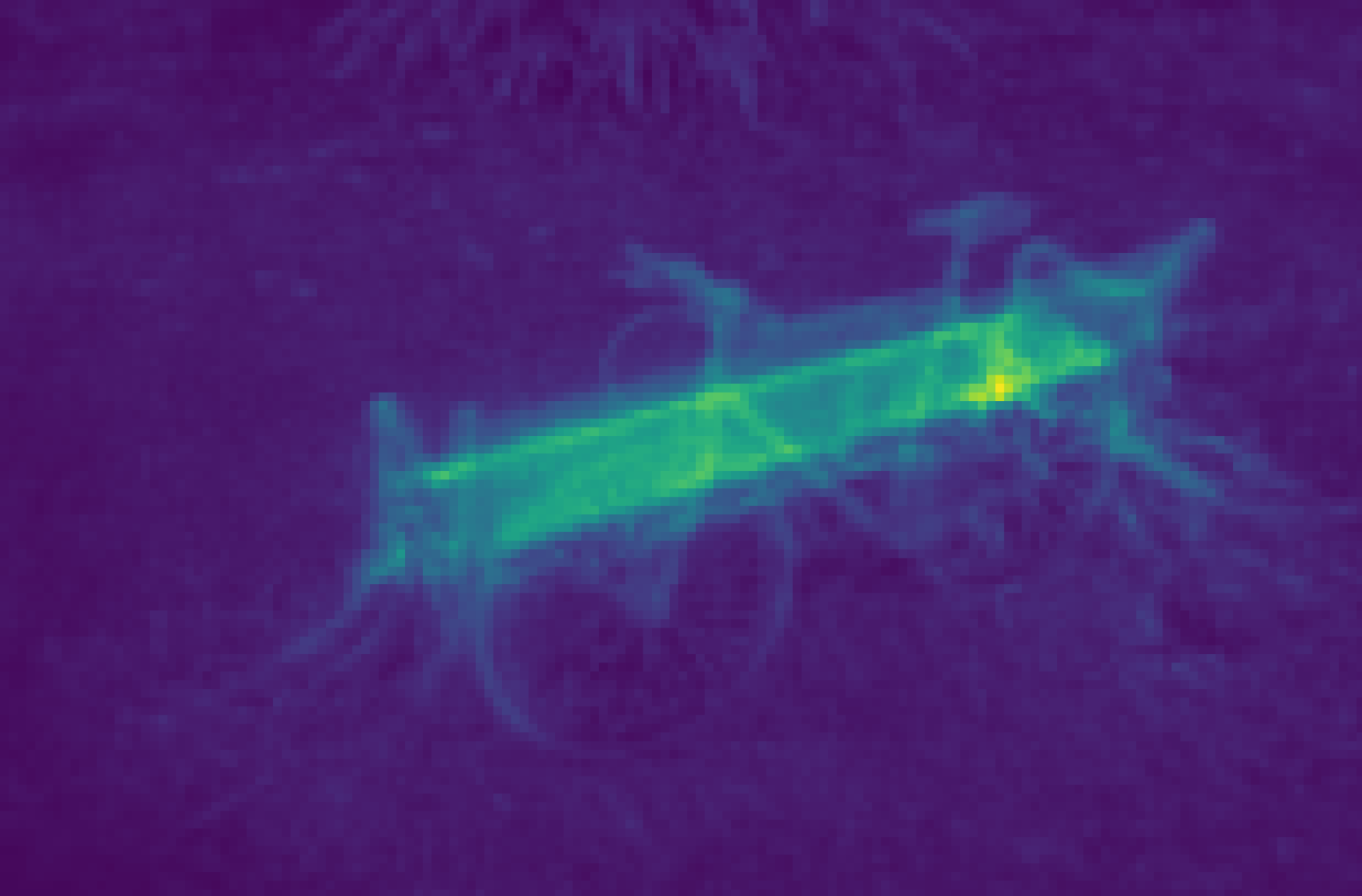}
    \end{subfigure}

    \begin{subfigure}[t]{0.115\textwidth}
        \centering
        \includegraphics[width=\linewidth]{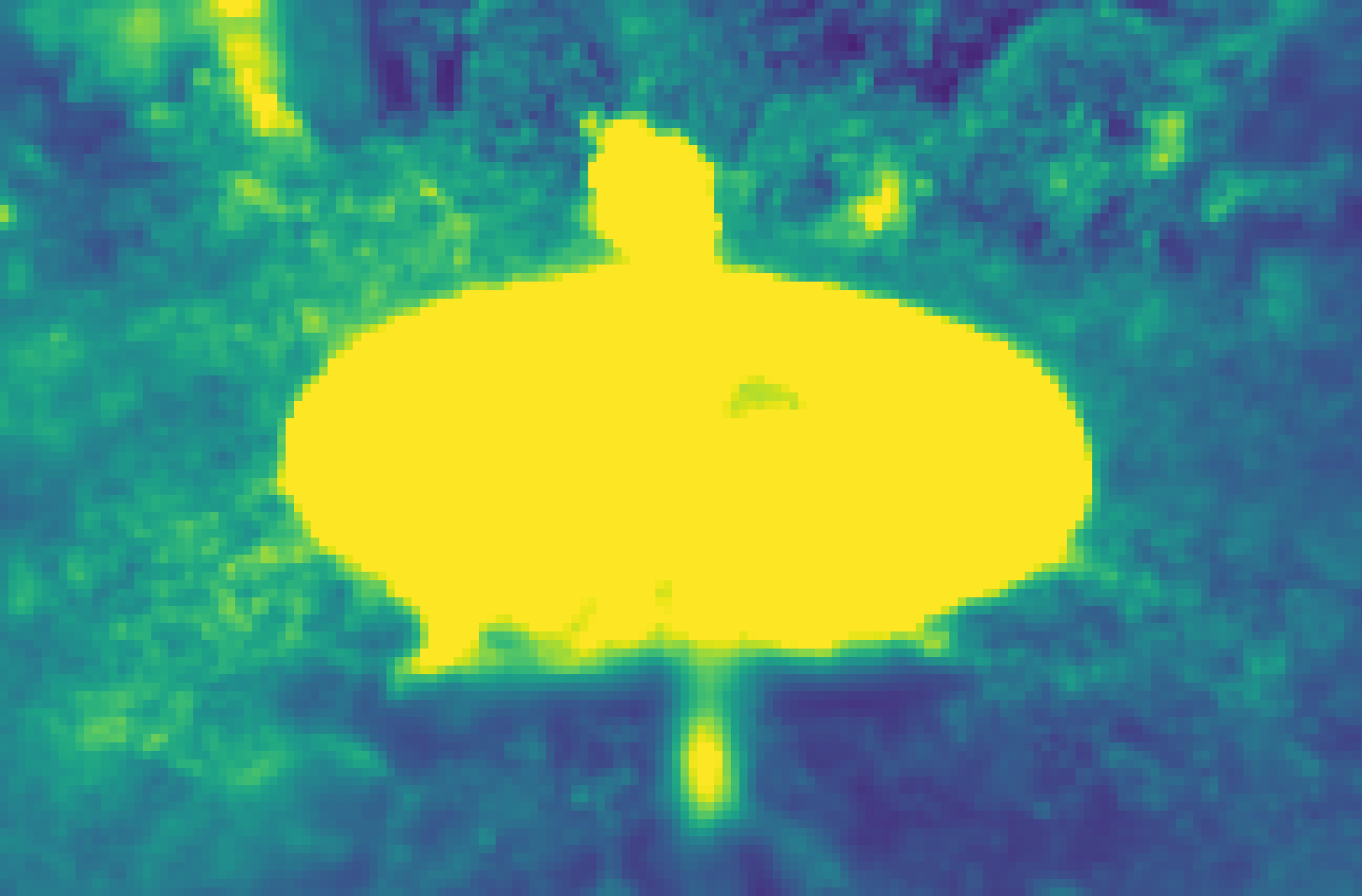}
        \caption*{\scriptsize 3DGS}
    \end{subfigure}
    \begin{subfigure}[t]{0.115\textwidth}
        \centering
        \includegraphics[width=\linewidth]{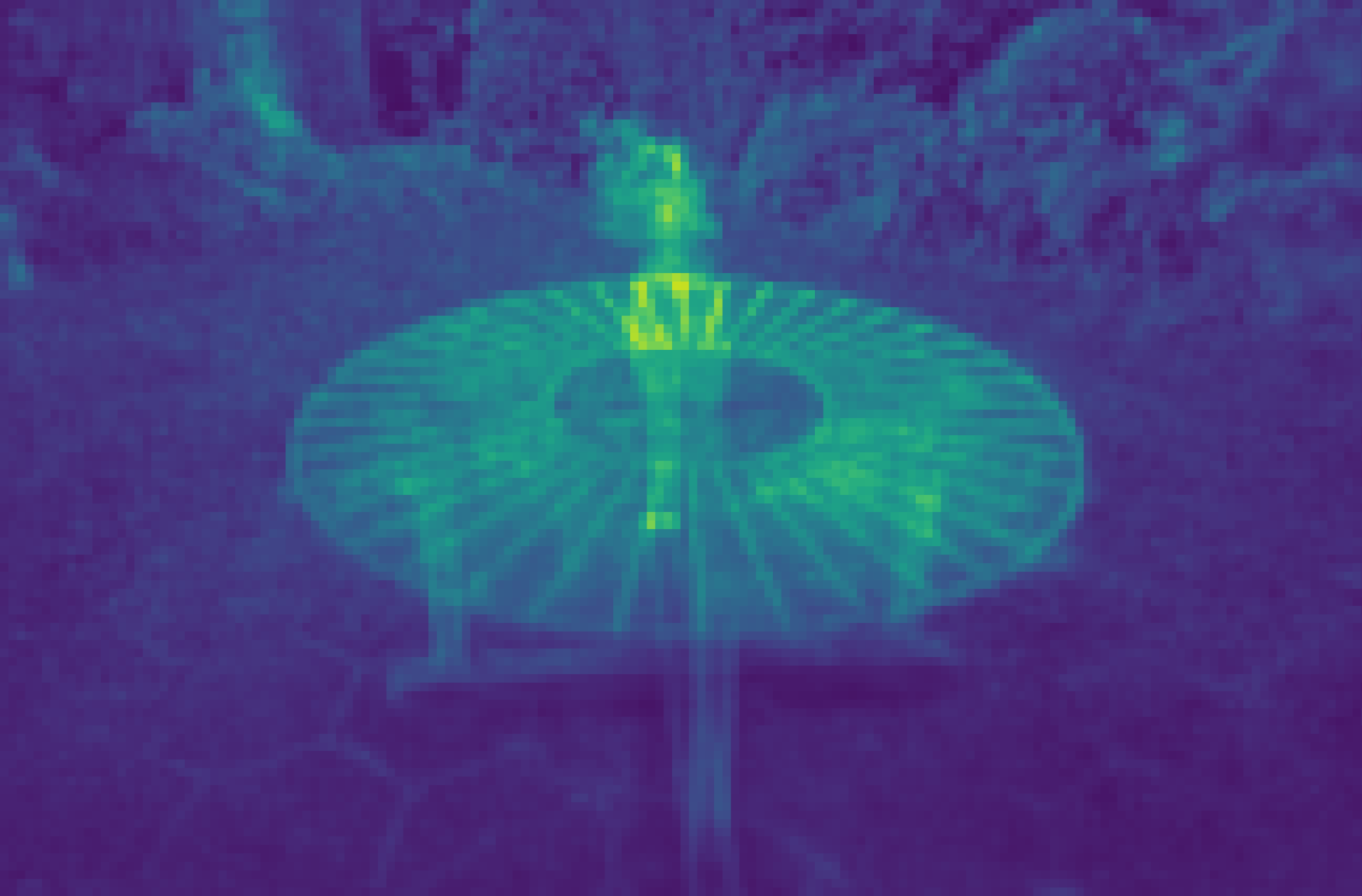}
        \caption*{\scriptsize 3DGS+Reset}
    \end{subfigure}
    \begin{subfigure}[t]{0.115\textwidth}
        \centering
        \includegraphics[width=\linewidth]{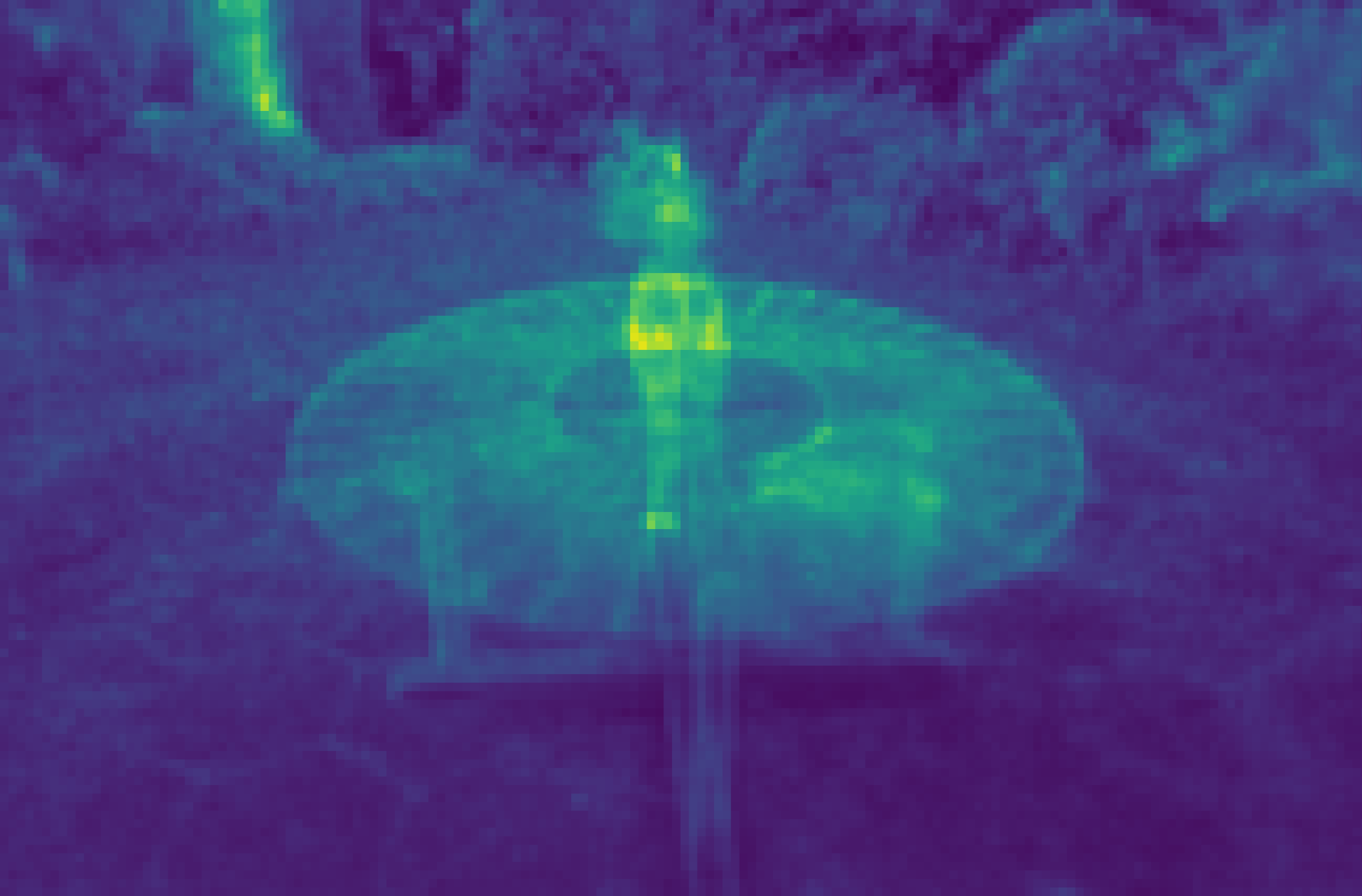}
        \caption*{\scriptsize 3DGS+Entropy}
    \end{subfigure}
    \begin{subfigure}[t]{0.115\textwidth}
        \centering
        \includegraphics[width=\linewidth]{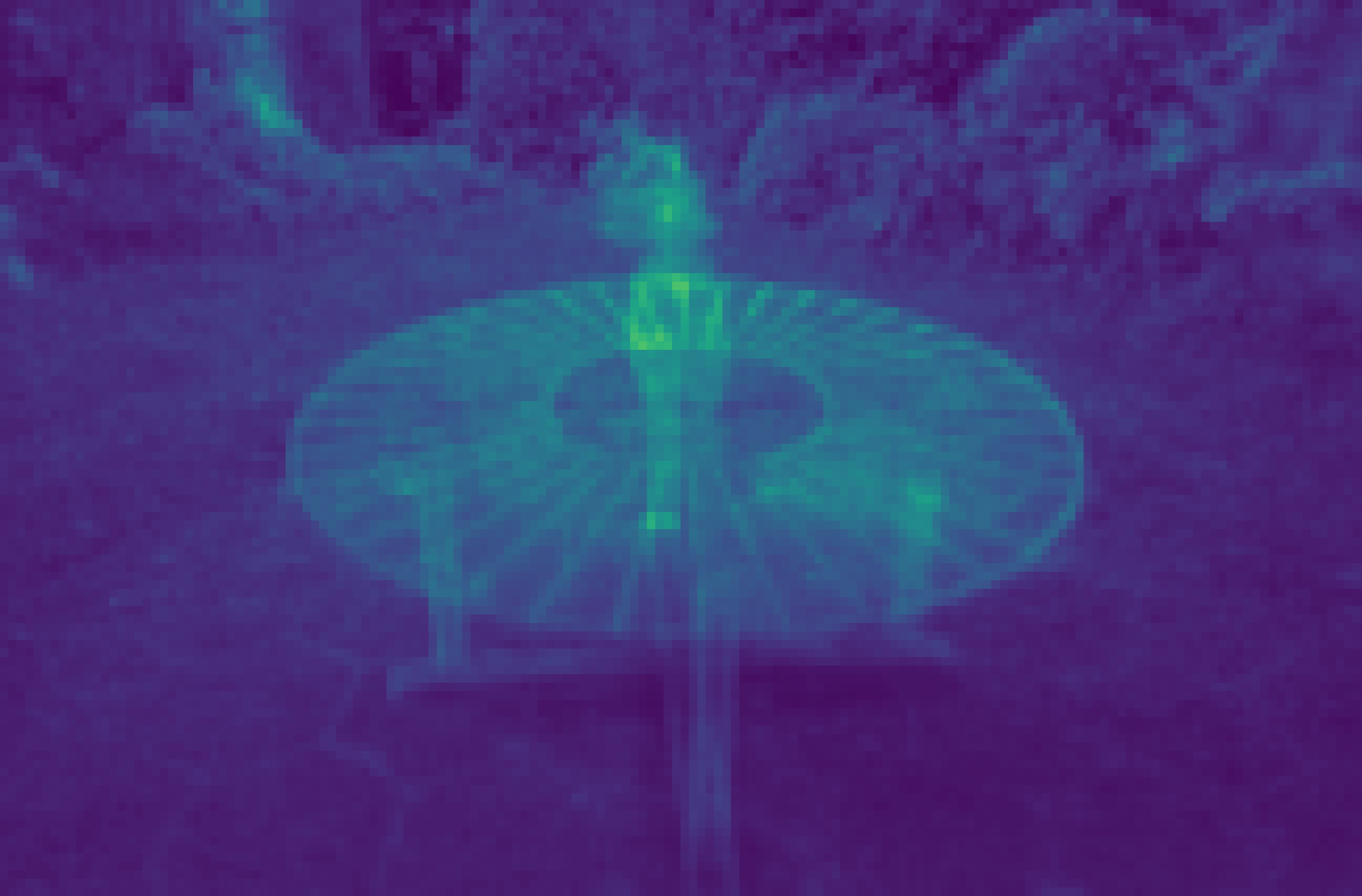}
        \caption*{\scriptsize 3DGS+R+E}
    \end{subfigure}
    
    \vspace{-3mm}
    \caption{Gaussian count heatmap comparison across modules. Colors represent per-tile Gaussian counts. See color bar in \cref{fig:gaussian_count_viz}.
    }
    \label{fig:gaussian_count_viz_ablation}
    \vspace{-4mm}
\end{figure}

\noindent\textbf{Length of Gaussian Lists. }We report Gaussian list lengths in \cref{fig:ablation_length}, measured as the average from the last 200 training iterations.
In this figure, results labeled as “3DGS” are obtained using LiteGS.
Both scale reset and entropy constraint significantly reduce list lengths, individually and jointly. 
Spatial visualizations are shown in \cref{fig:gaussian_count_viz_ablation}. Additional results in \cref{fig:supplementary_gaussian_count_viz_ablation} and per-scene distributions in \cref{fig:distributions_all_methods_all_metrics} are provided in the supplementary material, confirming consistent improvements across all scenes.

%% file: sec/stats_1.tex
\begin{table*}[t]
\centering
\caption{
Quantitative comparison across three datasets. 
We follow Taming-3DGS's target Gaussian count setup (3.3M, 2.8M, 1.8M for the three datasets respectively) for fair comparison.
For methods whose Gaussian number cannot be controlled, we report their default counts.
}\vspace{-1mm}
\label{tab:main_quantitative_comparison_33M}
\resizebox{\textwidth}{!}{
\begin{tabular}{l|c|ccccc|ccccc|ccccc}
\toprule
{\multirow{2}{*}{\textbf{Method}}} & \multirow{2}{*}{\textbf{Iters}} & \multicolumn{5}{c|}{\textbf{Mip-NeRF~360}} & \multicolumn{5}{c|}{\textbf{Deep Blending}} & \multicolumn{5}{c}{\textbf{Tanks and Temples}} \\
\cmidrule(lr){3-7} \cmidrule(lr){8-12} \cmidrule(lr){13-17}
& & $N_{\text{G}}$ & PSNR$\uparrow$ & SSIM$\uparrow$ & LPIPS$\downarrow$ & Time(s)$\downarrow$ & $N_{\text{G}}$ & PSNR$\uparrow$ & SSIM$\uparrow$ & LPIPS$\downarrow$ & Time(s)$\downarrow$ & $N_{\text{G}}$ & PSNR$\uparrow$ & SSIM$\uparrow$ & LPIPS$\downarrow$ & Time(s)$\downarrow$ \\
\midrule
AdR-Gaussian & 30K & 1.3M & 26.92 & 0.792 & 0.257 & 504.86 & 1.4M & 29.72 & 0.905 & 0.250 & 515.78 & 0.9M & 23.49 & 0.835 & 0.202 & 321.08 \\
EDGS & 5K & 3.5M & 26.46 & 0.817 & 0.205 & 318.06 & 2.3M & 29.08 & 0.898 & 0.260 & 214.76 & 2.2M & 22.23 & 0.831 & 0.184 & 219.06 \\
Mini-Splatting2-Dense & 18K & 3.6M & 27.56 & \textbf{0.827} & \textbf{0.184} & 220.22 & 4.2M & 29.78 & 0.908 & \textbf{0.222} & 162.38 & 2.6M & 23.43 & 0.852 & \textbf{0.152} & 162.24 \\
\midrule
3DGS & \multirow{4}{*}{\centering 30K} & \multirow{4}{*}{\centering 3.3M} & 27.55 & 0.819 & 0.209 & 919.51 & \multirow{4}{*}{\centering 2.8M} & 29.74 & 0.907 & 0.237 & 963.66 & \multirow{4}{*}{\centering 1.8M} & 23.77 & 0.854 & 0.166 & 560.52 \\
Taming-3DGS & & & \textbf{27.85} & 0.823 & 0.208 & 402.54 & & \textbf{30.07} & \textbf{0.910} & 0.234 & 322.06 & & \textbf{24.14} & \textbf{0.855} & 0.169 & 243.25 \\
DashGaussian & & & 27.84 & 0.824 & 0.203 & 218.85 & & 29.84 & 0.907 & 0.241 & 144.81 & & 23.97 & \textbf{0.855} & 0.167 & 173.41 \\
LiteGS & & & 27.75 & 0.822 & 0.208 & 191.17 & & 29.64 & 0.906 & 0.247 & 153.14 & & 23.93 & 0.848 & 0.178 & 157.56 \\
\midrule
\textbf{Ours} & 30K & 3.3M & 27.28 & 0.810 & 0.224 & \textbf{99.58} & 2.8M & 29.41 & 0.897 & 0.249 & \textbf{80.68} & 1.8M & 23.34 & 0.843 & 0.172 & \textbf{106.06} \\
\bottomrule
\end{tabular}
}
\end{table*}


\begin{table*}[t]
\centering
\vspace{-1mm}
\caption{
Quantitative comparison with 18K iterations and fewer Gaussians, following Mini-Splatting2's setup. 
The performance gap compared to \cref{tab:main_quantitative_comparison_33M} stems from LiteGS backbone limitations under constrained settings, not our acceleration strategy.
}
\vspace{-1mm}
\label{tab:main_quantitative_comparison_06M}
\resizebox{\textwidth}{!}{
\begin{tabular}{l|c|ccccc|ccccc|ccccc}
\toprule
{\multirow{2}{*}{\textbf{Method}}} & \multirow{2}{*}{\textbf{Iters}} & \multicolumn{5}{c|}{\textbf{Mip-NeRF~360}} & \multicolumn{5}{c|}{\textbf{Deep Blending}} & \multicolumn{5}{c}{\textbf{Tanks and Temples}} \\
\cmidrule(lr){3-7} \cmidrule(lr){8-12} \cmidrule(lr){13-17}
& & $N_{\text{G}}$ & PSNR$\uparrow$ & SSIM$\uparrow$ & LPIPS$\downarrow$ & Time(s)$\downarrow$ & $N_{\text{G}}$ & PSNR$\uparrow$ & SSIM$\uparrow$ & LPIPS$\downarrow$ & Time(s)$\downarrow$ & $N_{\text{G}}$ & PSNR$\uparrow$ & SSIM$\uparrow$ & LPIPS$\downarrow$ & Time(s)$\downarrow$ \\
\midrule
Mini-Splatting2 & \multirow{3}{*}{18K} & \multirow{3}{*}{0.6M} & \textbf{27.38} & \textbf{0.821} & \textbf{0.215} & 92.68 & \multirow{3}{*}{0.6M} & \textbf{30.11} & \textbf{0.912} & \textbf{0.239} & 78.87 & \multirow{3}{*}{0.4M} & 23.11 & \textbf{0.841} & \textbf{0.186} & 72.45 \\
DashGaussian & & & 26.89 & 0.777 & 0.284 & 65.01 & & 29.93 & 0.901 & 0.270 & 48.11 & & \textbf{23.42} & 0.823 & 0.229 & 57.94 \\
LiteGS & & & 26.75 & 0.765 & 0.299 & 66.33 & & 29.75 & 0.903 & 0.275 & 49.16 & & 23.07 & 0.816 & 0.234 & 49.74 \\
\midrule
\textbf{Ours} & 18K & 0.6M & 26.49 & 0.762 & 0.303 & \textbf{43.87} & 0.6M & 29.43 & 0.898 & 0.278 & \textbf{37.76} & 0.4M & 22.96 & 0.819 & 0.226 & \textbf{44.81} \\
\bottomrule
\end{tabular}
}
\vspace{-4mm}
\end{table*}


%% file: sec/ending.tex
\section{Conclusion}

We present a method to speed up the learning of 3D Gaussians with much shorter Gaussian lists. Without a need of reducing total Gaussian count, both our scale reset and entropy constraint successfully shorten the Gaussian lists used for rendering each pixel by resetting the size of Gaussians regularly and sharpening the weight distribution in alpha blending. Our evaluations and ablation studies on widely used benchmarks validate our approach and demonstrate significant advantages over other methods in efficiency with comparable rendering quality.

\section{Acknowledgment}

This project was partially supported by an NVIDIA academic award and a Richard Barber research award.

%% file: sec/X_suppl.tex
\clearpage
\setcounter{page}{1}
\maketitlesupplementary


\section{Sum of Weights Equals Unity}
\label{se:weights_sum_to_unity}
We proof the unity of the sum of weights along each ray. This ensures that we can calculate the entropy using weights as a distribution. In volumetric rendering with $N$ Gaussians and a background, the transmittance satisfies,
\begin{align}
    T_i = \prod_{j=1}^{i-1} (1-\alpha_j), \quad T_1 = 1,
\end{align}
and the blending weight is $w_i = T_i \alpha_i$ for $i \in [1,N]$. The background weight is $w_{N+1} = T_{N+1}$. We prove that $\sum_{i=1}^{N+1} w_i = 1$.

\paragraph{Proof.}
Equivalently, we show that $T_{N+1} = 1 - \sum_{i=1}^N w_i$ by mathematical induction.

Base case ($N=1$):
\begin{align}
    \sum_{i=1}^{1} w_i + T_2 = T_1\alpha_1 + T_1(1-\alpha_1) = T_1 = 1.
\end{align}

Inductive step: Assume $\sum_{i=1}^{k} w_i = 1 - T_{k+1}$.

For $N=k+1$:
\begin{align}
    \sum_{i=1}^{k+1} w_i &= \sum_{i=1}^{k} w_i + w_{k+1} 
    = (1 - T_{k+1}) + T_{k+1}\alpha_{k+1} \\
    &= 1 - T_{k+1}(1-\alpha_{k+1}) 
    = 1 - T_{k+2}.
\end{align}

Therefore, $\sum_{i=1}^{N+1} w_i = 1$.


\section{Gradient of Entropy with respect to Alpha}
\label{se:entropy_gradient}

\paragraph{Setup.}
For a pixel with $N$ Gaussians, recall that the transmittance and blending weights for the $i$-th Gaussian are:
\begin{align}
    T_i &= \prod_{j=1}^{i-1} (1-\alpha_j), \quad T_1 = 1, \\
    w_i &= T_i \alpha_i, \quad i \in [1,N].
\end{align}
Including the background weight $w_{N+1} = T_{N+1}$, we have $\sum_{i=1}^{N+1} w_i = 1$ (see \cref{se:weights_sum_to_unity}).

\paragraph{Entropy loss.}
The entropy for one pixel is
\begin{align}
    H &= - \sum_{i=1}^{N+1} w_i \log w_i,
\end{align}
with gradient
\begin{align}
    \frac{\partial H}{\partial w_i} &= -\log w_i - 1.
\end{align}
\paragraph{Chain rule.}
For each Gaussian, the learnable attributes (position, scale, rotation, opacity) are encapsulated in $\alpha_i$. We compute the gradient with respect to $\alpha_i$ using the chain rule:
\begin{align}
    \frac{\partial H}{\partial \alpha_i} 
    = 
    \sum_{k=i}^{N+1}
    \frac{\partial H}{\partial w_k} 
    \frac{\partial w_k}{\partial \alpha_i}.
\end{align}

\paragraph{Key components.}
We derive $\frac{\partial w_k}{\partial \alpha_i}$ for three cases.

\textbf{Case 1: $k = i$.}
\begin{align}
    \frac{\partial w_i}{\partial \alpha_i} 
    = \frac{\partial (T_i \alpha_i)}{\partial \alpha_i}
    = T_i.
\end{align}

\textbf{Case 2: $k > i$.}
Since $w_k = T_k \alpha_k = \prod_{j=1}^{k-1}(1-\alpha_j) \cdot \alpha_k$, we have
\begin{align}
    \frac{\partial w_k}{\partial \alpha_i}
    &= \alpha_k \cdot \frac{\partial}{\partial \alpha_i}\left[\prod_{j=1}^{k-1}(1-\alpha_j)\right] \\
    &= \alpha_k \cdot \prod_{j=1, j\neq i}^{k-1}(1-\alpha_j) \cdot (-1) \\
    &= -\frac{\prod_{j=1}^{k-1}(1-\alpha_j) \cdot \alpha_k}{1-\alpha_i} \\
    &= -\frac{w_k}{1 - \alpha_i}.
\end{align}

\textbf{Case 3: $k = N+1$.}
\begin{align}
    \frac{\partial w_{N+1}}{\partial \alpha_i}
    &= \frac{\partial T_{N+1}}{\partial \alpha_i}
    = -\frac{T_{N+1}}{1 - \alpha_i}
    = -\frac{w_{N+1}}{1 - \alpha_i}.
\end{align}

\paragraph{General formula.}
Substituting the three cases into the chain rule, we obtain:
\begin{align}
    \frac{\partial H}{\partial \alpha_i} 
    &= \frac{\partial H}{\partial w_i} \frac{\partial w_i}{\partial \alpha_i} + \sum_{k=i+1}^{N+1} \frac{\partial H}{\partial w_k} \frac{\partial w_k}{\partial \alpha_i} \\
    &= (-\log w_i - 1) T_i + \sum_{k=i+1}^{N+1} (-\log w_k - 1) \left(-\frac{w_k}{1-\alpha_i}\right) \\
    &= (-\log w_i - 1) T_i + \frac{1}{1-\alpha_i} \sum_{k=i+1}^{N+1} (\log w_k + 1) w_k.
\end{align}
Define the intermediate variable:
\begin{align}
    R_i = \sum_{k=i}^{N+1} (\log w_k + 1) w_k.
\end{align}
Then the gradient simplifies to:
\begin{align}
    \frac{\partial H}{\partial \alpha_i} = (-\log w_i - 1) T_i + \frac{R_{i+1}}{1-\alpha_i}.
\end{align}

\paragraph{Efficient computation.}
The intermediate values $R_i$ can be computed efficiently in reverse order:
\begin{align}
    R_{N+1} &= (\log w_{N+1} + 1) w_{N+1}, \\
    R_i &= (\log w_i + 1) w_i + R_{i+1}, \quad i = N, \ldots, 1.
\end{align}
This backward accumulation requires only $O(N)$ operations and can be parallelized across pixels in CUDA.


\section{Gradients with respect to Attributes}
\label{se:attribute_gradients}

Having derived $\frac{\partial H}{\partial \alpha_i}$, we now compute gradients with respect to the learnable Gaussian attributes using the chain rule.

\paragraph{Opacity.}
Recall that $\alpha_i = \sigma_i \cdot g_i$, where $\sigma_i \in [0,1]$ is the opacity and $g_i$ is the 2D Gaussian density. The gradient with respect to opacity is:
\begin{align}
    \frac{\partial H}{\partial \sigma_i} 
    = \frac{\partial H}{\partial \alpha_i} \frac{\partial \alpha_i}{\partial \sigma_i}
    = \frac{\partial H}{\partial \alpha_i} \cdot g_i.
\end{align}

\paragraph{Gaussian density.}
The 2D Gaussian density is given by:
\begin{align}
    g_i = \exp(d_i),
\end{align}
where $d_i$ is the exponent term (quadratic form):
\begin{align}
    d_i = -\frac{1}{2}(x - \mu_i')^T (\Sigma_i')^{-1}(x - \mu_i').
\end{align}
Here $\mu_i'$ and $\Sigma_i'$ denote the projected 2D mean and covariance of the $i$-th Gaussian.

The gradient with respect to $d_i$ is:
\begin{align}
    \frac{\partial H}{\partial d_i}
    &= \frac{\partial H}{\partial \alpha_i} \frac{\partial \alpha_i}{\partial g_i} \frac{\partial g_i}{\partial d_i} \\
    &= \frac{\partial H}{\partial \alpha_i} \cdot \sigma_i \cdot g_i \\
    &= \sigma_i g_i \cdot \frac{\partial H}{\partial \alpha_i}.
\end{align}

\paragraph{Inverse covariance matrix.}
The gradient with respect to the inverse covariance matrix is:
\begin{align}
    \frac{\partial H}{\partial (\Sigma_i')^{-1}}
    &= \frac{\partial H}{\partial d_i} \frac{\partial d_i}{\partial (\Sigma_i')^{-1}} \\
    &= \frac{\partial H}{\partial d_i} \cdot \left(-\frac{1}{2} (x - \mu_i')(x - \mu_i')^T\right).
\end{align}

\paragraph{Mean.}
The gradient with respect to the 2D mean is:
\begin{align}
    \frac{\partial H}{\partial \mu_i'}
    &= \frac{\partial H}{\partial d_i} \frac{\partial d_i}{\partial \mu_i'} \\
    &= \frac{\partial H}{\partial d_i} \cdot (\Sigma_i')^{-1}(x - \mu_i').
\end{align}


\section{Additional Results}

\begin{figure}[t]
  \centering
  \includegraphics[width=1.0\linewidth]{img/time_vs_iter.pdf}
  \includegraphics[width=1.0\linewidth]{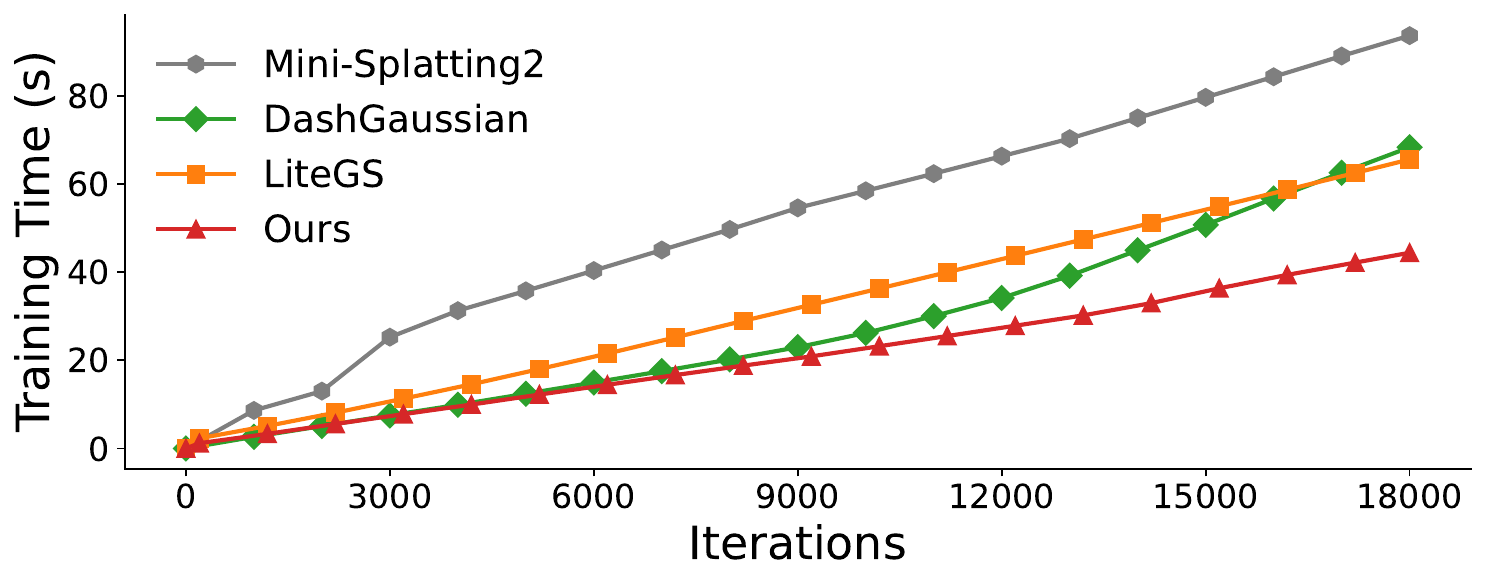}
  \caption{Training time versus iterations for full and compact models, corresponding to \cref{fig:time_vs_iter}.}
  \label{fig:supplementary_time_vs_iter}
\end{figure}

\begin{figure}[t]
  \centering
  \includegraphics[width=1.0\linewidth]{img/psnr_vs_iter.pdf}
  \includegraphics[width=1.0\linewidth]{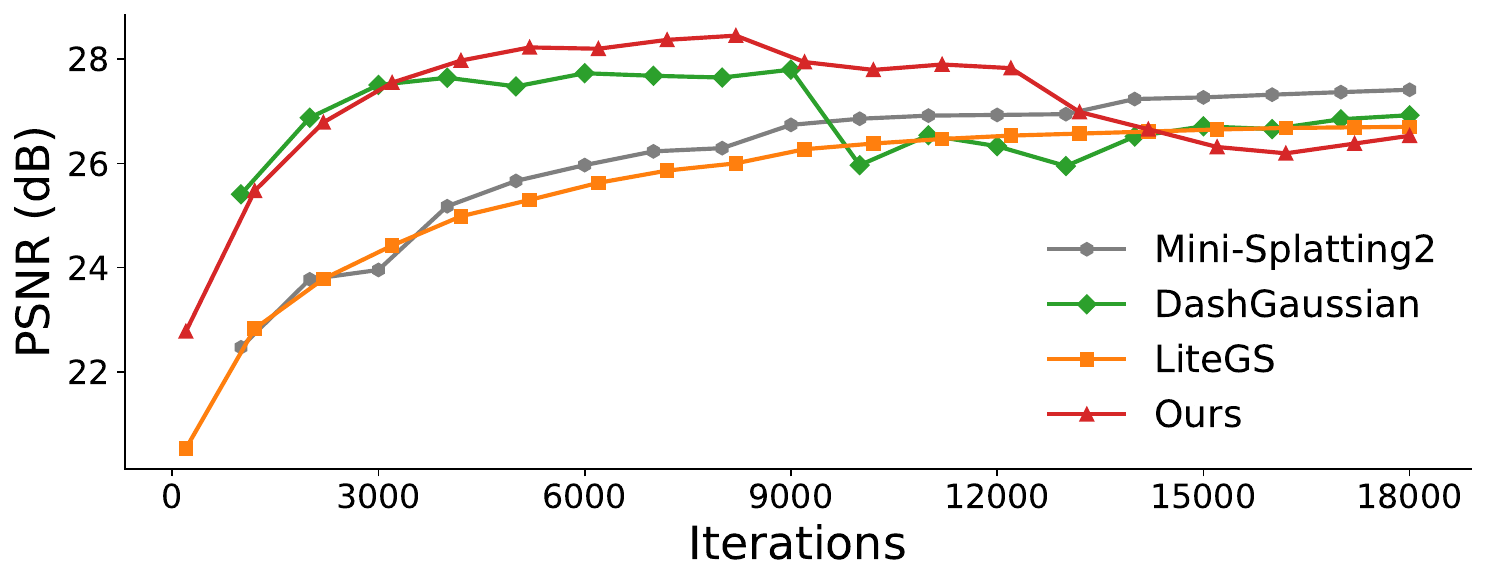}
  \caption{PSNR convergence across iterations for full and compact models, corresponding to \cref{fig:psnr_vs_iter}.}
  \label{fig:supplementary_psnr_vs_iter}
\end{figure}

\begin{figure}[t]
  \centering
   \includegraphics[width=1.0\linewidth]{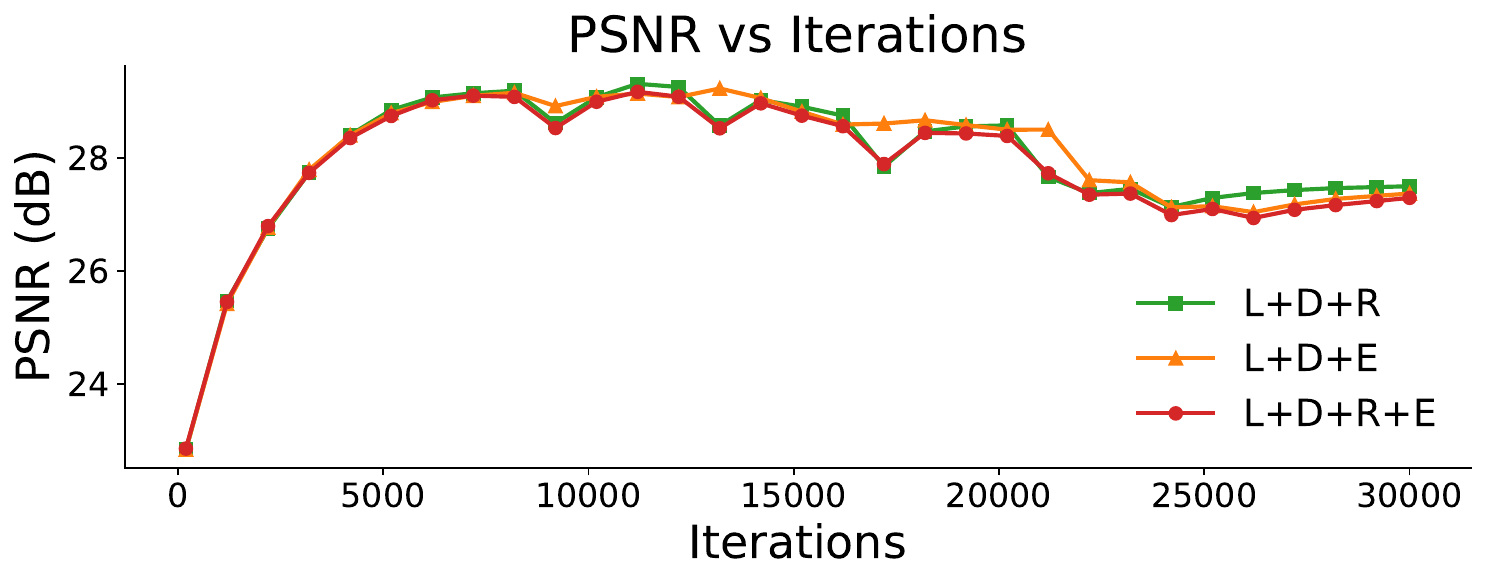}
   \includegraphics[width=1.0\linewidth]{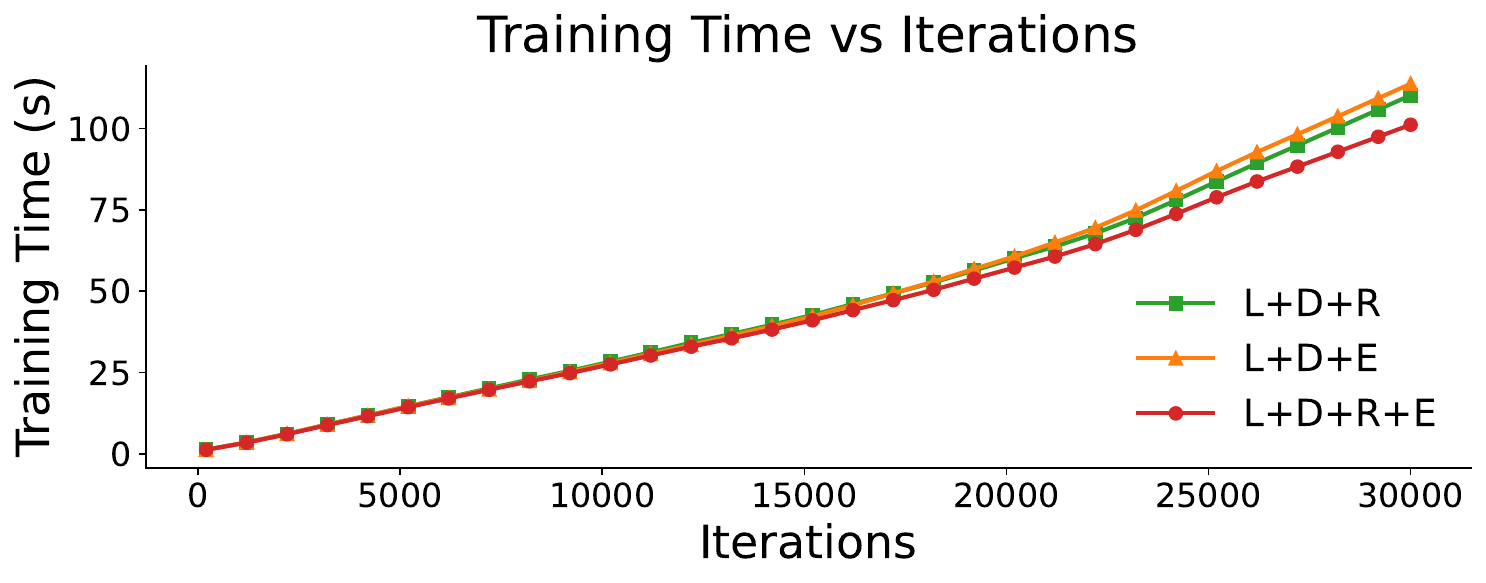}
   \caption{PSNR and time curves across iterations showing effects of individual modules. See \cref{sec:ablation_study} for more details.}
   \label{fig:supplementary_psnr_R_E_ablation}
\end{figure}

The training efficiency across iterations, as illustrated in \cref{fig:supplementary_time_vs_iter}, demonstrates that our method achieves the fastest convergence among all compared approaches. 
\cref{fig:supplementary_psnr_vs_iter} presents the PSNR progression throughout the training. Both DashGaussian and our method employ resolution scheduling, which results in higher PSNR values at lower resolutions compared to other methods. 
\cref{fig:supplementary_psnr_R_E_ablation} confirms the effectiveness of both proposed modules (scale reset and entropy regularization) across all evaluated scenes. The notation convention is detailed in \cref{sec:ablation_study}.

\begin{wraptable}{r}{0.48\linewidth}
\vspace{-2mm}
\centering
\caption{
Average testing FPS on the Mip-NeRF~360 dataset.
}\vspace{-2mm}
\label{tab:fps_mipnerf360_avg}
\scriptsize
\setlength{\tabcolsep}{10pt} 
\begin{tabular}{l|c}
\toprule
\textbf{Method} & \textbf{FPS}$\uparrow$ \\
\midrule
3DGS  & 140.91 \\
Taming & 219.60 \\
LiteGS & 233.76 \\
\midrule
\textbf{Ours} & \textbf{343.92} \\
\bottomrule
\end{tabular}
\vspace{-4mm}
\end{wraptable}
We also report the average testing FPS on Mip-NeRF~360 in \cref{tab:fps_mipnerf360_avg}, measured at the same rendering resolution for all methods, where our method achieves the highest rendering throughput among all compared approaches.

\begin{figure*}[t]
    \centering
    \begin{subfigure}[t]{0.31\textwidth}
        \centering
        \includegraphics[width=\linewidth]{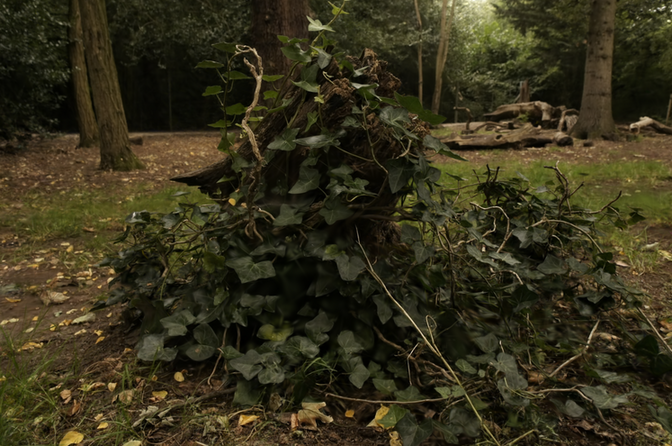}
        \caption{\scriptsize 3DGS, PSNR: 26.82, Time: 936.62s}
        \label{fig:rendered_comparison1_img1}
    \end{subfigure}
    \begin{subfigure}[t]{0.31\textwidth}
        \centering
        \includegraphics[width=\linewidth]{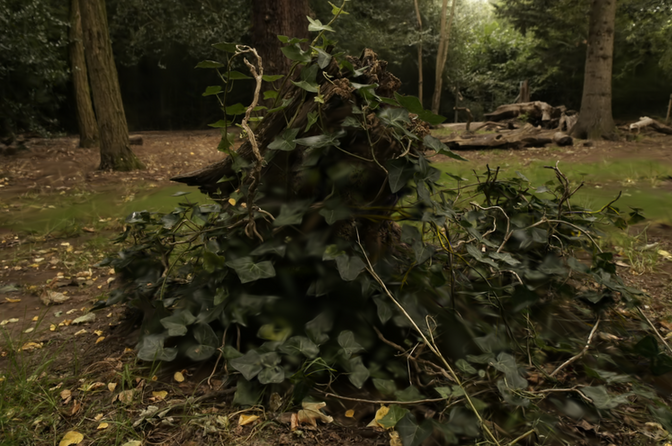}
        \caption{\scriptsize AdR-Gaussian, PSNR: 25.93, Time: 478.89s}
        \label{fig:rendered_comparison1_img2}
    \end{subfigure}
    \begin{subfigure}[t]{0.31\textwidth}
        \centering
        \includegraphics[width=\linewidth]{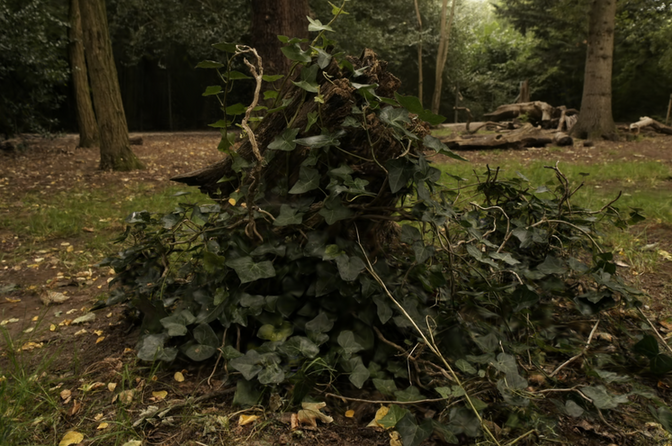}
        \caption{\scriptsize Taming 3DGS, PSNR: 26.88, Time: 436.33s}
        \label{fig:rendered_comparison1_img3}
    \end{subfigure}\\
    \begin{subfigure}[t]{0.31\textwidth}
        \centering
        \includegraphics[width=\linewidth]{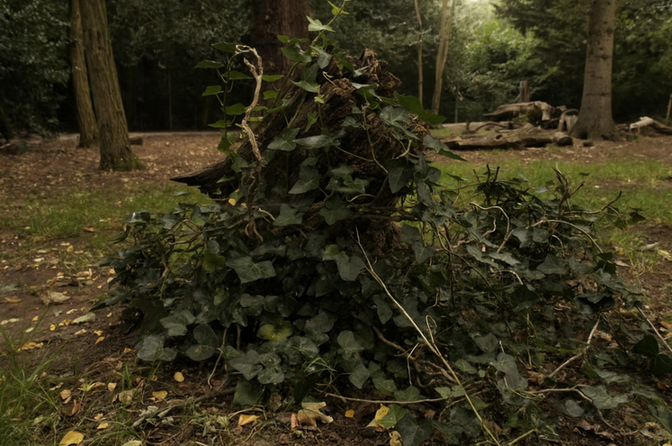}
        \caption{\scriptsize MiniSplatting2-D, PSNR: 27.03, Time: 169.71s}
        \label{fig:rendered_comparison1_img4}
    \end{subfigure}
    \begin{subfigure}[t]{0.31\textwidth}
        \centering
        \includegraphics[width=\linewidth]{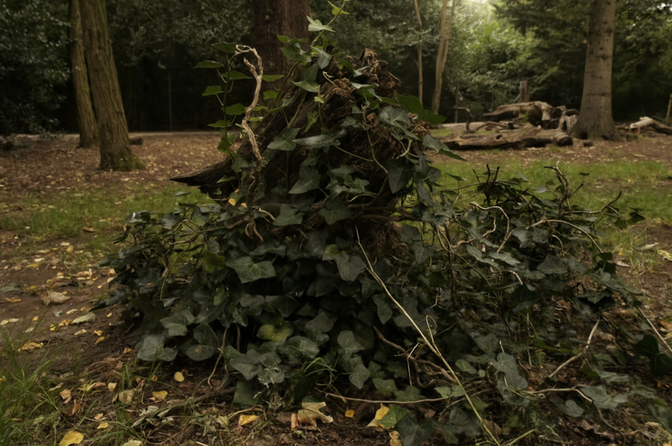}
        \caption{\scriptsize DashGaussian, PSNR: 27.37, Time: 185.38s}
        \label{fig:rendered_comparison1_img5}
    \end{subfigure}
    \begin{subfigure}[t]{0.31\textwidth}
        \centering
        \includegraphics[width=\linewidth]{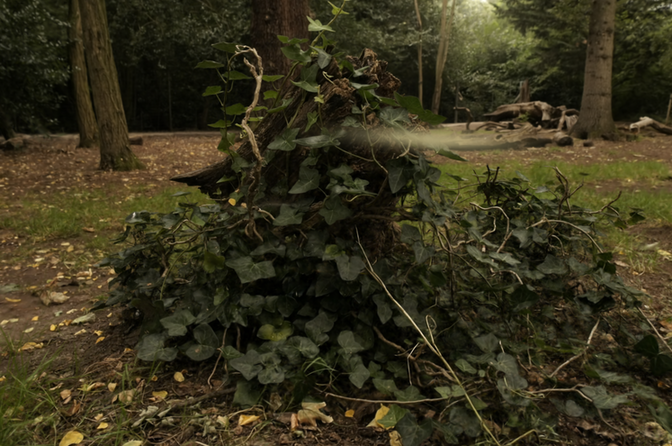}
        \caption{\scriptsize EDGS, PSNR: 25.66, Time: 196.35s}
        \label{fig:rendered_comparison1_img6}
    \end{subfigure}\\
    \begin{subfigure}[t]{0.31\textwidth}
        \centering
        \includegraphics[width=\linewidth]{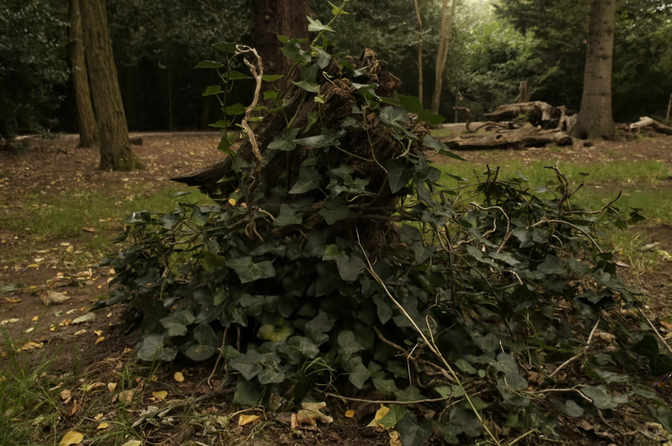}
        \caption{\scriptsize LiteGS, PSNR: 27.53, Time: 176.10s}
        \label{fig:rendered_comparison1_img7}
    \end{subfigure}
    \begin{subfigure}[t]{0.31\textwidth}
        \centering
        \includegraphics[width=\linewidth]{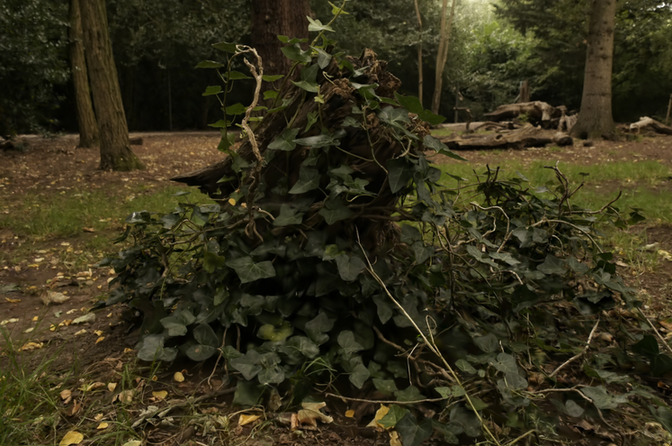}
        \caption{\scriptsize Ours, PSNR: 27.33, Time: 87.62s}
        \label{fig:rendered_comparison1_img8}
    \end{subfigure}
    \begin{subfigure}[t]{0.31\textwidth}
        \centering
        \includegraphics[width=\linewidth]{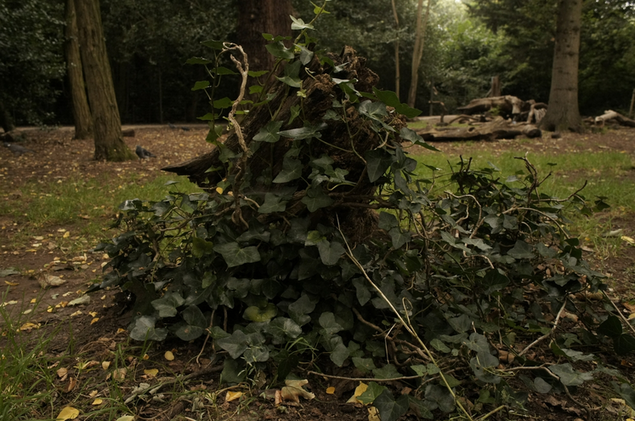}
        \caption{\scriptsize Ground Truth}
        \label{fig:rendered_comparison1_img9}
    \end{subfigure}
    
    \caption{Qualitative comparison of rendered results on scene Stump. Our method achieves comparable rendering quality (PSNR: 27.33) while being the fastest, with training time of only 87.62 seconds.}
    \label{fig:rendered_comparison1}
\end{figure*}

\begin{figure*}[t]
    \centering
    \begin{subfigure}[t]{0.31\textwidth}
        \centering
        \includegraphics[width=\linewidth]{img4/3dgs_DSCF6017_psnr30.908_ssim0.923.png}
        \caption{\scriptsize 3DGS, PSNR: 30.91, Time: 821.65s}
        \label{fig:rendered_comparison2_img1}
    \end{subfigure}
    \begin{subfigure}[t]{0.31\textwidth}
        \centering
        \includegraphics[width=\linewidth]{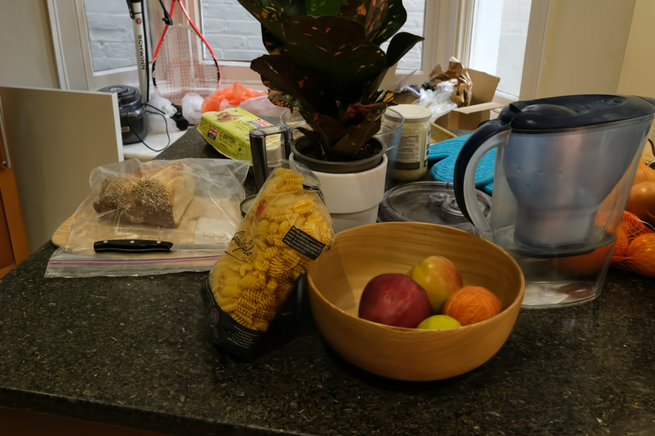}
        \caption{\scriptsize AdR-Gaussian, PSNR: 30.31, Time: 505.97s}
        \label{fig:rendered_comparison2_img2}
    \end{subfigure}
    \begin{subfigure}[t]{0.31\textwidth}
        \centering
        \includegraphics[width=\linewidth]{img4/taming_00020_psnr31.103_ssim0.924.png}
        \caption{\scriptsize Taming 3DGS, PSNR: 31.10, Time: 298.62s}
        \label{fig:rendered_comparison2_img3}
    \end{subfigure}\\
    \begin{subfigure}[t]{0.31\textwidth}
        \centering
        \includegraphics[width=\linewidth]{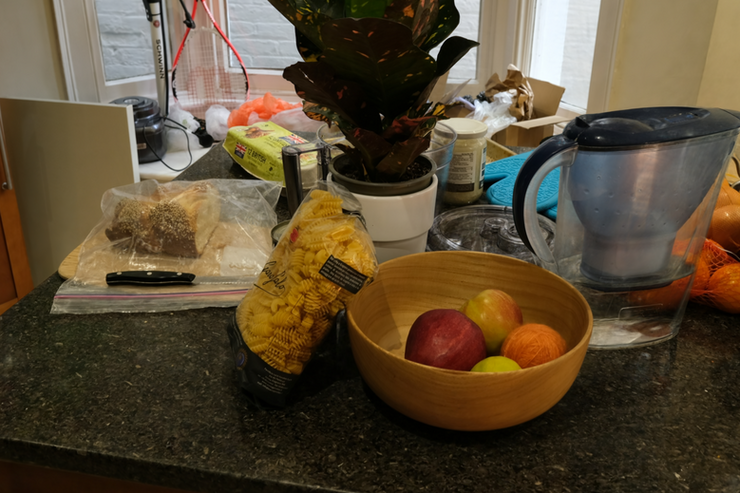}
        \caption{\scriptsize MiniSplatting2-D, PSNR: 30.89, Time: 261.92s}
        \label{fig:rendered_comparison2_img4}
    \end{subfigure}
    \begin{subfigure}[t]{0.31\textwidth}
        \centering
        \includegraphics[width=\linewidth]{img4/dash_00020_psnr30.689_ssim0.919.png}
        \caption{\scriptsize DashGaussian, PSNR: 30.69, Time: 158.53s}
        \label{fig:rendered_comparison2_img5}
    \end{subfigure}
    \begin{subfigure}[t]{0.31\textwidth}
        \centering
        \includegraphics[width=\linewidth]{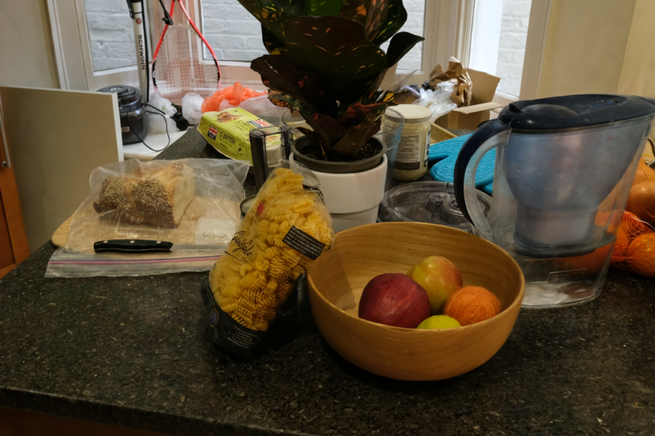}
        \caption{\scriptsize EDGS, PSNR: 29.38, Time: 404.50s}
        \label{fig:rendered_comparison2_img6}
    \end{subfigure}\\
    \begin{subfigure}[t]{0.31\textwidth}
        \centering
        \includegraphics[width=\linewidth]{img4/litegs_DSCF6017_psnr30.723_ssim0.922.png}
        \caption{\scriptsize LiteGS, PSNR: 30.72, Time: 179.39s}
        \label{fig:rendered_comparison2_img7}
    \end{subfigure}
    \begin{subfigure}[t]{0.31\textwidth}
        \centering
        \includegraphics[width=\linewidth]{img4/ours_DSCF6017_psnr30.381_ssim0.911.png}
        \caption{\scriptsize Ours, PSNR: 30.38, Time: 86.66s}
        \label{fig:rendered_comparison2_img8}
    \end{subfigure}
    \begin{subfigure}[t]{0.31\textwidth}
        \centering
        \includegraphics[width=\linewidth]{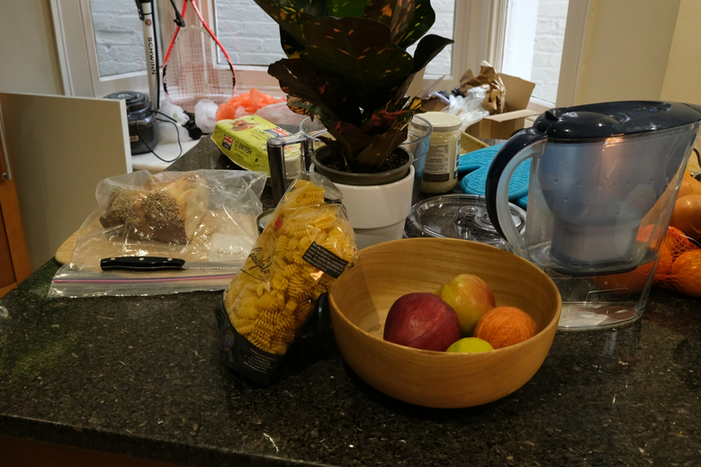}
        \caption{\scriptsize Ground Truth}
        \label{fig:rendered_comparison2_img9}
    \end{subfigure}
    
    \caption{Qualitative comparison of rendered results on scene Counter. Our method achieves comparable rendering quality (PSNR: 30.38) while being the fastest, with training time of only 86.66 seconds.}
    \label{fig:rendered_comparison2}
\end{figure*}

\begin{figure*}[t]
    \centering
    \begin{subfigure}[t]{0.31\textwidth}
        \centering
        \includegraphics[width=\linewidth]{img5/3dgs_DSC8703_psnr21.164_ssim0.693.png}
        \caption{\scriptsize 3DGS, PSNR: 21.16, Time: 1182.72s}
        \label{fig:rendered_comparison3_img1}
    \end{subfigure}
    \begin{subfigure}[t]{0.31\textwidth}
        \centering
        \includegraphics[width=\linewidth]{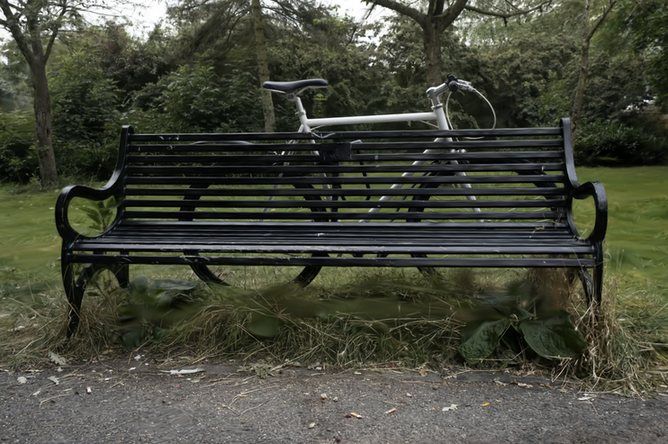}
        \caption{\scriptsize AdR-Gaussian, PSNR: 20.75, Time: 539.21s}
        \label{fig:rendered_comparison3_img2}
    \end{subfigure}
    \begin{subfigure}[t]{0.31\textwidth}
        \centering
        \includegraphics[width=\linewidth]{img5/taming_00003_psnr21.643_ssim0.707.png}
        \caption{\scriptsize Taming 3DGS, PSNR: 21.64, Time: 598.34s}
        \label{fig:rendered_comparison3_img3}
    \end{subfigure}\\
    \begin{subfigure}[t]{0.31\textwidth}
        \centering
        \includegraphics[width=\linewidth]{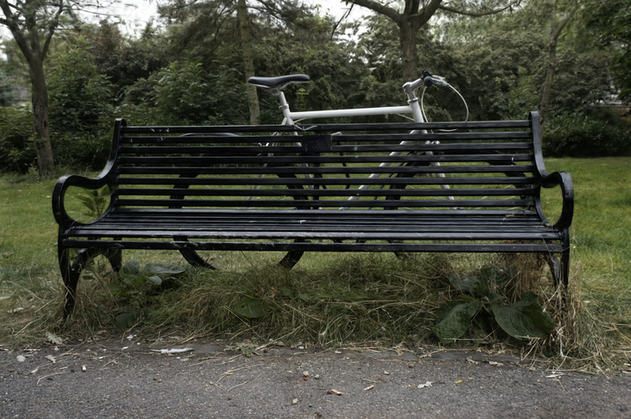}
        \caption{\scriptsize MiniSplatting2-D, PSNR: 21.18, Time: 193.93s}
        \label{fig:rendered_comparison3_img4}
    \end{subfigure}
    \begin{subfigure}[t]{0.31\textwidth}
        \centering
        \includegraphics[width=\linewidth]{img5/dash_00003_psnr21.156_ssim0.696.png}
        \caption{\scriptsize DashGaussian, PSNR: 21.16, Time: 293.70s}
        \label{fig:rendered_comparison3_img5}
    \end{subfigure}
    \begin{subfigure}[t]{0.31\textwidth}
        \centering
        \includegraphics[width=\linewidth]{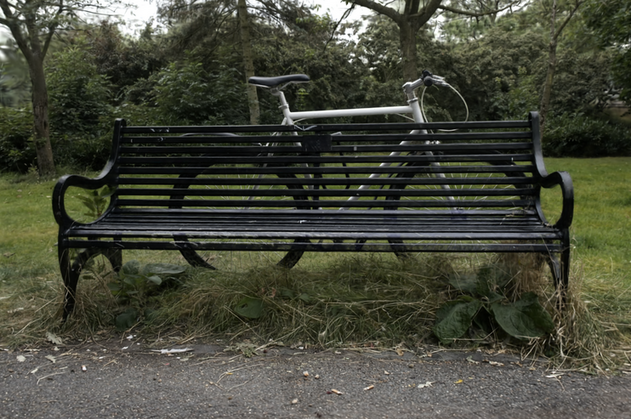}
        \caption{\scriptsize EDGS, PSNR: 20.87, Time: 298.30s}
        \label{fig:rendered_comparison3_img6}
    \end{subfigure}\\
    \begin{subfigure}[t]{0.31\textwidth}
        \centering
        \includegraphics[width=\linewidth]{img5/litegs_DSC8703_psnr21.273_ssim0.701.png}
        \caption{\scriptsize LiteGS, PSNR: 21.27, Time: 238.38s}
        \label{fig:rendered_comparison3_img7}
    \end{subfigure}
    \begin{subfigure}[t]{0.31\textwidth}
        \centering
        \includegraphics[width=\linewidth]{img5/ours_DSC8703_psnr21.287_ssim0.688.png}
        \caption{\scriptsize Ours, PSNR: 21.29, Time: 115.37s}
        \label{fig:rendered_comparison3_img8}
    \end{subfigure}
    \begin{subfigure}[t]{0.31\textwidth}
        \centering
        \includegraphics[width=\linewidth]{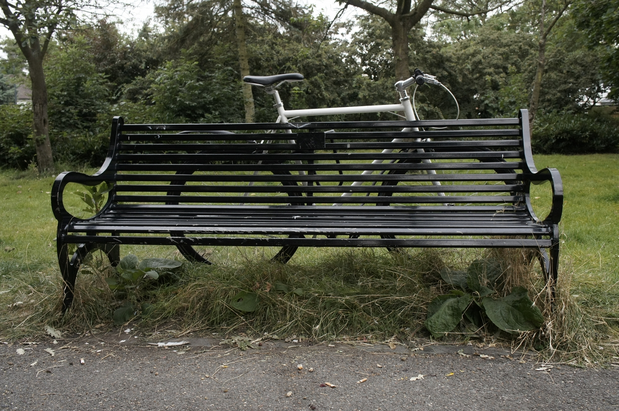}
        \caption{\scriptsize Ground Truth}
        \label{fig:rendered_comparison3_img9}
    \end{subfigure}
    
    \caption{Qualitative comparison of rendered results on scene Bicycle. Our method achieves comparable rendering quality (PSNR: 21.29) while being the fastest, with training time of only 115.37 seconds.}
    \label{fig:rendered_comparison3}
\end{figure*}

\begin{figure*}[t]
    \centering
    \begin{subfigure}[t]{0.31\textwidth}
        \centering
        \includegraphics[width=\linewidth]{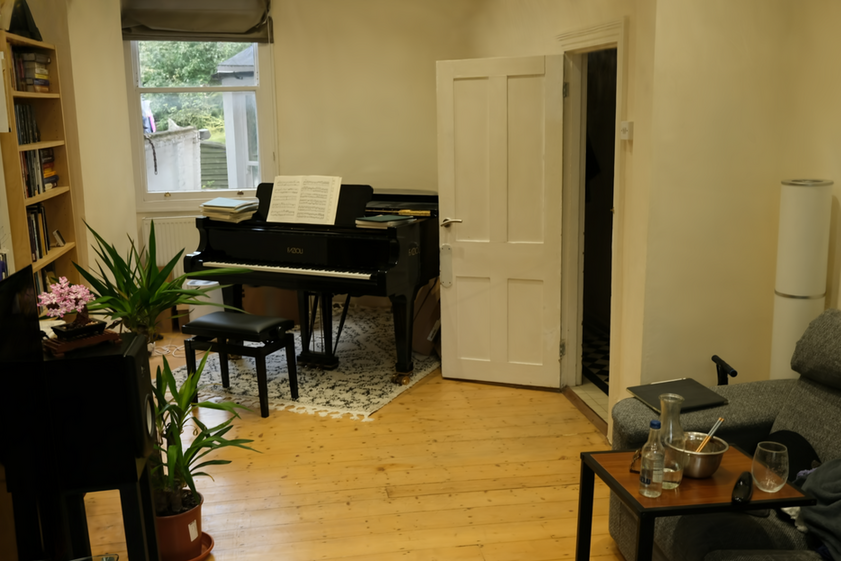}
        \caption{\scriptsize 3DGS, PSNR: 34.31, Time: 884.63s}
        \label{fig:rendered_comparison4_img1}
    \end{subfigure}
    \begin{subfigure}[t]{0.31\textwidth}
        \centering
        \includegraphics[width=\linewidth]{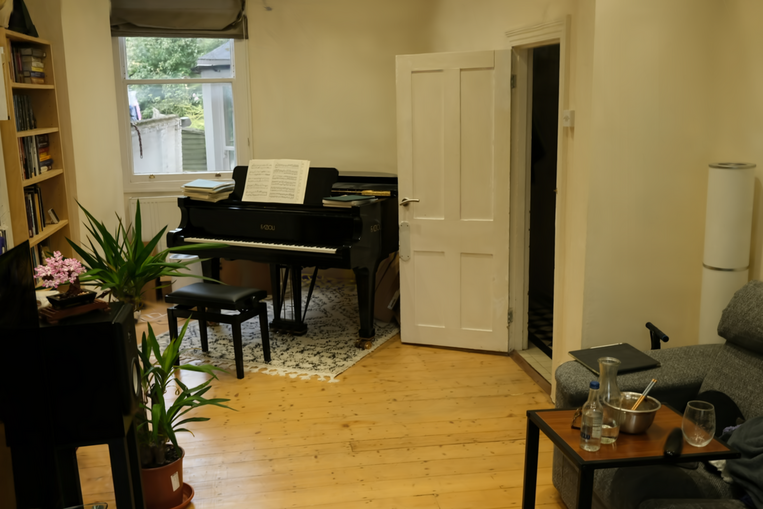}
        \caption{\scriptsize AdR-Gaussian, PSNR: 33.68, Time: 529.03s}
        \label{fig:rendered_comparison4_img2}
    \end{subfigure}
    \begin{subfigure}[t]{0.31\textwidth}
        \centering
        \includegraphics[width=\linewidth]{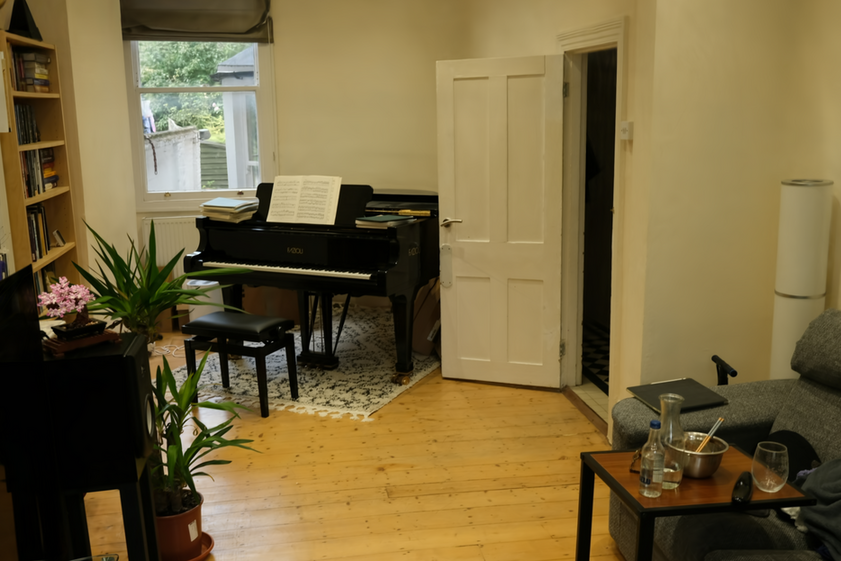}
        \caption{\scriptsize Taming 3DGS, PSNR: 34.83, Time: 307.47s}
        \label{fig:rendered_comparison4_img3}
    \end{subfigure}\\
    \begin{subfigure}[t]{0.31\textwidth}
        \centering
        \includegraphics[width=\linewidth]{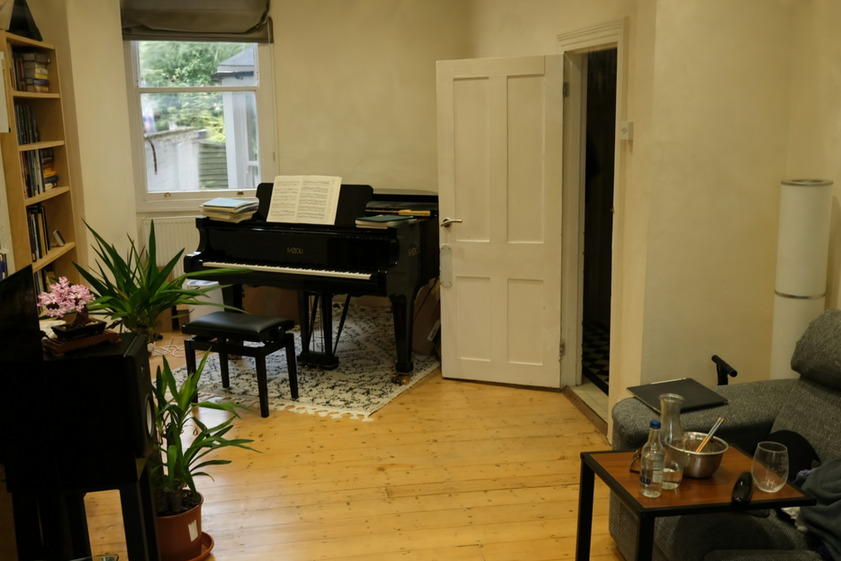}
        \caption{\scriptsize MiniSplatting2-D, PSNR: 33.56, Time: 213.22s}
        \label{fig:rendered_comparison4_img4}
    \end{subfigure}
    \begin{subfigure}[t]{0.31\textwidth}
        \centering
        \includegraphics[width=\linewidth]{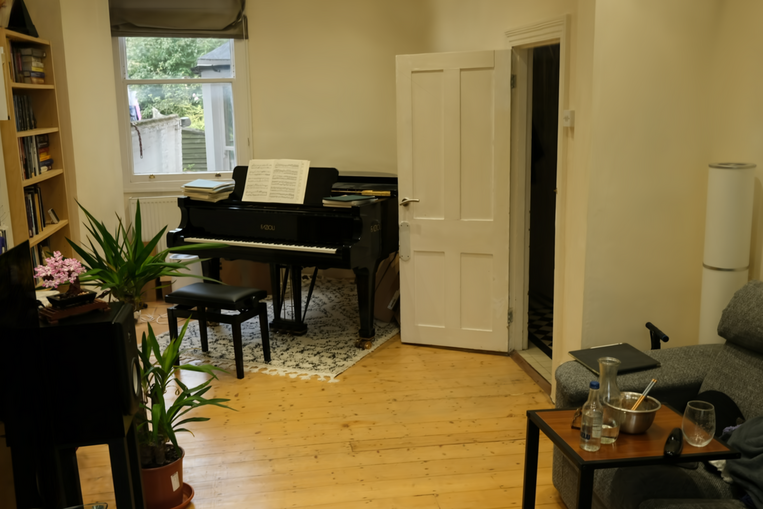}
        \caption{\scriptsize DashGaussian, PSNR: 34.59, Time: 162.80s}
        \label{fig:rendered_comparison4_img5}
    \end{subfigure}
    \begin{subfigure}[t]{0.31\textwidth}
        \centering
        \includegraphics[width=\linewidth]{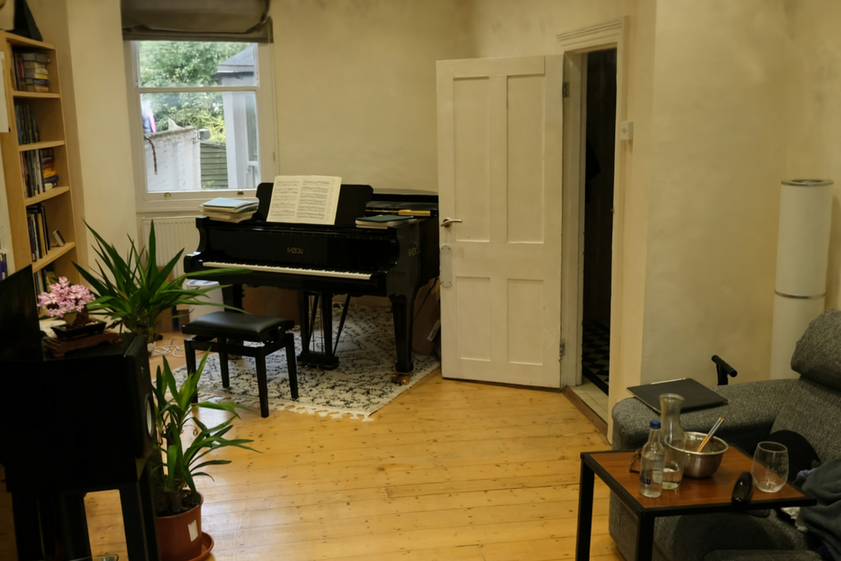}
        \caption{\scriptsize EDGS, PSNR: 32.85, Time: 362.45s}
        \label{fig:rendered_comparison4_img6}
    \end{subfigure}\\
    \begin{subfigure}[t]{0.31\textwidth}
        \centering
        \includegraphics[width=\linewidth]{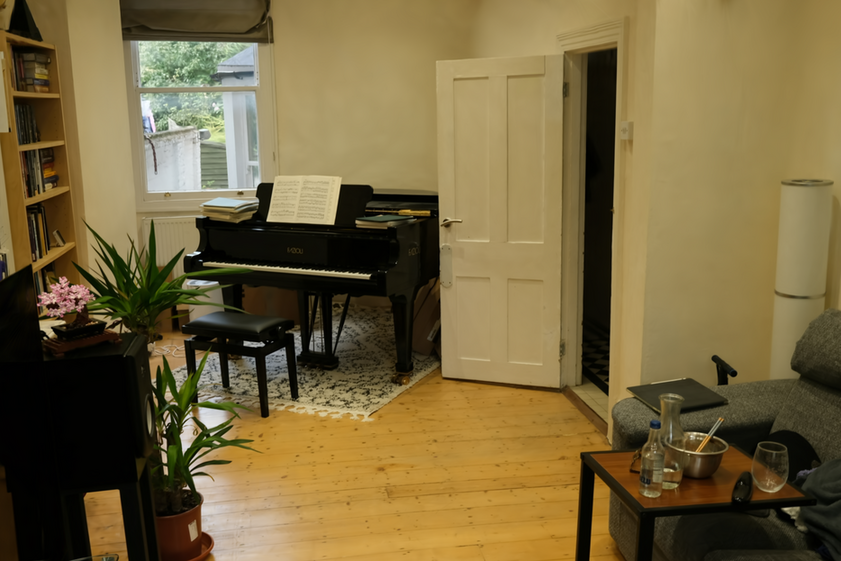}
        \caption{\scriptsize LiteGS, PSNR: 34.79, Time: 174.86s}
        \label{fig:rendered_comparison4_img7}
    \end{subfigure}
    \begin{subfigure}[t]{0.31\textwidth}
        \centering
        \includegraphics[width=\linewidth]{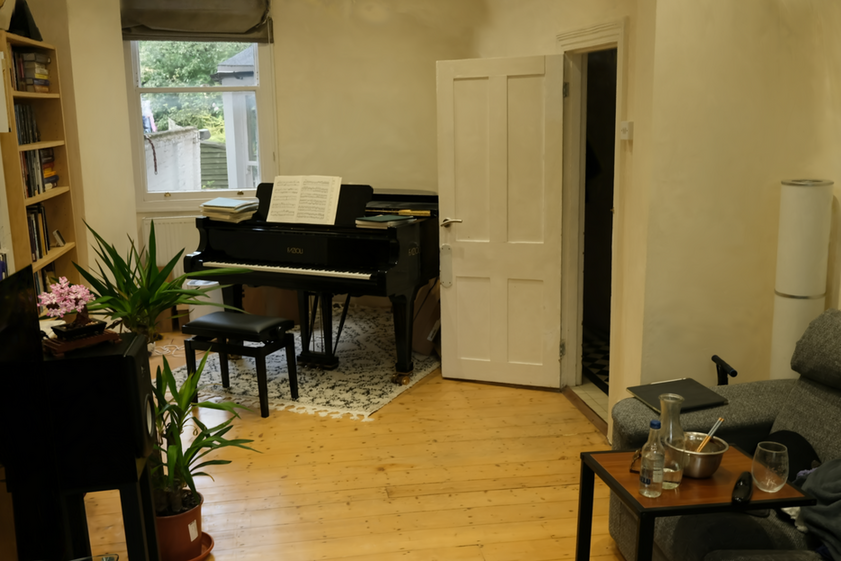}
        \caption{\scriptsize Ours, PSNR: 34.23, Time: 79.42s}
        \label{fig:rendered_comparison4_img8}
    \end{subfigure}
    \begin{subfigure}[t]{0.31\textwidth}
        \centering
        \includegraphics[width=\linewidth]{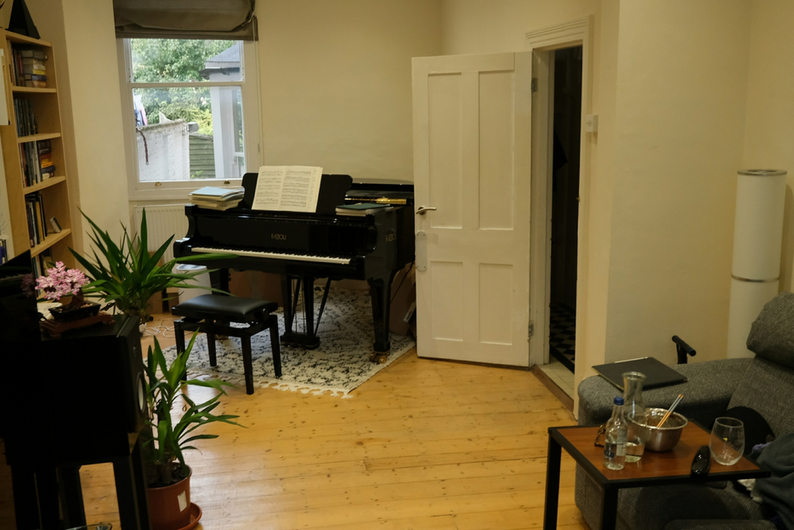}
        \caption{\scriptsize Ground Truth}
        \label{fig:rendered_comparison4_img9}
    \end{subfigure}
    
    \caption{Qualitative comparison of rendered results on scene Room. Our method achieves comparable rendering quality (PSNR: 34.23) while being the fastest, with training time of only 79.42 seconds.}
    \label{fig:rendered_comparison4}
\end{figure*}

Qualitative comparisons of the rendered results are presented in \cref{fig:rendered_comparison1}, \cref{fig:rendered_comparison2}, \cref{fig:rendered_comparison3}, and \cref{fig:rendered_comparison4}.
Additional qualitative comparisons of rendering with our method and DashGaussian at specific training times are provided in \cref{fig:render_process_p2} of the supplementary material.

\begin{figure*}[t]
    \centering
    
    \begin{subfigure}[t]{0.195\textwidth}
        \centering
        \includegraphics[width=\linewidth]{img12/8824_3dgs_8x8_iter_030000_psnr_29.00_time_1322s__DSC8824_heatmap.png}
    \end{subfigure}
    \begin{subfigure}[t]{0.195\textwidth}
        \centering
        \includegraphics[width=\linewidth]{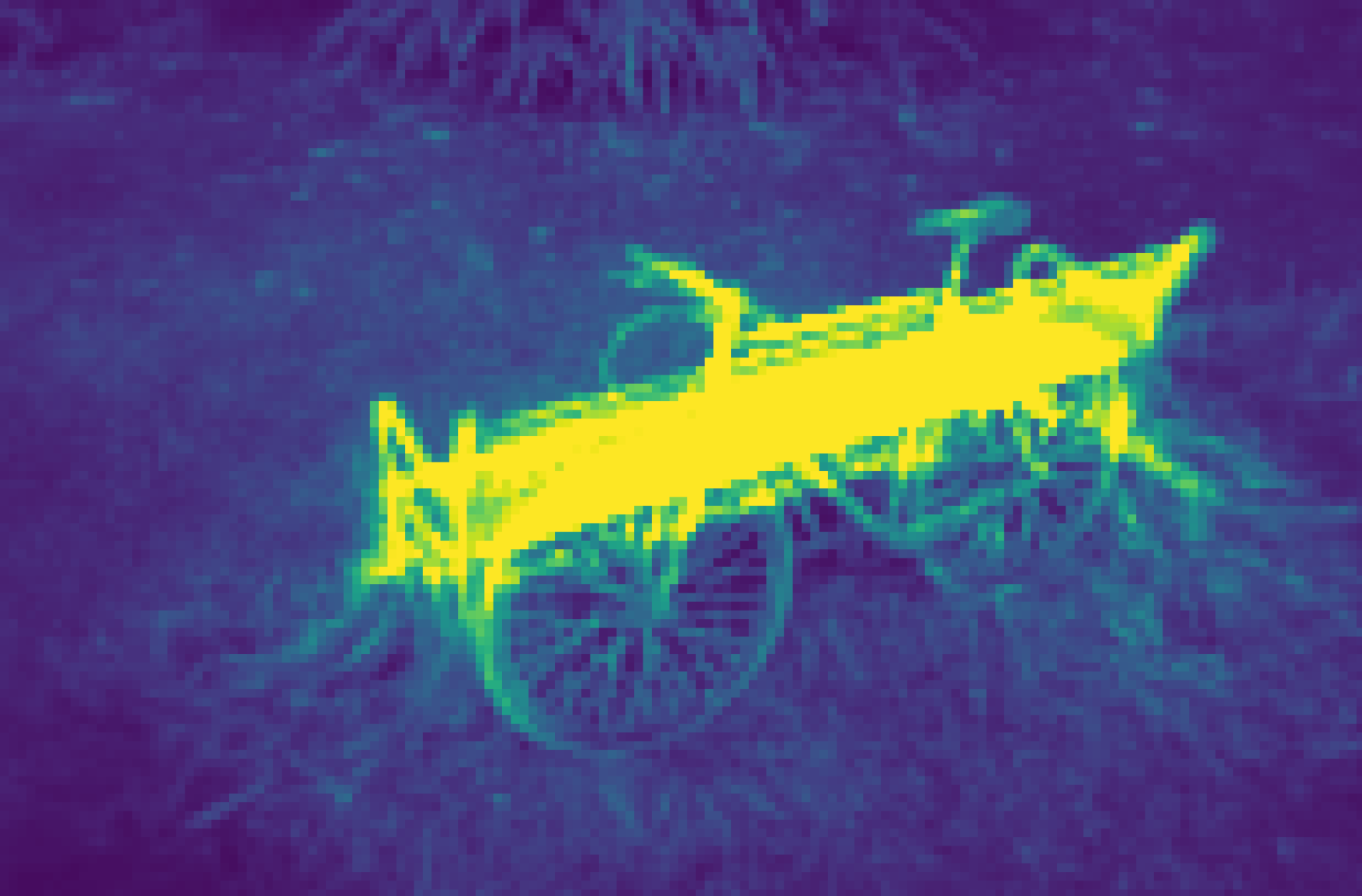}
    \end{subfigure}
    \begin{subfigure}[t]{0.195\textwidth}
        \centering
        \includegraphics[width=\linewidth]{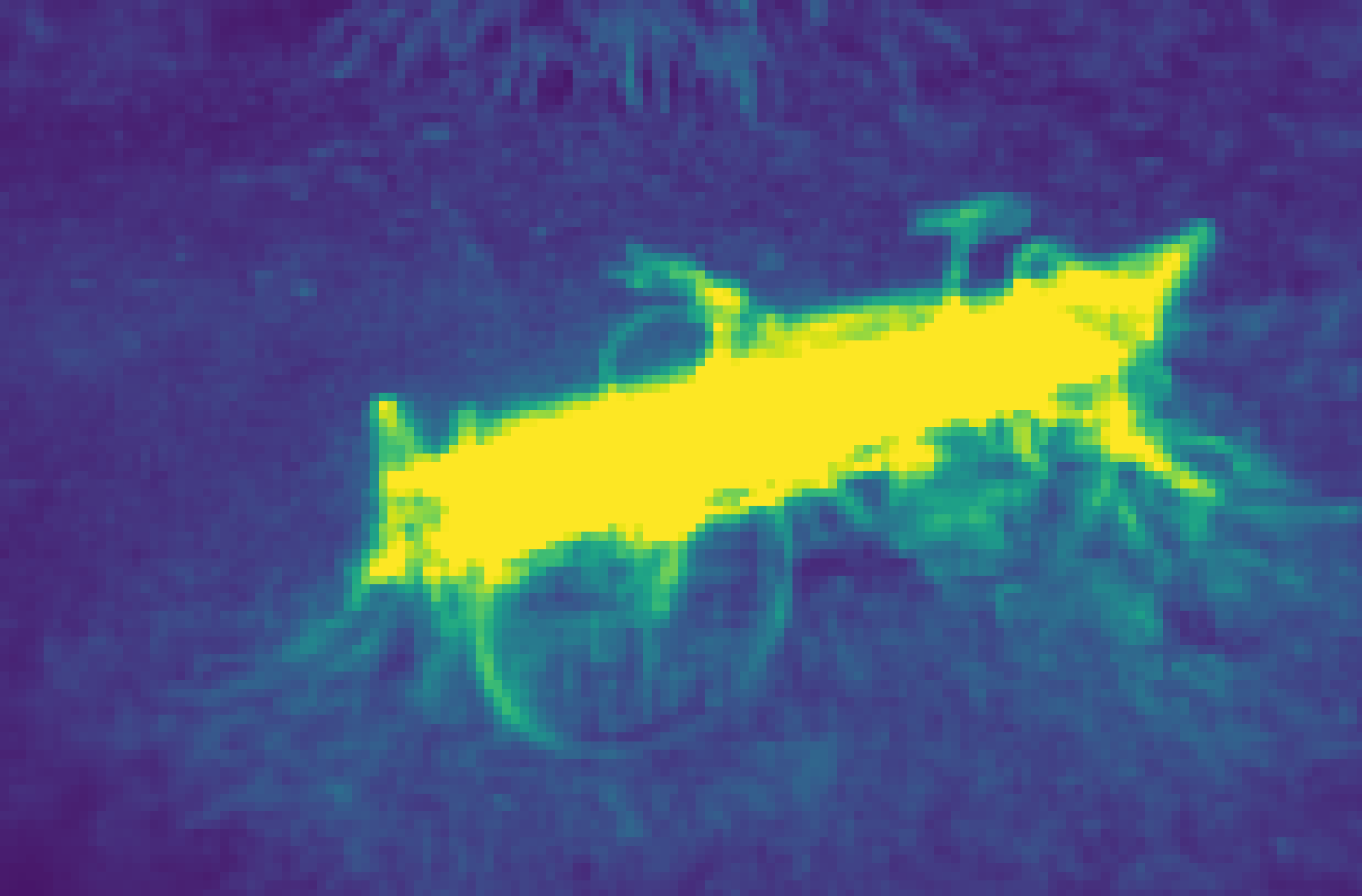}
    \end{subfigure}
    \begin{subfigure}[t]{0.195\textwidth}
        \centering
        \includegraphics[width=\linewidth]{img12/8824_ours_epoch_0149__DSC8824_gaussian_heatmap.png}
    \end{subfigure}
    \begin{subfigure}[t]{0.195\textwidth}
        \centering
        \includegraphics[width=\linewidth]{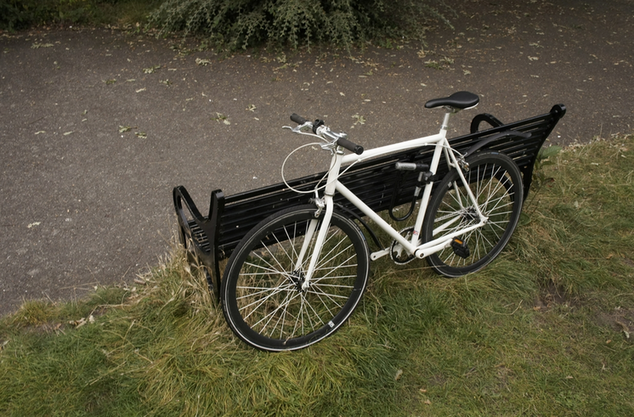}
    \end{subfigure}\\

    \begin{subfigure}[t]{0.195\textwidth}
        \centering
        \includegraphics[width=\linewidth]{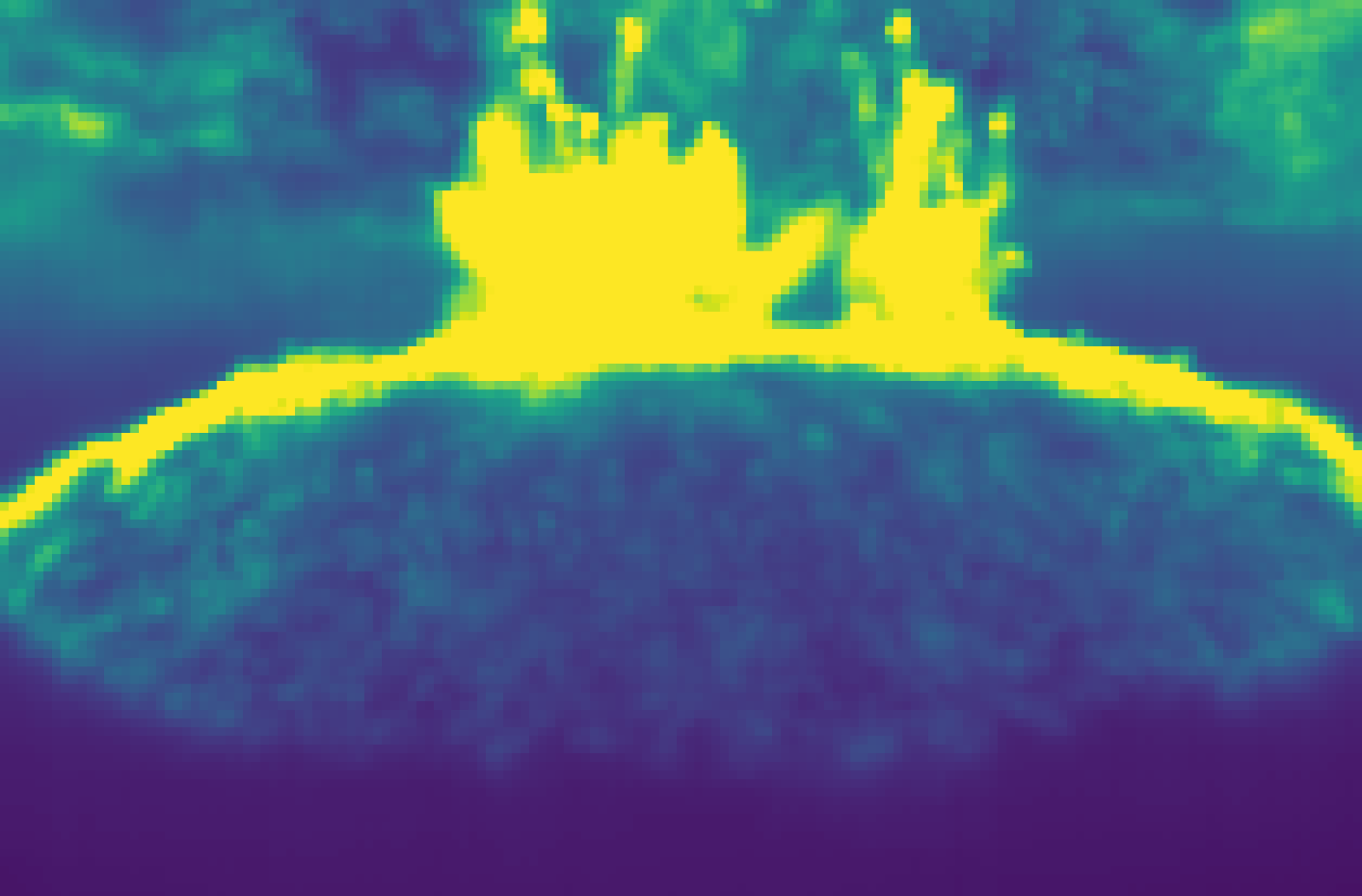}
    \end{subfigure}
    \begin{subfigure}[t]{0.195\textwidth}
        \centering
        \includegraphics[width=\linewidth]{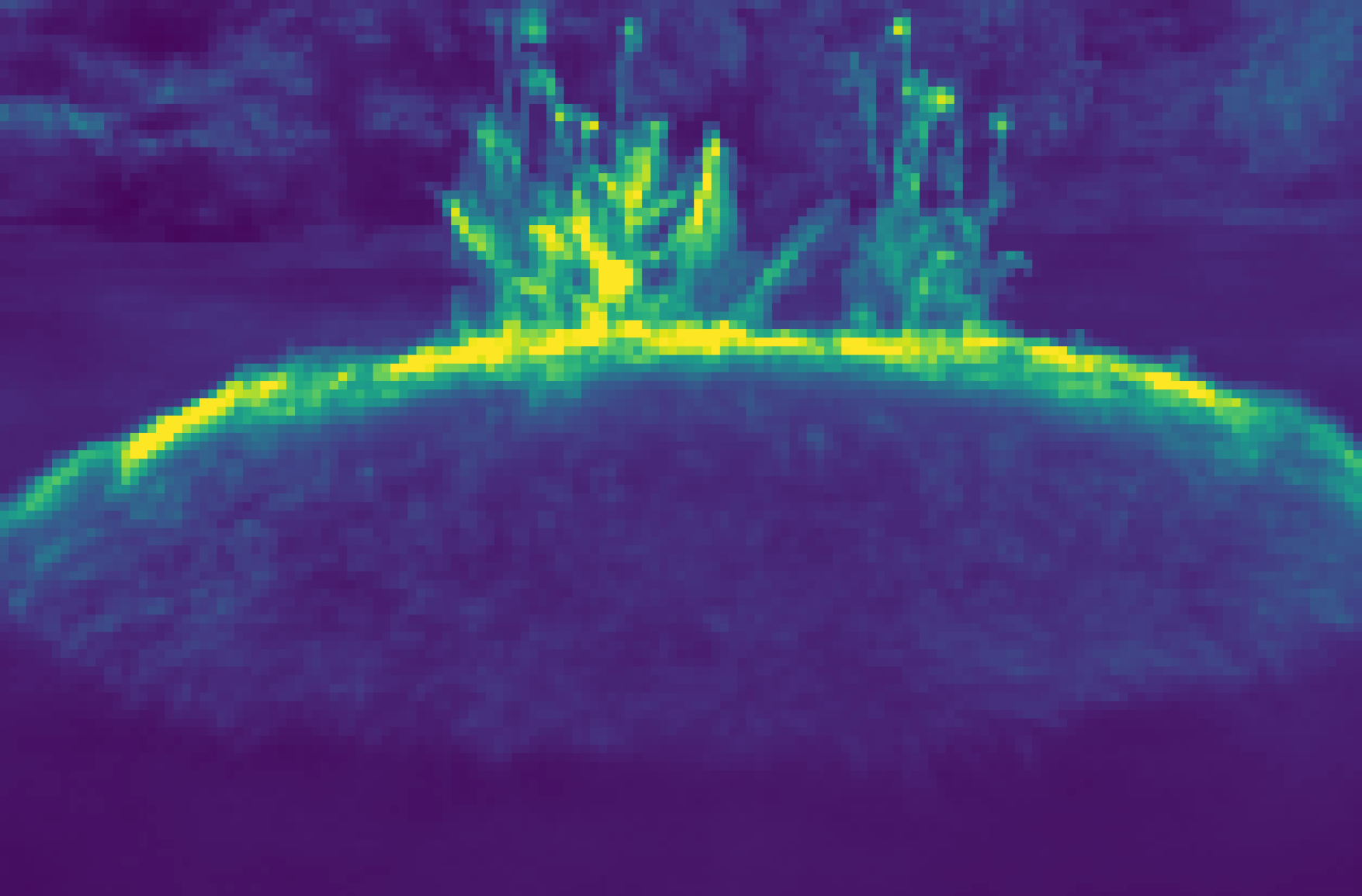}
    \end{subfigure}
    \begin{subfigure}[t]{0.195\textwidth}
        \centering
        \includegraphics[width=\linewidth]{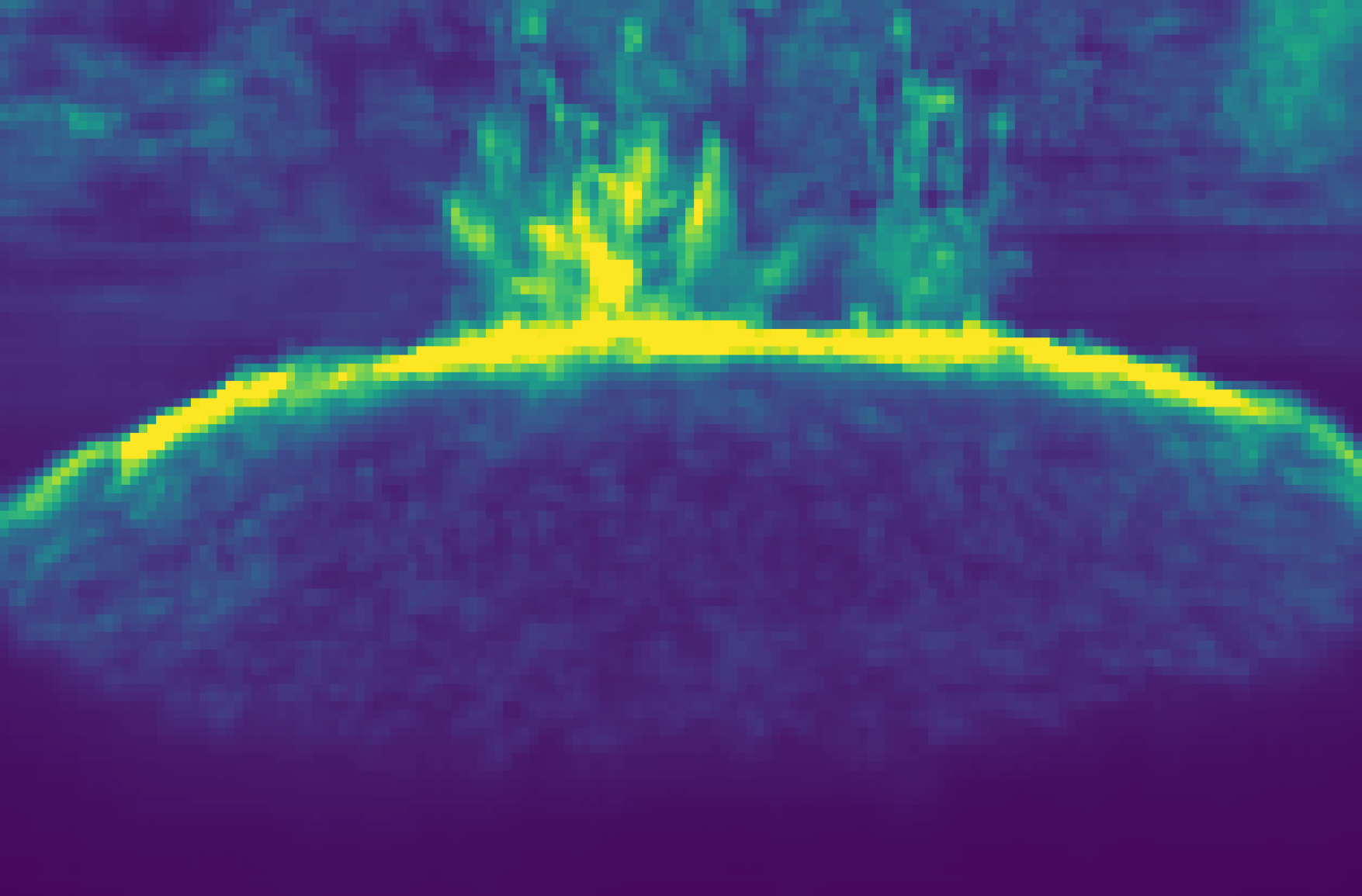}
    \end{subfigure}
    \begin{subfigure}[t]{0.195\textwidth}
        \centering
        \includegraphics[width=\linewidth]{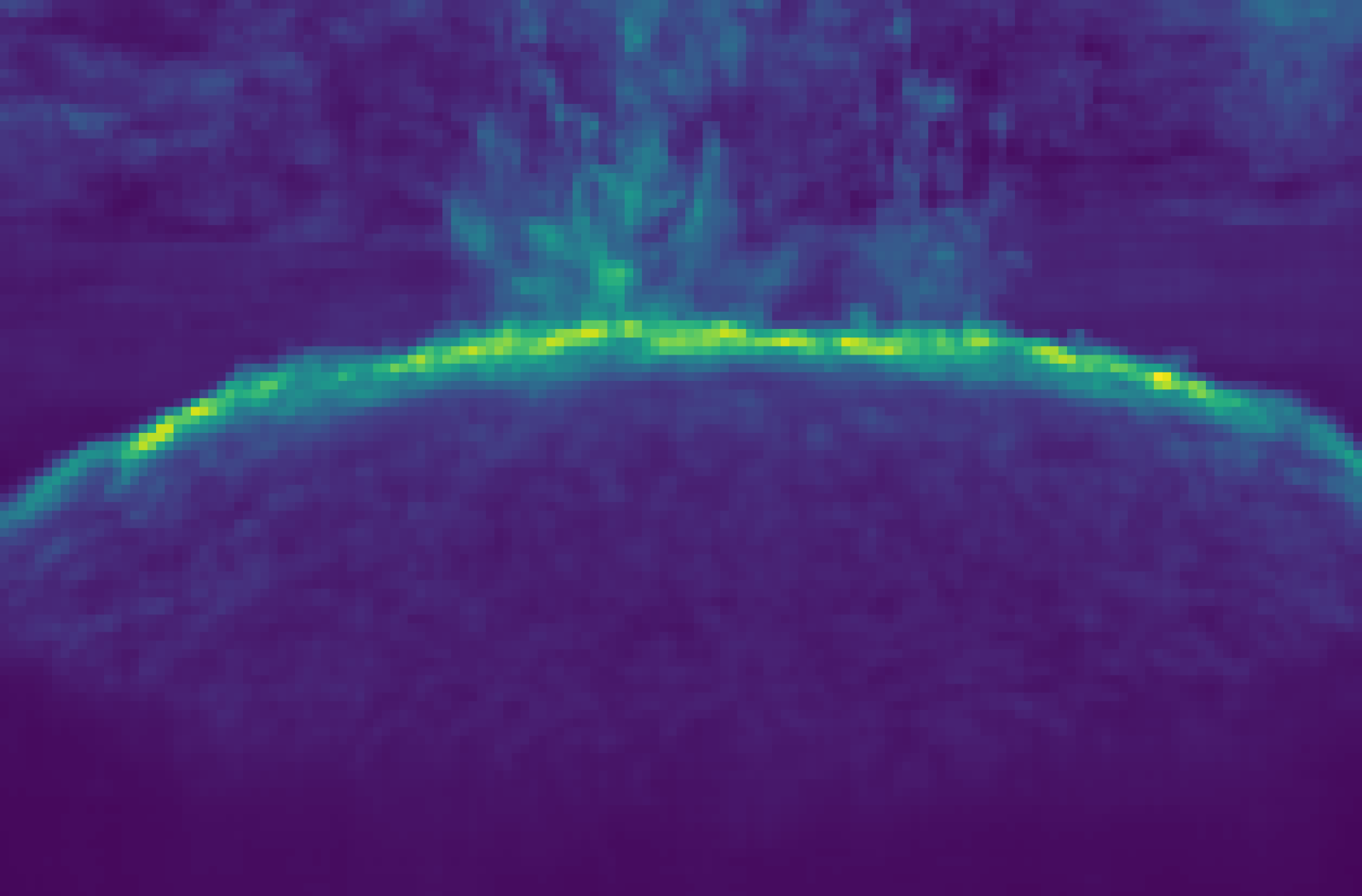}
    \end{subfigure}
    \begin{subfigure}[t]{0.195\textwidth}
        \centering
        \includegraphics[width=\linewidth]{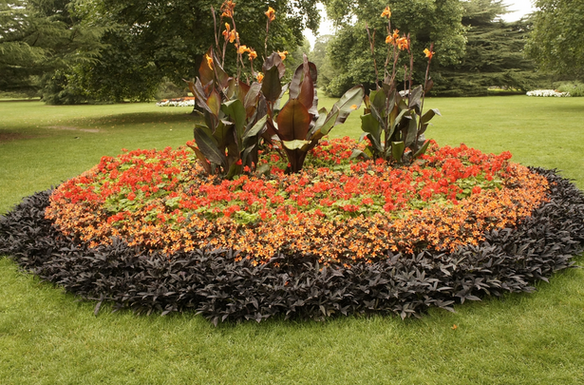}
    \end{subfigure}\\

    \begin{subfigure}[t]{0.195\textwidth}
        \centering
        \includegraphics[width=\linewidth]{img12/8052_3dgs_8x8_iter_030000_psnr_27.46_time_1209s_DSC08052_heatmap.png}
    \end{subfigure}
    \begin{subfigure}[t]{0.195\textwidth}
        \centering
        \includegraphics[width=\linewidth]{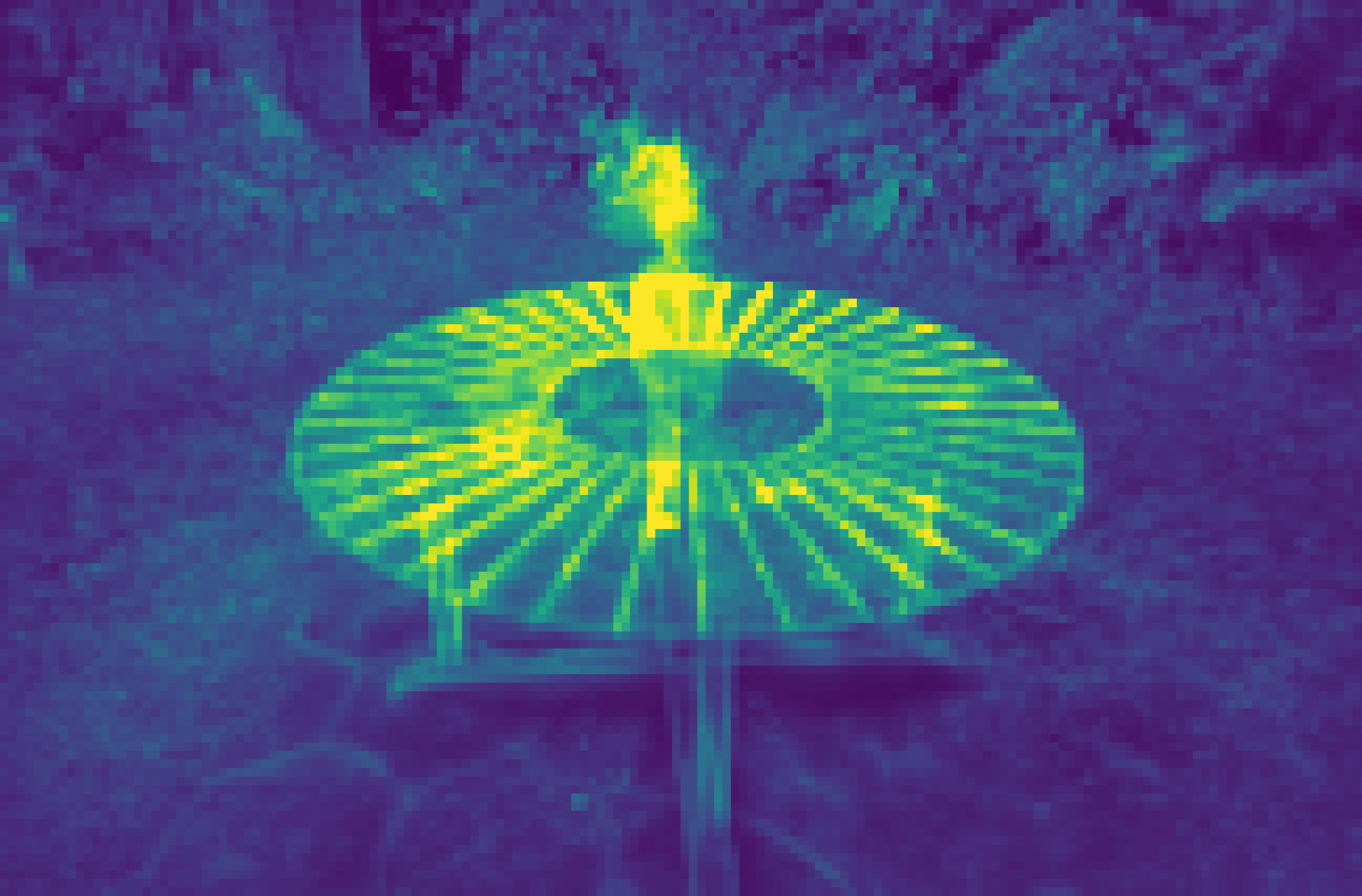}
    \end{subfigure}
    \begin{subfigure}[t]{0.195\textwidth}
        \centering
        \includegraphics[width=\linewidth]{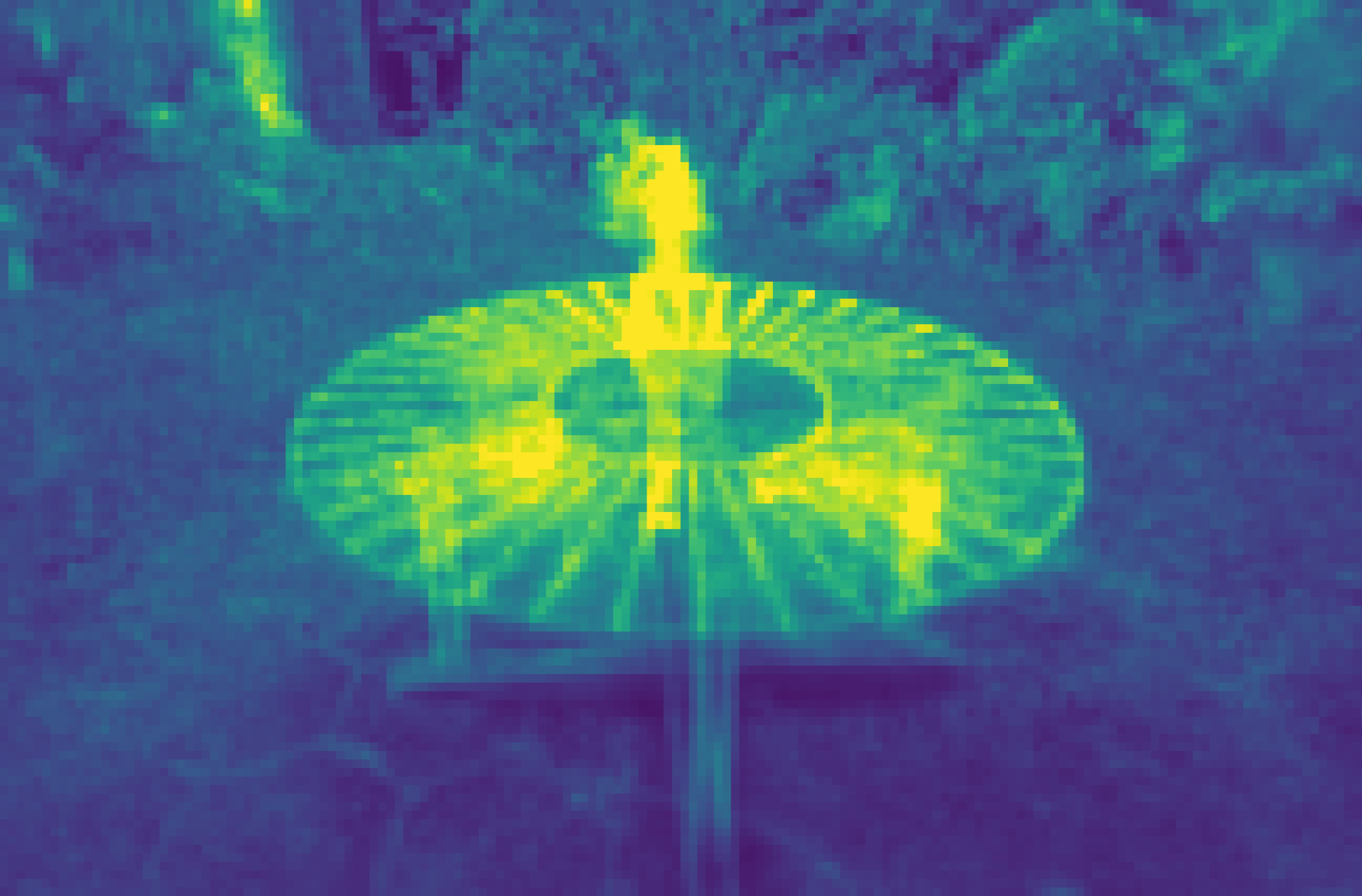}
    \end{subfigure}
    \begin{subfigure}[t]{0.195\textwidth}
        \centering
        \includegraphics[width=\linewidth]{img12/8052_ours_epoch_0149_DSC08052_gaussian_heatmap.png}
    \end{subfigure}
    \begin{subfigure}[t]{0.195\textwidth}
        \centering
        \includegraphics[width=\linewidth]{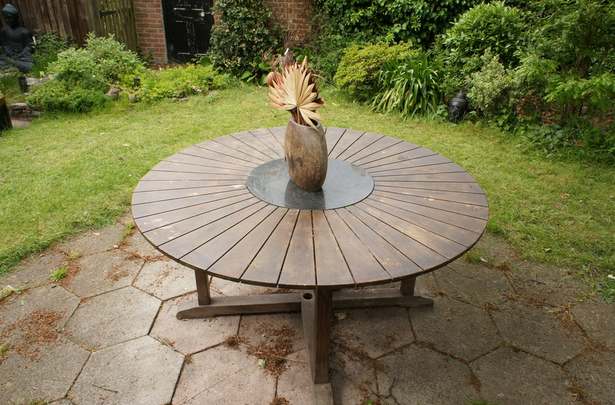}
    \end{subfigure}\\
    
    \begin{subfigure}[t]{0.195\textwidth}
        \centering
        \includegraphics[width=\linewidth]{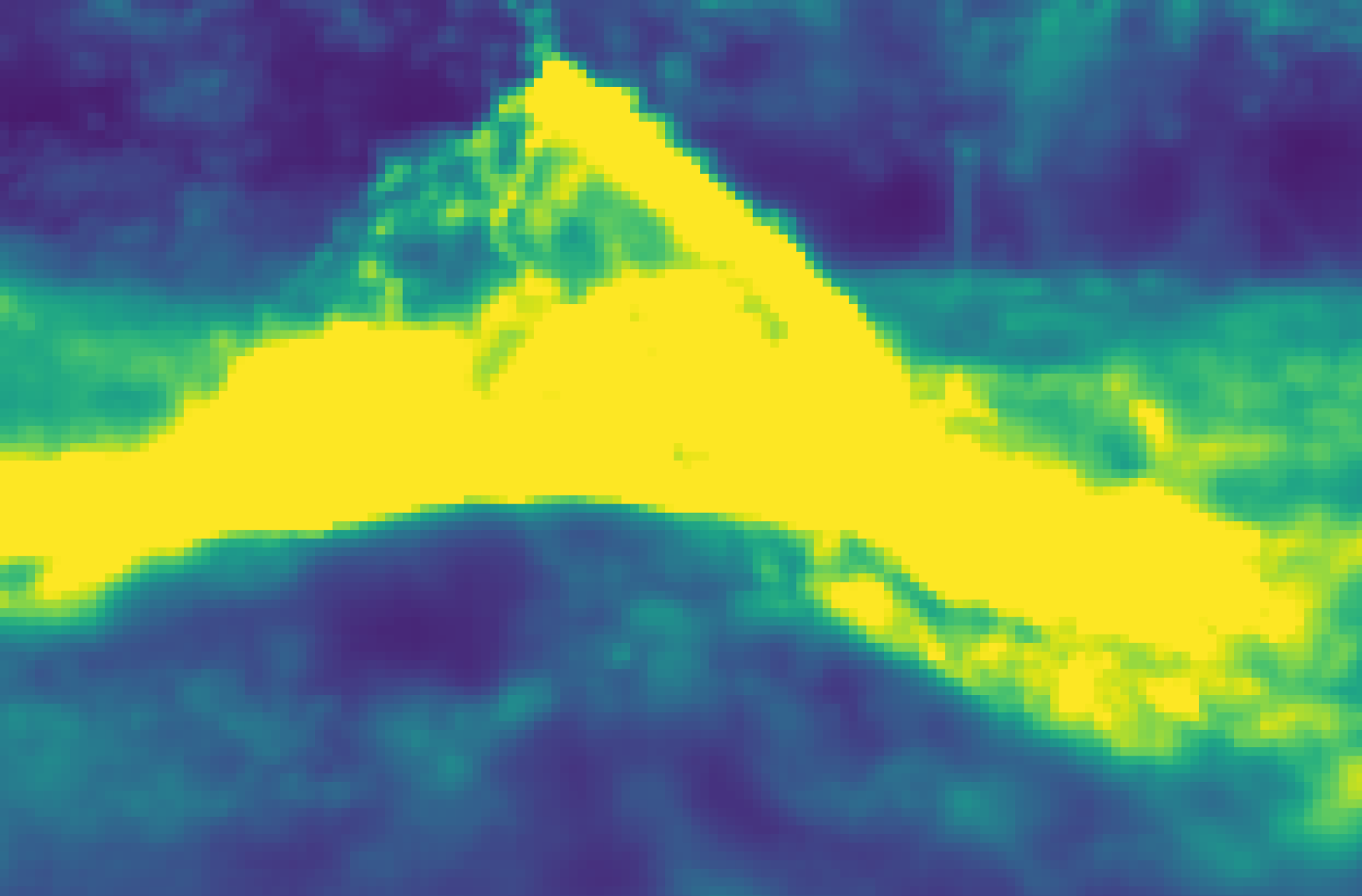}
    \end{subfigure}
    \begin{subfigure}[t]{0.195\textwidth}
        \centering
        \includegraphics[width=\linewidth]{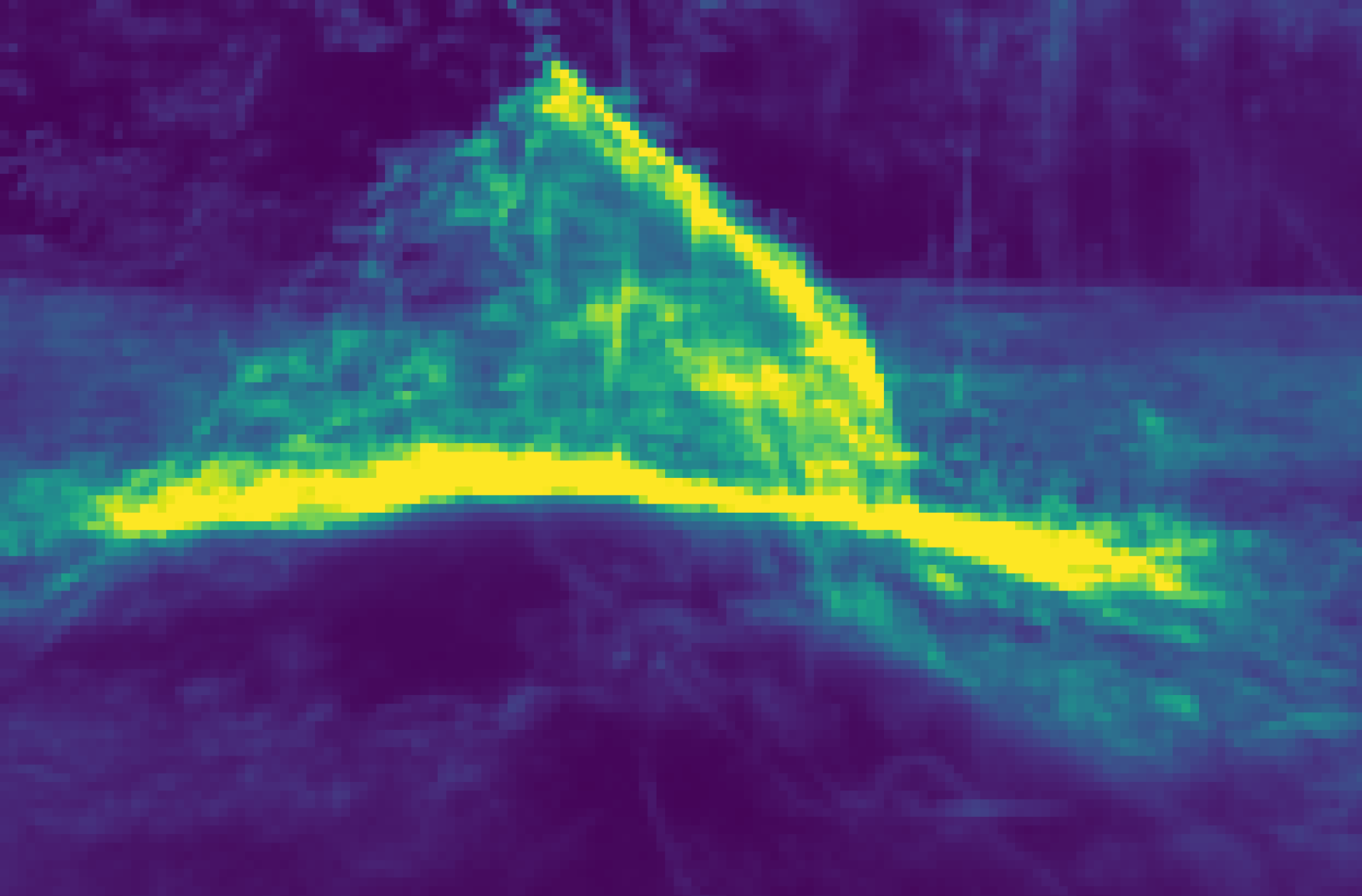}
    \end{subfigure}
    \begin{subfigure}[t]{0.195\textwidth}
        \centering
        \includegraphics[width=\linewidth]{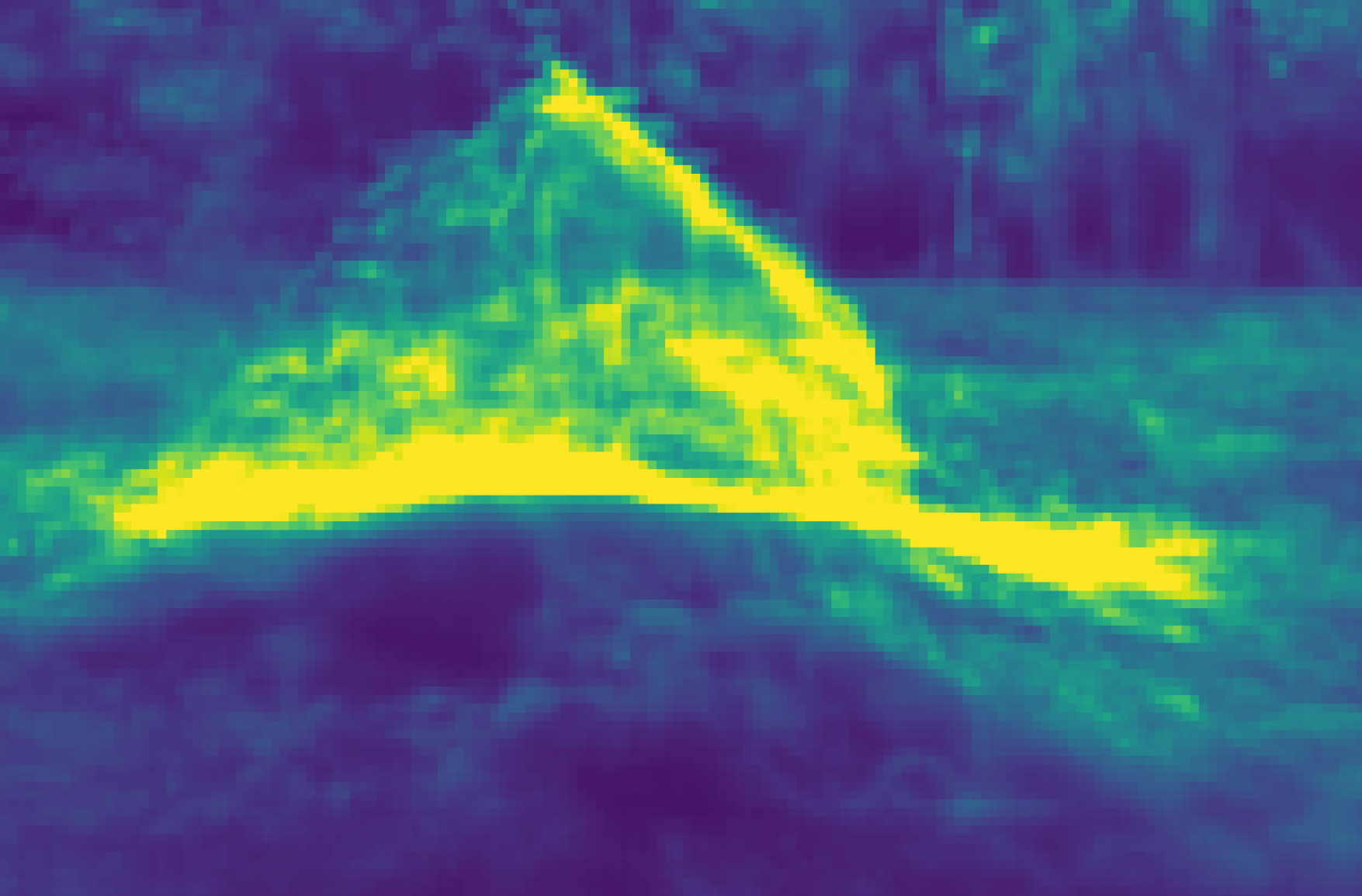}
    \end{subfigure}
    \begin{subfigure}[t]{0.195\textwidth}
        \centering
        \includegraphics[width=\linewidth]{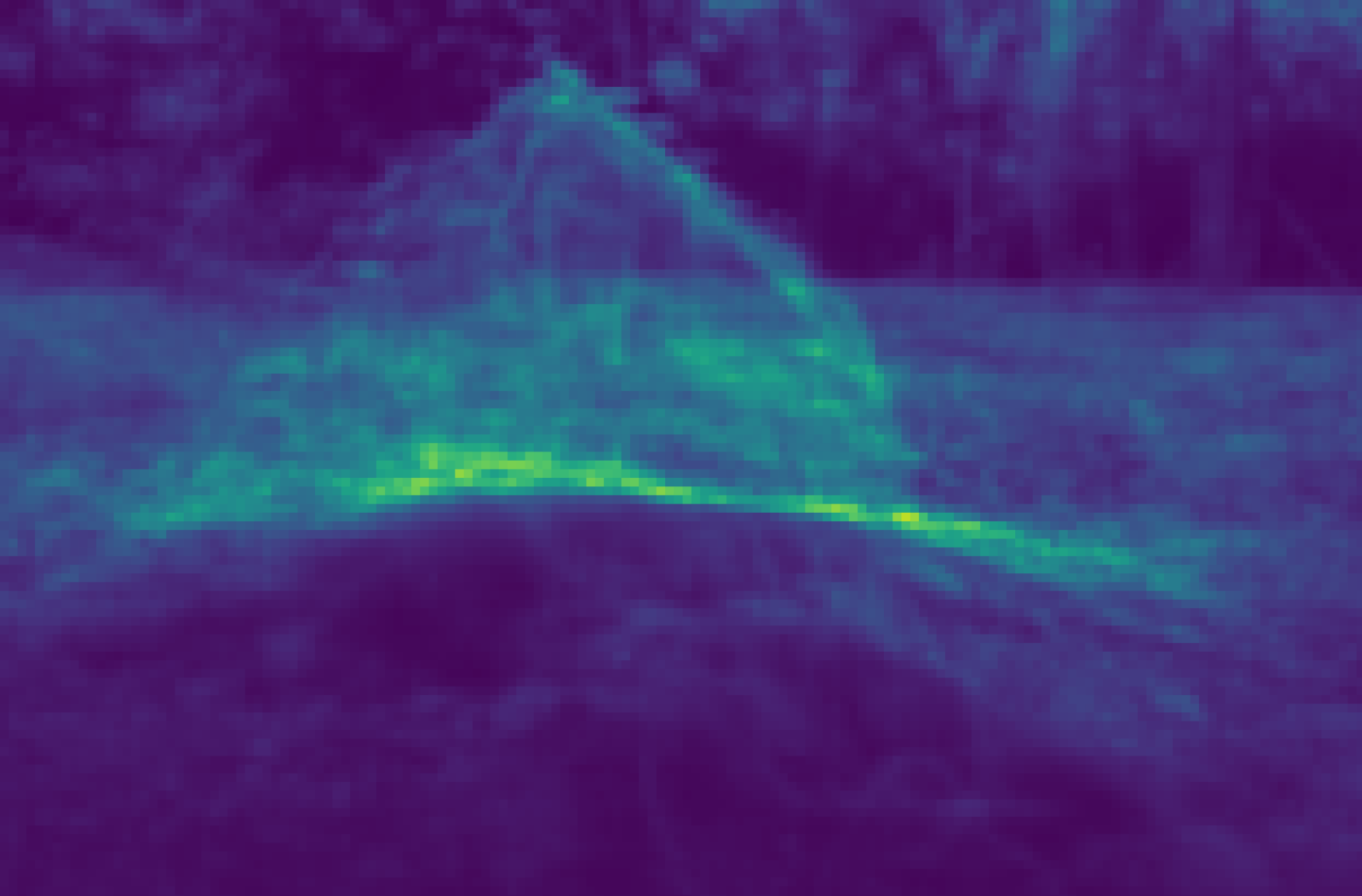}
    \end{subfigure}
    \begin{subfigure}[t]{0.195\textwidth}
        \centering
        \includegraphics[width=\linewidth]{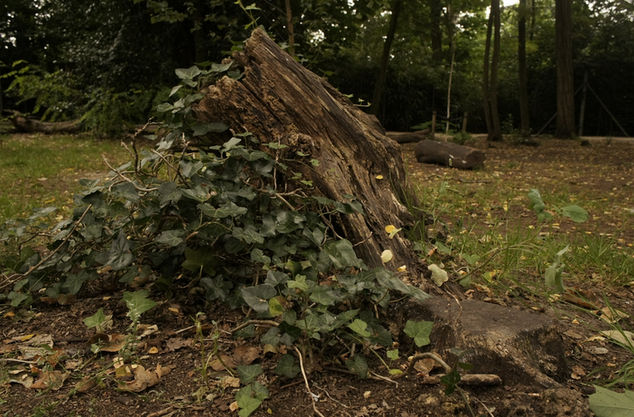}
    \end{subfigure}\\

    \begin{subfigure}[t]{0.195\textwidth}
        \centering
        \includegraphics[width=\linewidth]{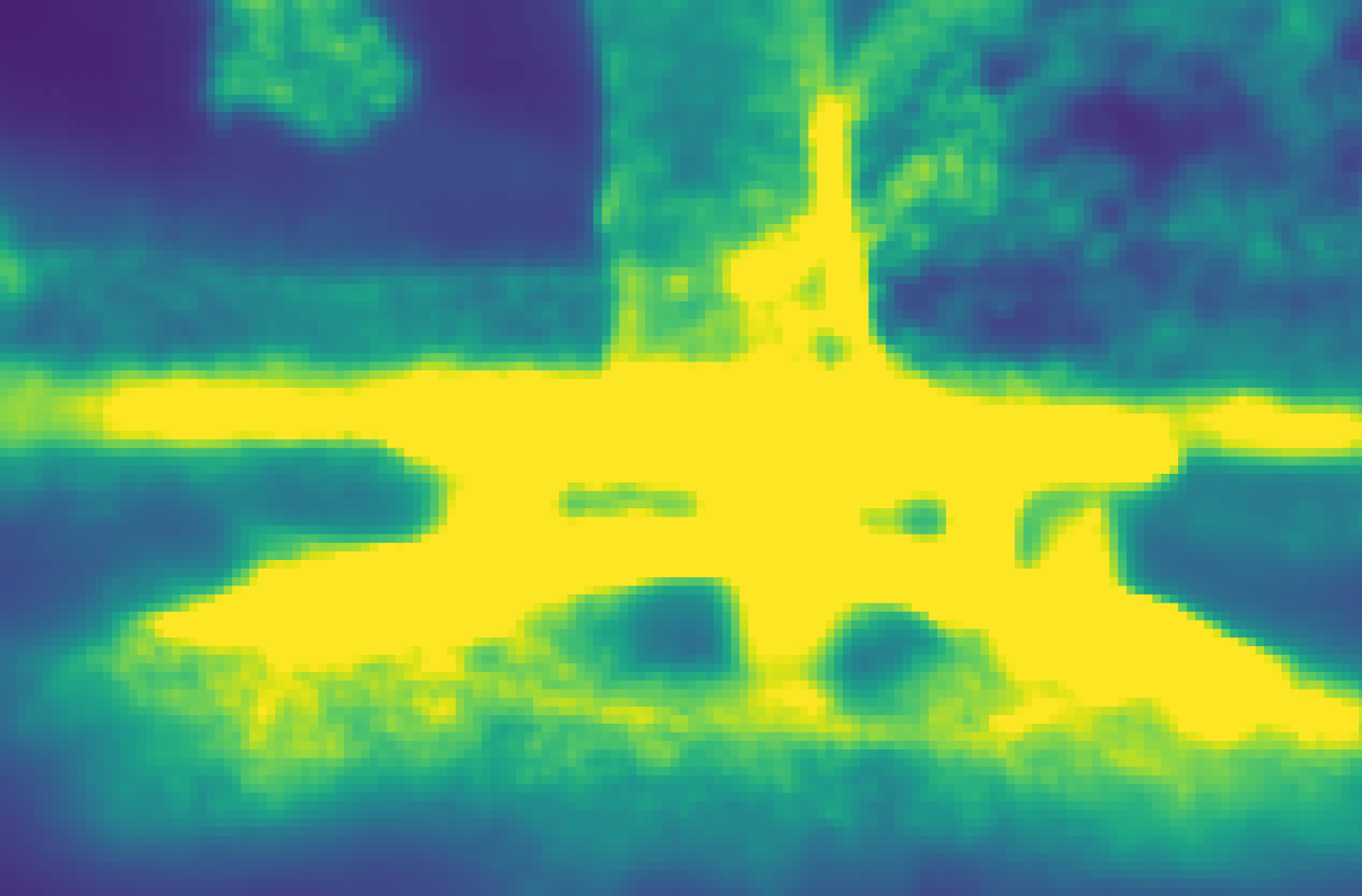}
    \end{subfigure}
    \begin{subfigure}[t]{0.195\textwidth}
        \centering
        \includegraphics[width=\linewidth]{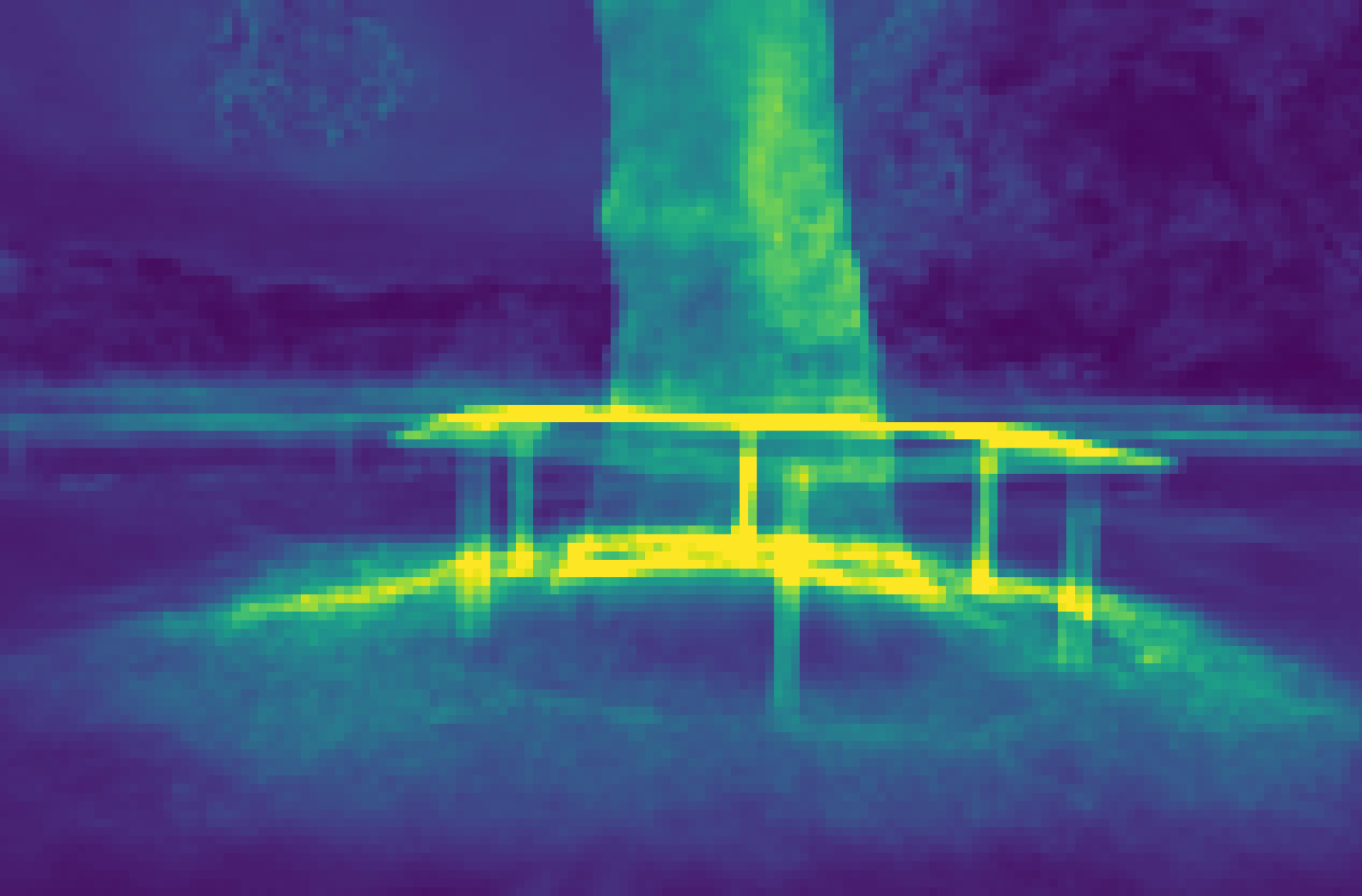}
    \end{subfigure}
    \begin{subfigure}[t]{0.195\textwidth}
        \centering
        \includegraphics[width=\linewidth]{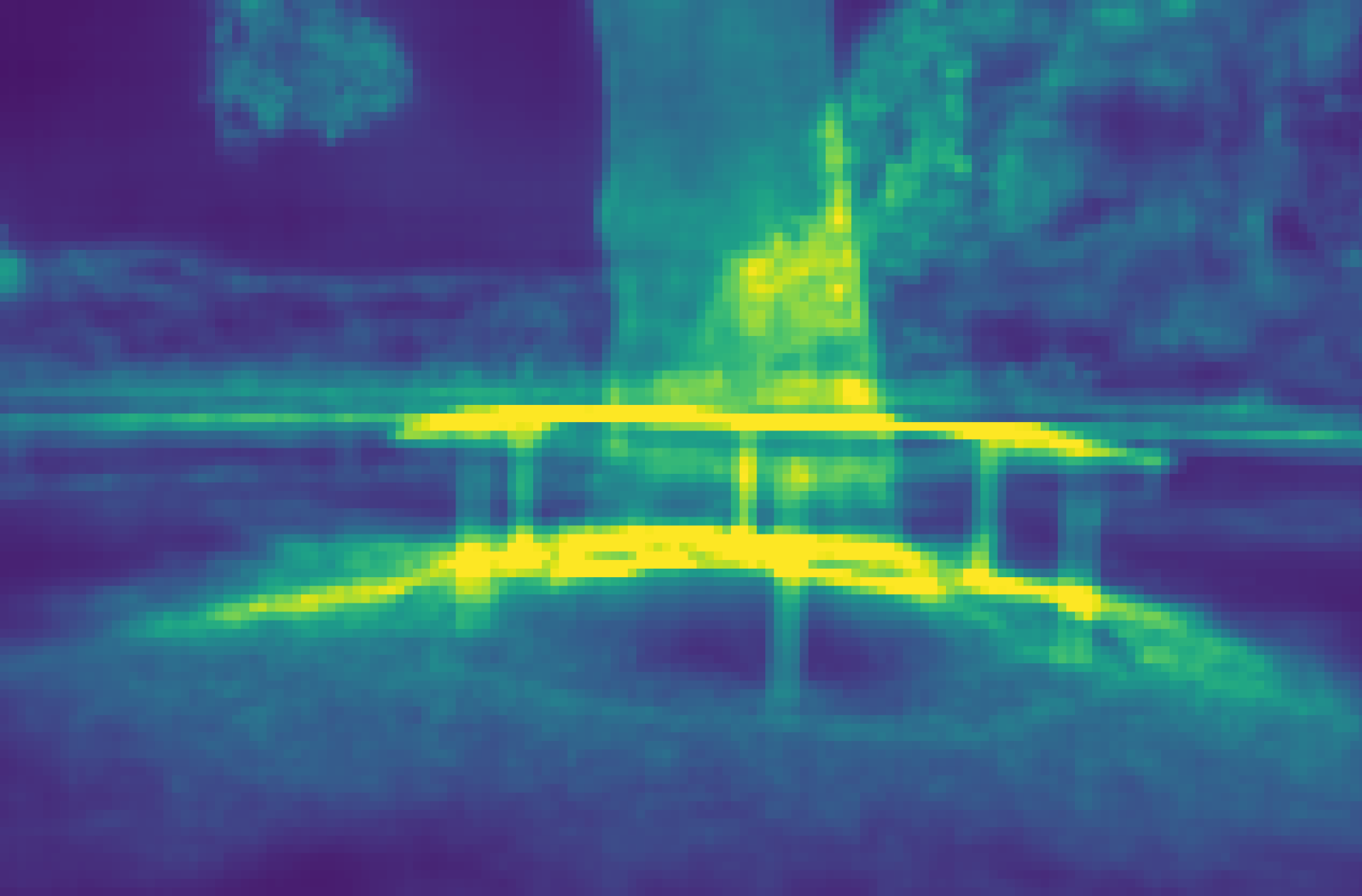}
    \end{subfigure}
    \begin{subfigure}[t]{0.195\textwidth}
        \centering
        \includegraphics[width=\linewidth]{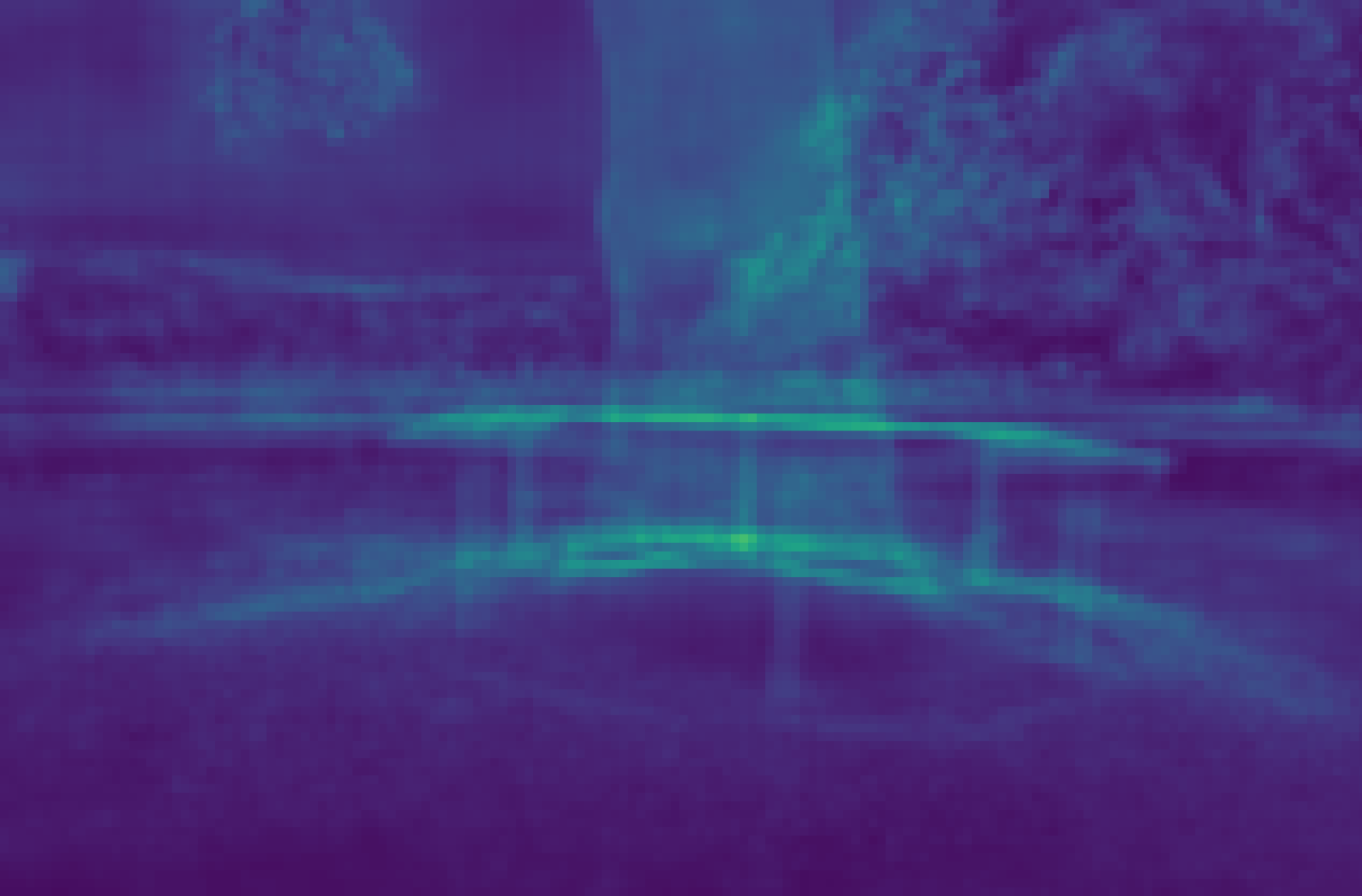}
    \end{subfigure}
    \begin{subfigure}[t]{0.195\textwidth}
        \centering
        \includegraphics[width=\linewidth]{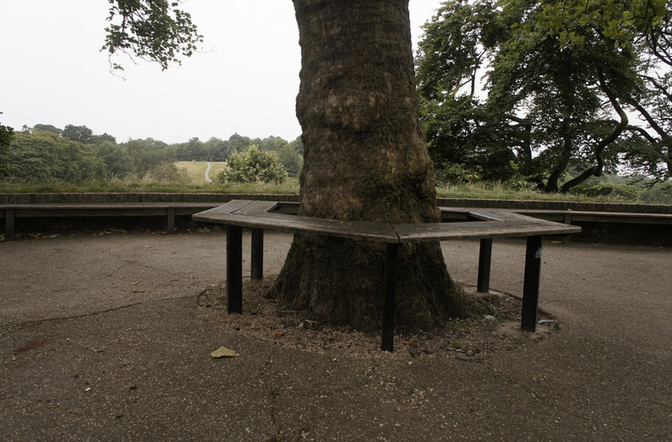}
    \end{subfigure}\\
    
    \begin{subfigure}[t]{0.195\textwidth}
        \centering
        \includegraphics[width=\linewidth]{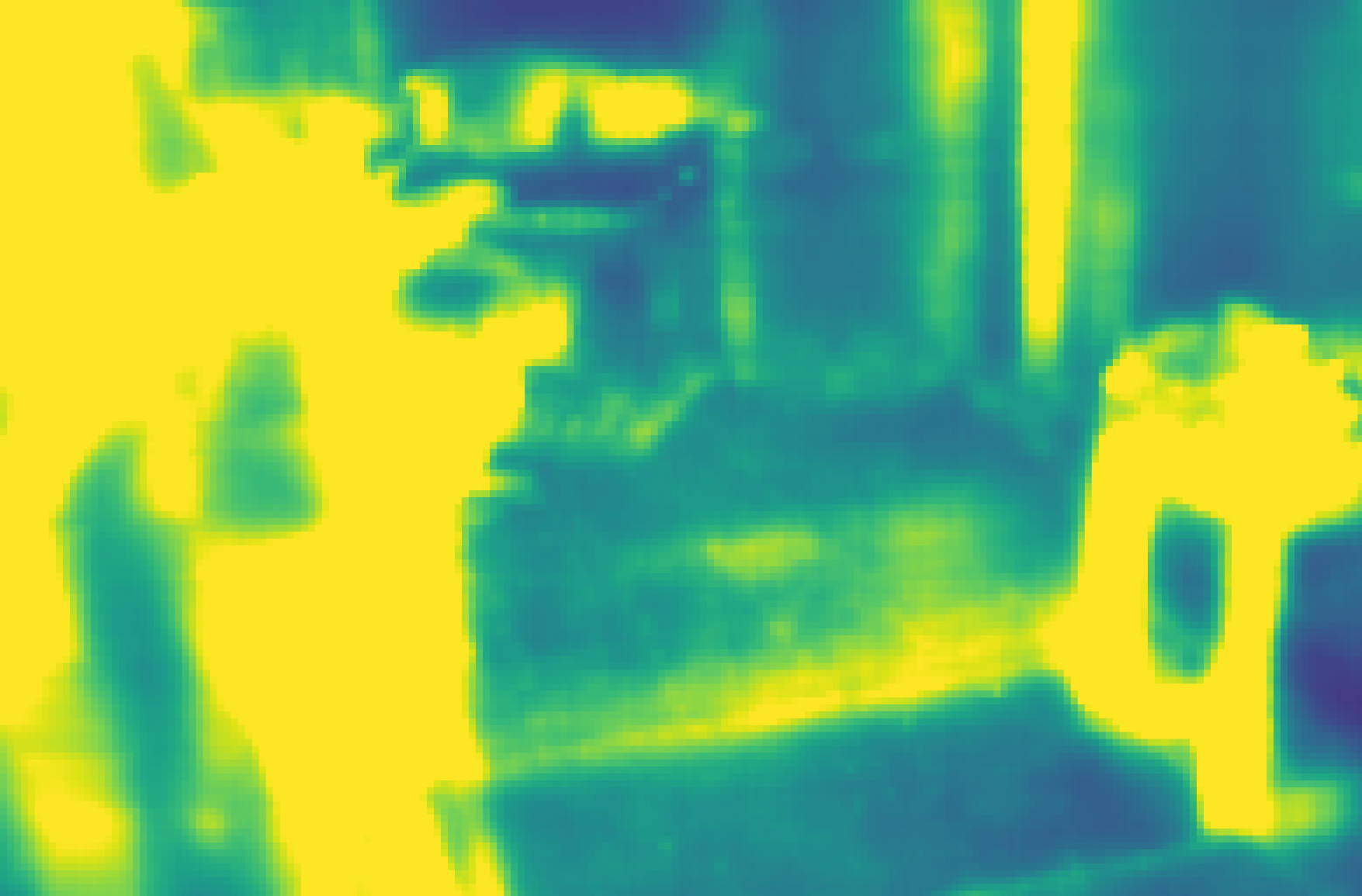}
    \end{subfigure}
    \begin{subfigure}[t]{0.195\textwidth}
        \centering
        \includegraphics[width=\linewidth]{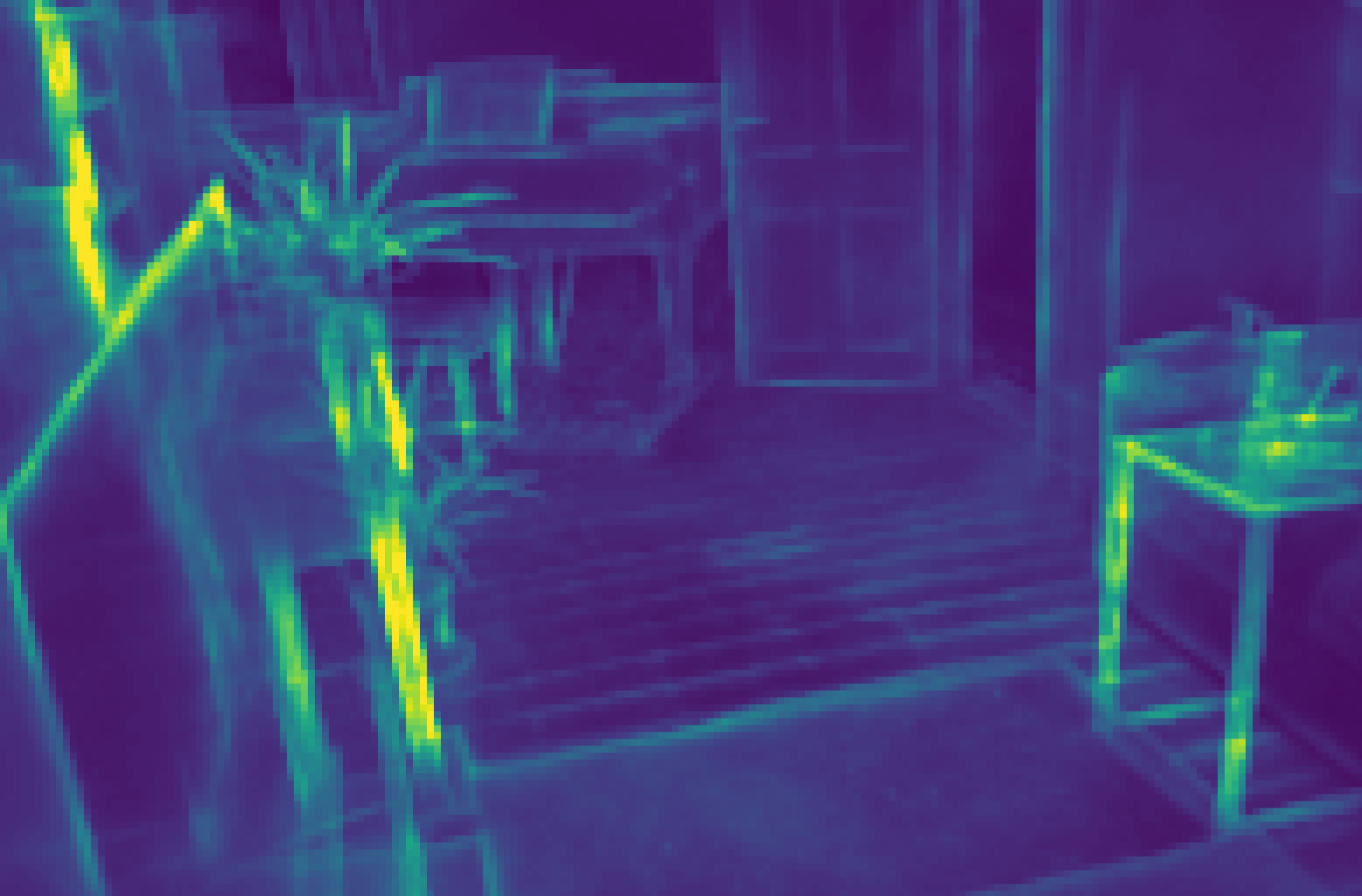}
    \end{subfigure}
    \begin{subfigure}[t]{0.195\textwidth}
        \centering
        \includegraphics[width=\linewidth]{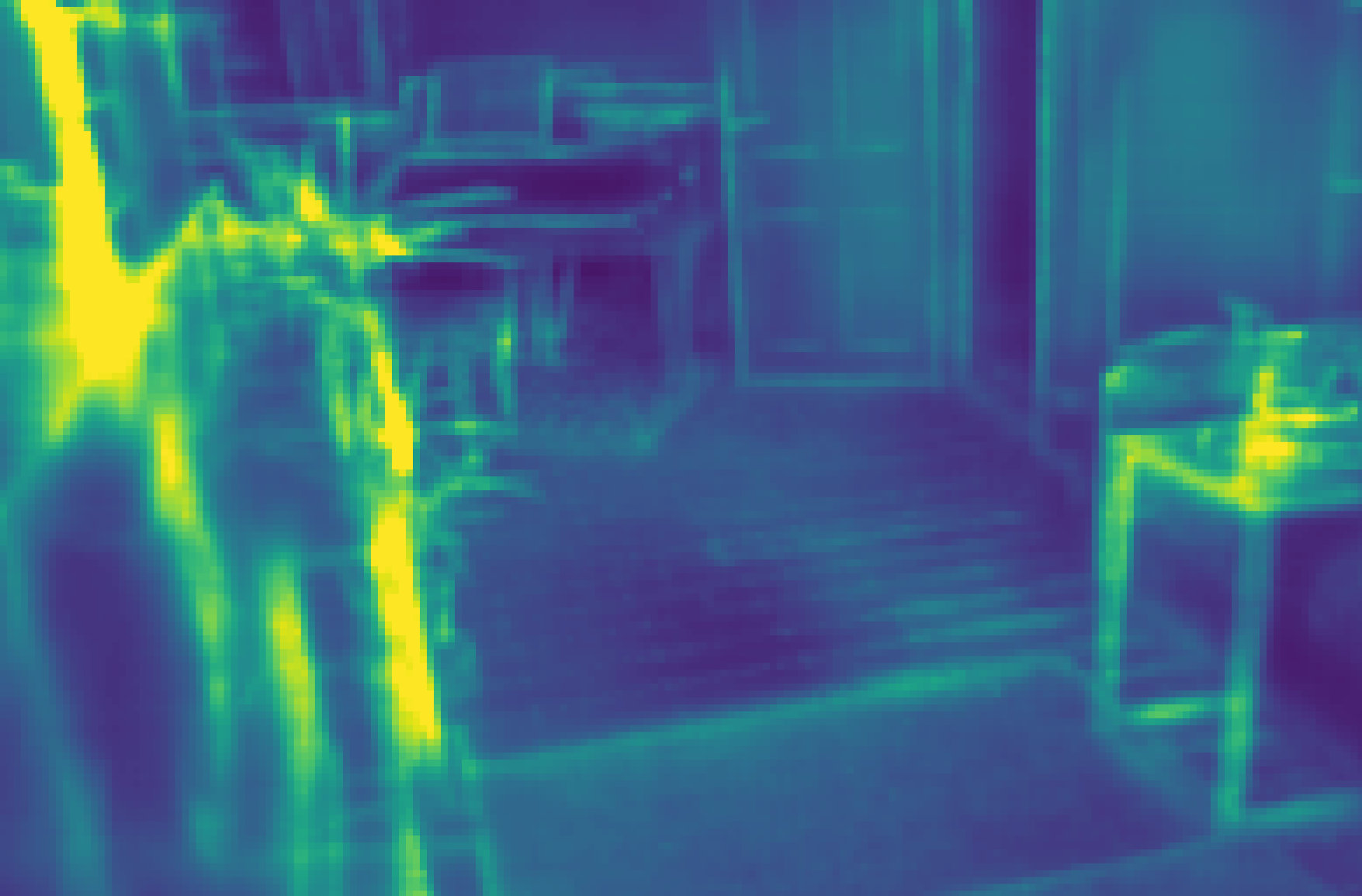}
    \end{subfigure}
    \begin{subfigure}[t]{0.195\textwidth}
        \centering
        \includegraphics[width=\linewidth]{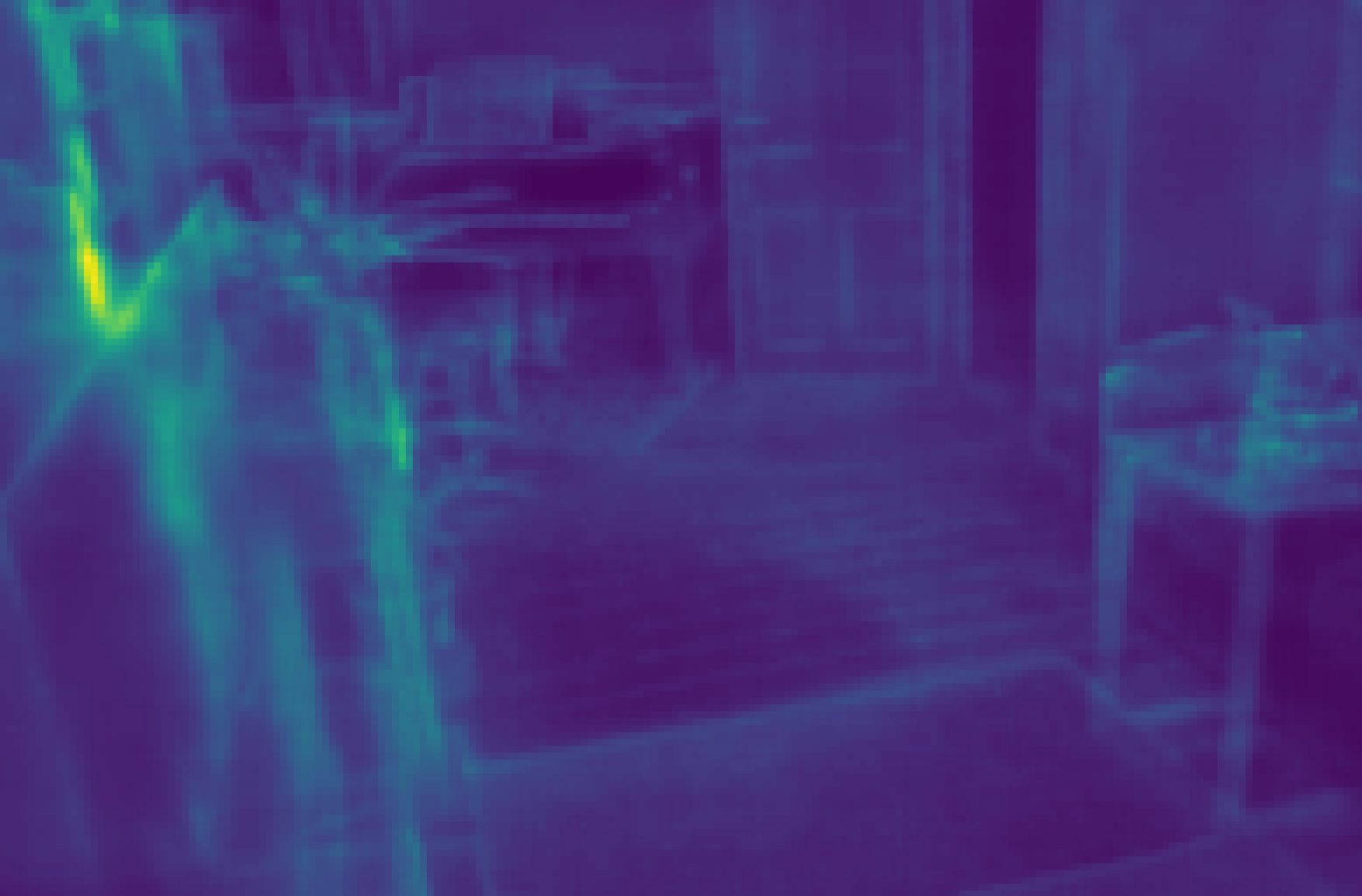}
    \end{subfigure}
    \begin{subfigure}[t]{0.195\textwidth}
        \centering
        \includegraphics[width=\linewidth]{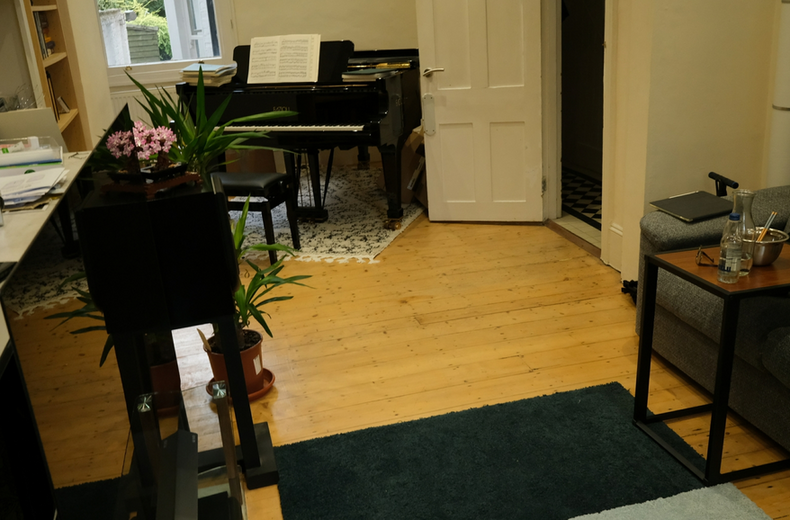}
    \end{subfigure}\\

    \begin{subfigure}[t]{0.195\textwidth}
        \centering
        \includegraphics[width=\linewidth]{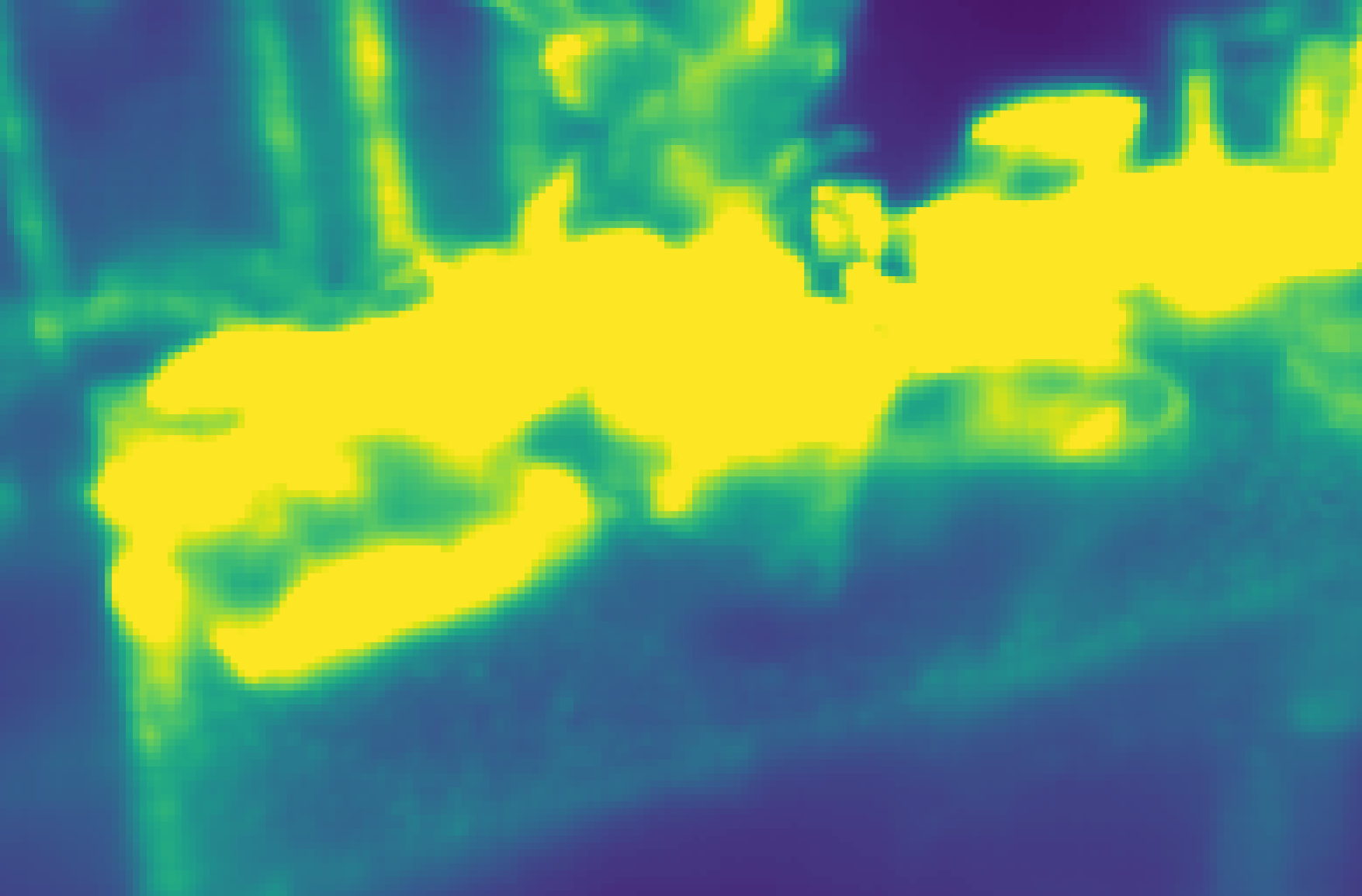}
    \end{subfigure}
    \begin{subfigure}[t]{0.195\textwidth}
        \centering
        \includegraphics[width=\linewidth]{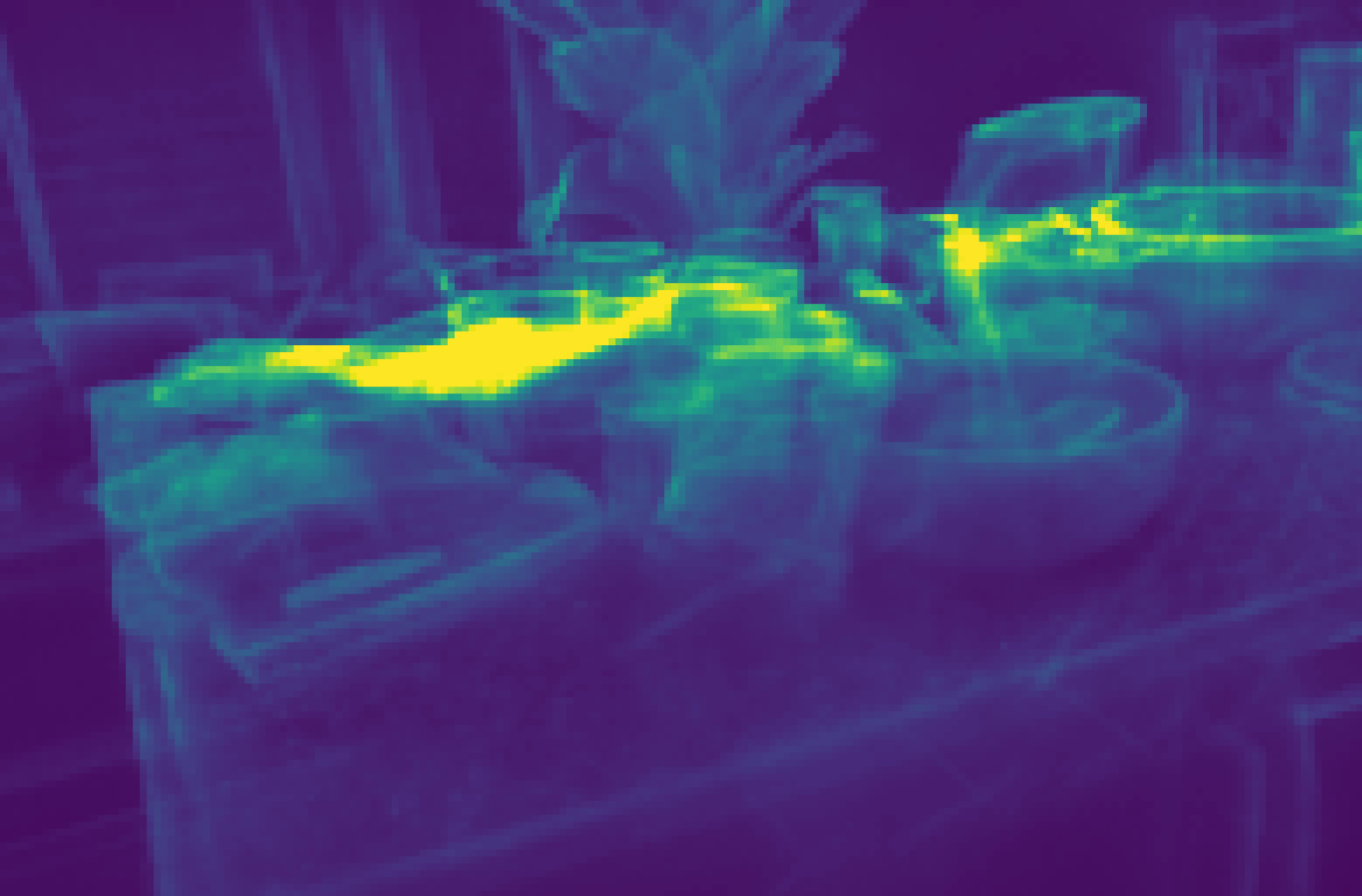}
    \end{subfigure}
    \begin{subfigure}[t]{0.195\textwidth}
        \centering
        \includegraphics[width=\linewidth]{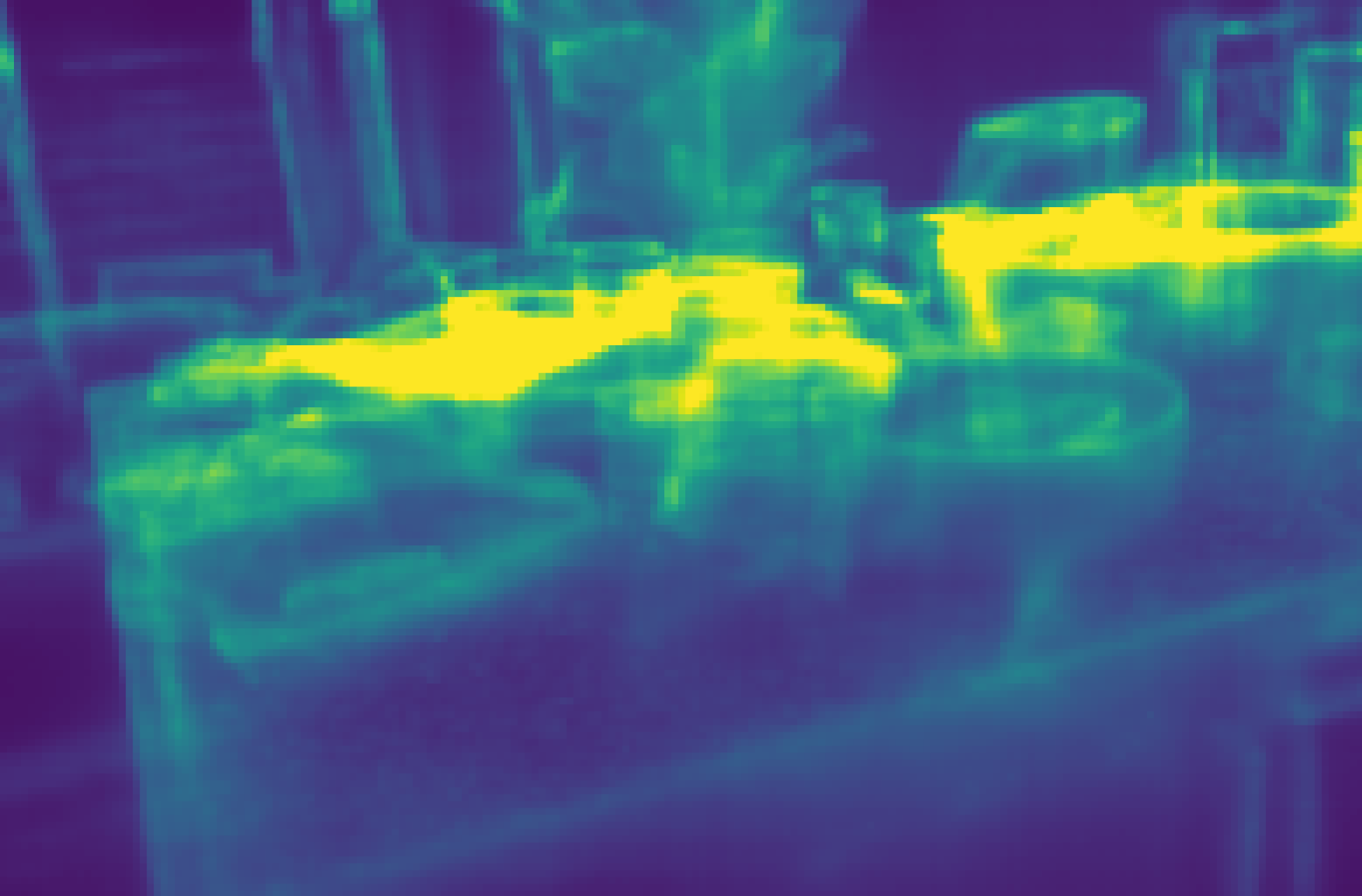}
    \end{subfigure}
    \begin{subfigure}[t]{0.195\textwidth}
        \centering
        \includegraphics[width=\linewidth]{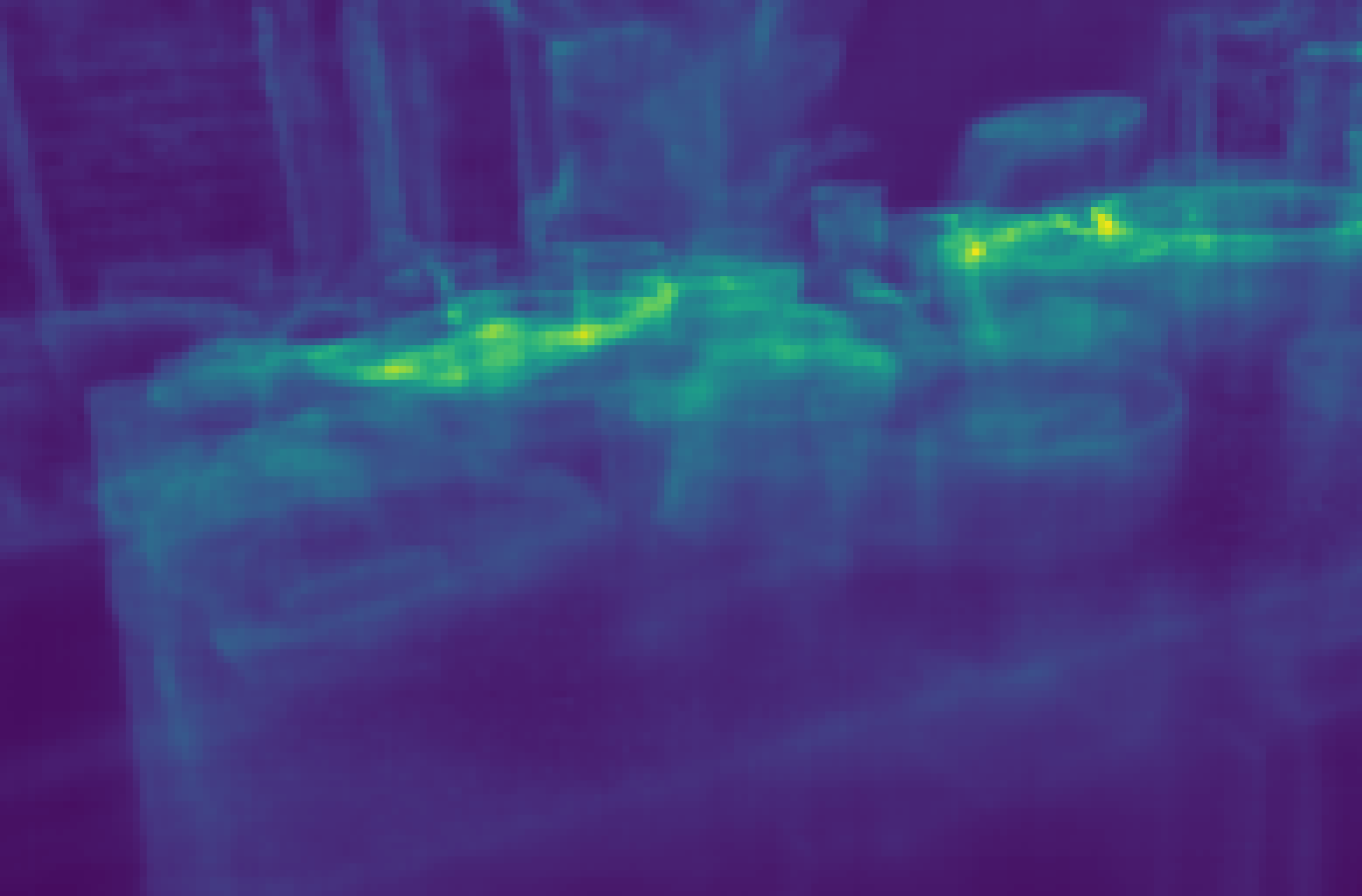}
    \end{subfigure}
    \begin{subfigure}[t]{0.195\textwidth}
        \centering
        \includegraphics[width=\linewidth]{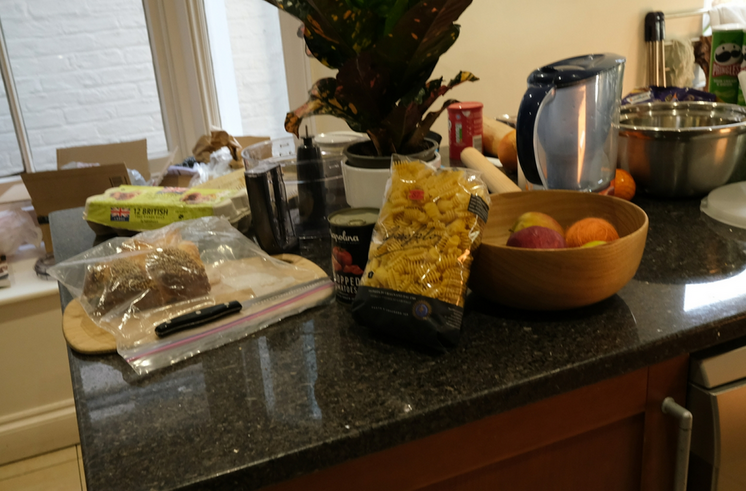}
    \end{subfigure}\\

    \begin{subfigure}[t]{0.195\textwidth}
        \centering
        \includegraphics[width=\linewidth]{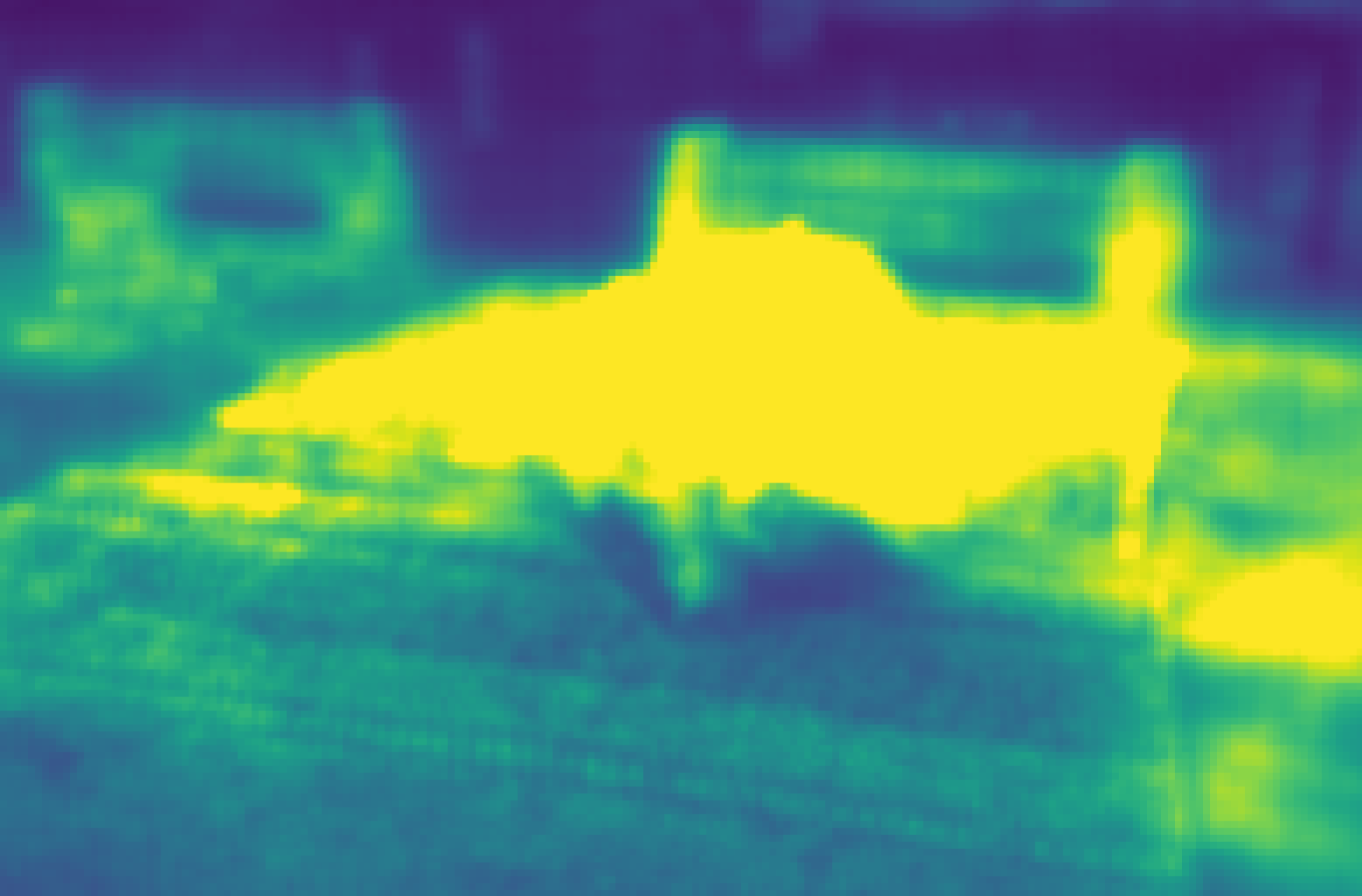}
    \end{subfigure}
    \begin{subfigure}[t]{0.195\textwidth}
        \centering
        \includegraphics[width=\linewidth]{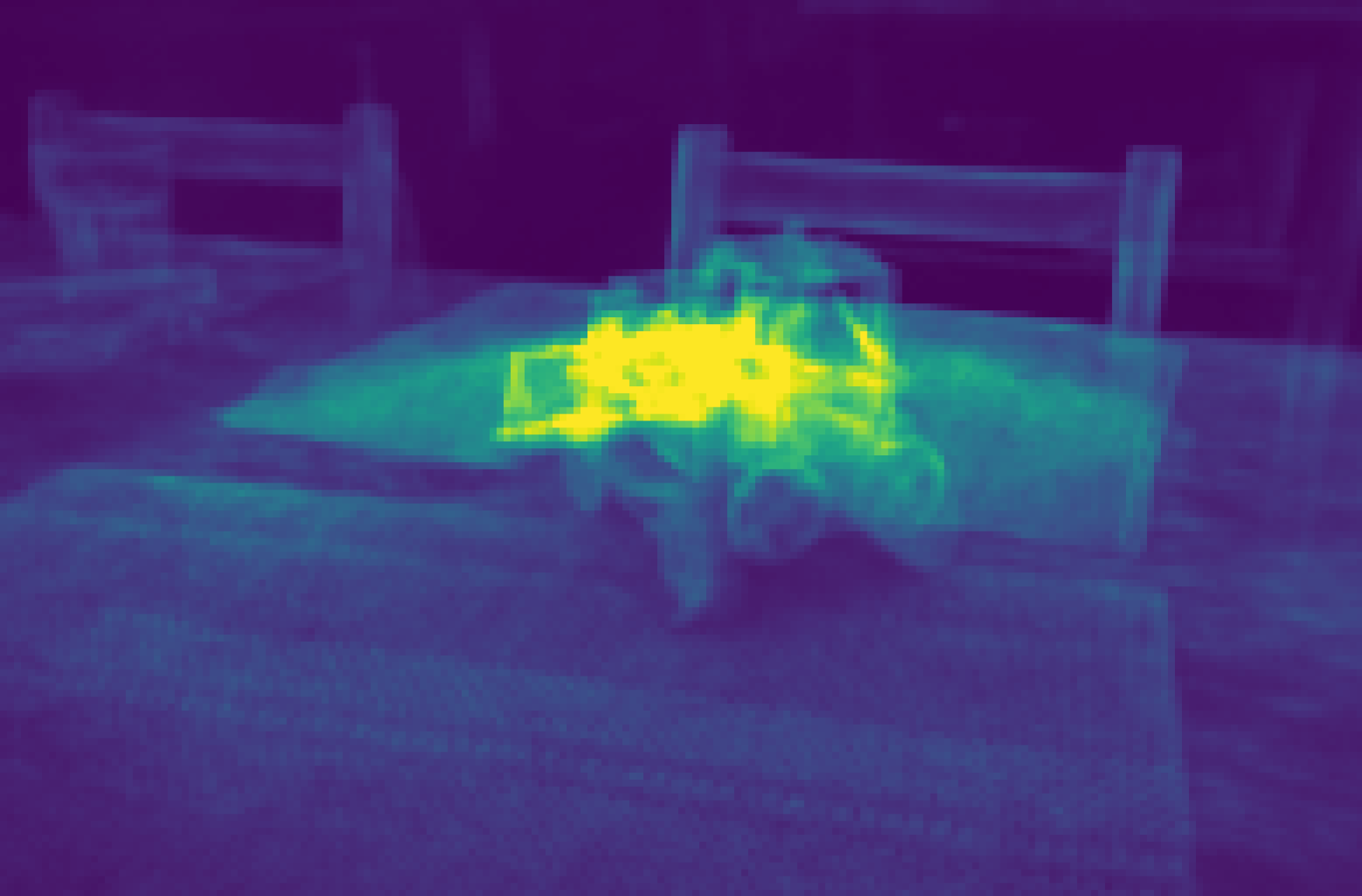}
    \end{subfigure}
    \begin{subfigure}[t]{0.195\textwidth}
        \centering
        \includegraphics[width=\linewidth]{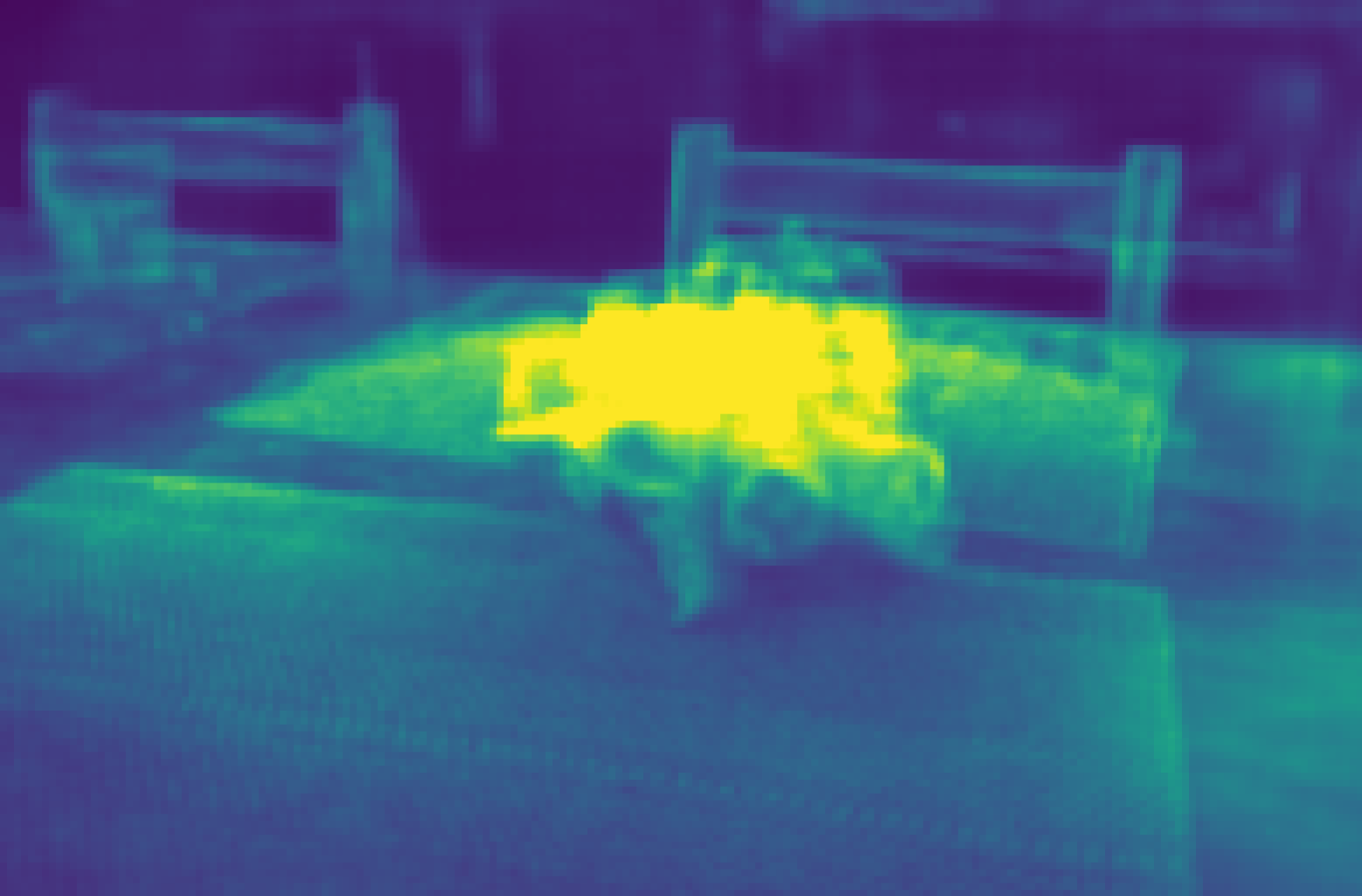}
    \end{subfigure}
    \begin{subfigure}[t]{0.195\textwidth}
        \centering
        \includegraphics[width=\linewidth]{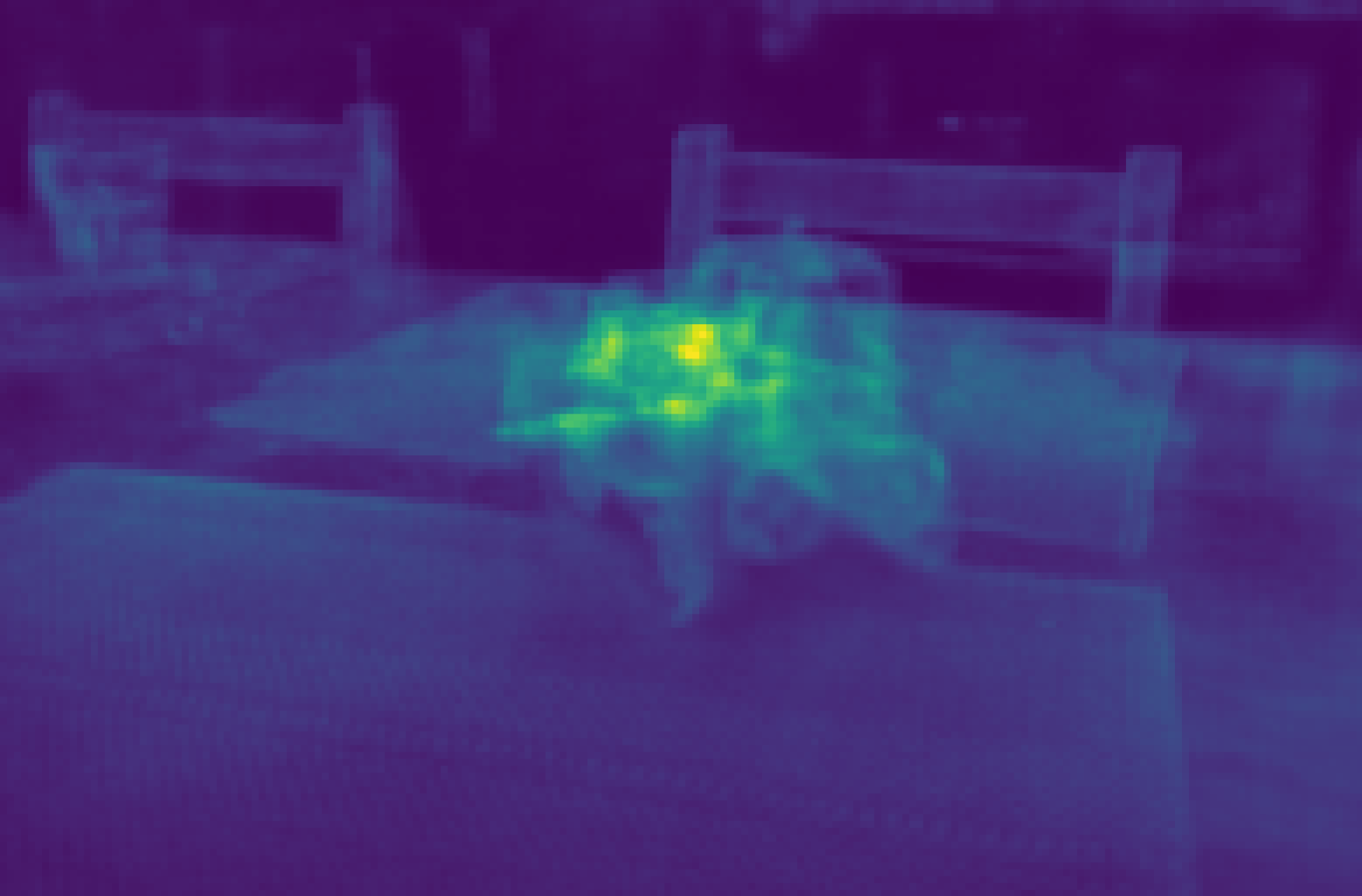}
    \end{subfigure}
    \begin{subfigure}[t]{0.195\textwidth}
        \centering
        \includegraphics[width=\linewidth]{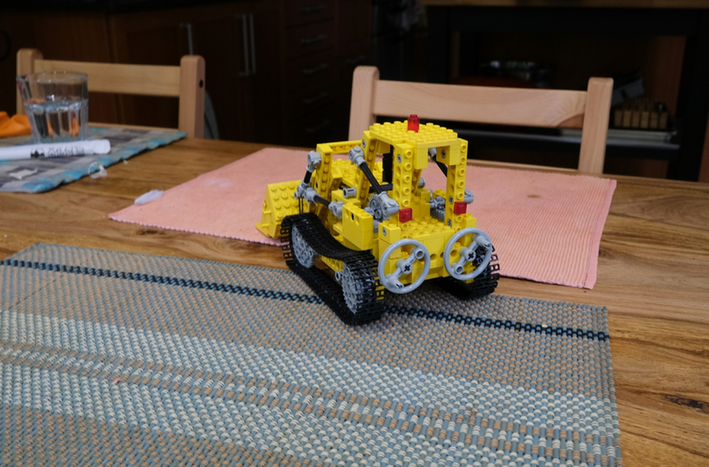}
    \end{subfigure}\\

    \begin{subfigure}[t]{0.195\textwidth}
        \centering
        \includegraphics[width=\linewidth]{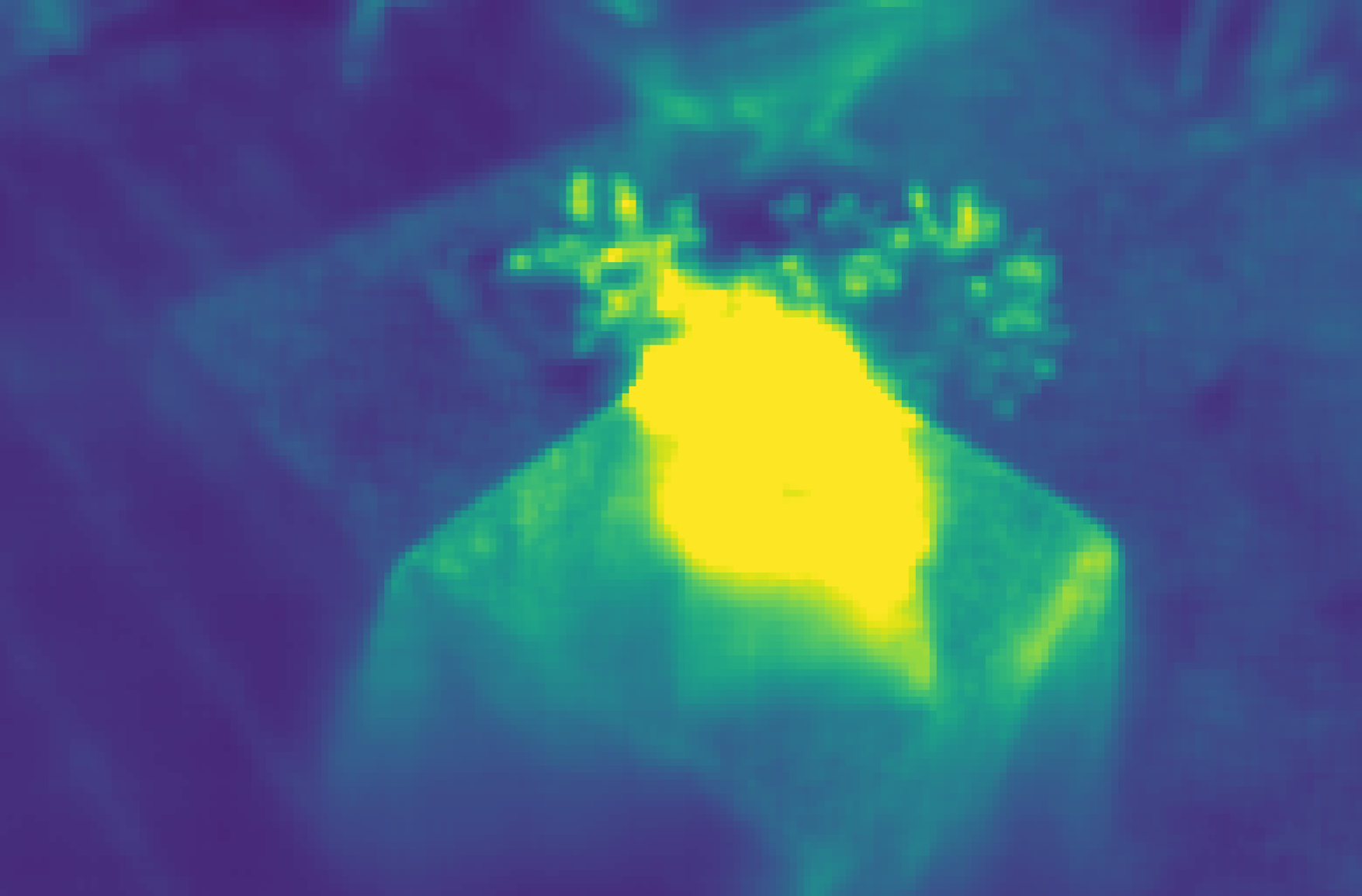}
        \caption*{\scriptsize 3DGS}
    \end{subfigure}
    \begin{subfigure}[t]{0.195\textwidth}
        \centering
        \includegraphics[width=\linewidth]{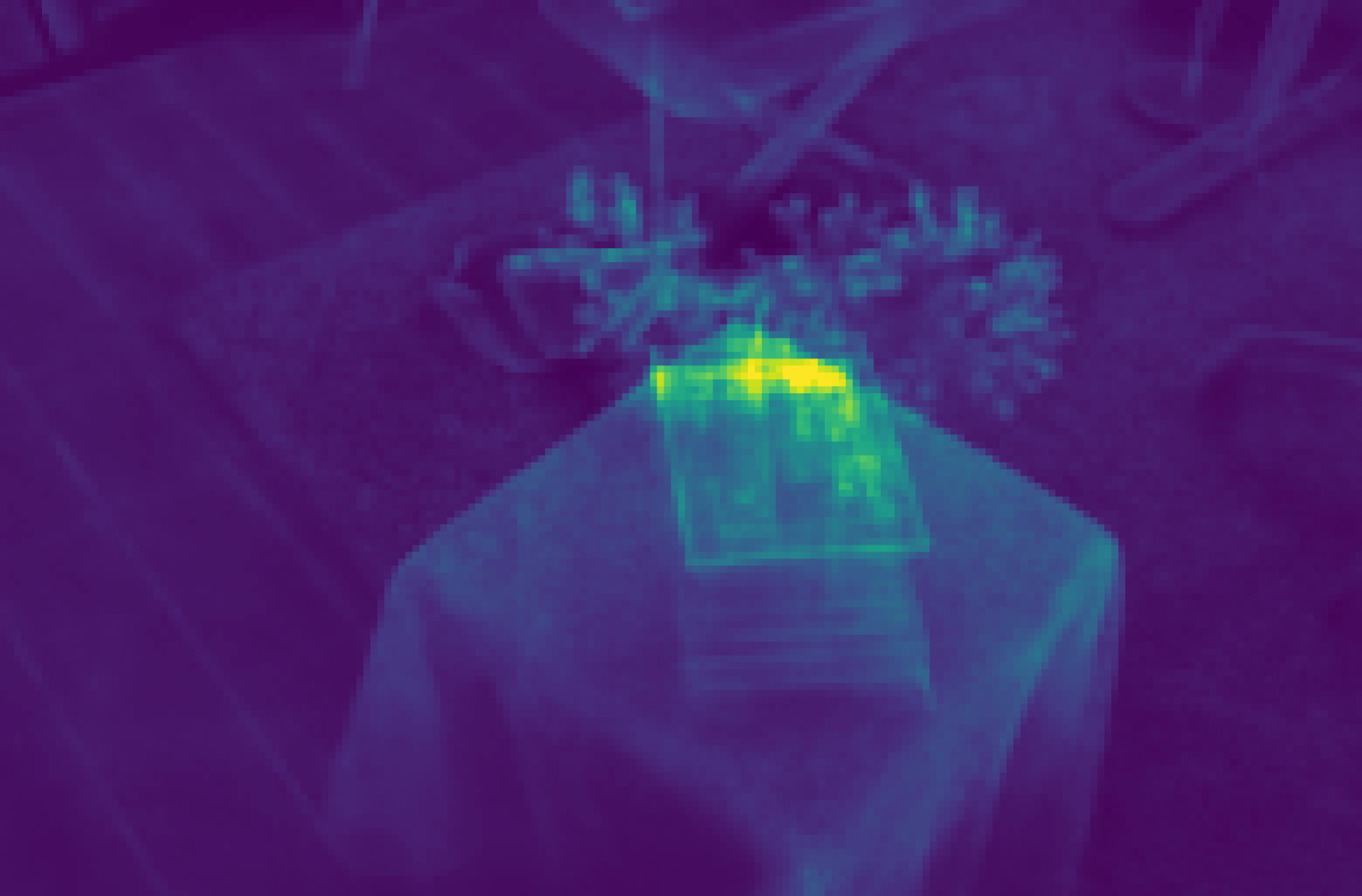}
        \caption*{\scriptsize Taming-3DGS}
    \end{subfigure}
    \begin{subfigure}[t]{0.195\textwidth}
        \centering
        \includegraphics[width=\linewidth]{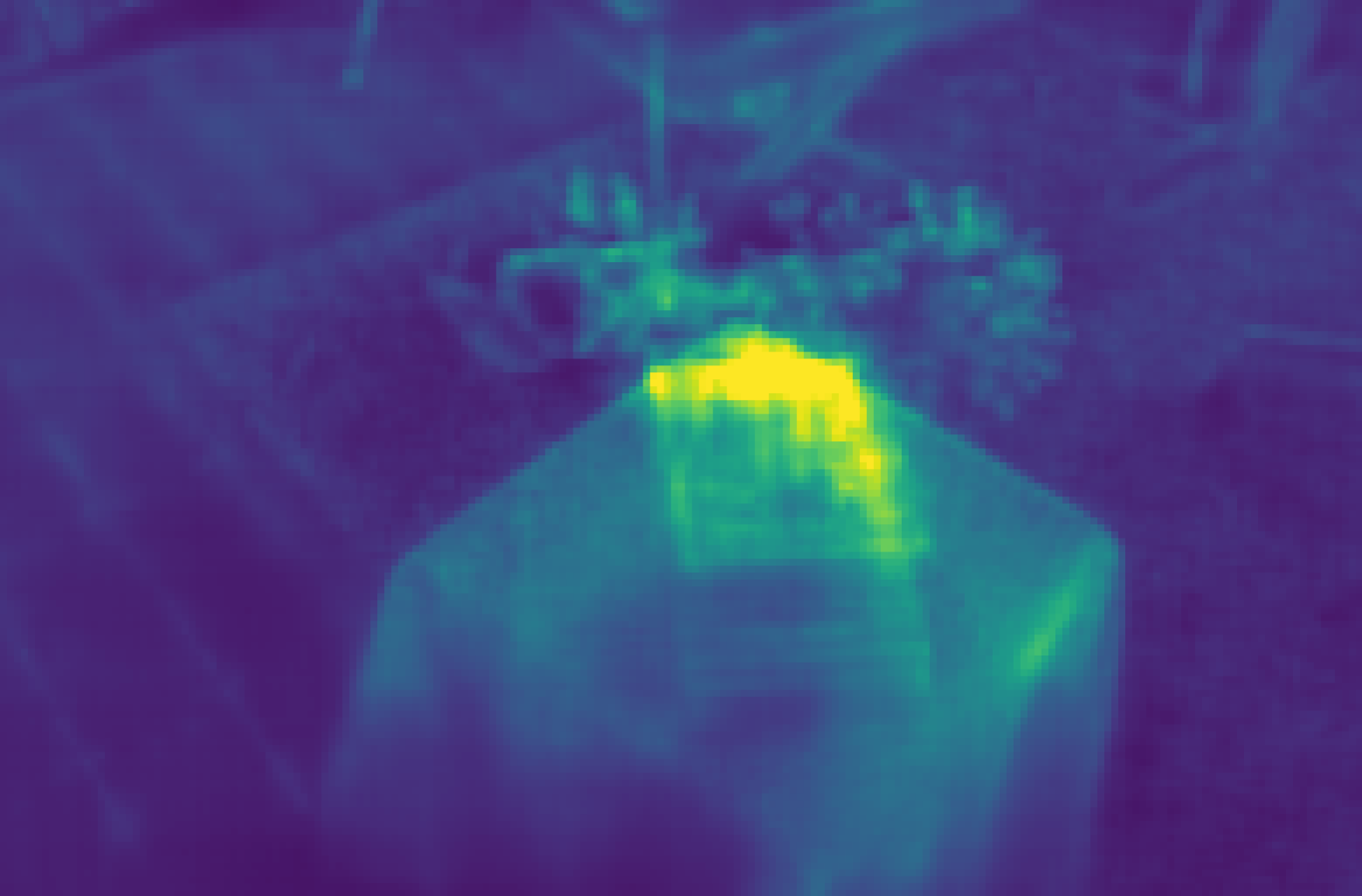}
        \caption*{\scriptsize LiteGS}
    \end{subfigure}
    \begin{subfigure}[t]{0.195\textwidth}
        \centering
        \includegraphics[width=\linewidth]{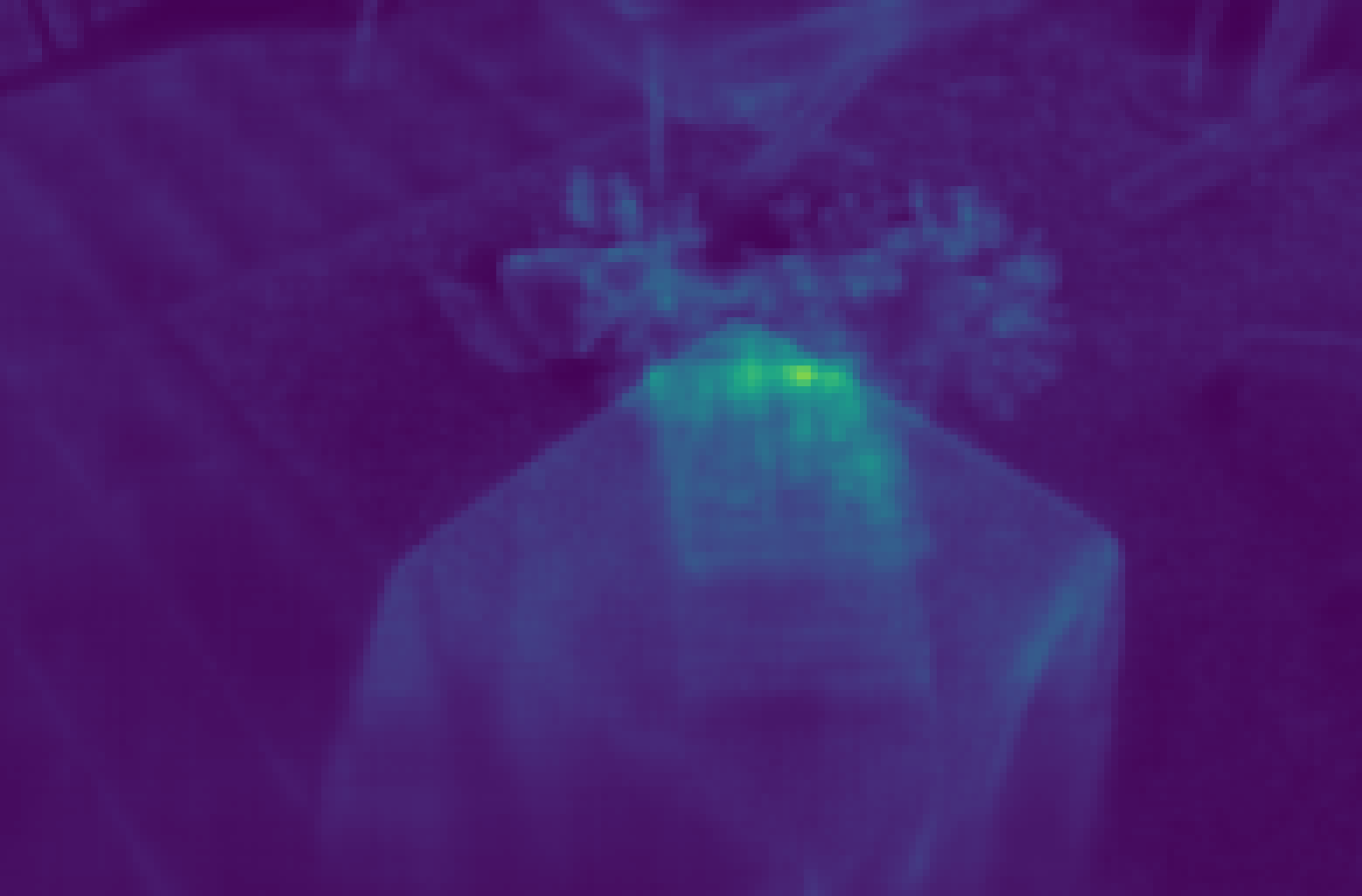}
        \caption*{\scriptsize Ours}
    \end{subfigure}
    \begin{subfigure}[t]{0.195\textwidth}
        \centering
        \includegraphics[width=\linewidth]{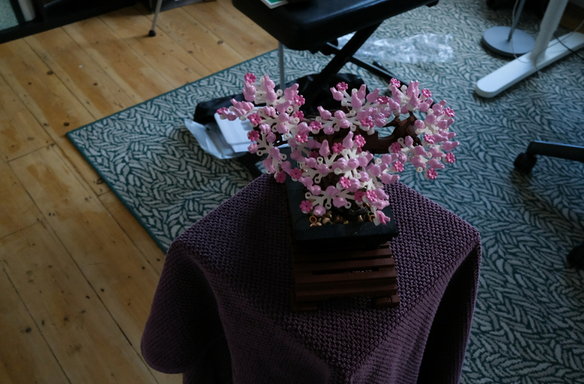}
        \caption*{\scriptsize Ground Truth}
    \end{subfigure}\\
    
    \caption{Gaussian count heatmap comparison across scenes in Mip-NeRF~360 dataset, corresponding to \cref{fig:gaussian_count_viz}. Each row shows ground truth and heatmaps for 3DGS, Taming-3DGS, LiteGS, and our method. Colors represent per-tile Gaussian counts, where purple indicates low counts and yellow indicates high counts. Our method consistently achieves the lowest Gaussian counts (darker colors). }
    \label{fig:supplementary_gaussian_count_viz}
\end{figure*}

\begin{figure*}[t]
    \centering
    
    \begin{subfigure}[t]{0.195\textwidth}
        \centering
        \includegraphics[width=\linewidth]{img12/8824_3dgs_8x8_iter_030000_psnr_29.00_time_1322s__DSC8824_heatmap.png}
    \end{subfigure}
    \begin{subfigure}[t]{0.195\textwidth}
        \centering
        \includegraphics[width=\linewidth]{img12/8824_reset_epoch_0149__DSC8824_gaussian_heatmap.png}
    \end{subfigure}
    \begin{subfigure}[t]{0.195\textwidth}
        \centering
        \includegraphics[width=\linewidth]{img12/8824_entropy_epoch_0149__DSC8824_gaussian_heatmap.png}
    \end{subfigure}
    \begin{subfigure}[t]{0.195\textwidth}
        \centering
        \includegraphics[width=\linewidth]{img12/8824_ours_epoch_0149__DSC8824_gaussian_heatmap.png}
    \end{subfigure}
    \begin{subfigure}[t]{0.195\textwidth}
        \centering
        \includegraphics[width=\linewidth]{img12/8824_gt__DSC8824_resized.png}
    \end{subfigure}\\

    \begin{subfigure}[t]{0.195\textwidth}
        \centering
        \includegraphics[width=\linewidth]{img12/9136_3dgs_8x8_iter_030000_psnr_22.45_time_845s__DSC9136_heatmap.png}
    \end{subfigure}
    \begin{subfigure}[t]{0.195\textwidth}
        \centering
        \includegraphics[width=\linewidth]{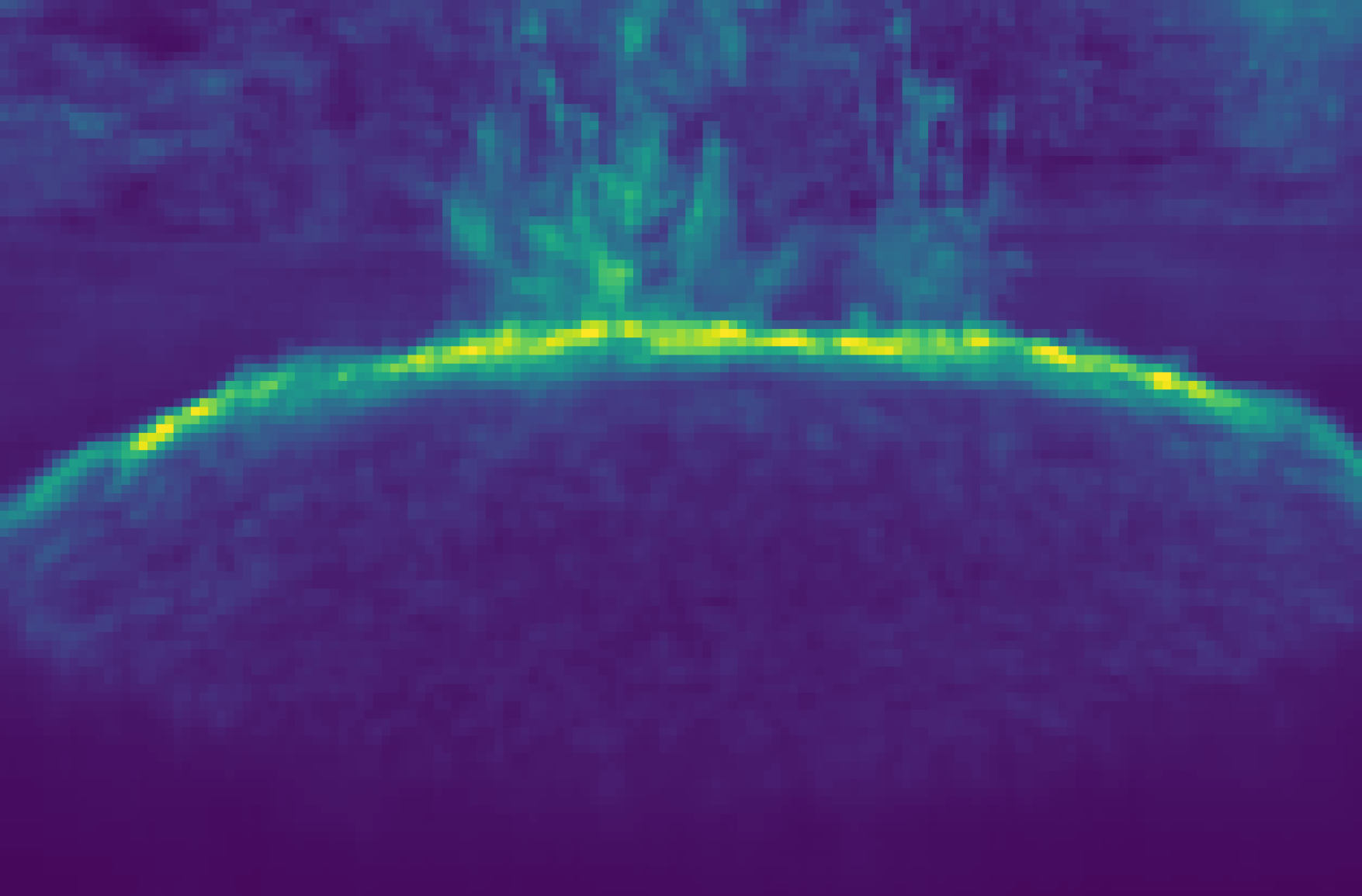}
    \end{subfigure}
    \begin{subfigure}[t]{0.195\textwidth}
        \centering
        \includegraphics[width=\linewidth]{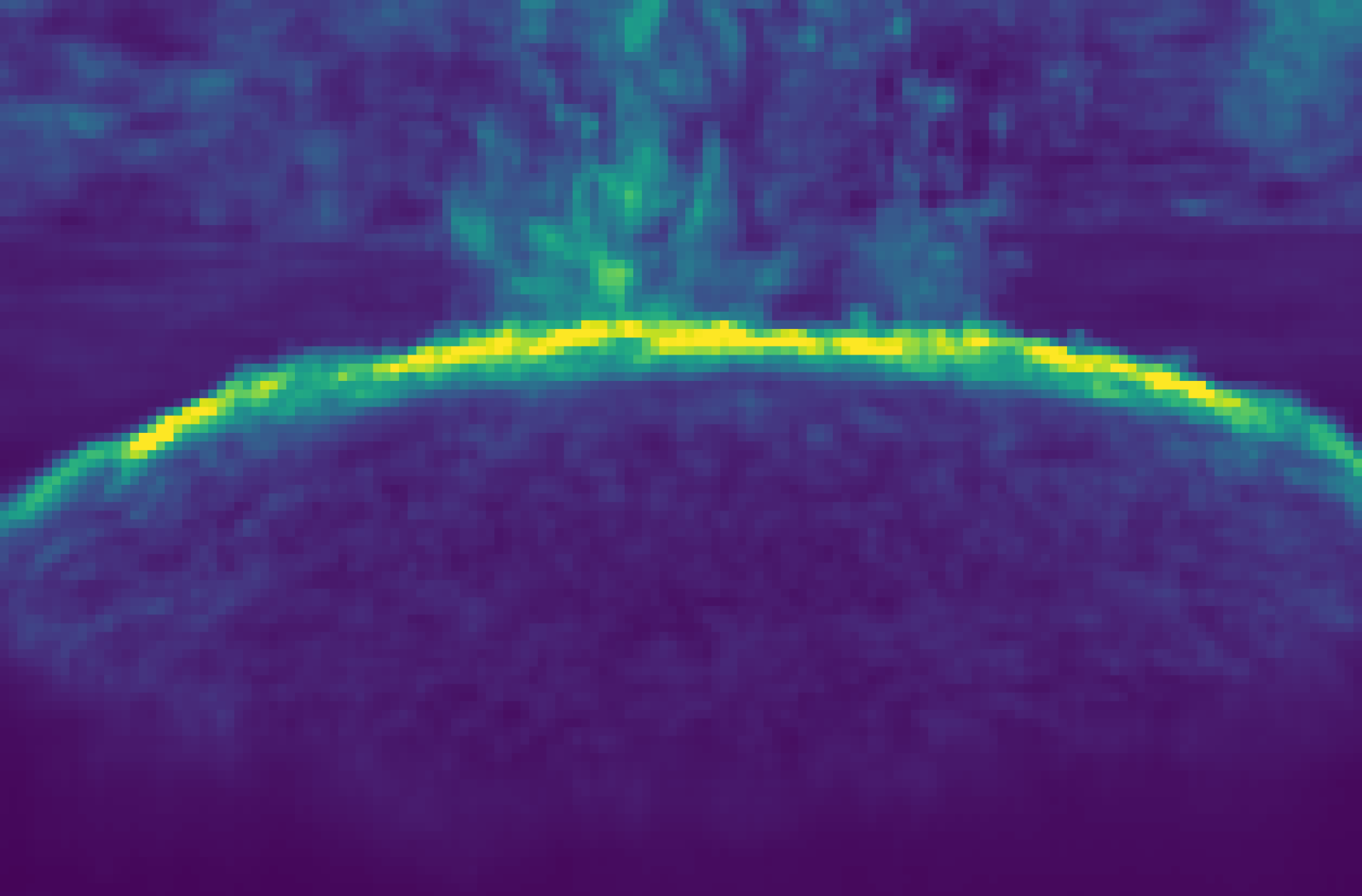}
    \end{subfigure}
    \begin{subfigure}[t]{0.195\textwidth}
        \centering
        \includegraphics[width=\linewidth]{img12/9136_ours_epoch_0149__DSC9136_gaussian_heatmap.png}
    \end{subfigure}
    \begin{subfigure}[t]{0.195\textwidth}
        \centering
        \includegraphics[width=\linewidth]{img12/9136_gt__DSC9136_resized.png}
    \end{subfigure}\\

    \begin{subfigure}[t]{0.195\textwidth}
        \centering
        \includegraphics[width=\linewidth]{img12/8052_3dgs_8x8_iter_030000_psnr_27.46_time_1209s_DSC08052_heatmap.png}
    \end{subfigure}
    \begin{subfigure}[t]{0.195\textwidth}
        \centering
        \includegraphics[width=\linewidth]{img12/8052_reset_epoch_0149_DSC08052_gaussian_heatmap.png}
    \end{subfigure}
    \begin{subfigure}[t]{0.195\textwidth}
        \centering
        \includegraphics[width=\linewidth]{img12/8052_entropy_epoch_0149_DSC08052_gaussian_heatmap.png}
    \end{subfigure}
    \begin{subfigure}[t]{0.195\textwidth}
        \centering
        \includegraphics[width=\linewidth]{img12/8052_ours_epoch_0149_DSC08052_gaussian_heatmap.png}
    \end{subfigure}
    \begin{subfigure}[t]{0.195\textwidth}
        \centering
        \includegraphics[width=\linewidth]{img12/8052_gt_DSC08052_resized.png}
    \end{subfigure}\\
    
    \begin{subfigure}[t]{0.195\textwidth}
        \centering
        \includegraphics[width=\linewidth]{img12/9221_3dgs_8x8_iter_030000_psnr_22.36_time_1033s__DSC9221_heatmap.png}
    \end{subfigure}
    \begin{subfigure}[t]{0.195\textwidth}
        \centering
        \includegraphics[width=\linewidth]{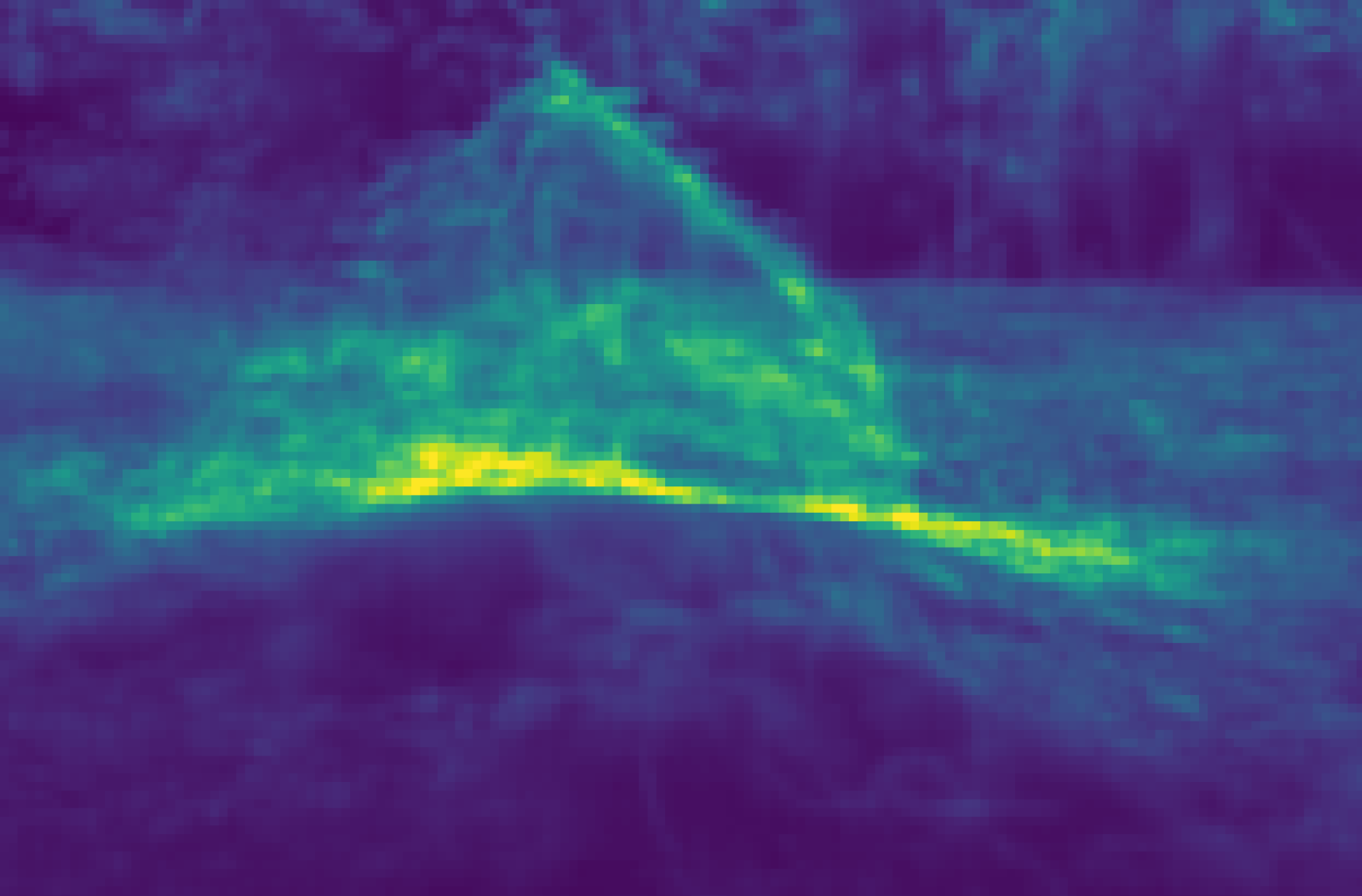}
    \end{subfigure}
    \begin{subfigure}[t]{0.195\textwidth}
        \centering
        \includegraphics[width=\linewidth]{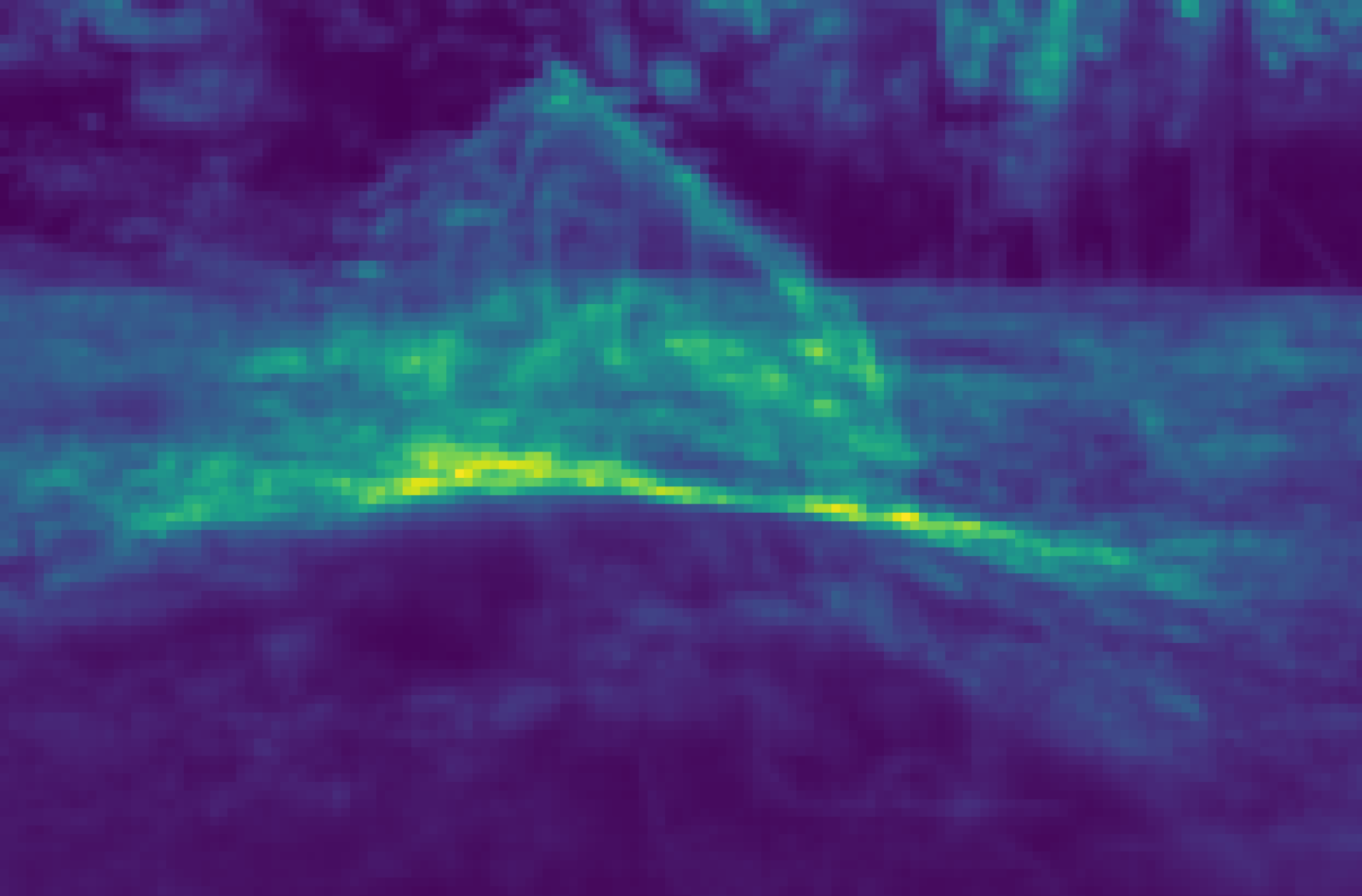}
    \end{subfigure}
    \begin{subfigure}[t]{0.195\textwidth}
        \centering
        \includegraphics[width=\linewidth]{img12/9221_ours_epoch_0149__DSC9221_gaussian_heatmap.png}
    \end{subfigure}
    \begin{subfigure}[t]{0.195\textwidth}
        \centering
        \includegraphics[width=\linewidth]{img12/9221_gt__DSC9221_resized.png}
    \end{subfigure}\\

    \begin{subfigure}[t]{0.195\textwidth}
        \centering
        \includegraphics[width=\linewidth]{img12/8906_3dgs_8x8_iter_030000_psnr_20.15_time_951s__DSC8906_heatmap.png}
    \end{subfigure}
    \begin{subfigure}[t]{0.195\textwidth}
        \centering
        \includegraphics[width=\linewidth]{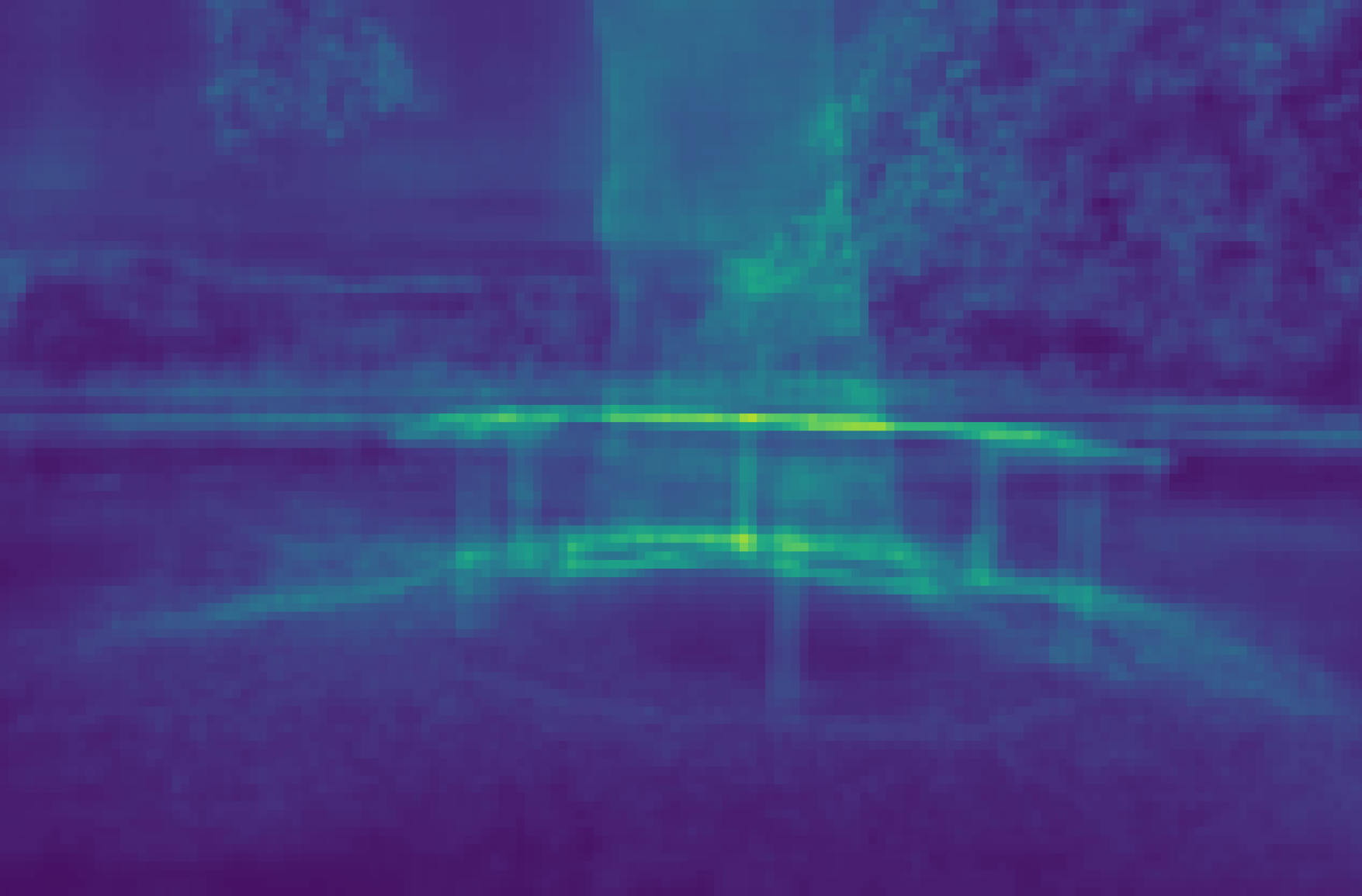}
    \end{subfigure}
    \begin{subfigure}[t]{0.195\textwidth}
        \centering
        \includegraphics[width=\linewidth]{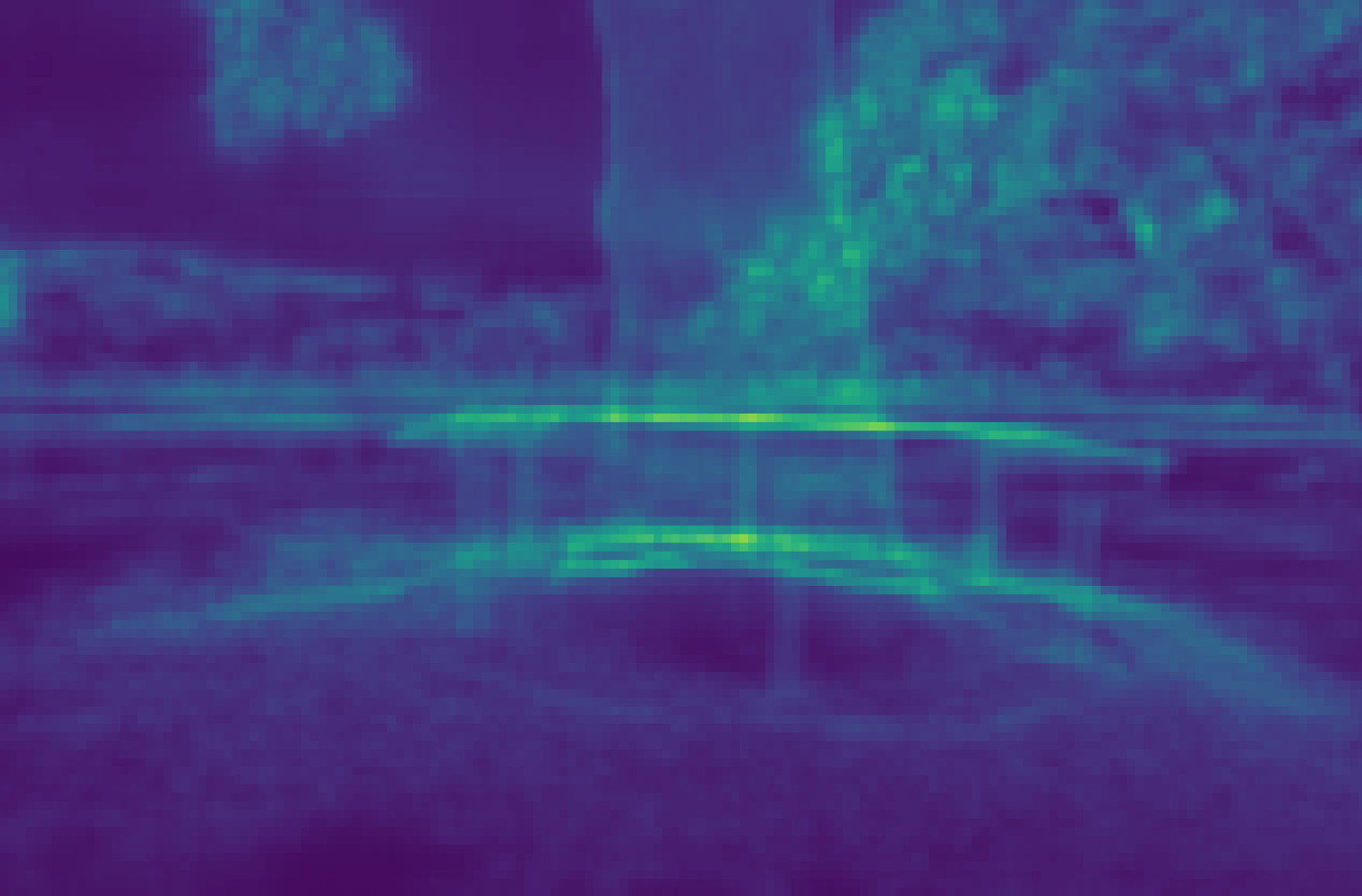}
    \end{subfigure}
    \begin{subfigure}[t]{0.195\textwidth}
        \centering
        \includegraphics[width=\linewidth]{img12/8906_ours_epoch_0149__DSC8906_gaussian_heatmap.png}
    \end{subfigure}
    \begin{subfigure}[t]{0.195\textwidth}
        \centering
        \includegraphics[width=\linewidth]{img12/8906_gt__DSC8906_resized.png}
    \end{subfigure}\\
    
    \begin{subfigure}[t]{0.195\textwidth}
        \centering
        \includegraphics[width=\linewidth]{img12/4707_3dgs_8x8_iter_030000_psnr_30.51_time_974s_DSCF4707_heatmap.png}
    \end{subfigure}
    \begin{subfigure}[t]{0.195\textwidth}
        \centering
        \includegraphics[width=\linewidth]{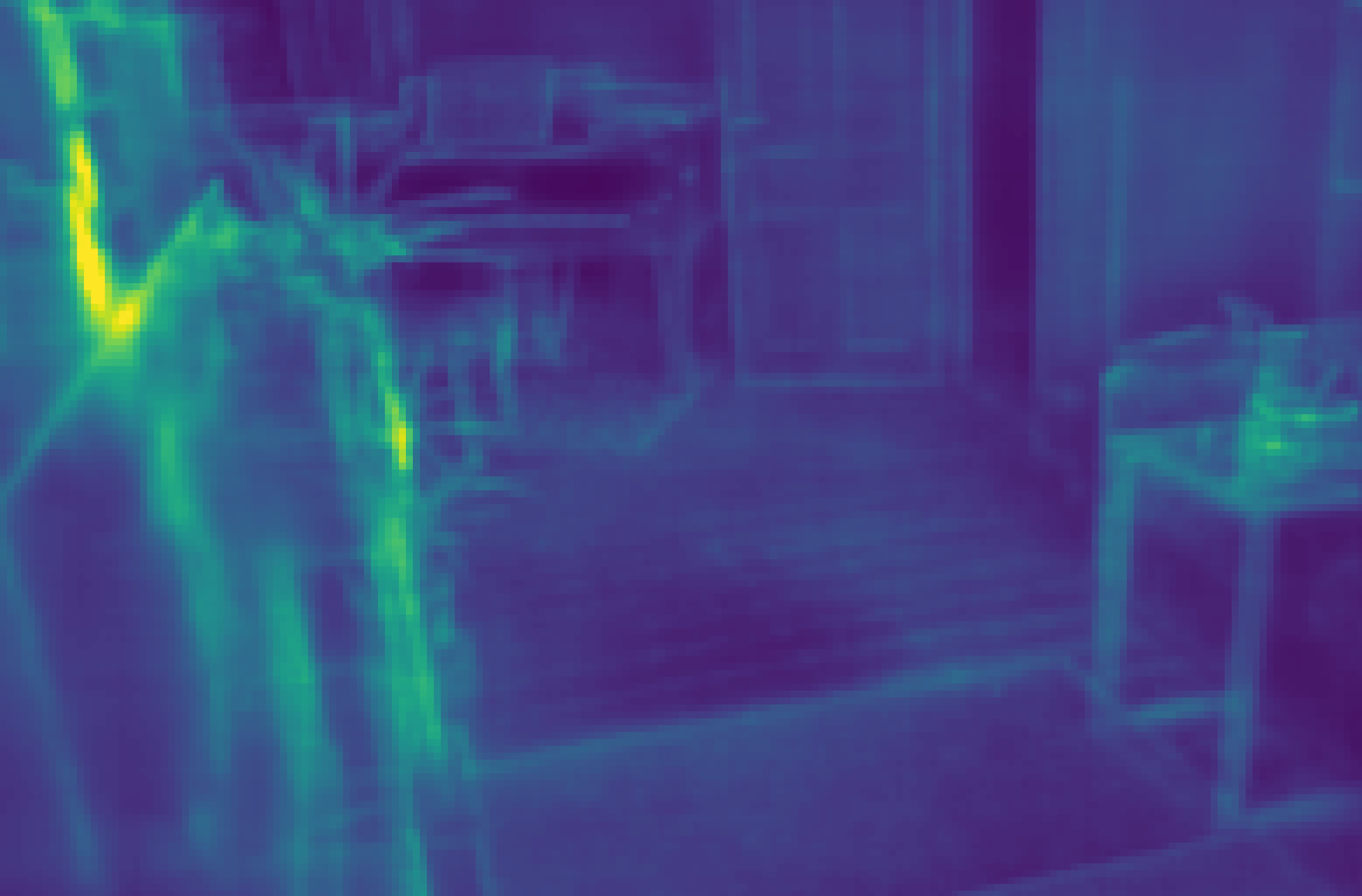}
    \end{subfigure}
    \begin{subfigure}[t]{0.195\textwidth}
        \centering
        \includegraphics[width=\linewidth]{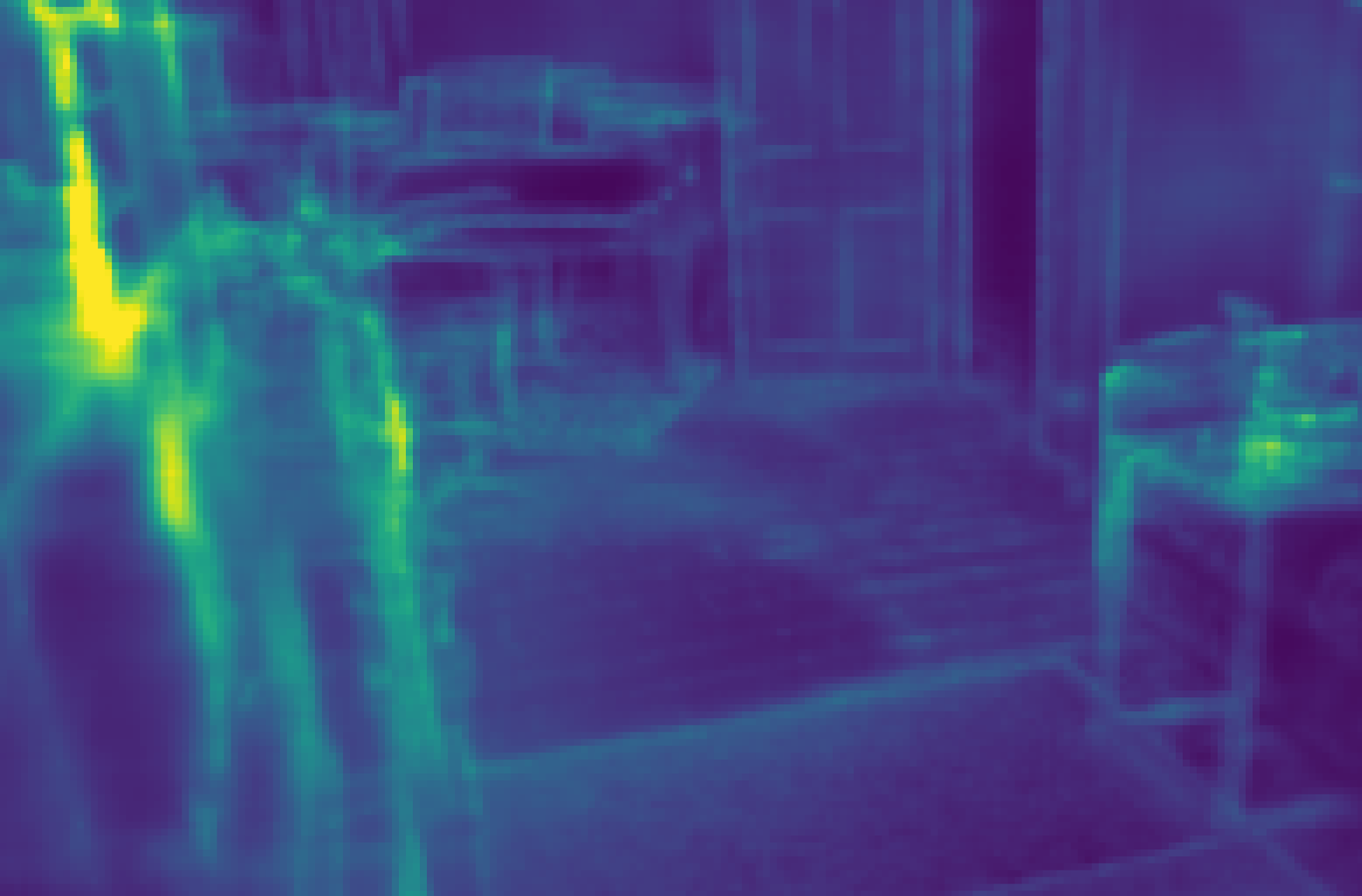}
    \end{subfigure}
    \begin{subfigure}[t]{0.195\textwidth}
        \centering
        \includegraphics[width=\linewidth]{img12/4707_ours_epoch_0149_DSCF4707_gaussian_heatmap.png}
    \end{subfigure}
    \begin{subfigure}[t]{0.195\textwidth}
        \centering
        \includegraphics[width=\linewidth]{img12/4707_gt_DSCF4707_resized.png}
    \end{subfigure}\\

    \begin{subfigure}[t]{0.195\textwidth}
        \centering
        \includegraphics[width=\linewidth]{img12/5993_3dgs_8x8_iter_030000_psnr_27.41_time_902s_DSCF5993_heatmap.png}
    \end{subfigure}
    \begin{subfigure}[t]{0.195\textwidth}
        \centering
        \includegraphics[width=\linewidth]{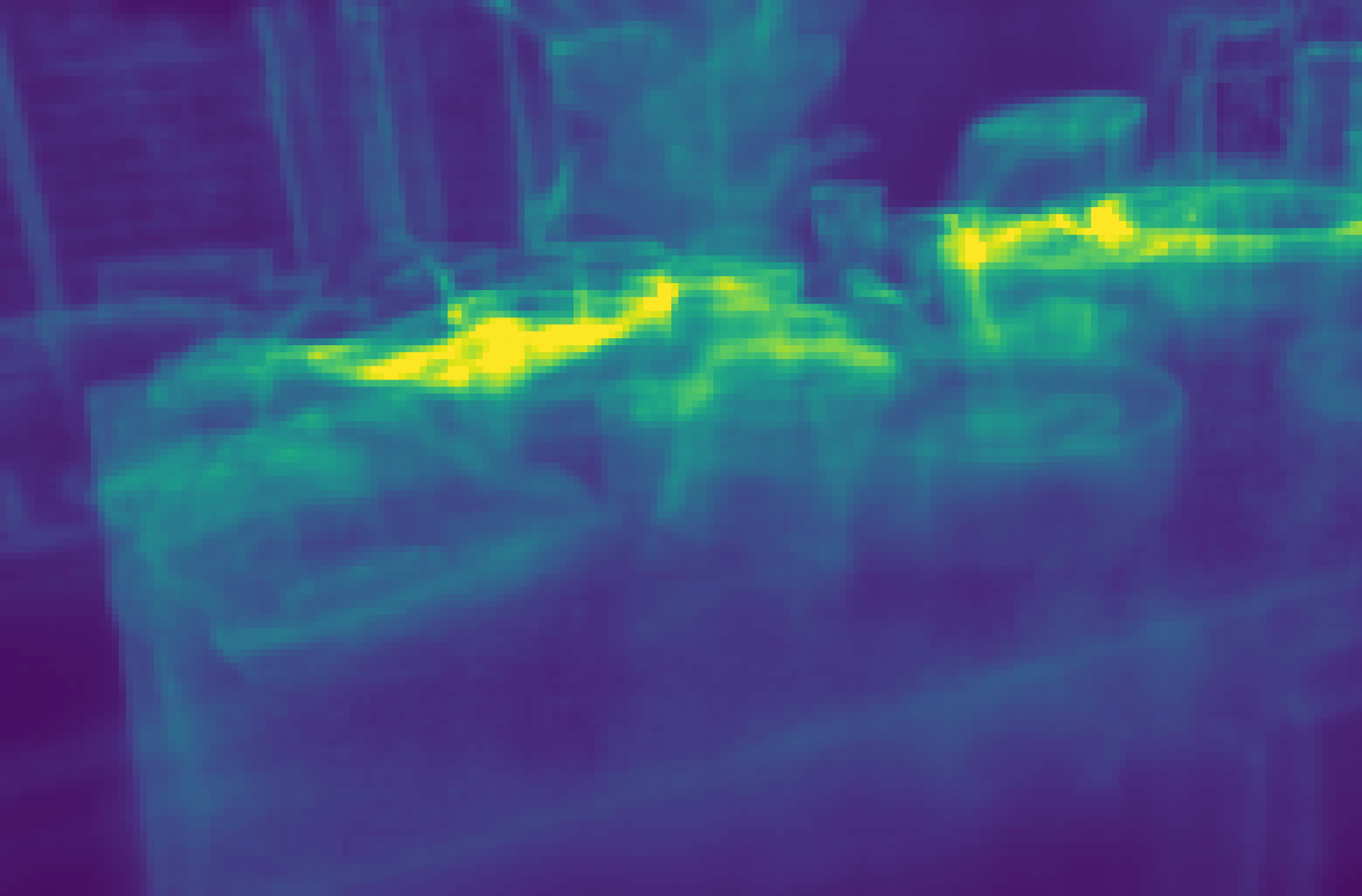}
    \end{subfigure}
    \begin{subfigure}[t]{0.195\textwidth}
        \centering
        \includegraphics[width=\linewidth]{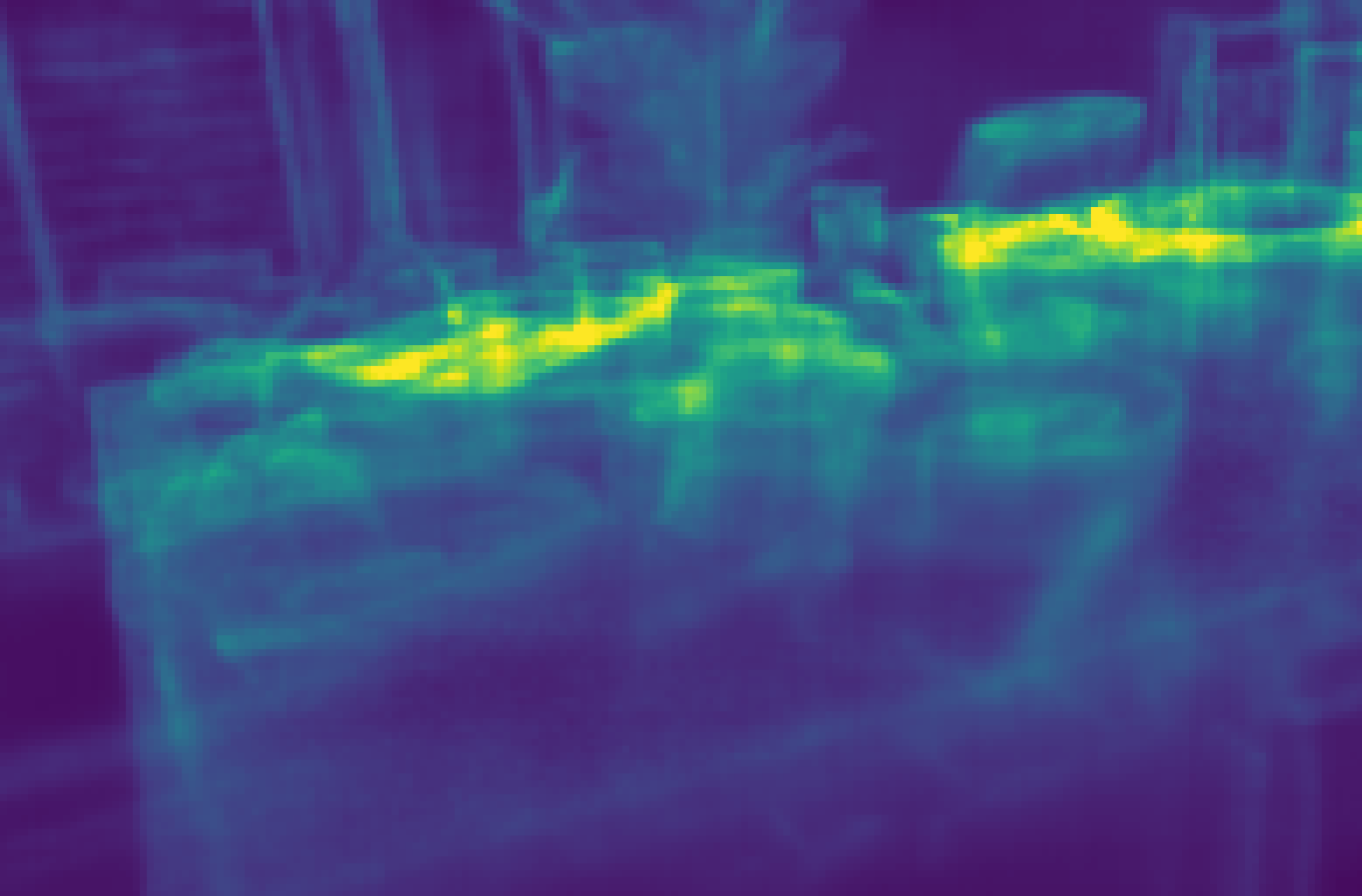}
    \end{subfigure}
    \begin{subfigure}[t]{0.195\textwidth}
        \centering
        \includegraphics[width=\linewidth]{img12/5993_ours_epoch_0149_DSCF5993_gaussian_heatmap.png}
    \end{subfigure}
    \begin{subfigure}[t]{0.195\textwidth}
        \centering
        \includegraphics[width=\linewidth]{img12/5993_gt_DSCF5993_resized.png}
    \end{subfigure}\\

    \begin{subfigure}[t]{0.195\textwidth}
        \centering
        \includegraphics[width=\linewidth]{img12/0840_3dgs_8x8_iter_030000_psnr_33.67_time_1079s_DSCF0840_heatmap.png}
    \end{subfigure}
    \begin{subfigure}[t]{0.195\textwidth}
        \centering
        \includegraphics[width=\linewidth]{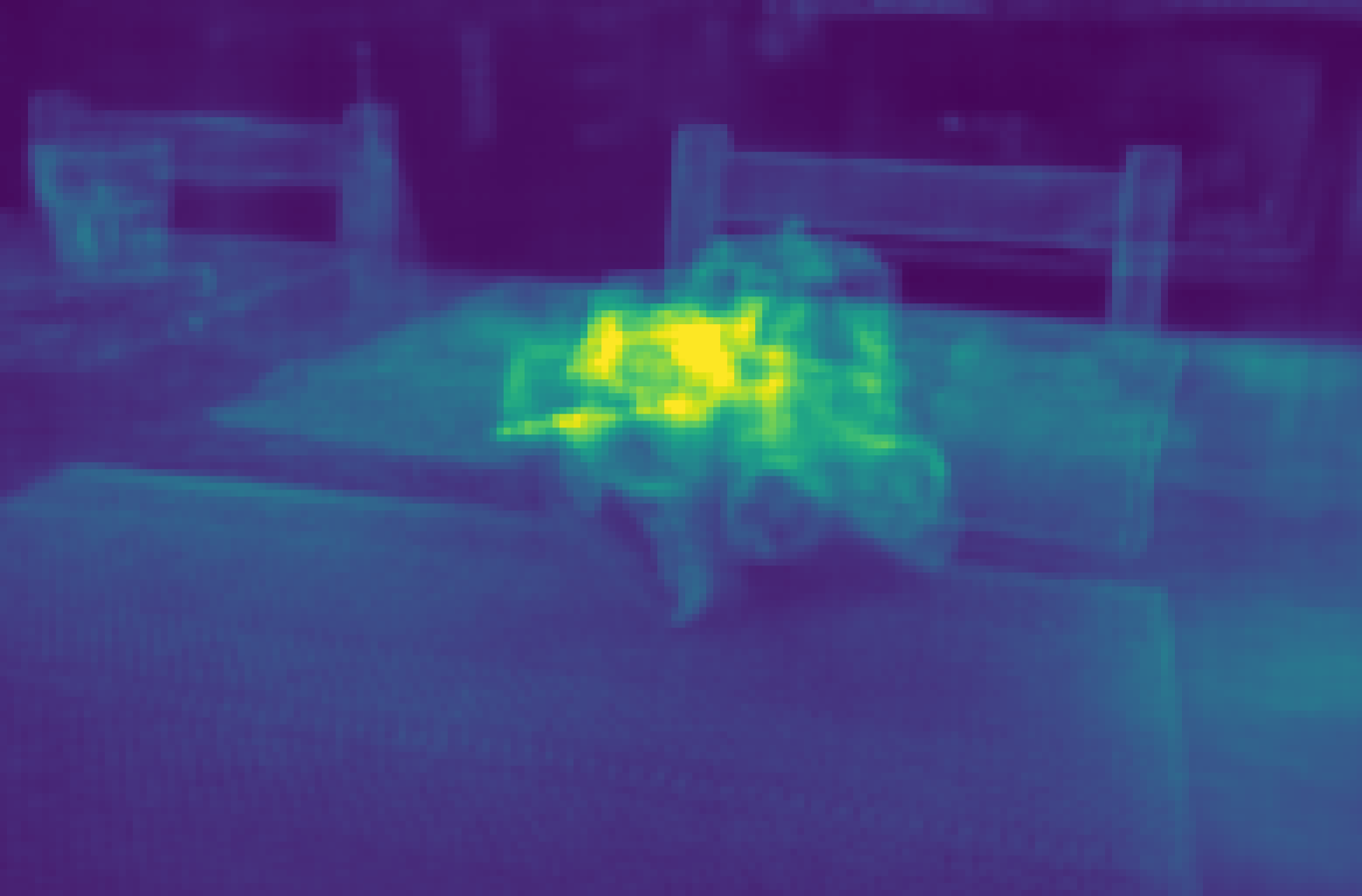}
    \end{subfigure}
    \begin{subfigure}[t]{0.195\textwidth}
        \centering
        \includegraphics[width=\linewidth]{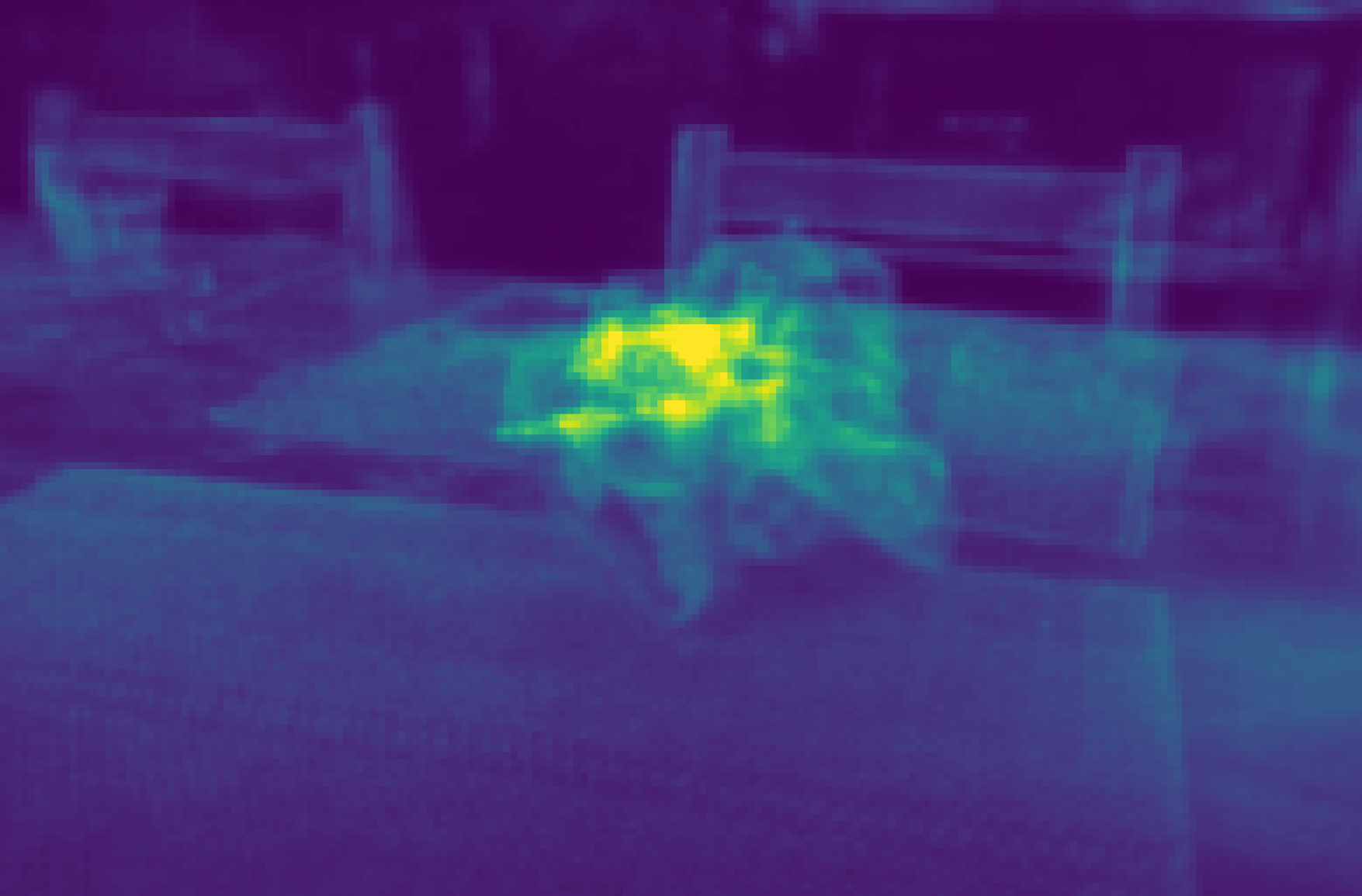}
    \end{subfigure}
    \begin{subfigure}[t]{0.195\textwidth}
        \centering
        \includegraphics[width=\linewidth]{img12/0840_ours_epoch_0149_DSCF0840_gaussian_heatmap.png}
    \end{subfigure}
    \begin{subfigure}[t]{0.195\textwidth}
        \centering
        \includegraphics[width=\linewidth]{img12/0840_gt_DSCF0840_resized.png}
    \end{subfigure}\\

    \begin{subfigure}[t]{0.195\textwidth}
        \centering
        \includegraphics[width=\linewidth]{img12/5701_3dgs_8x8_iter_030000_psnr_34.20_time_747s_DSCF5701_heatmap.png}
        \caption*{\scriptsize 3DGS}
    \end{subfigure}
    \begin{subfigure}[t]{0.195\textwidth}
        \centering
        \includegraphics[width=\linewidth]{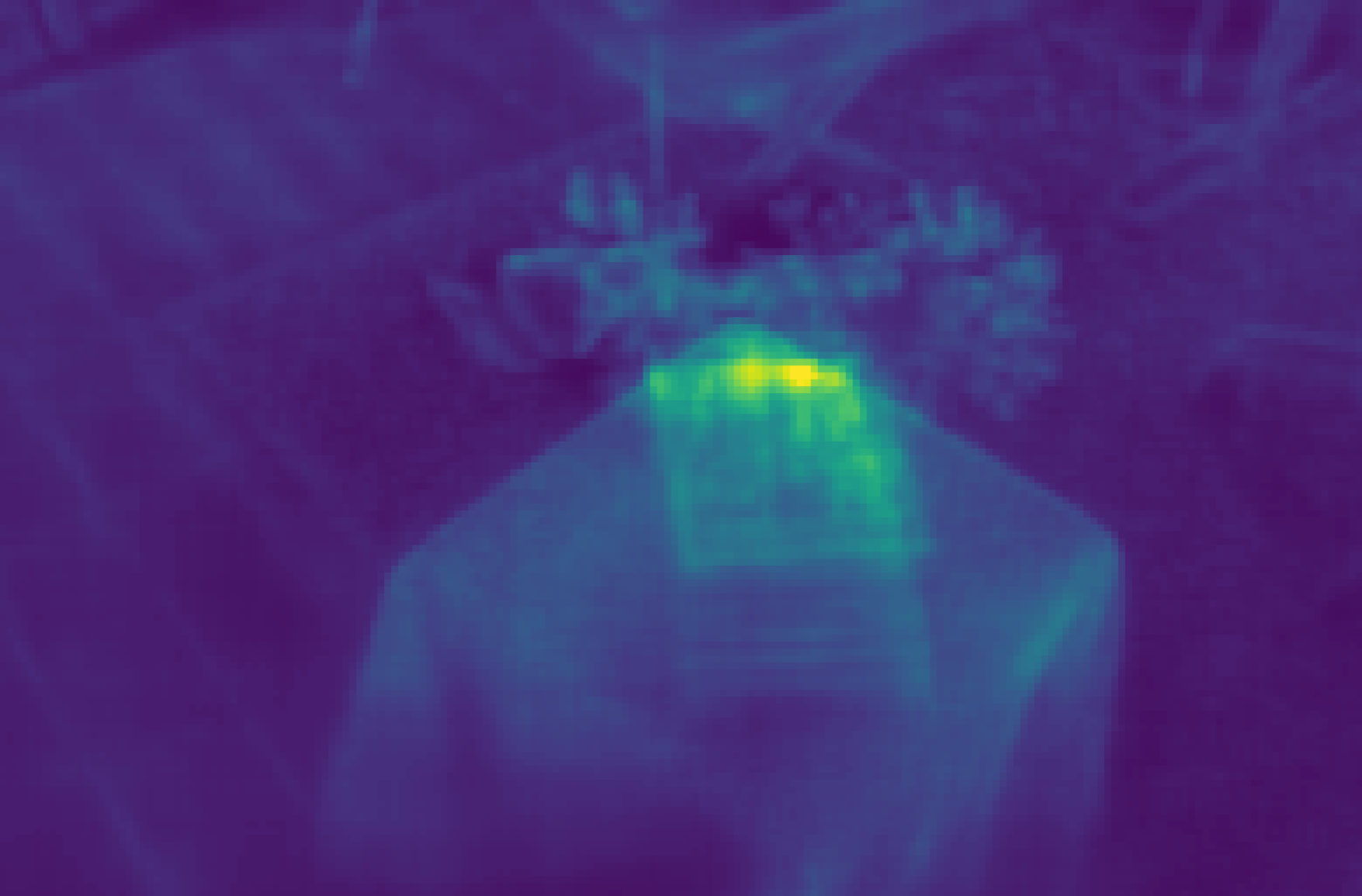}
        \caption*{\scriptsize 3DGS+Scale Reset}
    \end{subfigure}
    \begin{subfigure}[t]{0.195\textwidth}
        \centering
        \includegraphics[width=\linewidth]{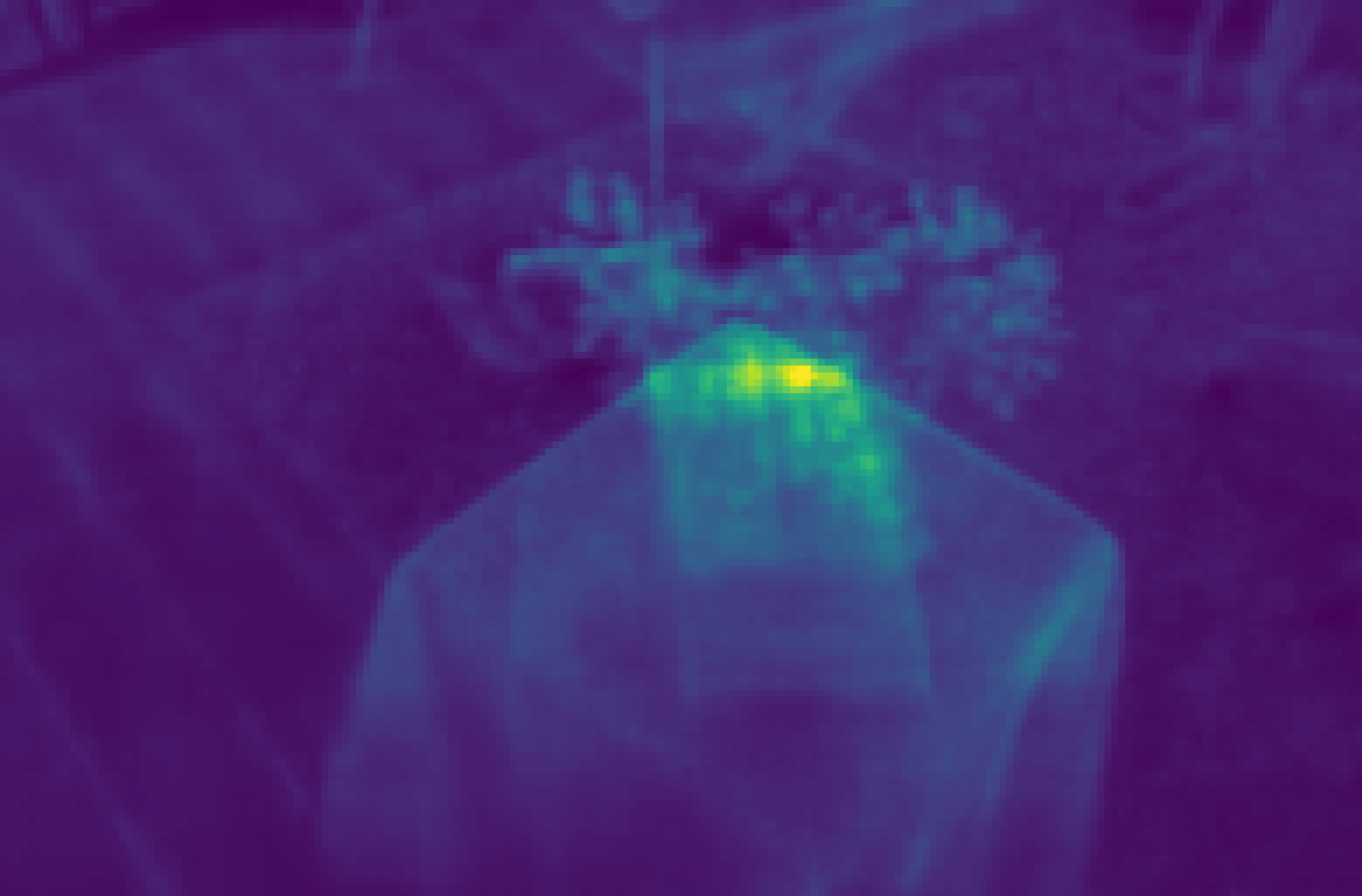}
        \caption*{\scriptsize 3DGS+Entropy Constraint}
    \end{subfigure}
    \begin{subfigure}[t]{0.195\textwidth}
        \centering
        \includegraphics[width=\linewidth]{img12/5701_ours_epoch_0149_DSCF5701_gaussian_heatmap.png}
        \caption*{\scriptsize 3DGS+Reset+Entropy}
    \end{subfigure}
    \begin{subfigure}[t]{0.195\textwidth}
        \centering
        \includegraphics[width=\linewidth]{img12/5701_gt_DSCF5701_resized.png}
        \caption*{\scriptsize Ground Truth}
    \end{subfigure}\\
    
    \caption{Gaussian count heatmap comparison across scenes in Mip-NeRF~360 dataset for ablation study, corresponding to \cref{fig:gaussian_count_viz_ablation}. Each row shows ground truth and heatmaps for 3DGS, 3DGS+Scale Reset, 3DGS+Entropy Constraint, and 3DGS+Reset+Entropy. Colors represent per-tile Gaussian counts, where purple indicates low counts and yellow indicates high counts.}
    \label{fig:supplementary_gaussian_count_viz_ablation}
\end{figure*}

\begin{figure*}[t]
    \centering
    
    \rotatebox{90}{\small{bicycle}}\hspace{1pt}
    \begin{subfigure}[t]{0.31\textwidth}
        \centering
        \begin{tikzpicture}
            \node[anchor=south west, inner sep=0] (image) at (0,0) {
                \includegraphics[width=\linewidth]{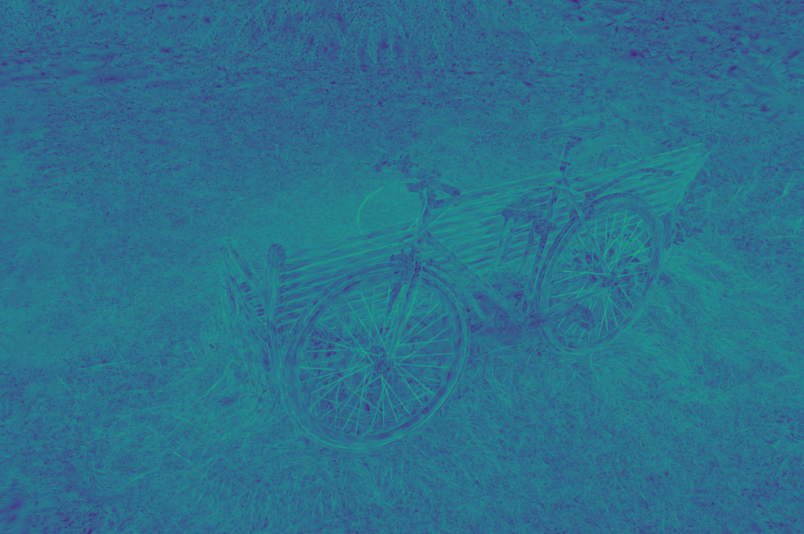}
            };
            \node[anchor=south east, text=white, font=\footnotesize, inner sep=2pt] 
                at (image.south east) {mean entropy: 2.744};
        \end{tikzpicture}
    \end{subfigure}
    \begin{subfigure}[t]{0.31\textwidth}
        \centering
        \begin{tikzpicture}
            \node[anchor=south west, inner sep=0] (image) at (0,0) {
                \includegraphics[width=\linewidth]{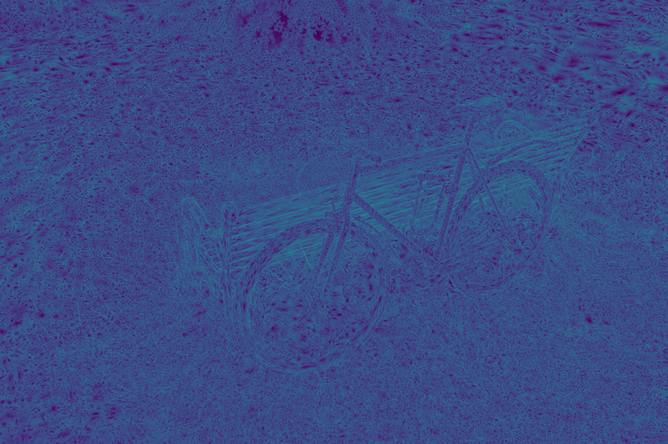}
            };
            \node[anchor=south east, text=white, font=\footnotesize, inner sep=2pt] 
                at (image.south east) {mean entropy: 1.739};
        \end{tikzpicture}
    \end{subfigure}
    \begin{subfigure}[t]{0.31\textwidth}
        \centering
        \includegraphics[width=\linewidth]{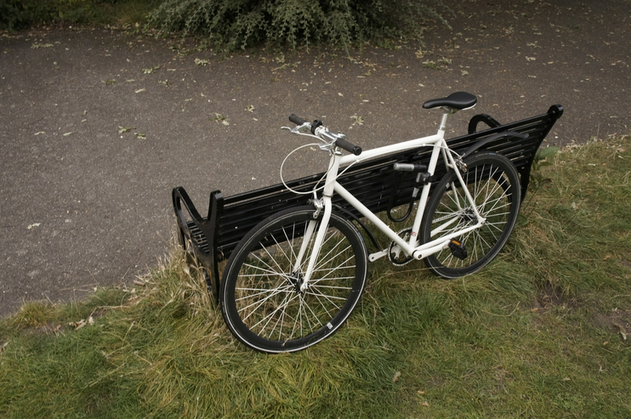}
    \end{subfigure}\\[2pt]
    \rotatebox{90}{\small{garden}}\hspace{1pt}
    \begin{subfigure}[t]{0.31\textwidth}
        \centering
        \begin{tikzpicture}
            \node[anchor=south west, inner sep=0] (image) at (0,0) {
                \includegraphics[width=\linewidth]{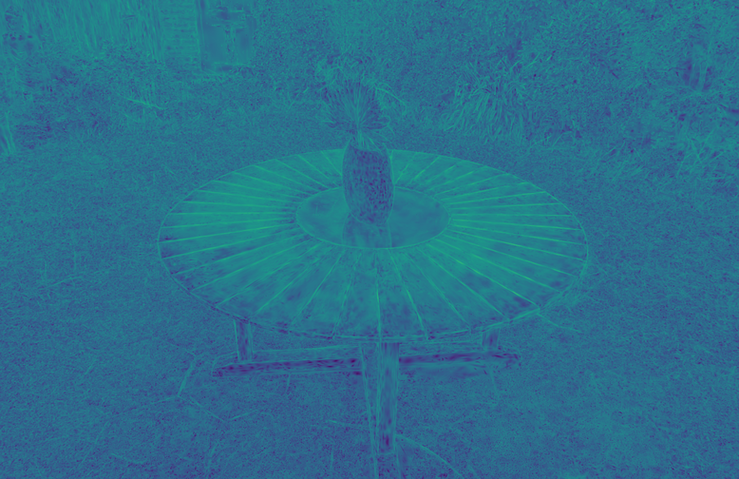}
            };
            \node[anchor=south east, text=white, font=\footnotesize, inner sep=2pt] 
                at (image.south east) {mean entropy: 2.758};
        \end{tikzpicture}
    \end{subfigure}
    \begin{subfigure}[t]{0.31\textwidth}
        \centering
        \begin{tikzpicture}
            \node[anchor=south west, inner sep=0] (image) at (0,0) {
                \includegraphics[width=\linewidth]{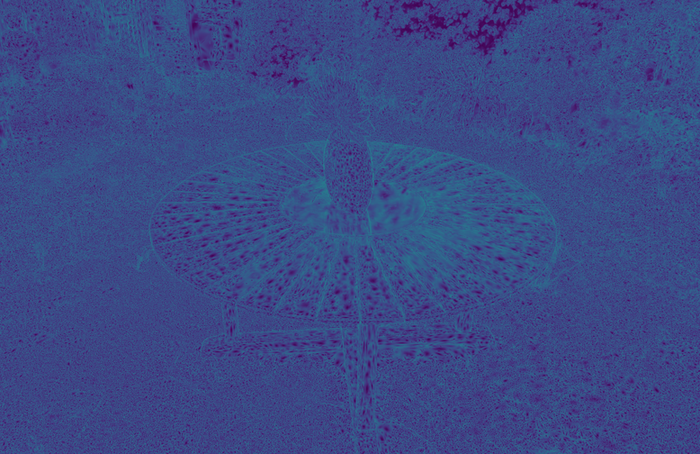}
            };
            \node[anchor=south east, text=white, font=\footnotesize, inner sep=2pt] 
                at (image.south east) {mean entropy: 1.704};
        \end{tikzpicture}
    \end{subfigure}
    \begin{subfigure}[t]{0.31\textwidth}
        \centering
        \includegraphics[width=\linewidth]{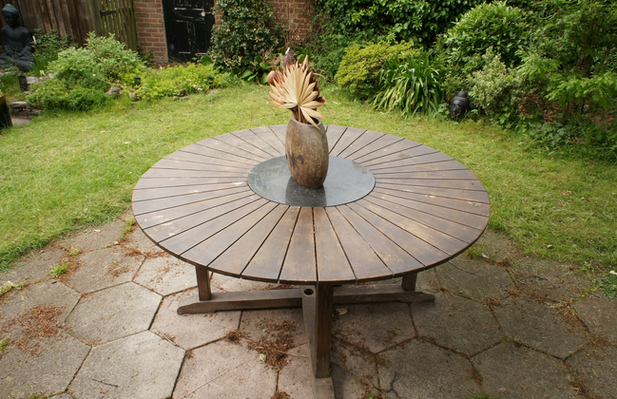}
    \end{subfigure}\\[2pt]
    \rotatebox{90}{\small{treehill}}\hspace{1pt}
    \begin{subfigure}[t]{0.31\textwidth}
        \centering
        \begin{tikzpicture}
            \node[anchor=south west, inner sep=0] (image) at (0,0) {
                \includegraphics[width=\linewidth]{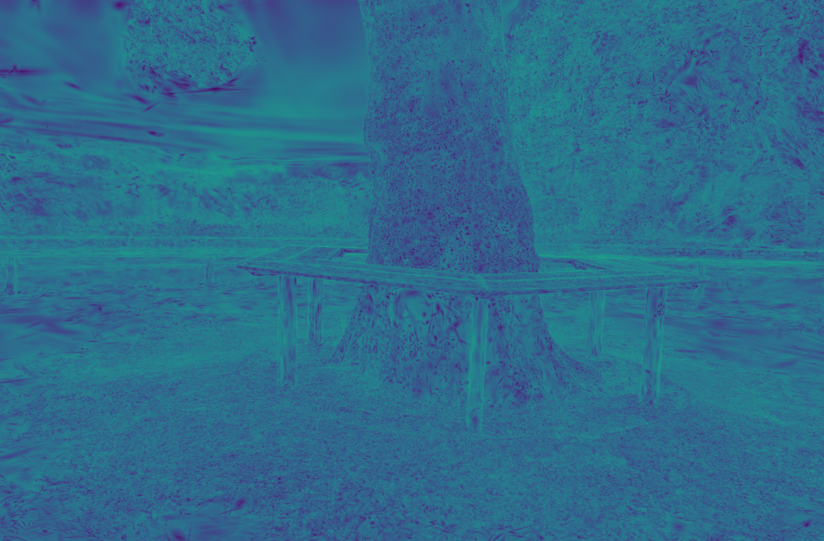}
            };
            \node[anchor=south east, text=white, font=\footnotesize, inner sep=2pt] 
                at (image.south east) {mean entropy: 2.643};
        \end{tikzpicture}
    \end{subfigure}
    \begin{subfigure}[t]{0.31\textwidth}
        \centering
        \begin{tikzpicture}
            \node[anchor=south west, inner sep=0] (image) at (0,0) {
                \includegraphics[width=\linewidth]{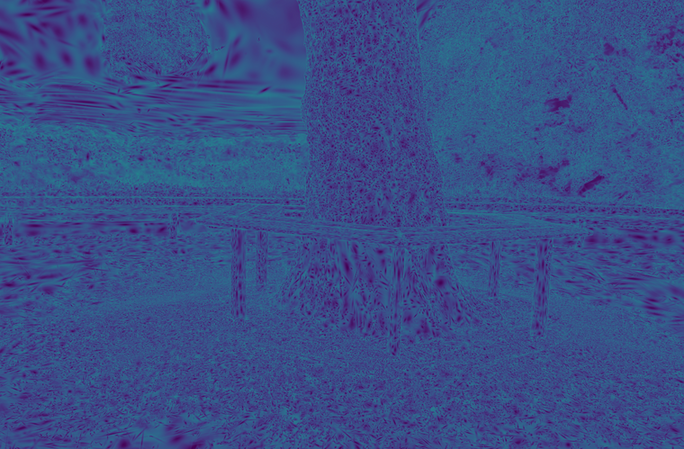}
            };
            \node[anchor=south east, text=white, font=\footnotesize, inner sep=2pt] 
                at (image.south east) {mean entropy: 1.825};
        \end{tikzpicture}
    \end{subfigure}
    \begin{subfigure}[t]{0.31\textwidth}
        \centering
        \includegraphics[width=\linewidth]{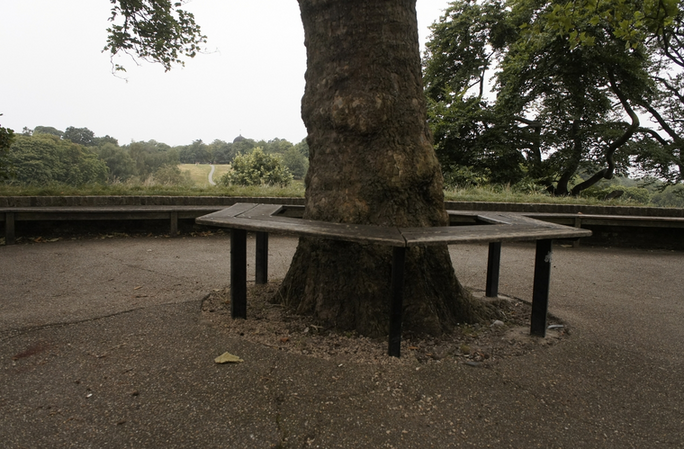}
    \end{subfigure}\\[2pt]
    \rotatebox{90}{\small{counter}}\hspace{1pt}
    \begin{subfigure}[t]{0.31\textwidth}
        \centering
        \begin{tikzpicture}
            \node[anchor=south west, inner sep=0] (image) at (0,0) {
                \includegraphics[width=\linewidth]{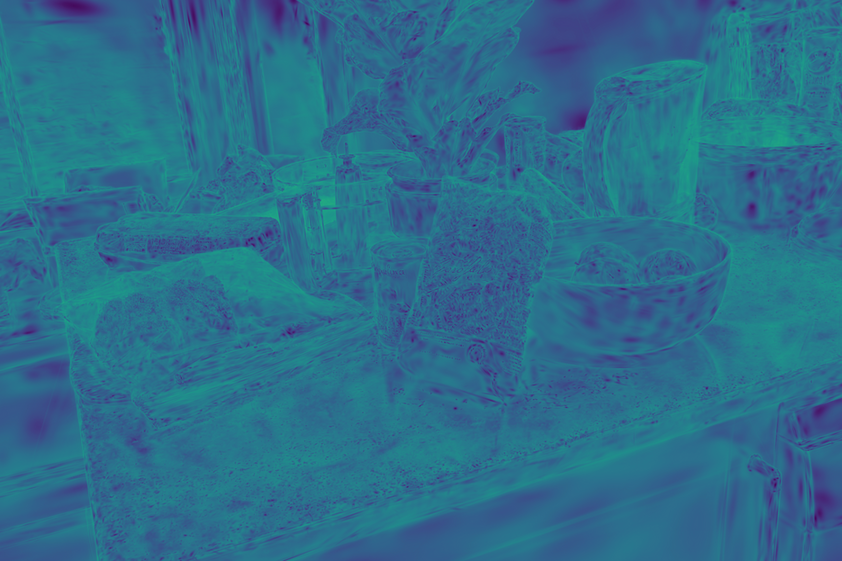}
            };
            \node[anchor=south east, text=white, font=\footnotesize, inner sep=2pt] 
                at (image.south east) {mean entropy: 2.728};
        \end{tikzpicture}
    \end{subfigure}
    \begin{subfigure}[t]{0.31\textwidth}
        \centering
        \begin{tikzpicture}
            \node[anchor=south west, inner sep=0] (image) at (0,0) {
                \includegraphics[width=\linewidth]{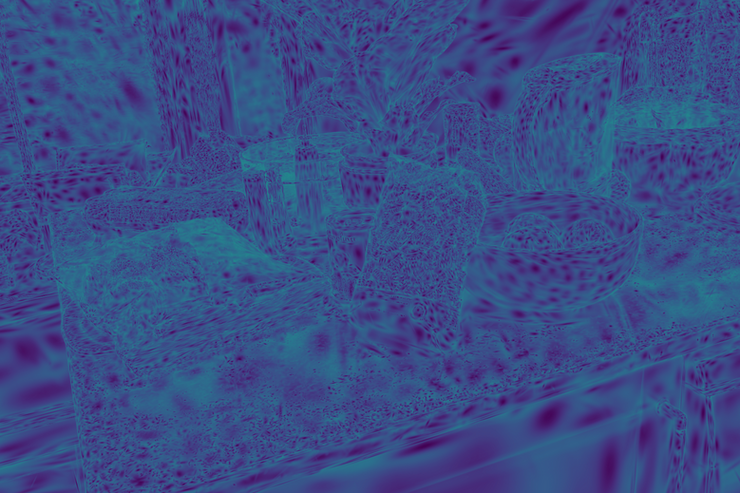}
            };
            \node[anchor=south east, text=white, font=\footnotesize, inner sep=2pt] 
                at (image.south east) {mean entropy: 1.784};
        \end{tikzpicture}
    \end{subfigure}
    \begin{subfigure}[t]{0.31\textwidth}
        \centering
        \includegraphics[width=\linewidth]{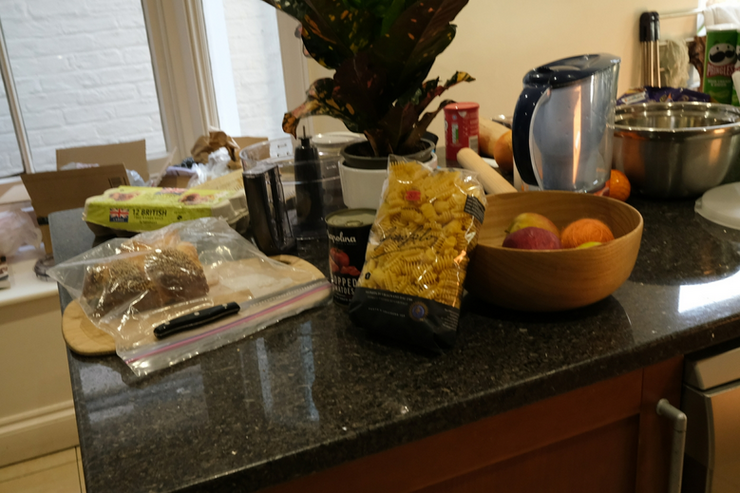}
    \end{subfigure}\\[2pt]
    \rotatebox{90}{\small{bonsai}}\hspace{1pt}
    \begin{subfigure}[t]{0.31\textwidth}
        \centering
        \begin{tikzpicture}
            \node[anchor=south west, inner sep=0] (image) at (0,0) {
                \includegraphics[width=\linewidth]{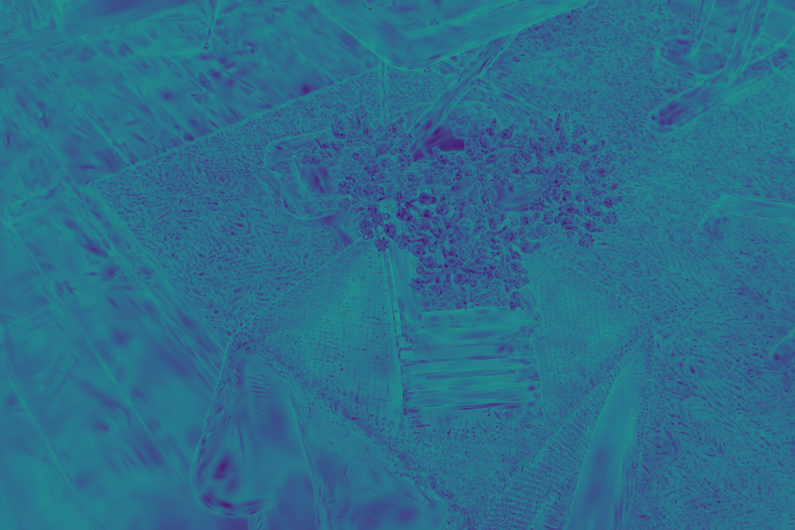}
            };
            \node[anchor=south east, text=white, font=\footnotesize, inner sep=2pt] 
                at (image.south east) {mean entropy: 2.696};
        \end{tikzpicture}
        \caption*{\small LiteGS}
    \end{subfigure}
    \begin{subfigure}[t]{0.31\textwidth}
        \centering
        \begin{tikzpicture}
            \node[anchor=south west, inner sep=0] (image) at (0,0) {
                \includegraphics[width=\linewidth]{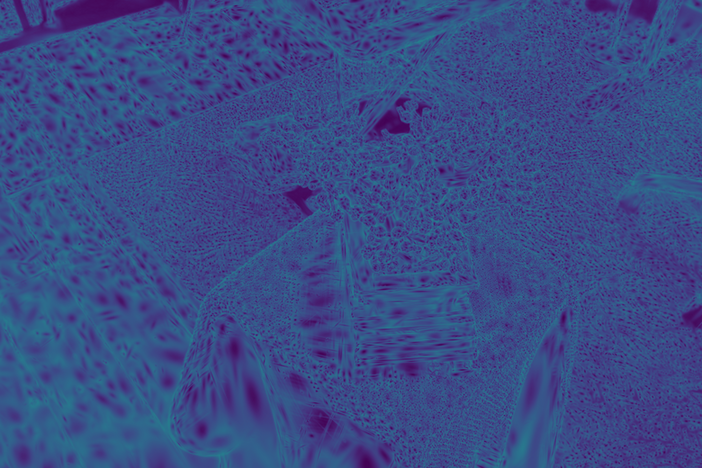}
            };
            \node[anchor=south east, text=white, font=\footnotesize, inner sep=2pt] 
                at (image.south east) {mean entropy: 1.758};
        \end{tikzpicture}
        \caption*{\small Ours}
    \end{subfigure}
    \begin{subfigure}[t]{0.31\textwidth}
        \centering
        \includegraphics[width=\linewidth]{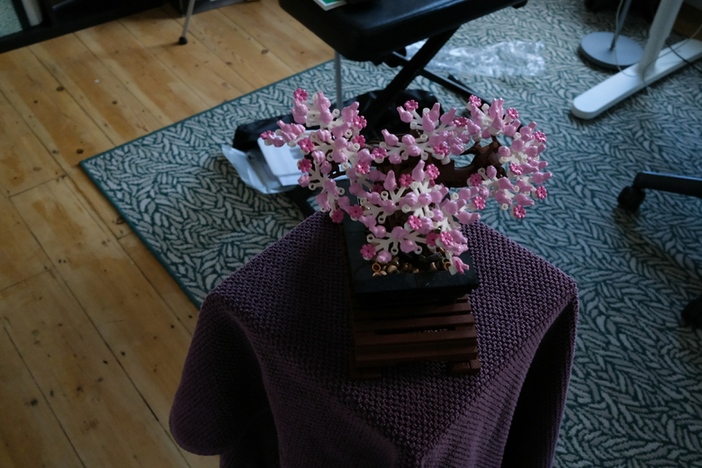}
        \caption*{\small Ground Truth}
    \end{subfigure}\\

    \caption{Per-pixel entropy heatmap comparison between LiteGS and our method across scenes in the Mip-NeRF~360 dataset. Darker colors correspond to lower entropy.}
    \label{fig:entropy_comparison}
\end{figure*}
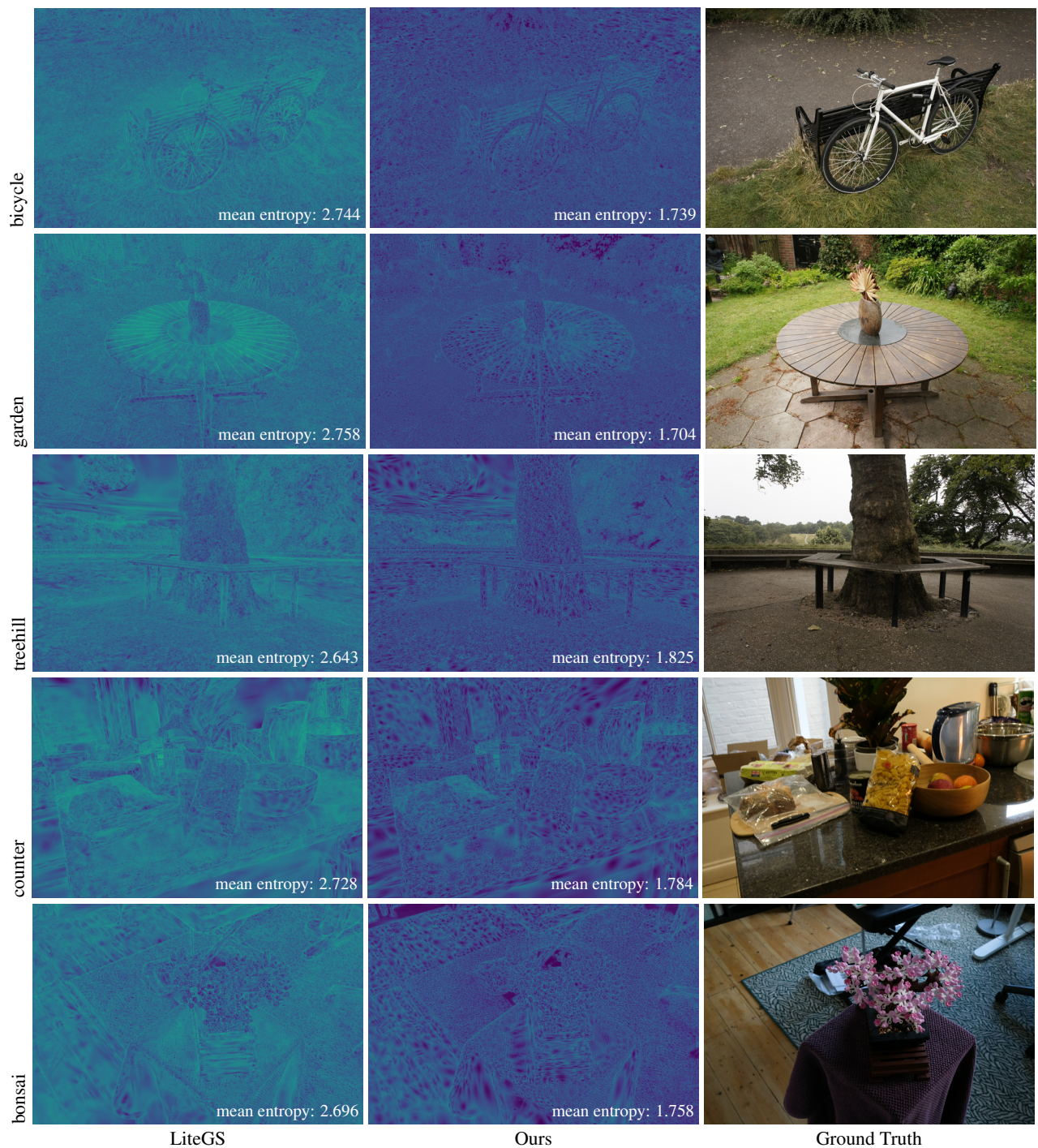

\begin{figure*}[p]
\centering
\renewcommand{\arraystretch}{0.8} 
\tiny 
\resizebox{\textwidth}{!}{
\begin{tabular}{lccc}
\toprule
\textbf{Scene} & \textbf{Length} & \textbf{Scale} & \textbf{Opacity} \\
\midrule
Room &
\includegraphics[width=0.13\textwidth]{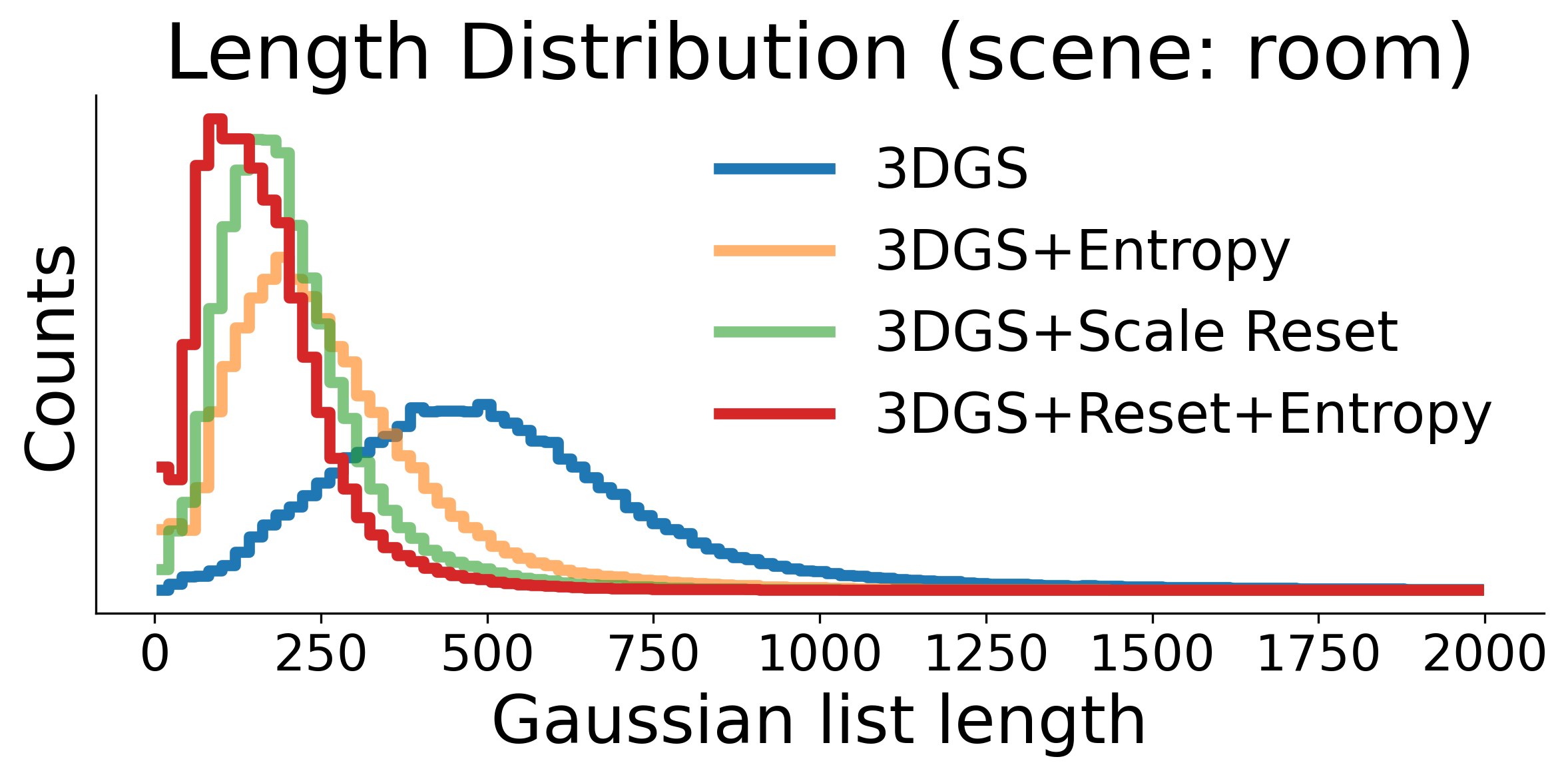} &
\includegraphics[width=0.13\textwidth]{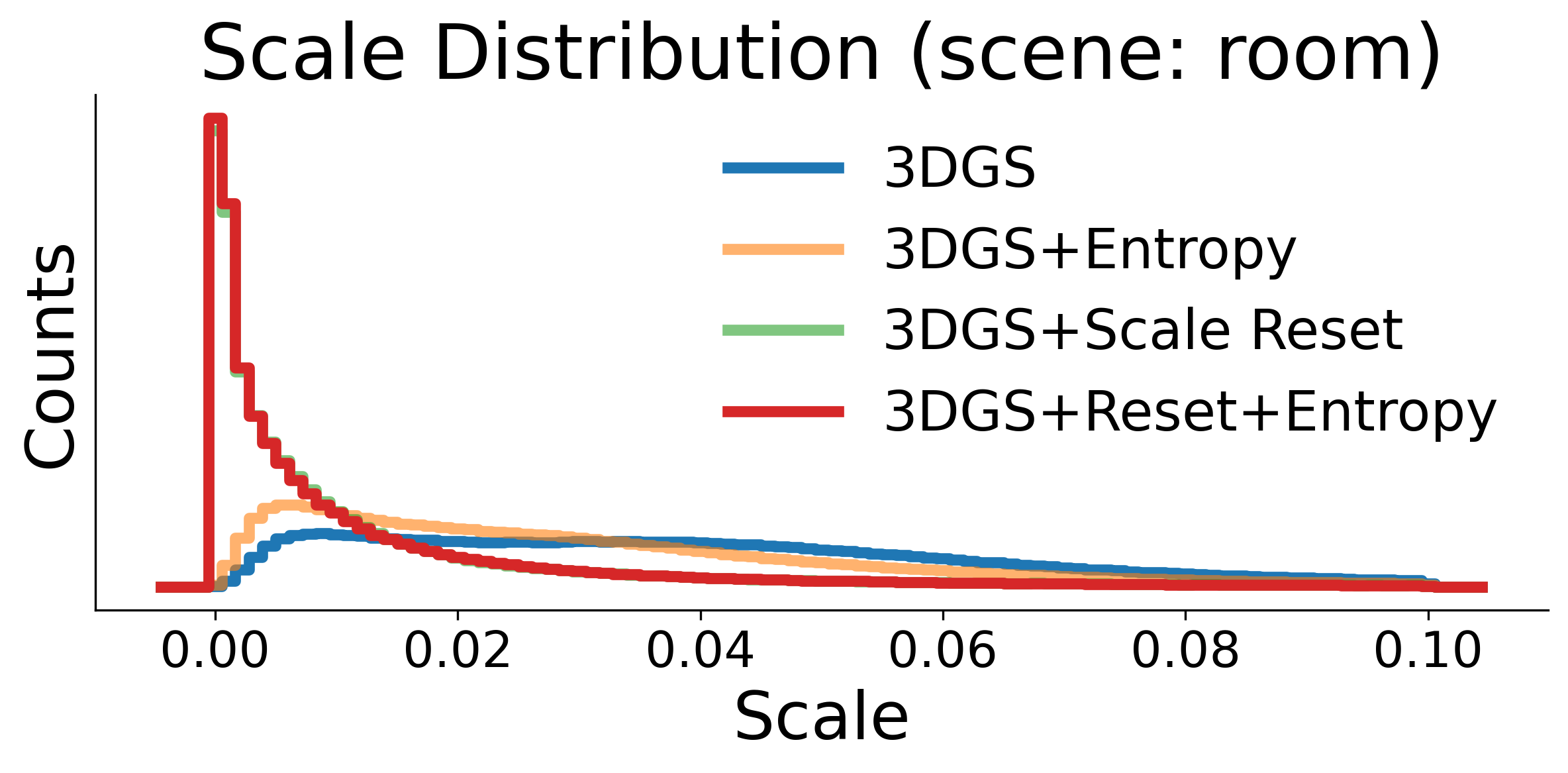} &
\includegraphics[width=0.13\textwidth]{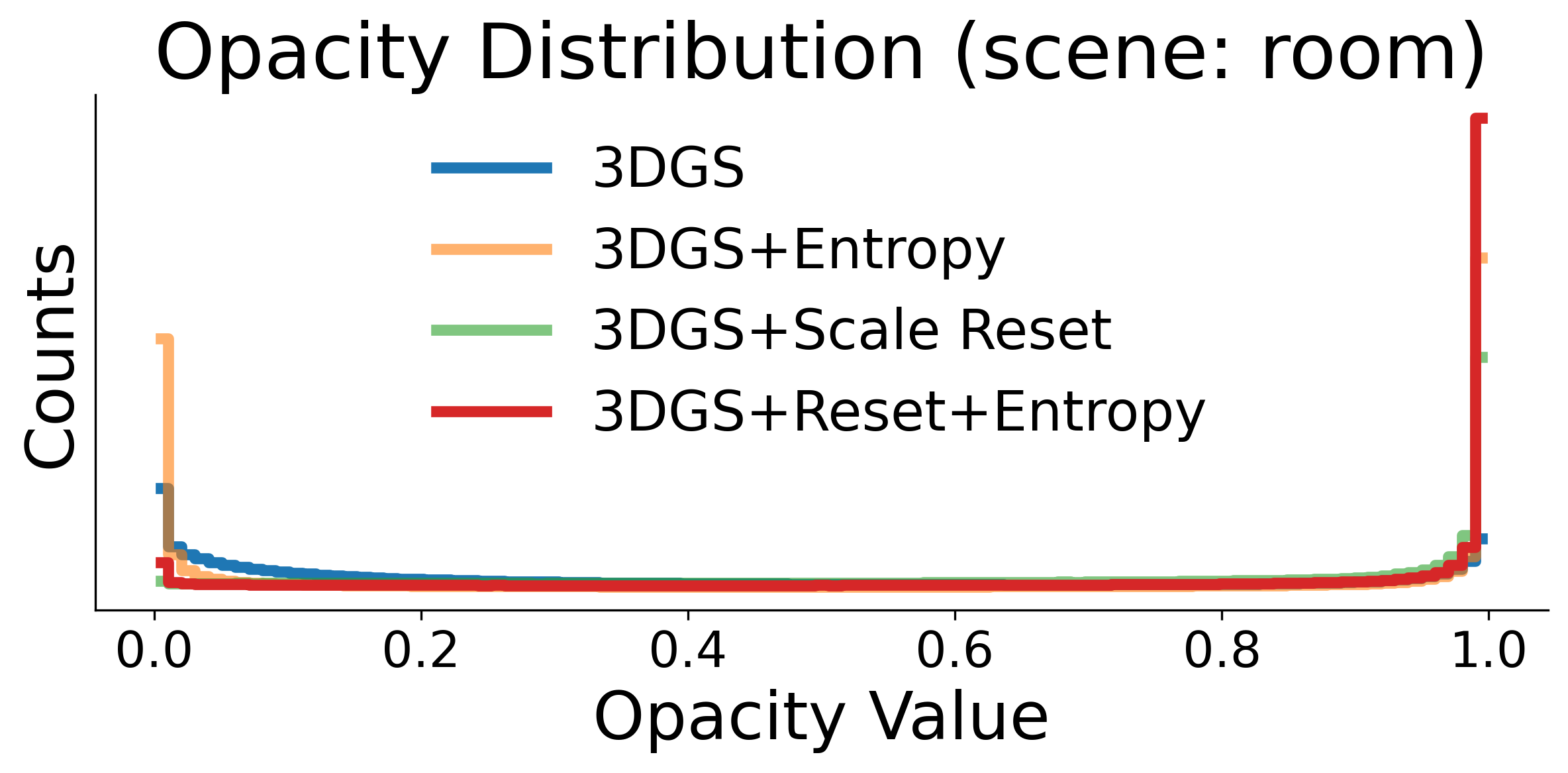} \\

Bonsai &
\includegraphics[width=0.13\textwidth]{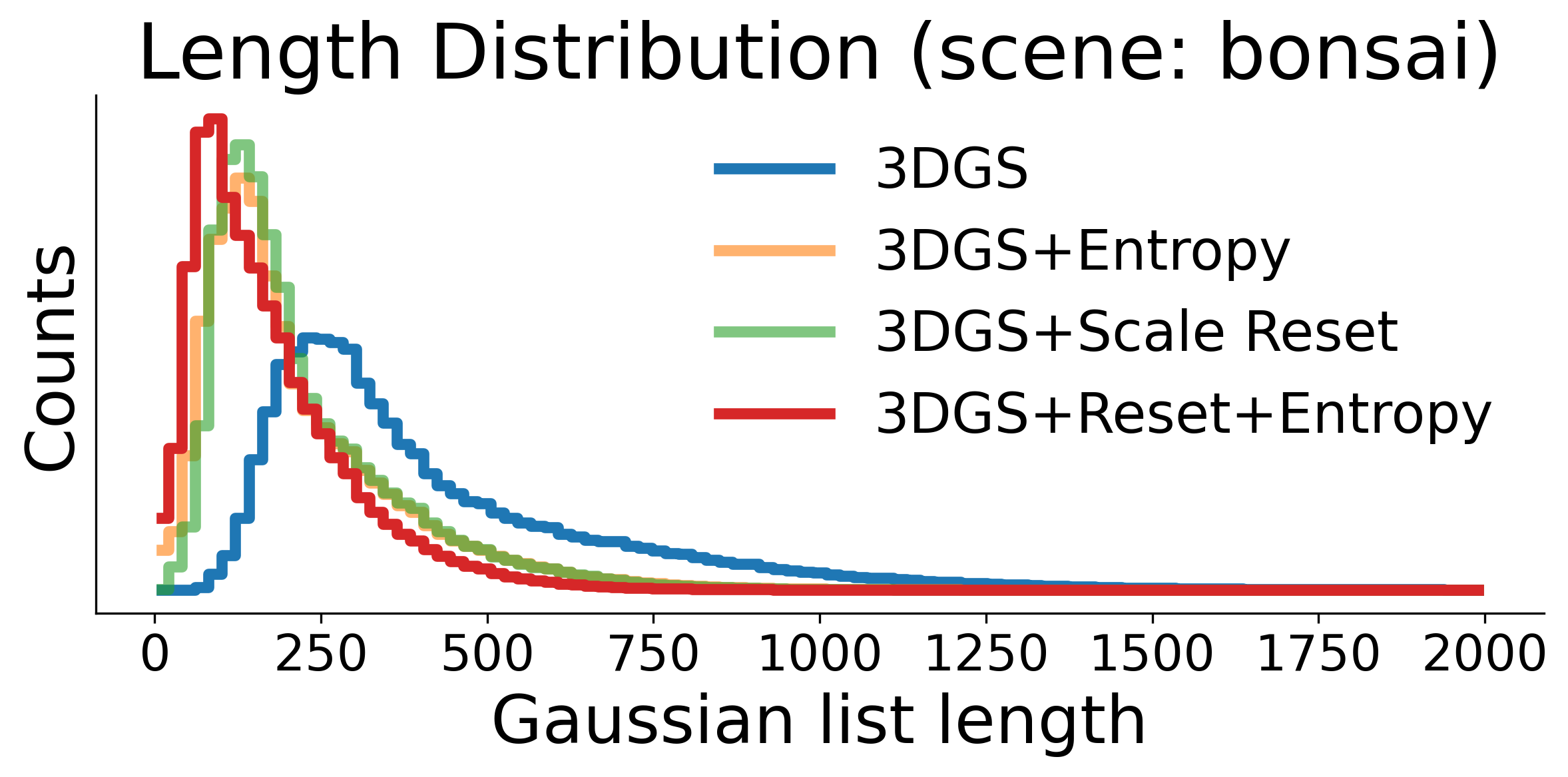} &
\includegraphics[width=0.13\textwidth]{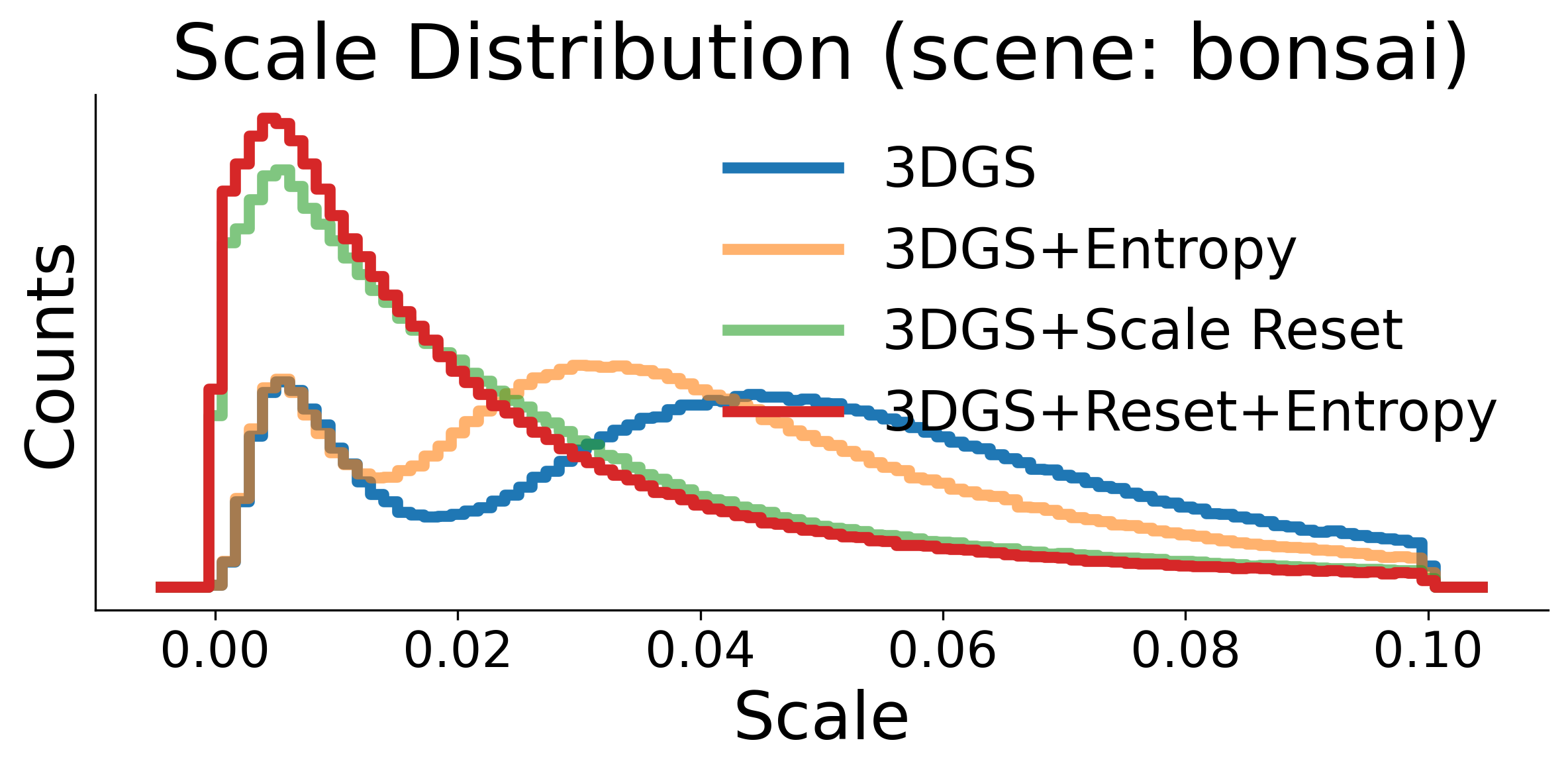} &
\includegraphics[width=0.13\textwidth]{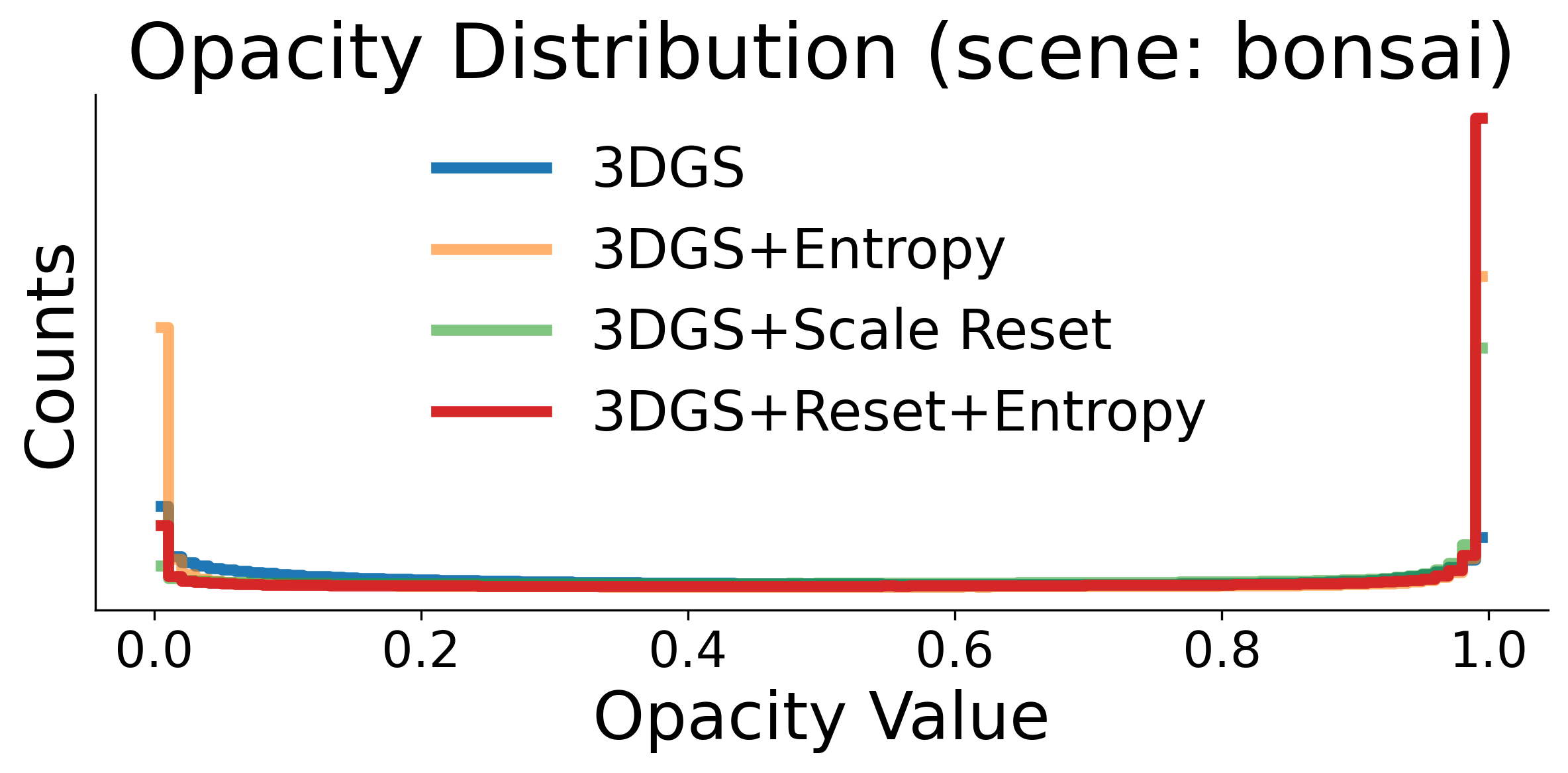} \\

Counter &
\includegraphics[width=0.13\textwidth]{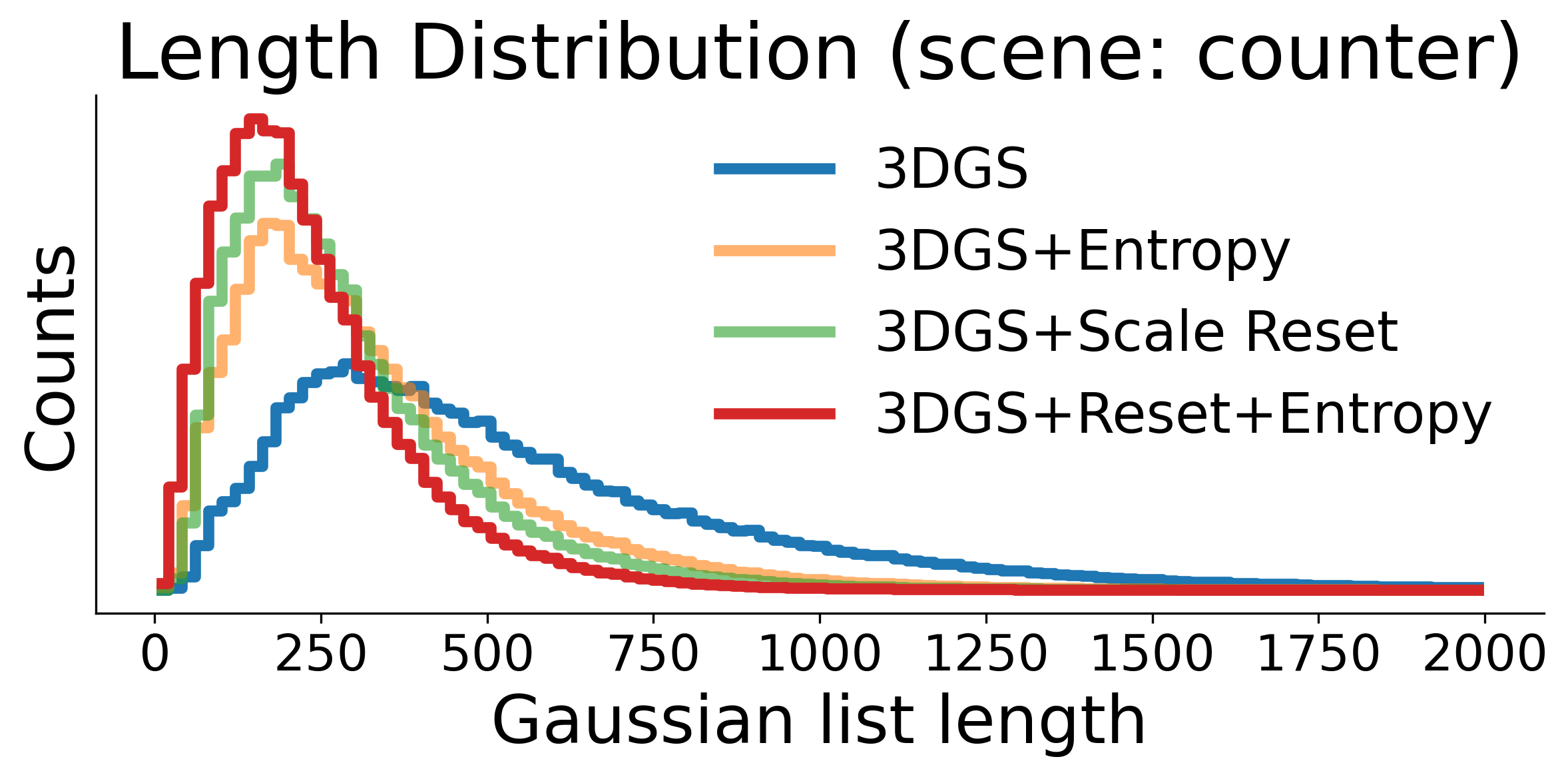} &
\includegraphics[width=0.13\textwidth]{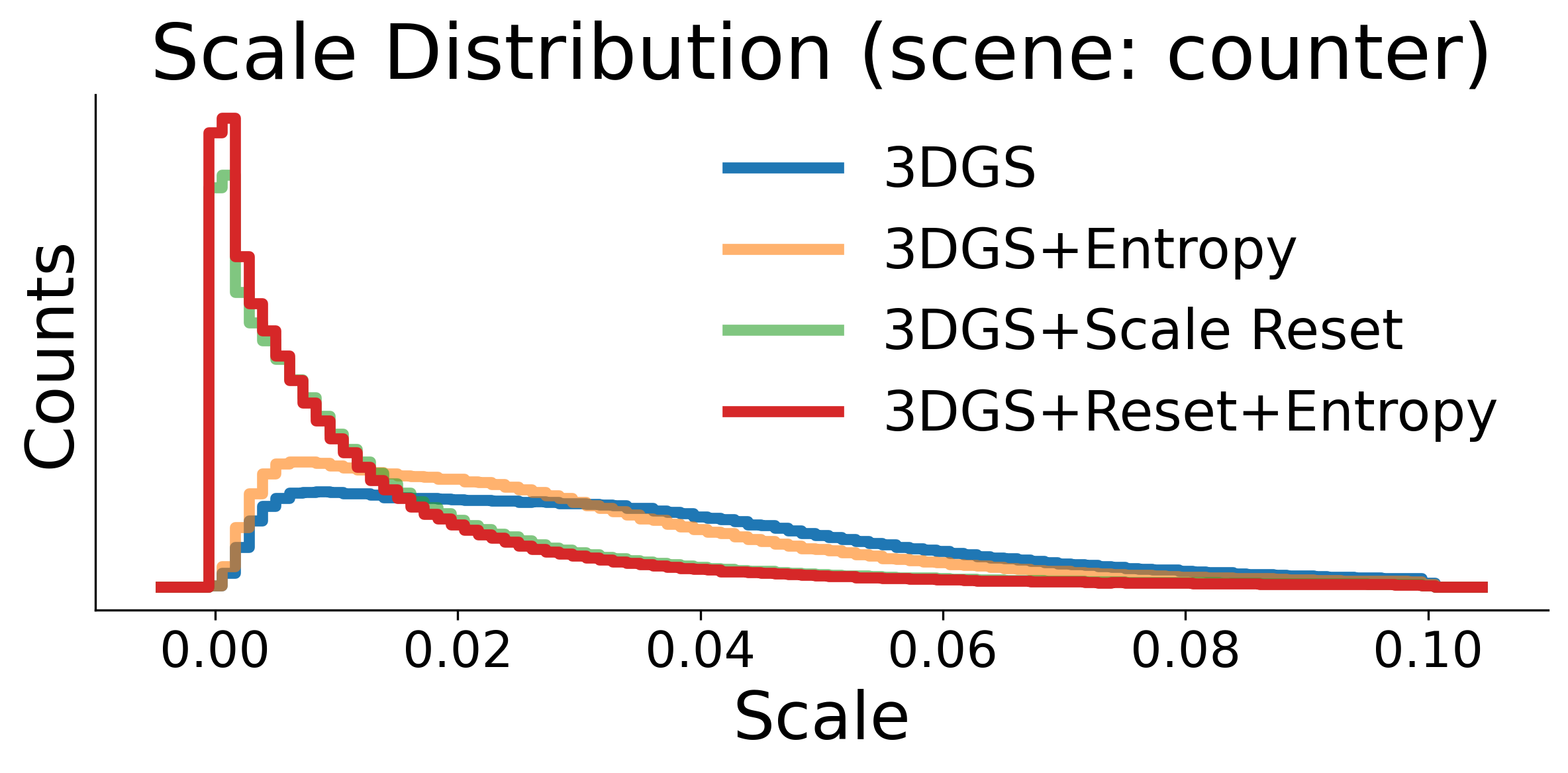} &
\includegraphics[width=0.13\textwidth]{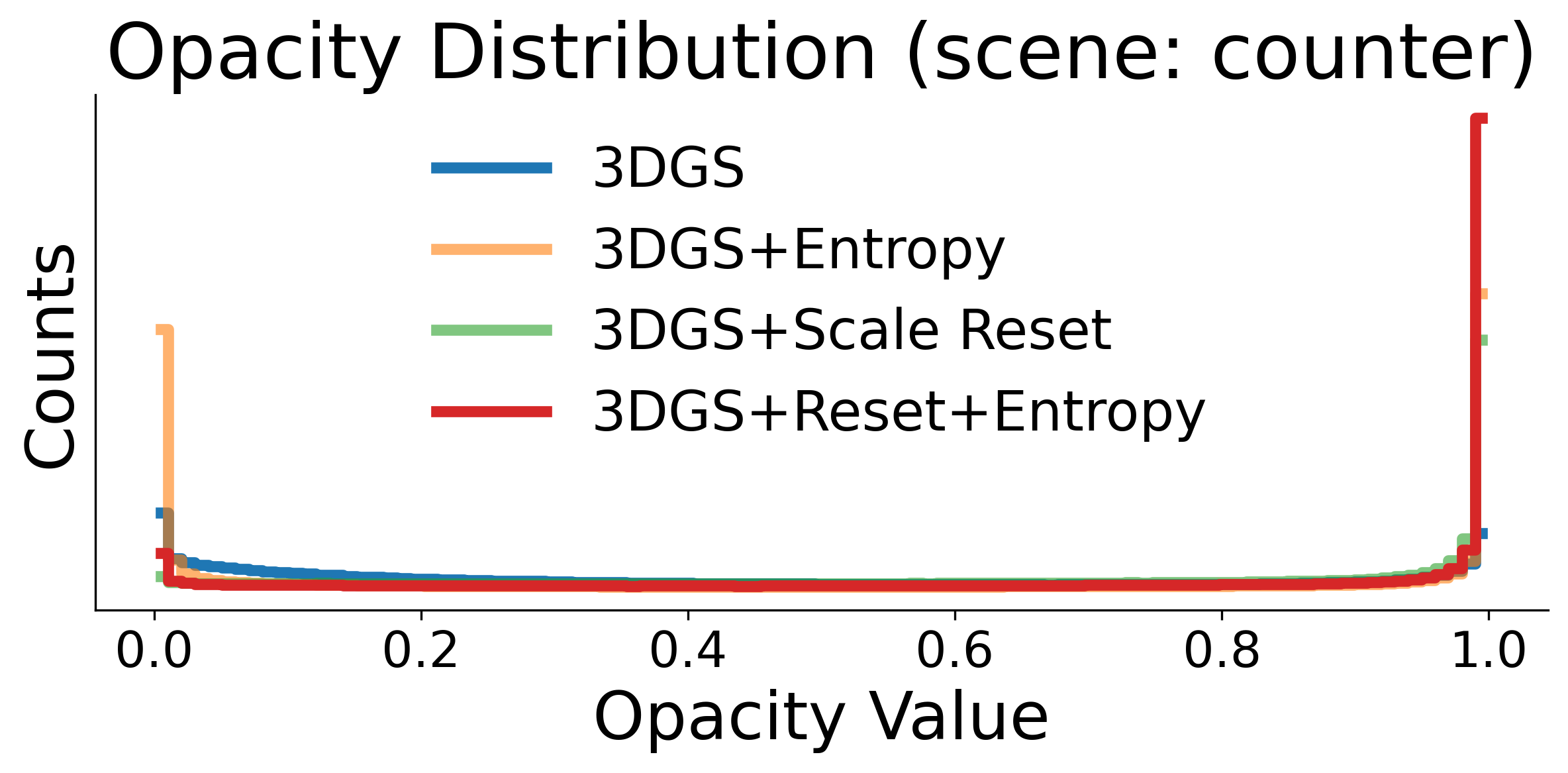} \\

Kitchen &
\includegraphics[width=0.13\textwidth]{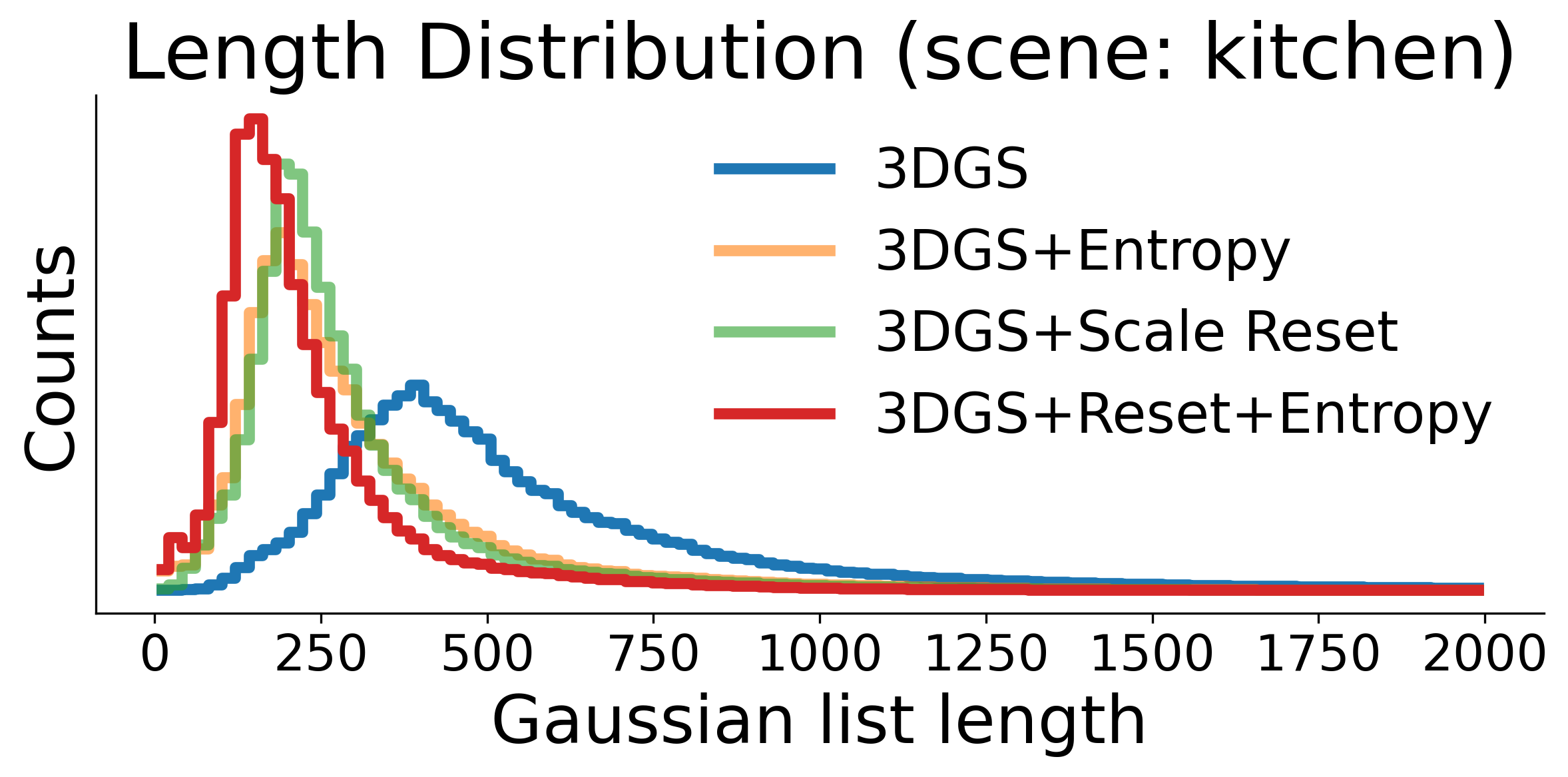} &
\includegraphics[width=0.13\textwidth]{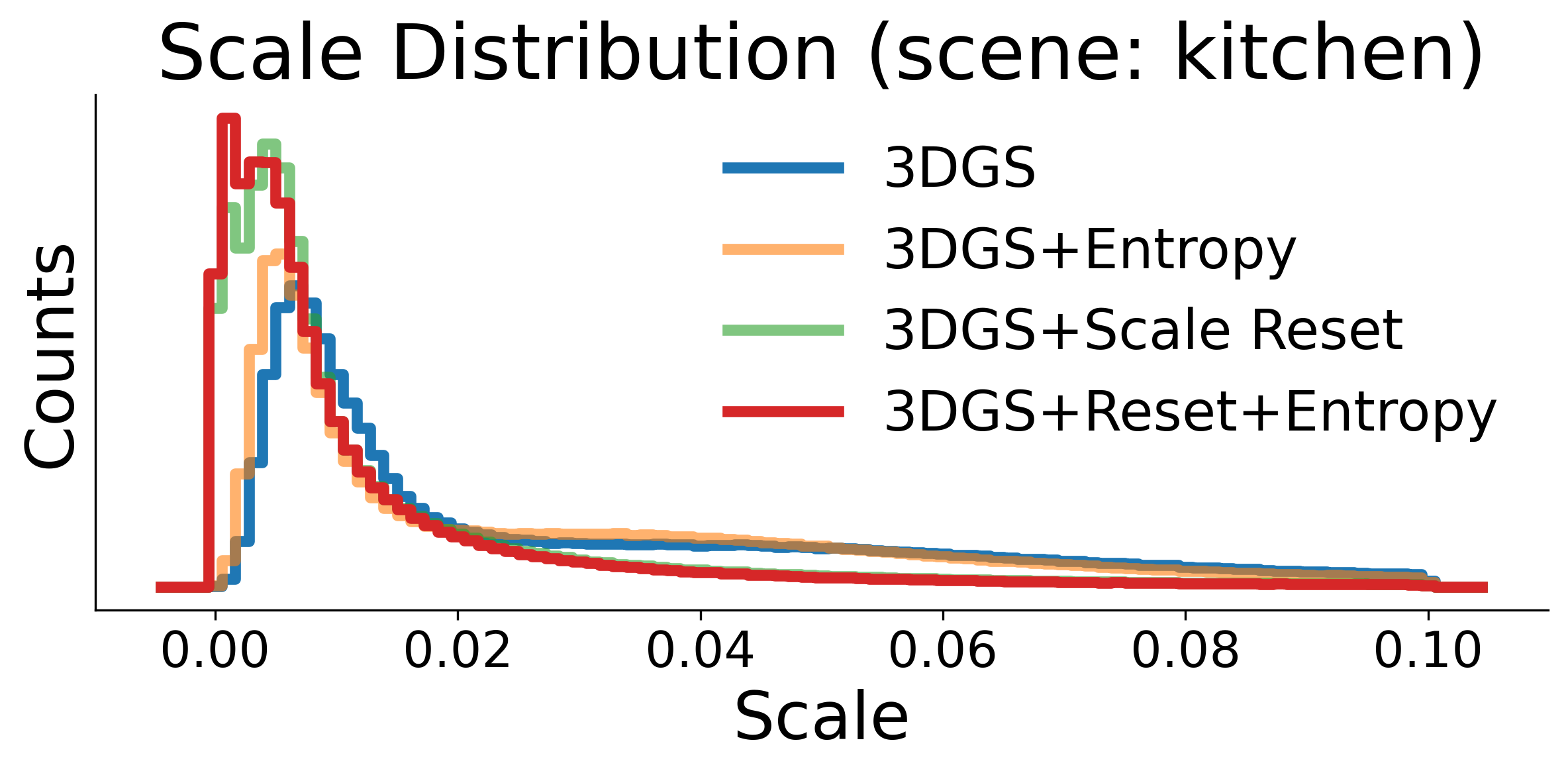} &
\includegraphics[width=0.13\textwidth]{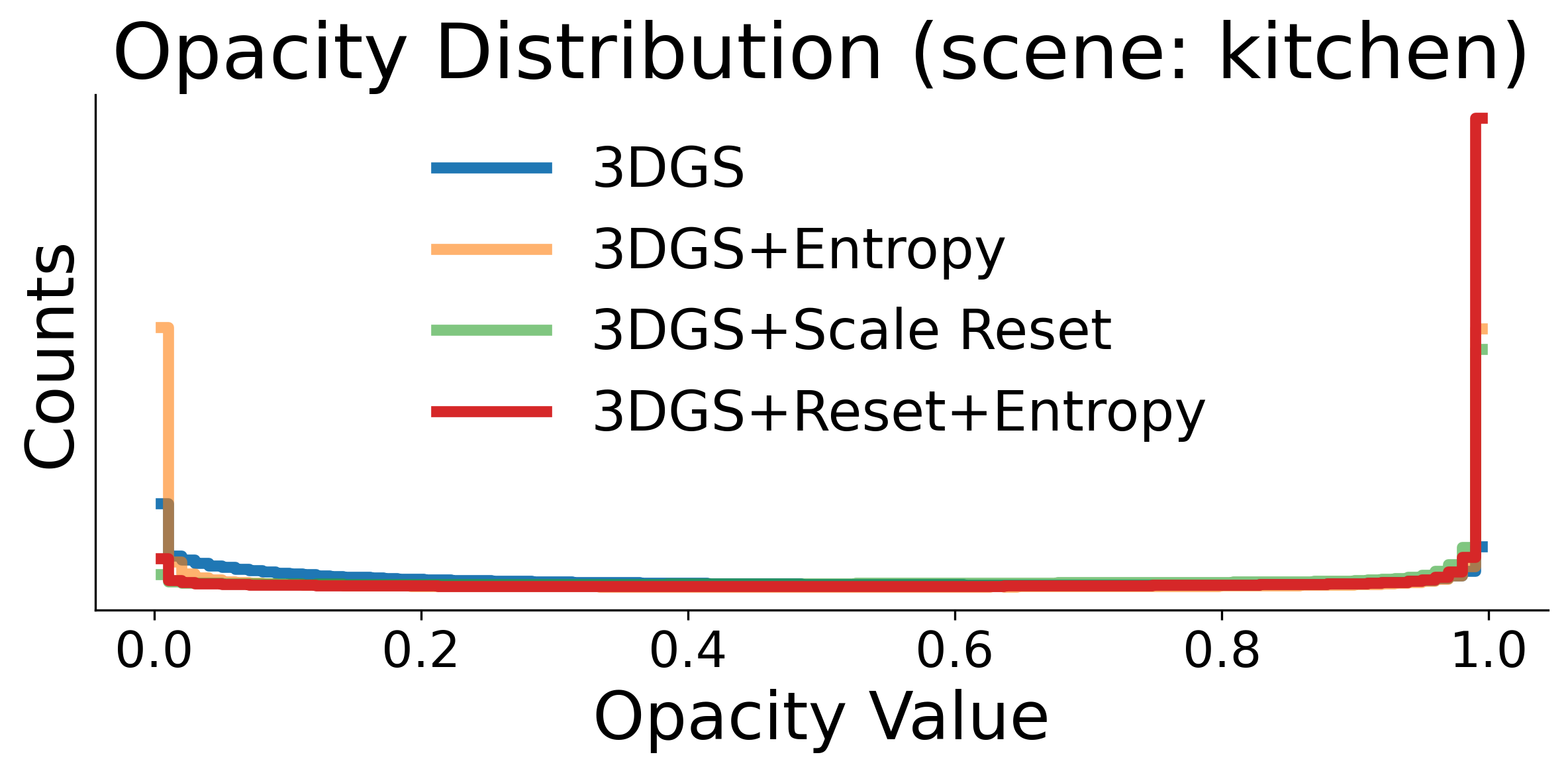} \\

Treehill &
\includegraphics[width=0.13\textwidth]{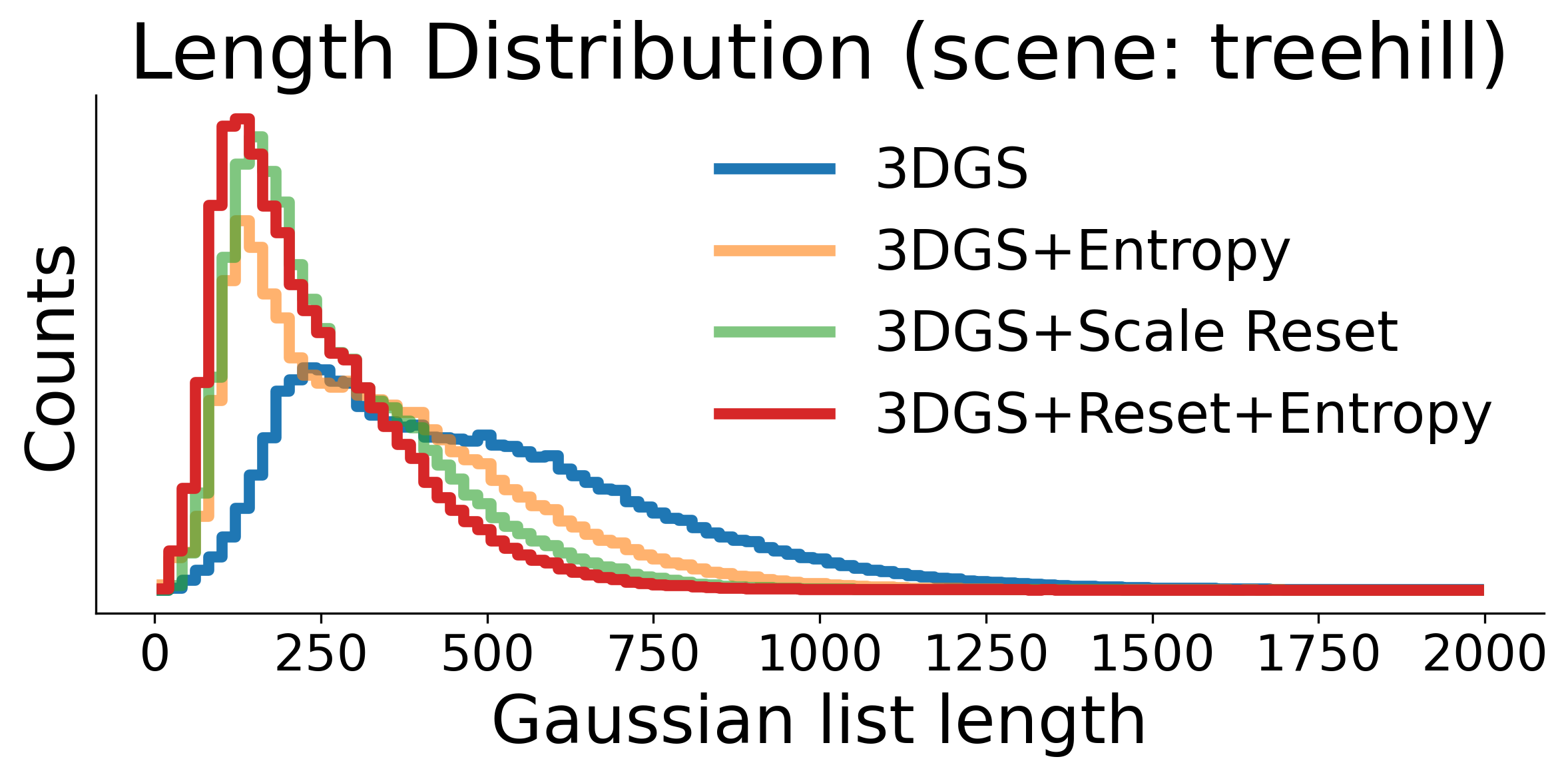} &
\includegraphics[width=0.13\textwidth]{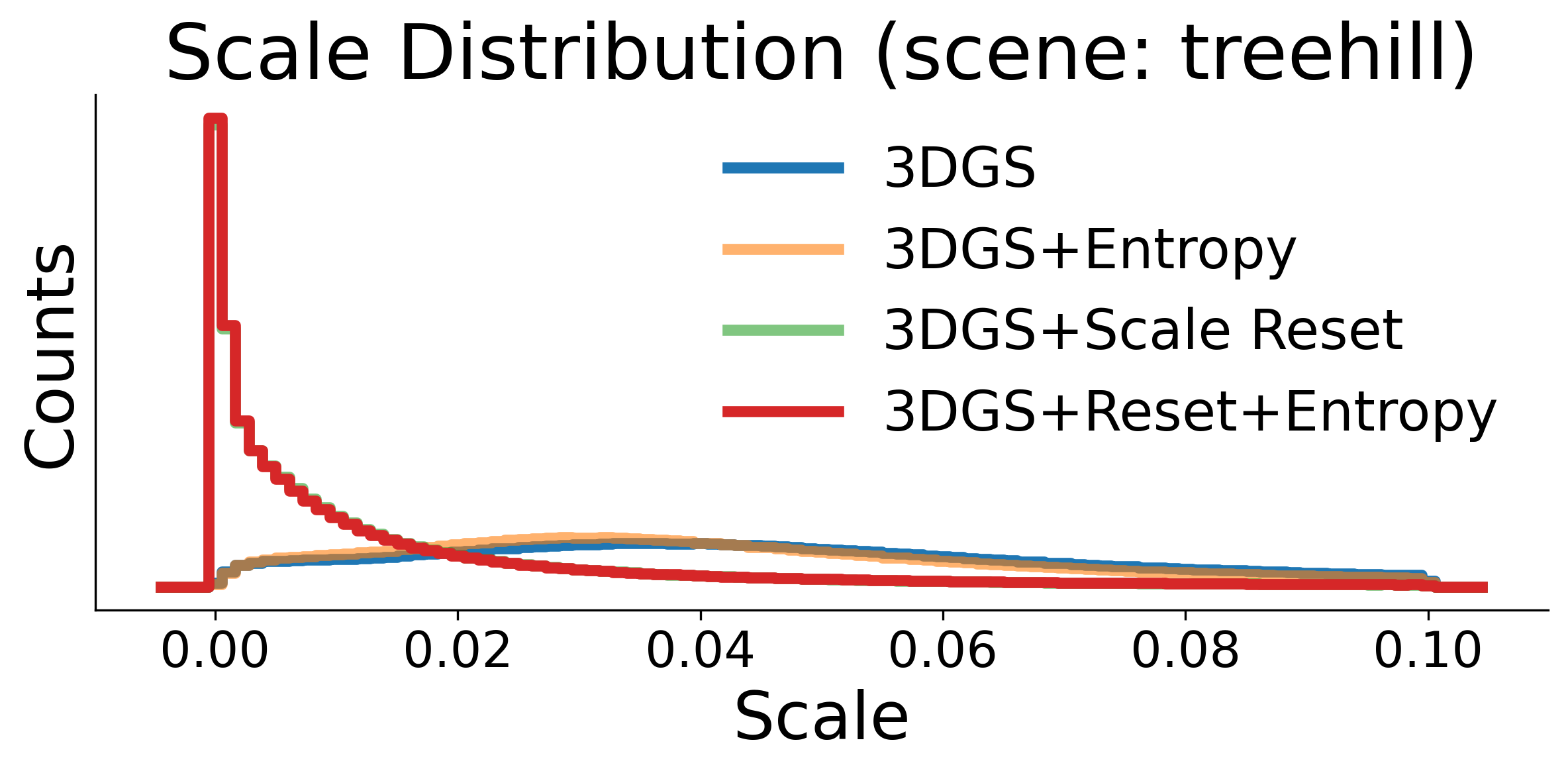} &
\includegraphics[width=0.13\textwidth]{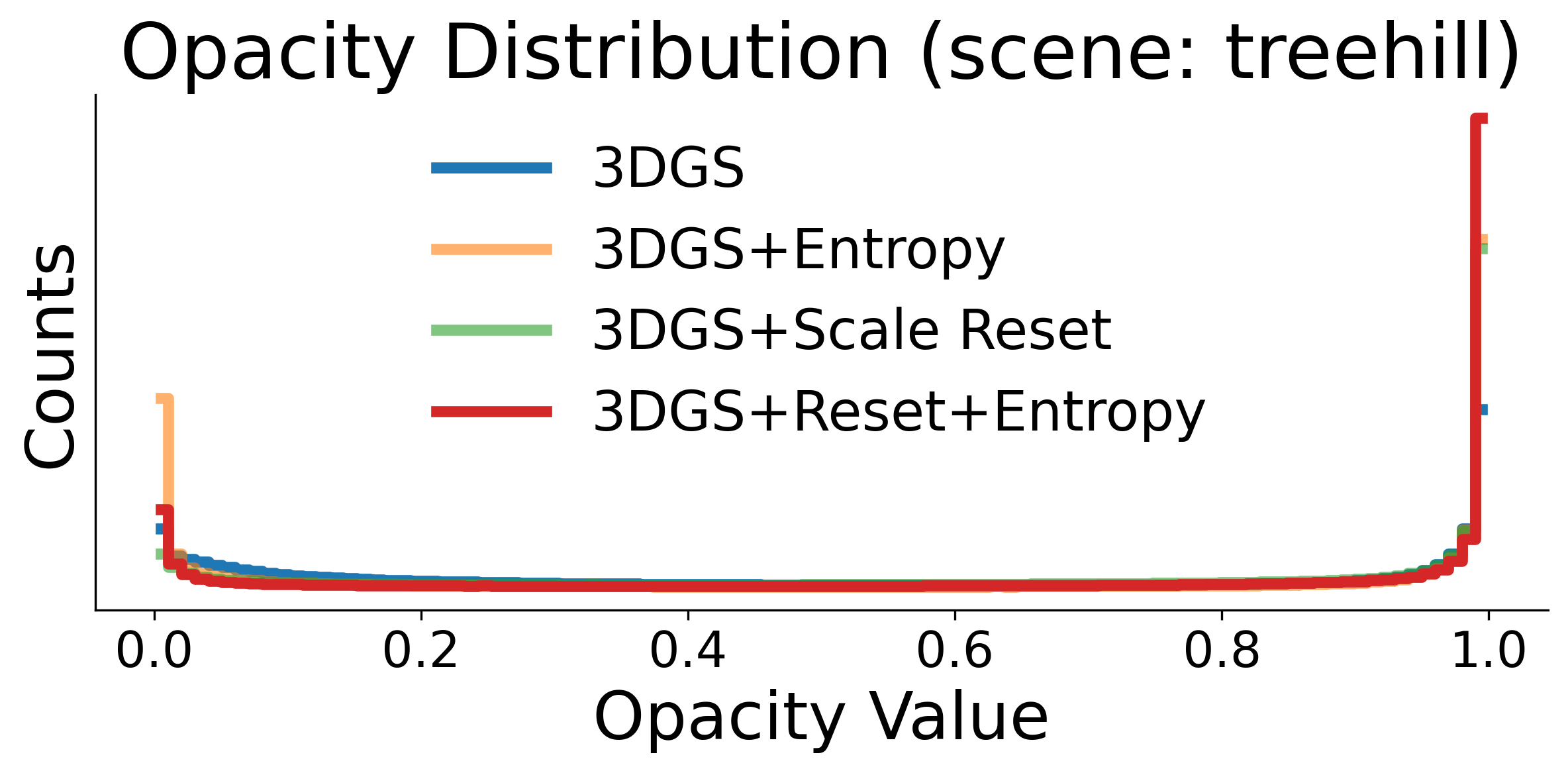} \\

Bicycle &
\includegraphics[width=0.13\textwidth]{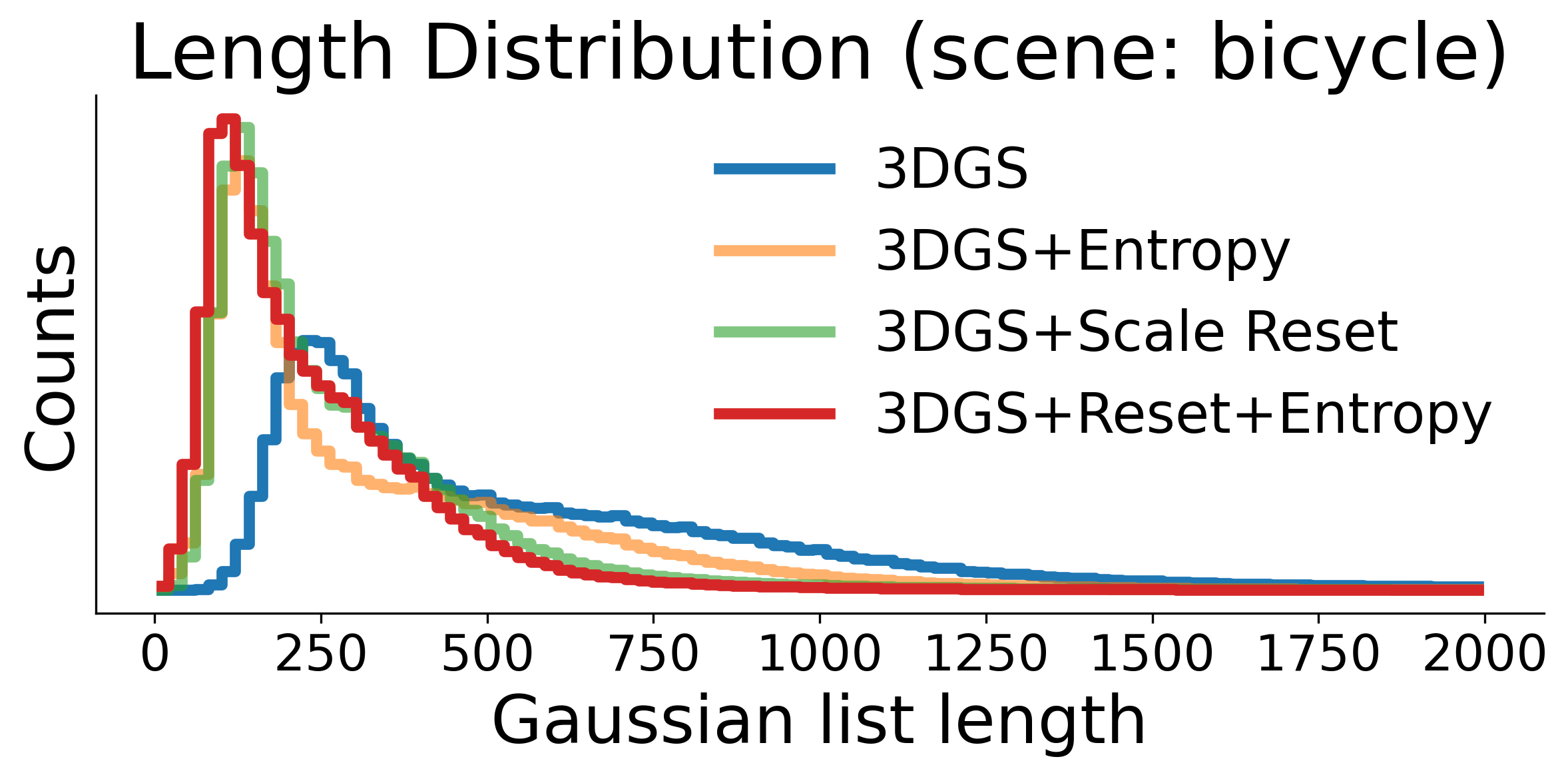} &
\includegraphics[width=0.13\textwidth]{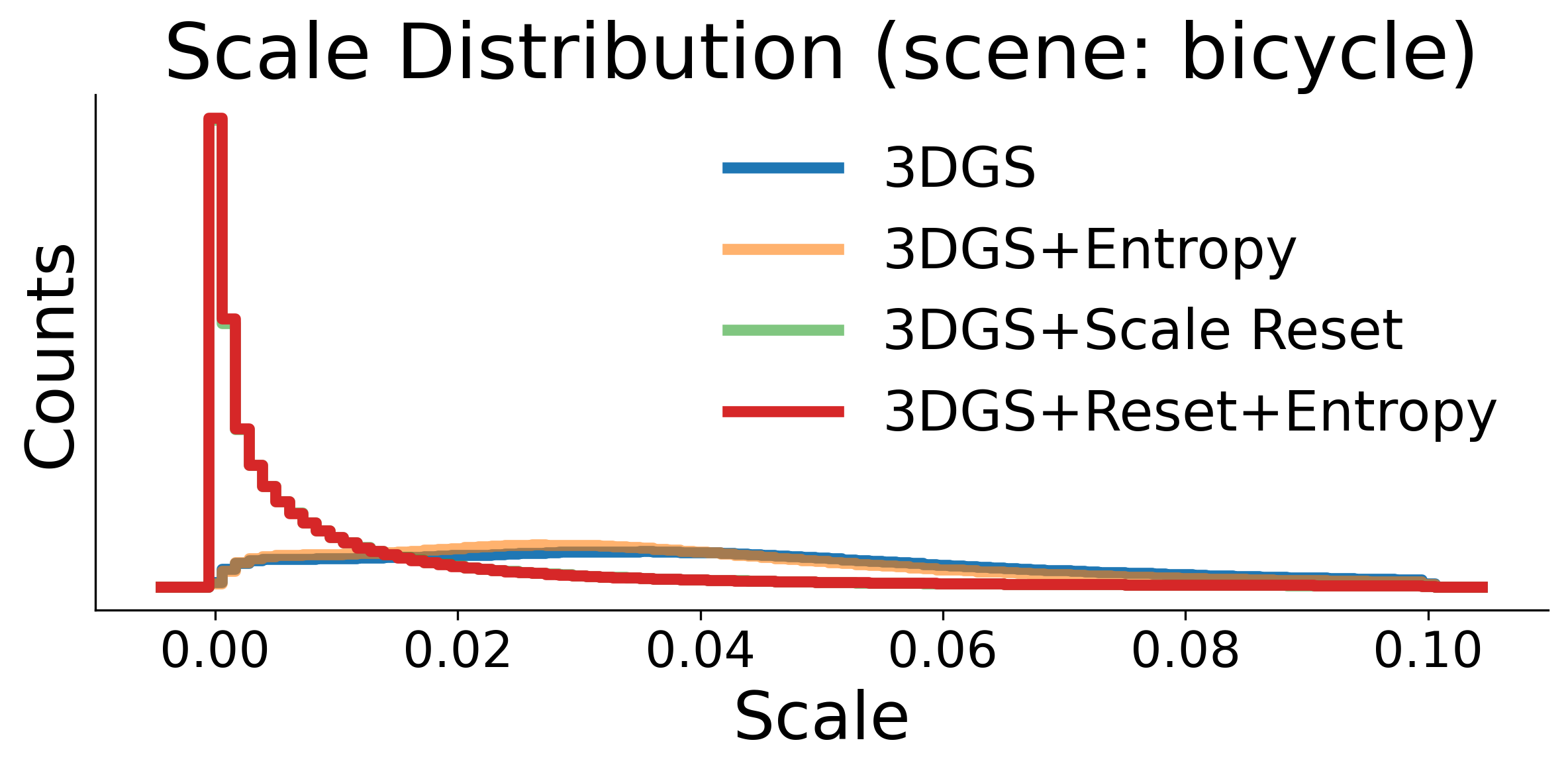} &
\includegraphics[width=0.13\textwidth]{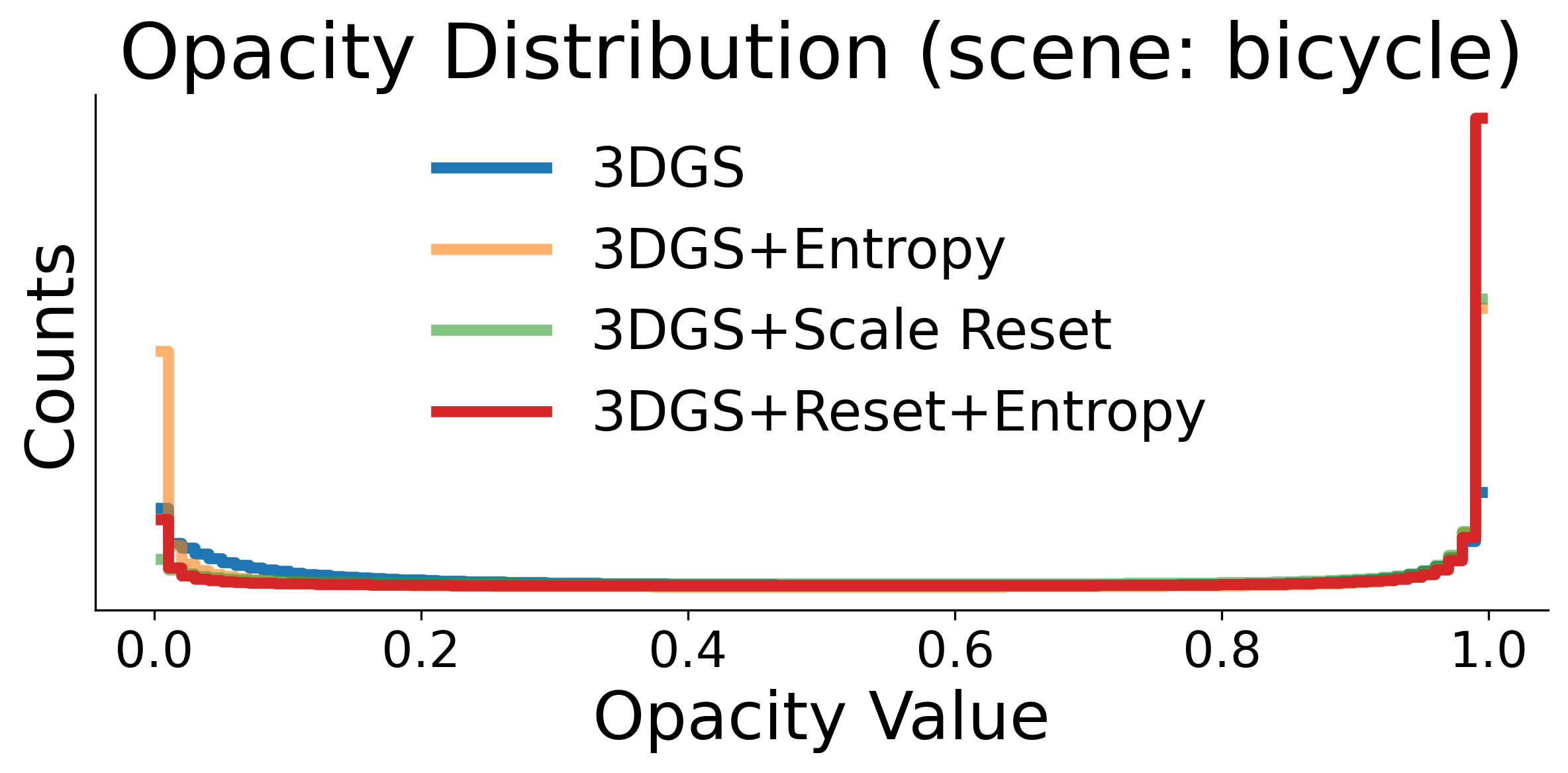} \\

Garden &
\includegraphics[width=0.13\textwidth]{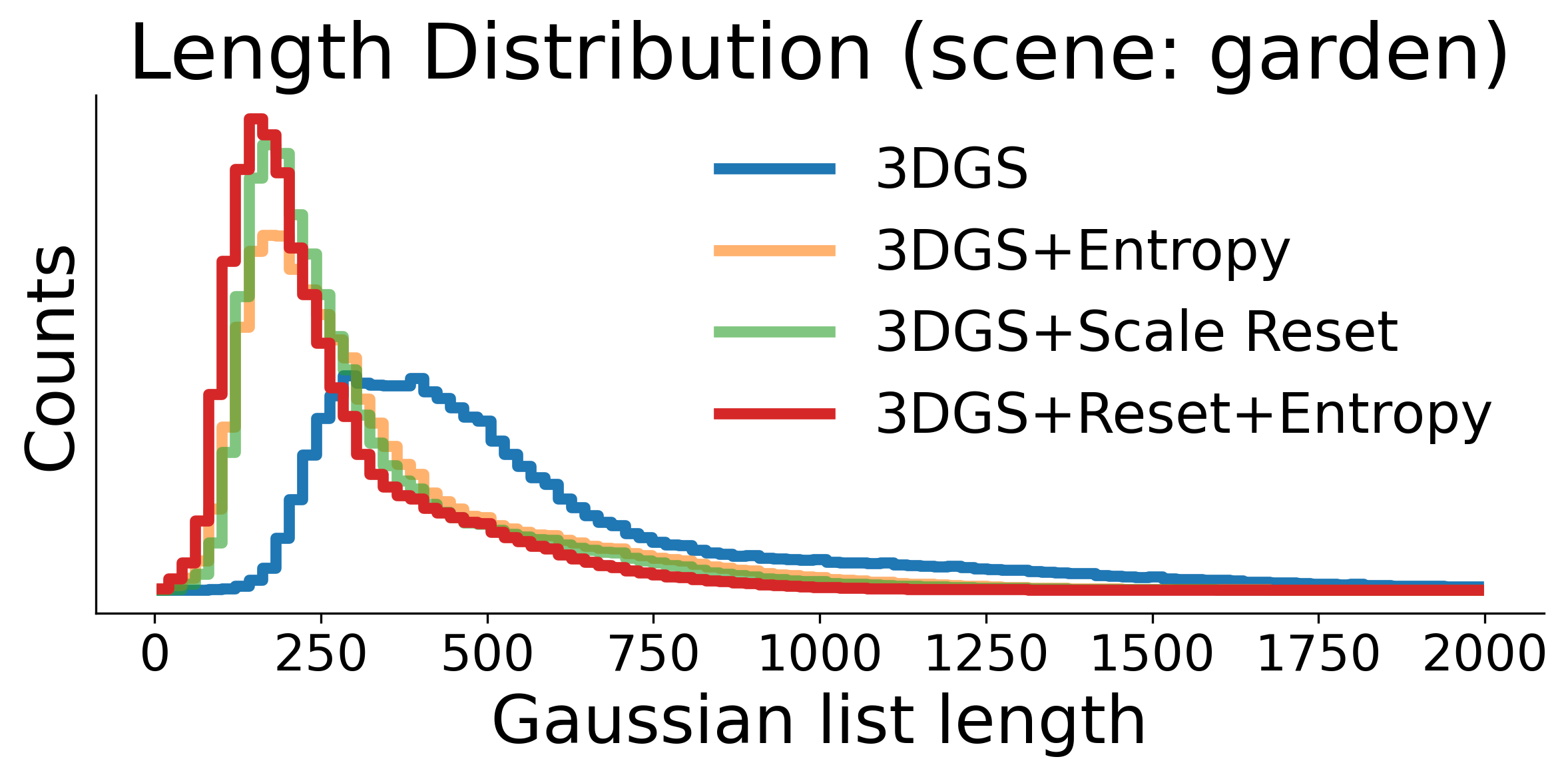} &
\includegraphics[width=0.13\textwidth]{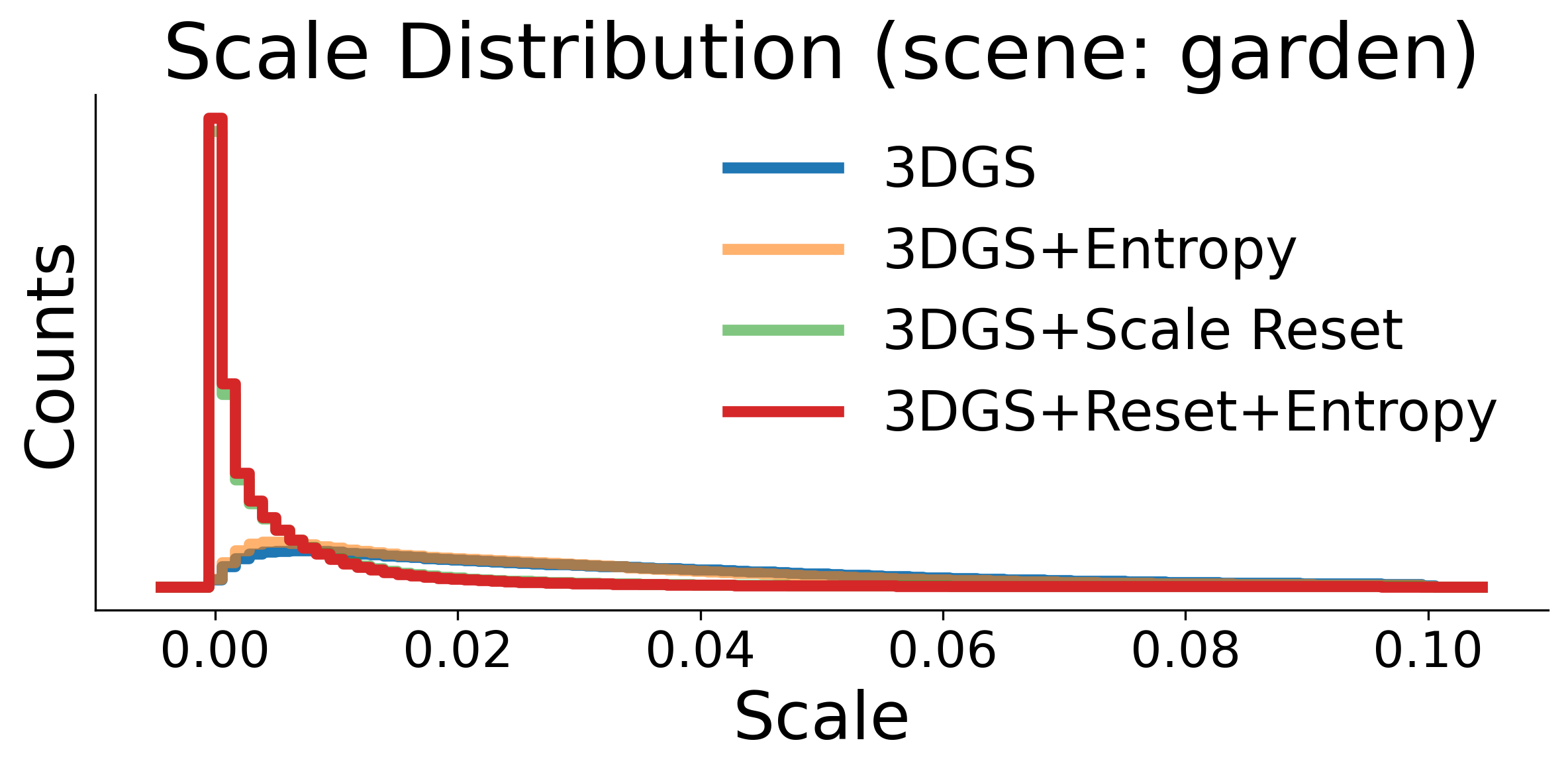} &
\includegraphics[width=0.13\textwidth]{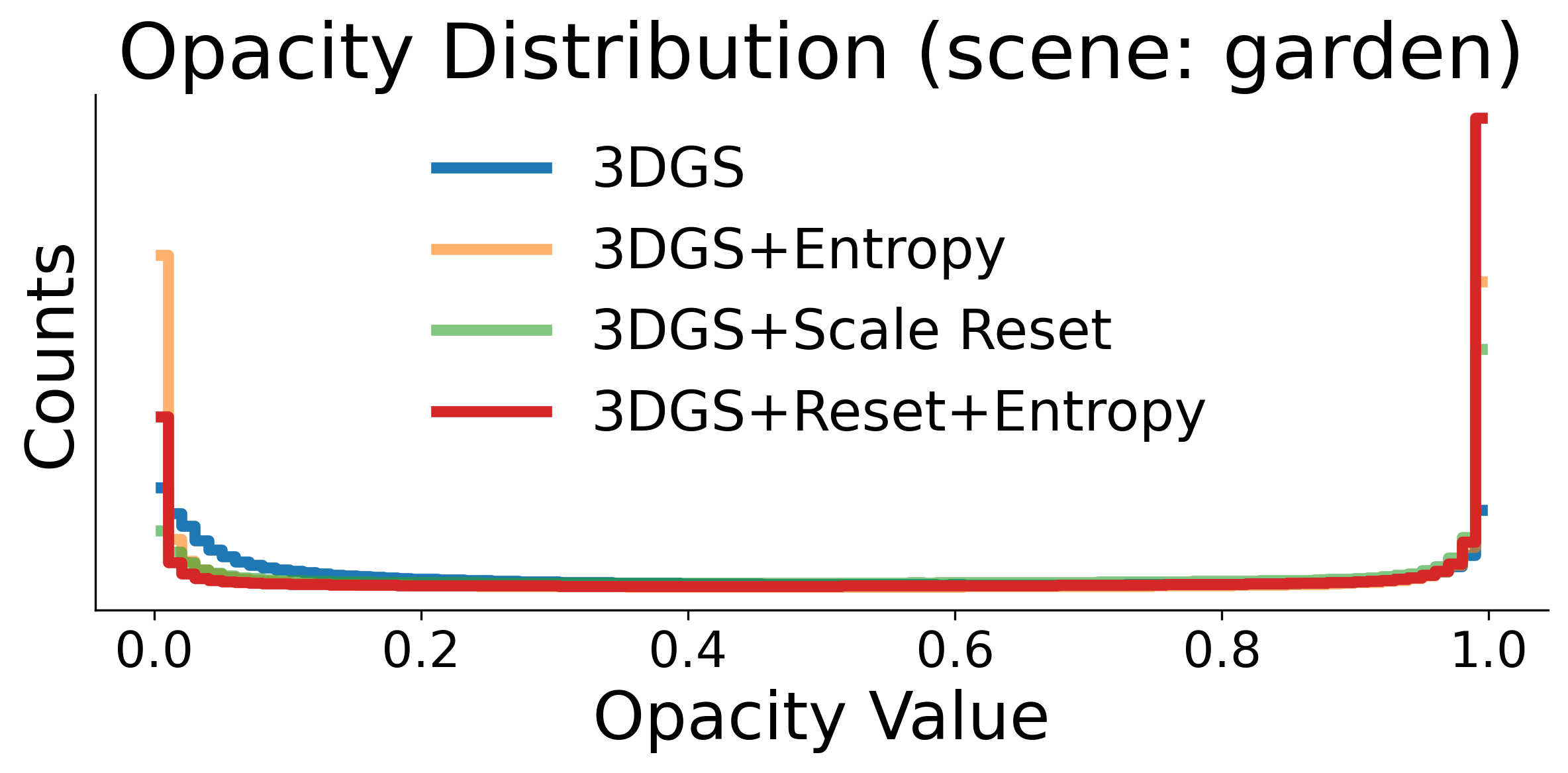} \\

Flowers &
\includegraphics[width=0.13\textwidth]{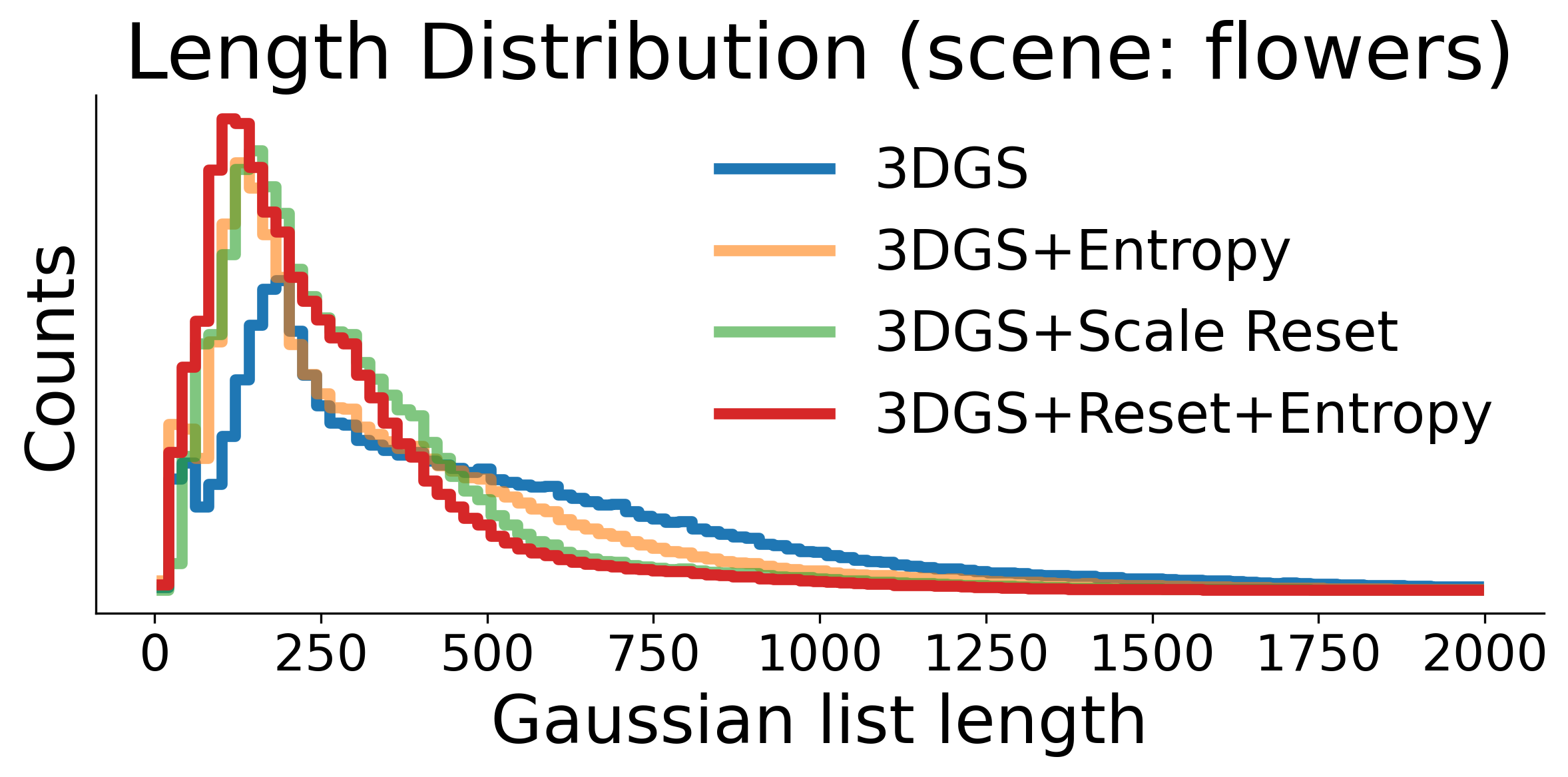} &
\includegraphics[width=0.13\textwidth]{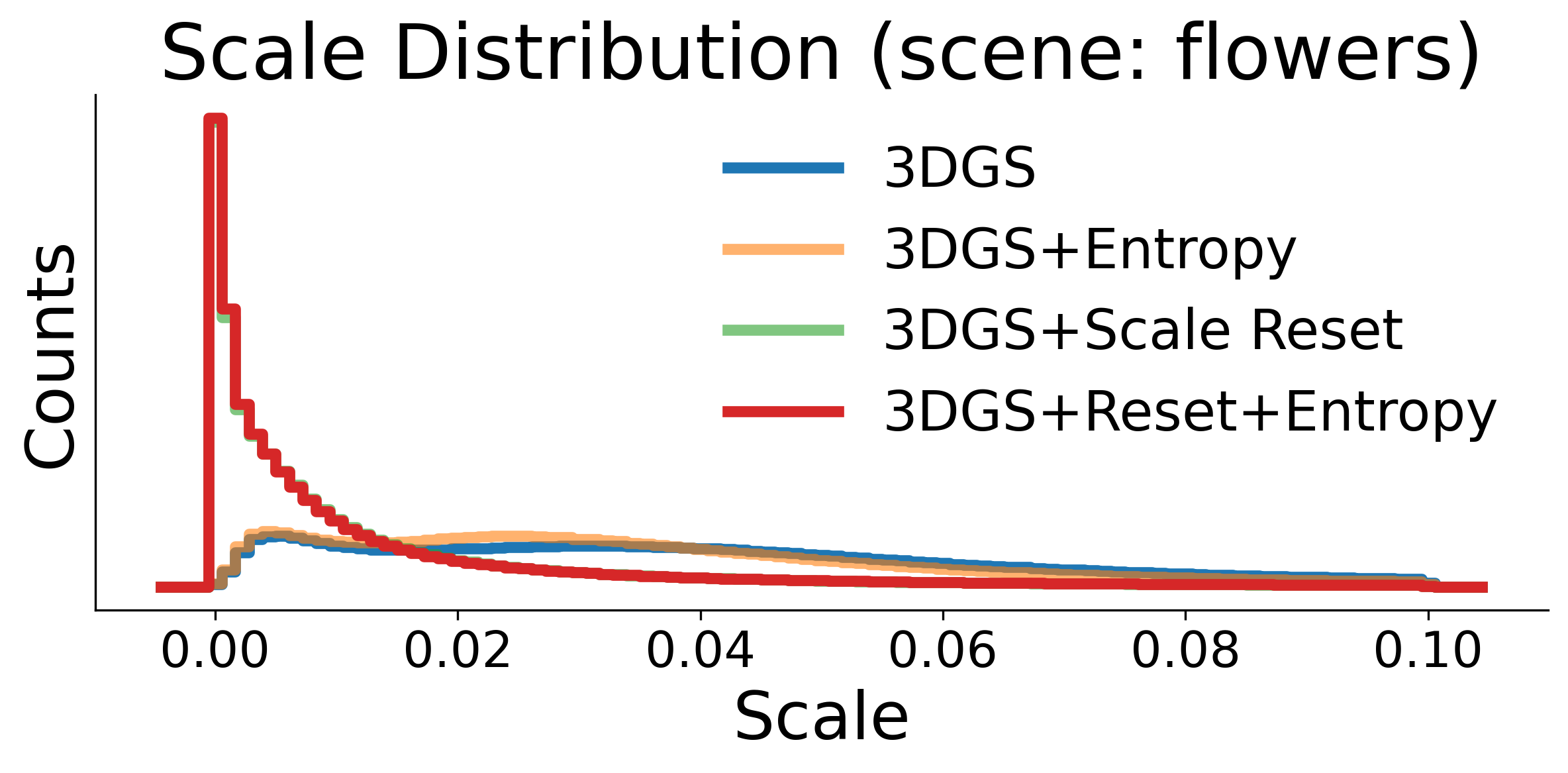} &
\includegraphics[width=0.13\textwidth]{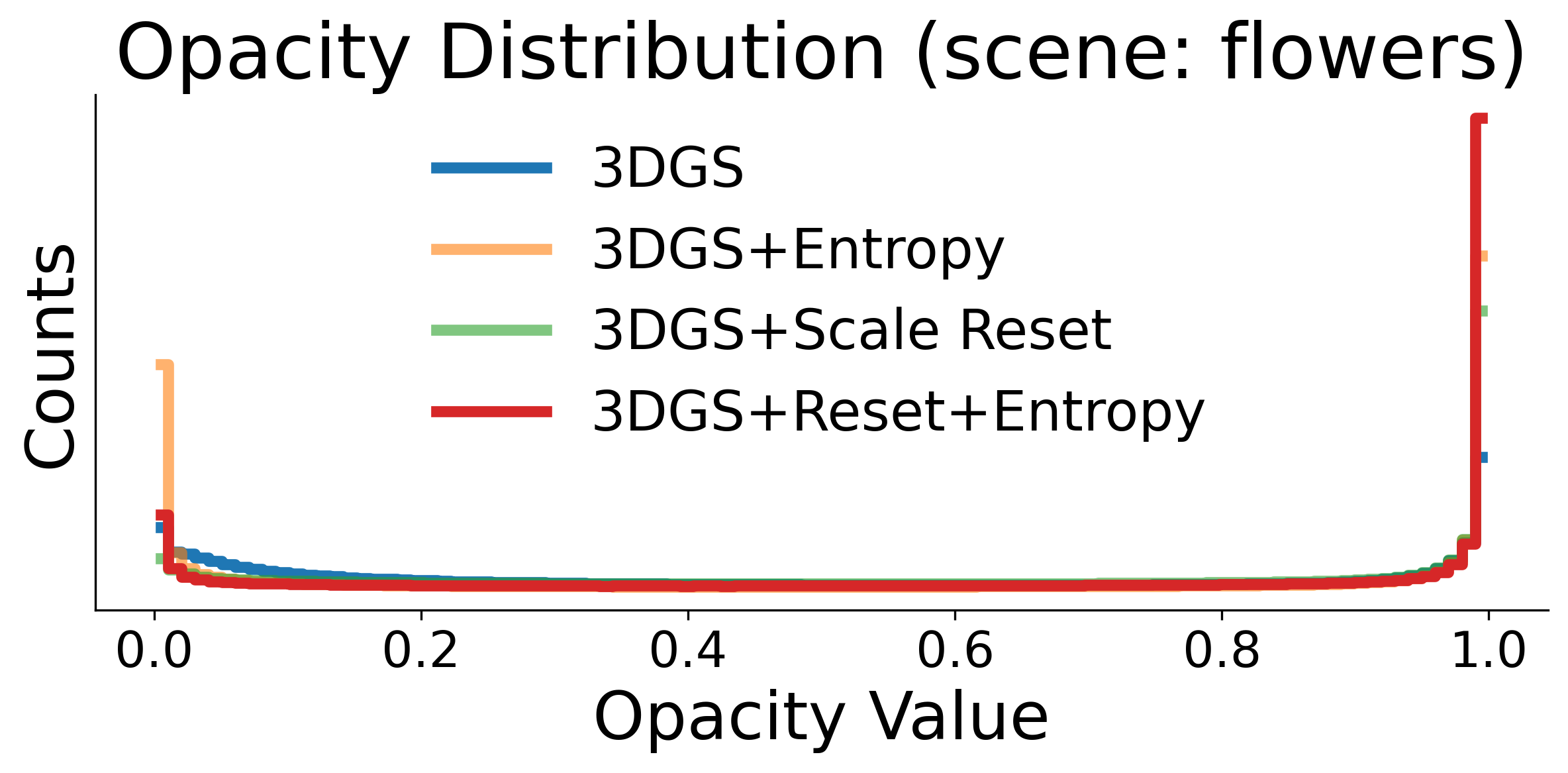} \\

Stump &
\includegraphics[width=0.13\textwidth]{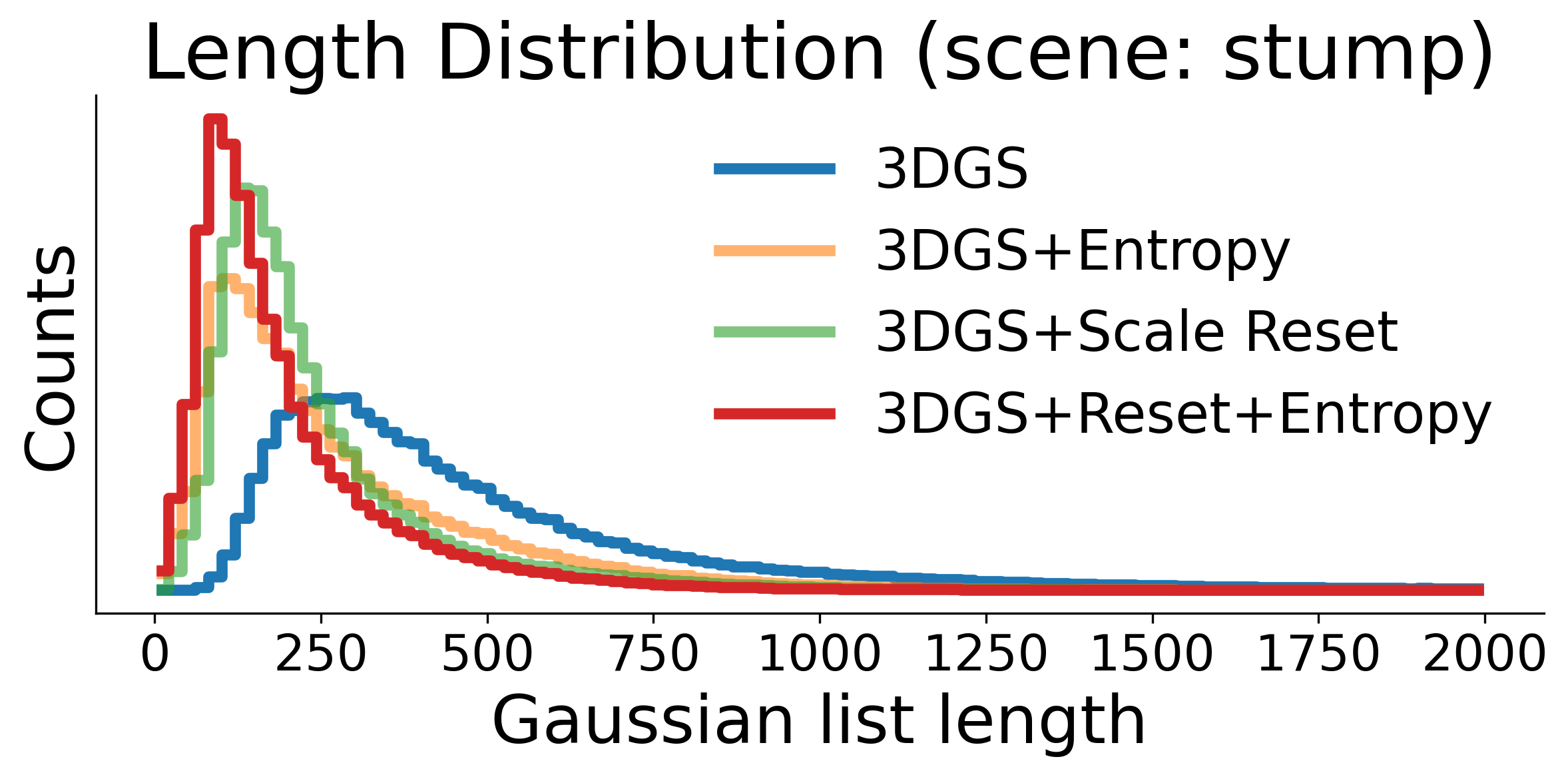} &
\includegraphics[width=0.13\textwidth]{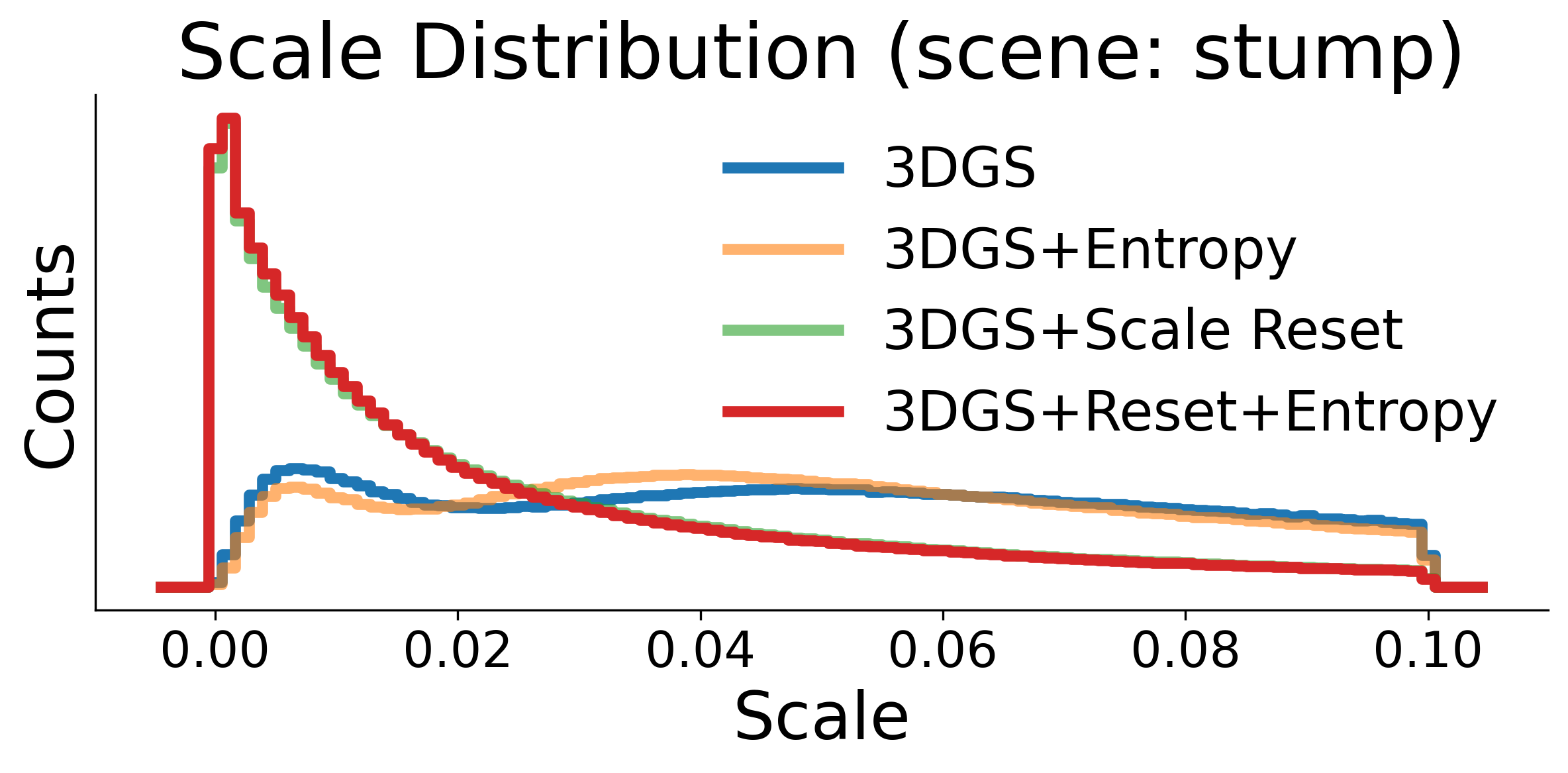} &
\includegraphics[width=0.13\textwidth]{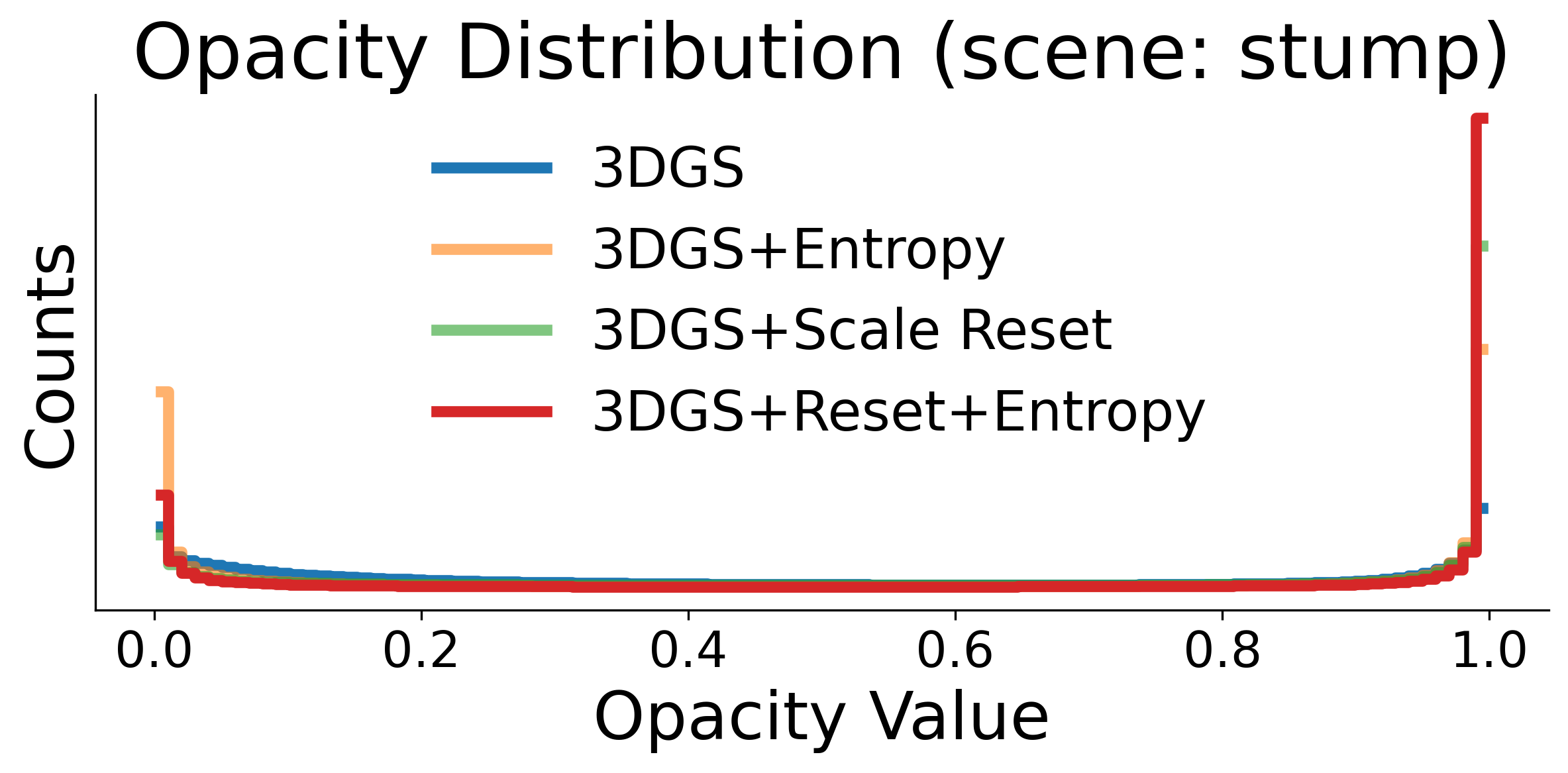} \\
\bottomrule
\end{tabular}
}
\caption{Distribution comparison per scene across all methods on length, scale, and opacity. 
Results labeled as “3DGS” are obtained using LiteGS.
See \cref{sec:ablation_study} for more details.}
\label{fig:distributions_all_methods_all_metrics}
\end{figure*}

\begin{figure*}[htbp]
\centering
\setlength{\tabcolsep}{2pt}
\begin{tabular}{@{}c cc@{}}
    & 3DGS & Ours \\[2pt]
    
    \rotatebox{90}{\small Bonsai} &
    \includegraphics[width=0.45\textwidth]{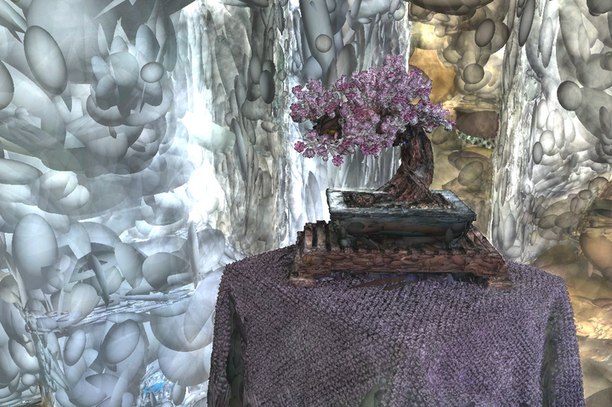} &
    \includegraphics[width=0.45\textwidth]{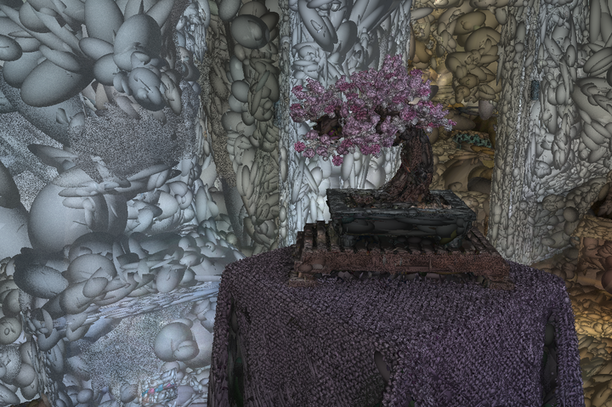} \\[4pt]
    
    \rotatebox{90}{\small Counter} &
    \includegraphics[width=0.45\textwidth]{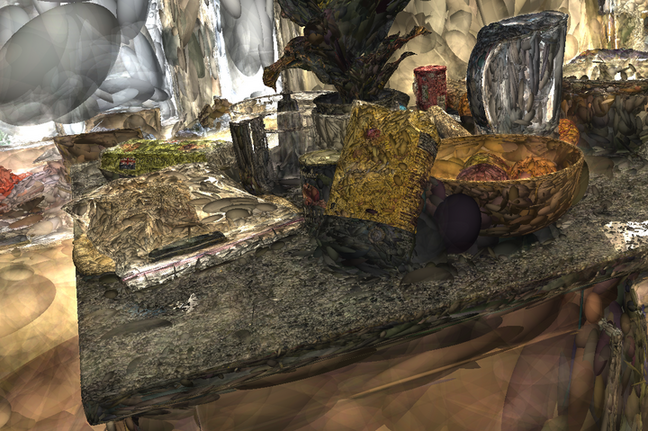} &
    \includegraphics[width=0.45\textwidth]{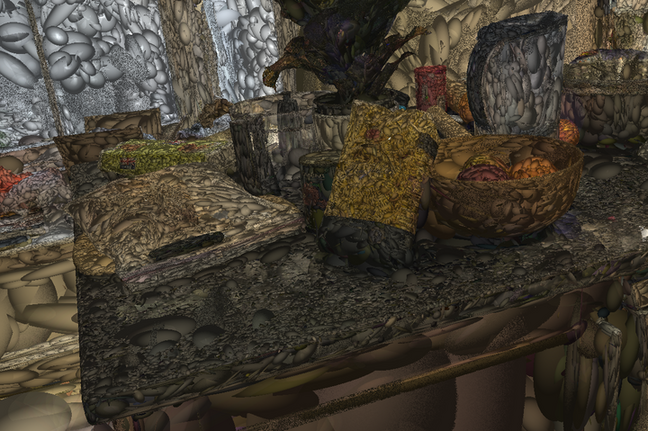} \\[4pt]
    
    \rotatebox{90}{\small Garden} &
    \includegraphics[width=0.45\textwidth]{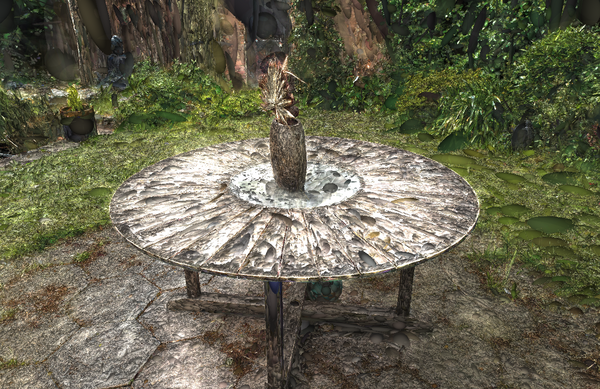} &
    \includegraphics[width=0.45\textwidth]{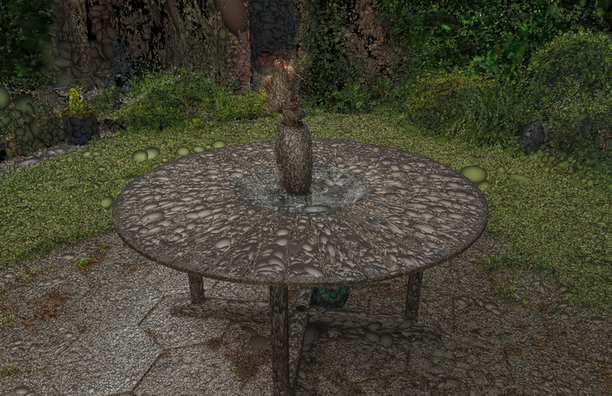} \\[4pt]
    
    \rotatebox{90}{\small Treehill} &
    \includegraphics[width=0.45\textwidth]{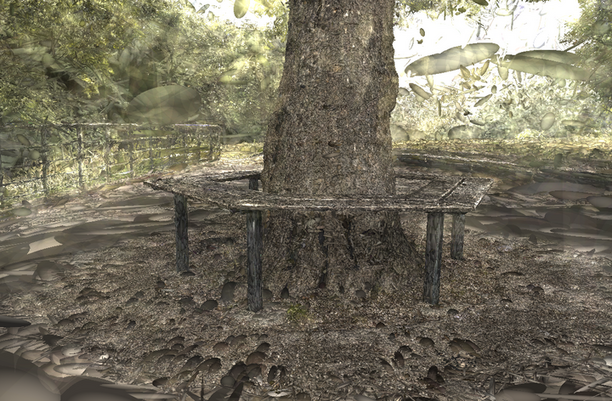} &
    \includegraphics[width=0.45\textwidth]{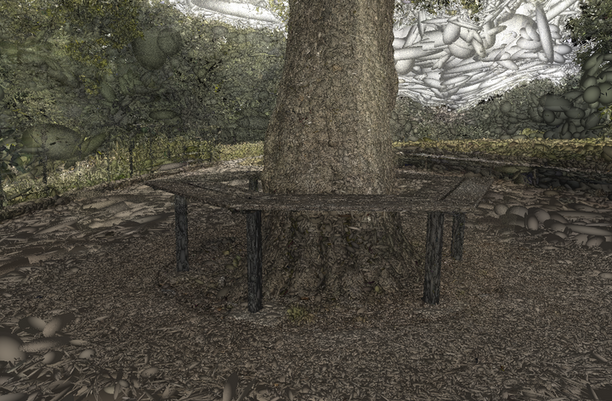} \\
\end{tabular}
\caption{Visualization of Gaussians on selected scenes from the Mip-NeRF~360 dataset. Compared to 3DGS, our method produces smaller and more compact Gaussians.}
\label{fig:comparison_gaussians_viz}
\end{figure*}

Heatmaps measuring the per-tile Gaussian list lengths in color are presented in \cref{fig:supplementary_gaussian_count_viz}. 
Please also see our video for additional visualizations of Gaussian list lengths from multiple testing perspectives and how they evolve during training.
The ablation study examining the effect of individual modules is detailed in \cref{fig:supplementary_gaussian_count_viz_ablation}.
Per-pixel entropy heatmap comparison between LiteGS and our method across scenes in the Mip-NeRF~360 dataset is shown in \cref{fig:entropy_comparison}.
Per-scene distributions of Gaussian length, scale, and opacity are presented in \cref{fig:distributions_all_methods_all_metrics}, demonstrating consistent improvements across all scenes.
Gaussians are visualized in \cref{fig:comparison_gaussians_viz}, showing our method produces smaller Gaussians than the original 3DGS.


\section{Code}

We provide a piece of demonstration code for reference. We will release our code upon the acceptance. 

\section{Video}

We provide a video to show more visualizations and details.


\clearpage

\clearpage

\clearpage

\begin{figure*}[t]
    \centering
    \includegraphics[width=\textwidth]{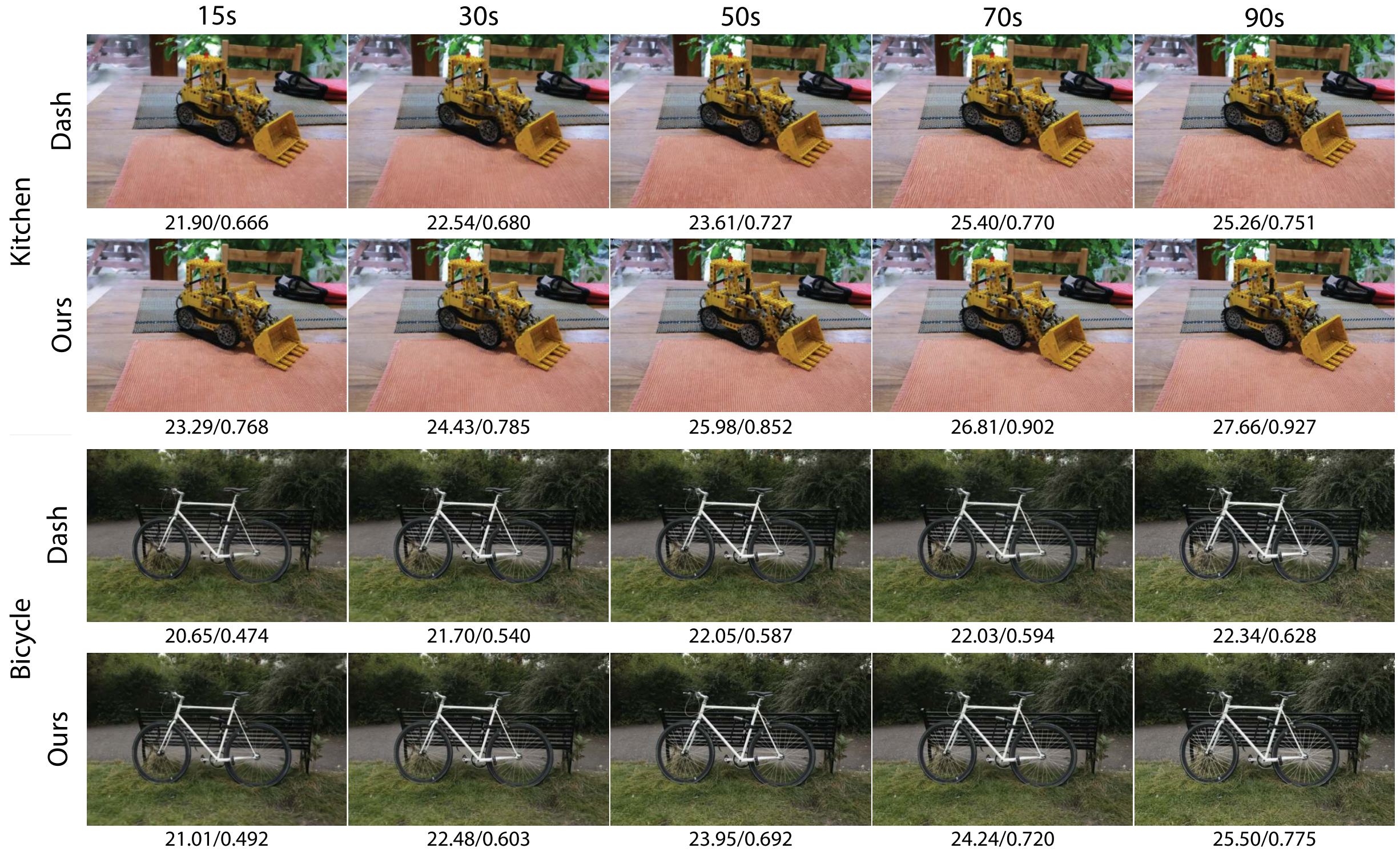}
    \caption{
    Qualitative comparison of rendering quality across two scenes from Mip-NeRF~360, comparing our method with DashGaussian. Each row shows results at training times of 15s, 30s, 50s, 70s, and 90s. PSNR and SSIM metrics are displayed in each image. 
    }
    \label{fig:render_process_p2}
\end{figure*}

\clearpage

\clearpage

\newpage

\clearpage

\newpage

%% file: main.bib
@String(CVPR= {IEEE Conf. Comput. Vis. Pattern Recog.})

@String(ICCV= {Int. Conf. Comput. Vis.})

@String(ECCV= {Eur. Conf. Comput. Vis.})

@String(AAAI = {AAAI})

@String(CVPR  = {CVPR})

@String(ICCV  = {ICCV})

@String(ECCV  = {ECCV})

@Article{kerbl3Dgaussians,
      author       = {Kerbl, Bernhard and Kopanas, Georgios and Leimk{\"u}hler, Thomas and Drettakis, George},
      title        = {3D Gaussian Splatting for Real-Time Radiance Field Rendering},
      journal      = {ACM Transactions on Graphics},
      number       = {4},
      volume       = {42},
      month        = {July},
      year         = {2023},
      url          = {https://repo-sam.inria.fr/fungraph/3d-gaussian-splatting/}
}

@misc{kotovenko2025edgs,
      title={EDGS: Eliminating Densification for Efficient Convergence of 3DGS}, 
      author={Dmytro Kotovenko and Olga Grebenkova and Björn Ommer},
      year={2025},
      eprint={2504.13204},
      archivePrefix={arXiv},
      primaryClass={cs.GR},
      url={https://arxiv.org/abs/2504.13204}, 
}

@misc{fan2024instantsplat,
    title={InstantSplat: Sparse-view Gaussian Splatting in Seconds},
    author={Zhiwen Fan and Kairun Wen and Wenyan Cong and Kevin Wang and Jian Zhang and Xinghao Ding and Danfei Xu and Boris Ivanovic and Marco Pavone and Georgios Pavlakos and Zhangyang Wang and Yue Wang},
    year={2024},
    eprint={2403.20309},
    archivePrefix={arXiv},
    primaryClass={cs.CV}
}

@inproceedings{liu2024mvsgaussian,
    title={MVSGaussian: Fast Generalizable Gaussian Splatting Reconstruction from Multi-View Stereo},
    author={Liu, Tianqi and Wang, Guangcong and Hu, Shoukang and Shen, Liao and Ye, Xinyi and Zang, Yuhang and Cao, Zhiguo and Li, Wei and Liu, Ziwei},
    booktitle={European Conference on Computer Vision},
    pages={37--53},
    year={2025},
    organization={Springer}
}

@article{fang2024miniv2,
  title={Mini-splatting2: Building 360 scenes within minutes via aggressive gaussian densification},
  author={Fang, Guangchi and Wang, Bing},
  journal={arXiv preprint arXiv:2411.12788},
  year={2024}
}

@InProceedings{hoellein_2025_3dgslm,
    title={3DGS-LM: Faster Gaussian-Splatting Optimization with Levenberg-Marquardt},
    author={H{\"o}llein, Lukas and Bo\v{z}i\v{c}, Alja\v{z} and Zollh{\"o}fer, Michael and Nie{\ss}ner, Matthias},
    booktitle={Proceedings of the IEEE/CVF International Conference on Computer Vision (ICCV)},
    year={2025}
}

@inproceedings{lan20253dgs2,
  title={3dgs2: Near second-order converging 3d gaussian splatting},
  author={Lan, Lei and Shao, Tianjia and Lu, Zixuan and Zhang, Yu and Jiang, Chenfanfu and Yang, Yin},
  booktitle={Proceedings of the Special Interest Group on Computer Graphics and Interactive Techniques Conference Conference Papers},
  pages={1--10},
  year={2025}
}

@article{pehlivan2025second,
  title={Second-order Optimization of Gaussian Splats with Importance Sampling},
  author={Pehlivan, Hamza and Boscolo Camiletto, Andrea and Foo, Lin Geng and Habermann, Marc and Theobalt, Christian},
  journal={arXiv preprint arXiv:2504.12905},
  year={2025}
}

@InProceedings{kim2024color,
    author    = {Kim, Sieun and Lee, Kyungjin and Lee, Youngki},
    title     = {Color-cued Efficient Densification Method for 3D Gaussian Splatting},
    booktitle = {Proceedings of the IEEE/CVF Conference on Computer Vision and Pattern Recognition (CVPR) Workshops},
    month     = {June},
    year      = {2024},
    pages     = {775-783}
}

@inproceedings{wang2024adr,
  title={AdR-Gaussian: Accelerating Gaussian Splatting with Adaptive Radius},
  author={Wang, Xinzhe and Yi, Ran and Ma, Lizhuang},
  booktitle={ACM SIGGRAPH Asia 2024 Conference Proceedings},
  pages={1--10},
  year={2024}
}

@inproceedings{chen2025dashgaussian,
  title     = {DashGaussian: Optimizing 3D Gaussian Splatting in 200 Seconds},
  author    = {Chen, Youyu and Jiang, Junjun and Jiang, Kui and Tang, Xiao and Li, Zhihao and Liu, Xianming and Nie, Yinyu},
  booktitle = {CVPR},
  year      = {2025}
}

@inproceedings{girish2024eagles,
  title={Eagles: Efficient accelerated 3d gaussians with lightweight encodings},
  author={Girish, Sharath and Gupta, Kamal and Shrivastava, Abhinav},
  booktitle={European Conference on Computer Vision},
  pages={54--71},
  year={2024},
  organization={Springer}
}

@misc{armagan2025gsta,
    title={GSta: Efficient Training Scheme with Siestaed Gaussians for Monocular 3D Scene Reconstruction},
    author={Anil Armagan and Albert Saà-Garriga and Bruno Manganelli and Kyuwon Kim and M. Kerim Yucel},
    year={2025},
    eprint={2504.06716},
    archivePrefix={arXiv},
    primaryClass={cs.CV}
}

@InProceedings{hanson2025speedy,
    author    = {Hanson, Alex and Tu, Allen and Lin, Geng and Singla, Vasu and Zwicker, Matthias and Goldstein, Tom},
    title     = {Speedy-Splat: Fast 3D Gaussian Splatting with Sparse Pixels and Sparse Primitives},
    booktitle = {Proceedings of the Computer Vision and Pattern Recognition Conference (CVPR)},
    month     = {June},
    year      = {2025},
    pages     = {21537-21546},
    url       = {https://speedysplat.github.io/}
}

@inproceedings{mallick2024taming,
    author = {Mallick, Saswat Subhajyoti and Goel, Rahul and Kerbl, Bernhard and Steinberger, Markus and Carrasco, Francisco Vicente and De La Torre, Fernando},
    title = {Taming 3DGS: High-Quality Radiance Fields with Limited Resources},
    year = {2024},
    isbn = {9798400711312},
    publisher = {Association for Computing Machinery},
    address = {New York, NY, USA},
    url = {https://doi.org/10.1145/3680528.3687694},
    doi = {10.1145/3680528.3687694},
    booktitle = {SIGGRAPH Asia 2024 Conference Papers},
    articleno = {2},
    numpages = {11},
    location = {
    },
    series = {SA '24}
    }

@misc{liao2025litegshighperformanceframeworktrain,
  title={LiteGS: A High-performance Framework to Train 3DGS in Subminutes via System and Algorithm Codesign}, 
  author={Kaimin Liao and Hua Wang and Zhi Chen and Luchao Wang and Yaohua Tang},
  year={2025},
  eprint={2503.01199},
  archivePrefix={arXiv},
  primaryClass={cs.CV},
  url={https://arxiv.org/abs/2503.01199}, 
}

@article{ye2025gsplat,
  title={gsplat: An open-source library for Gaussian splatting},
  author={Ye, Vickie and Li, Ruilong and Kerr, Justin and Turkulainen, Matias and Yi, Brent and Pan, Zhuoyang and Seiskari, Otto and Ye, Jianbo and Hu, Jeffrey and Tancik, Matthew and others},
  journal={Journal of Machine Learning Research},
  volume={26},
  number={34},
  pages={1--17},
  year={2025}
}

@article{durvasula2023distwar,
  title={Distwar: Fast differentiable rendering on raster-based rendering pipelines},
  author={Durvasula, Sankeerth and Zhao, Adrian and Chen, Fan and Liang, Ruofan and Sanjaya, Pawan Kumar and Vijaykumar, Nandita},
  journal={arXiv preprint arXiv:2401.05345},
  year={2023}
}

@article{wang2024faster,
title={Faster and better 3d splatting via group training},
author={Wang, Chengbo and Ma, Guozheng and Xue, Yifei and Lao, Yizhen},
journal={arXiv preprint arXiv:2412.07608},year={2024}
}

@misc{xu2025gaussian,
      title={Gaussian On-the-Fly Splatting: A Progressive Framework for Robust Near Real-Time 3DGS Optimization}, 
      author={Yiwei Xu and Yifei Yu and Wentian Gan and Tengfei Wang and Zongqian Zhan and Hao Cheng and Xin Wang},
      year={2025},
      eprint={2503.13086},
      archivePrefix={arXiv},
      primaryClass={cs.CV},
      url={https://arxiv.org/abs/2503.13086}, 
}

@misc{gui2024balanced,
      title={Balanced 3DGS: Gaussian-wise Parallelism Rendering with Fine-Grained Tiling}, 
      author={Hao Gui and Lin Hu and Rui Chen and Mingxiao Huang and Yuxin Yin and Jin Yang and Yong Wu and Chen Liu and Zhongxu Sun and Xueyang Zhang and Kun Zhan},
      year={2025},
      eprint={2412.17378},
      archivePrefix={arXiv},
      primaryClass={cs.CV},
      url={https://arxiv.org/abs/2412.17378}, 
}

@misc{mast3r_eccv24,
      title={Grounding Image Matching in 3D with MASt3R}, 
      author={Vincent Leroy and Yohann Cabon and Jerome Revaud},
      booktitle = {ECCV},
      year = {2024}
}

@inproceedings{zwicker2001ewa,
  title={Ewa volume splatting},
  author={Zwicker, Matthias and Pfister, Hanspeter and Van Baar, Jeroen and Gross, Markus},
  booktitle={Proceedings Visualization, 2001. VIS'01.},
  pages={29--538},
  year={2001},
  organization={IEEE}
}

@article{barron2022mipnerf360,
    title={Mip-NeRF 360: Unbounded Anti-Aliased Neural Radiance Fields},
    author={Jonathan T. Barron and Ben Mildenhall and 
            Dor Verbin and Pratul P. Srinivasan and Peter Hedman},
    journal={CVPR},
    year={2022}
}

@article{Knapitsch2017,
    author    = {Arno Knapitsch and Jaesik Park and Qian-Yi Zhou and Vladlen Koltun},
    title     = {Tanks and Temples: Benchmarking Large-Scale Scene Reconstruction},
    journal   = {ACM Transactions on Graphics},
    volume    = {36},
    number    = {4},
    year      = {2017},
}

@Article{HPPFDB18,
  author       = "Hedman, Peter and Philip, Julien and Price, True and Frahm, Jan-Michael and Drettakis, George and Brostow, Gabriel",
  title        = "Deep Blending for Free-Viewpoint Image-Based Rendering",
  journal      = "ACM Transactions on Graphics (SIGGRAPH Asia Conference Proceedings)",
  number       = "6",
  volume       = "37",
  month        = "November",
  year         = "2018",
  url          = "http://www-sop.inria.fr/reves/Basilic/2018/HPPFDB18"
}

@inproceedings{Dai2024GaussianSurfels,
  author = {Dai, Pinxuan and Xu, Jiamin and Xie, Wenxiang and Liu, Xinguo and Wang, Huamin and Xu, Weiwei},
  title = {High-quality Surface Reconstruction using Gaussian Surfels},
  publisher = {Association for Computing Machinery},
  booktitle = {ACM SIGGRAPH 2024 Conference Papers},
  year = {2024},
  articleno = {22},
  numpages = {11}
}

@inproceedings{mildenhall2020nerf,
 title={NeRF: Representing Scenes as Neural Radiance Fields for View Synthesis},
 author={Ben Mildenhall and Pratul P. Srinivasan and Matthew Tancik and Jonathan T. Barron and Ravi Ramamoorthi and Ren Ng},
 year={2020},
 booktitle={ECCV},
}

@misc{KheradmandVicini2025stochasticsplats,
  title         = {StochasticSplats: Stochastic Rasterization for Sorting-Free 3D Gaussian Splatting},
  author        = {Shakiba Kheradmand and Delio Vicini and George Kopanas and Dmitry Lagun
                    and Kwang Moo Yi and Mark Matthews and Andrea Tagliasacchi},
  year          = {2025},
  url           = {https://arxiv.org/abs/2503.24366},
  eprint        = {2503.24366},
  archiveprefix = {arXiv},
  primaryclass  = {cs.CV}
}

@article{radl2024stopthepop,
  author    = {Radl, Lukas and Steiner, Michael and Parger, Mathias and Weinrauch, Alexander and Kerbl, Bernhard and Steinberger, Markus},
  title     = {{StopThePop: Sorted Gaussian Splatting for View-Consistent Real-time Rendering}},
  journal   = {ACM Transactions on Graphics},
  number    = {4},
  volume    = {43},
  articleno = {64},
  year      = {2024},
}

@article{zhang2025advances,
  title={Advances in feed-forward 3d reconstruction and view synthesis: A survey},
  author={Zhang, Jiahui and Li, Yuelei and Chen, Anpei and Xu, Muyu and Liu, Kunhao and Wang, Jianyuan and Long, Xiao-Xiao and Liang, Hanxue and Xu, Zexiang and Su, Hao and others},
  journal={arXiv preprint arXiv:2507.14501},
  year={2025}
}

@article{irshad2024neural,
  title={Neural fields in robotics: A survey},
  author={Irshad, Muhammad Zubair and Comi, Mauro and Lin, Yen-Chen and Heppert, Nick and Valada, Abhinav and Ambrus, Rares and Kira, Zsolt and Tremblay, Jonathan},
  journal={arXiv preprint arXiv:2410.20220},
  year={2024}
}

@article{ren2025fastgs,
  title={FastGS: Training 3D Gaussian Splatting in 100 Seconds},
  author={Ren, Shiwei and Wen, Tianci and Fang, Yongchun and Lu, Biao},
  journal={arXiv preprint arXiv:2511.04283},
  year={2025}
}

@inproceedings{sun2025svraster,
  title={Sparse Voxels Rasterization: Real-time High-fidelity Radiance Field Rendering},
  author={Sun, Cheng and Choe, Jaesung and Loop, Charles and Ma, Wei-Chiu and Wang, Yu-Chiang Frank},
  booktitle={CVPR},
  year={2025}
}

@article{zoomers2025nvgs,
  title={NVGS: Neural Visibility for Occlusion Culling in 3D Gaussian Splatting},
  author={Zoomers, Brent and Hahlbohm, Florian and Vanherck, Joni and Jorissen, Lode and Magnor, Marcus and Michiels, Nick},
  journal={arXiv preprint arXiv:2511.19202},
  year={2025}
}

@inproceedings{MaterialRefGS,
  title     = {MaterialRefGS: Reflective Gaussian Splatting with Multi-view Consistent Material Inference},
  author    = {Zhang, Wenyuan and Tang, Jimin and Zhang, Weiqi and Fang, Yi and Liu, Yu-Shen and Han, Zhizhong},
  booktitle = {Advances in Neural Information Processing Systems},
  year      = {2025}
}

@inproceedings{MonoInstance,
  title     = {MonoInstance: Enhancing Monocular Priors via Multi-view Instance Alignment for Neural Rendering and Reconstruction},
  author    = {Zhang, Wenyuan and Yang, Yixiao and Huang, Han and Han, Liang and Shi, Kanle and Liu, Yu-Shen and Han, Zhizhong},
  booktitle = {Proceedings of the IEEE/CVF Conference on Computer Vision and Pattern Recognition},
  year      = {2025}
}

@inproceedings{Binocular3DGS,
  title     = {Binocular-Guided 3D Gaussian Splatting with View Consistency for Sparse View Synthesis},
  author    = {Han, Liang and Zhou, Junsheng and Liu, Yu-Shen and Han, Zhizhong},
  booktitle = {Advances in Neural Information Processing Systems},
  year      = {2024}
}

@inproceedings{SelfConstrainedPriors3DGS,
  title     = {3D Gaussian Splatting with Self-Constrained Priors for High Fidelity Surface Reconstruction},
  author    = {Noda, Takeshi and Liu, Yu-Shen and Han, Zhizhong},
  booktitle = {Proceedings of the IEEE/CVF Conference on Computer Vision and Pattern Recognition},
  year      = {2026}
}

@inproceedings{VGGS,
  title     = {VGGS: VGGT-guided Gaussian Splatting for Efficient and Faithful Sparse-View Surface Reconstruction},
  author    = {Xiang, Peng and Han, Liang and Zhang, Hui and Liu, Yu-Shen and Han, Zhizhong},
  booktitle = {Proceedings of the AAAI Conference on Artificial Intelligence},
  year      = {2026}
}

@inproceedings{GaussianUDF,
  title     = {GaussianUDF: Inferring Unsigned Distance Functions through 3D Gaussian Splatting},
  author    = {Li, Shujuan and Liu, Yu-Shen and Han, Zhizhong},
  booktitle = {Proceedings of the IEEE/CVF Conference on Computer Vision and Pattern Recognition},
  year      = {2025}
}

@inproceedings{GSPull,
  title     = {Neural Signed Distance Function Inference through Splatting 3D Gaussians Pulled on Zero-Level Set},
  author    = {Zhang, Wenyuan and Liu, Yu-Shen and Han, Zhizhong},
  booktitle = {Advances in Neural Information Processing Systems},
  year      = {2024}
}

@inproceedings{SGADSLAM,
  title     = {SGAD-SLAM: Splatting Gaussians at Adjusted Depth for Better Radiance Fields in RGB-D SLAM},
  author    = {Hu, Pengchong and Han, Zhizhong},
  booktitle = {Proceedings of the IEEE/CVF Conference on Computer Vision and Pattern Recognition},
  year      = {2026}
}

@inproceedings{VTGaussianSLAM,
  title     = {VTGaussian-SLAM: RGBD SLAM for Large Scale Scenes with Splatting View-Tied 3D Gaussians},
  author    = {Hu, Pengchong and Han, Zhizhong},
  booktitle = {International Conference on Machine Learning},
  year      = {2025}
}

@inproceedings{QQSLAM,
  title     = {Query Quantized Neural SLAM},
  author    = {Jiang, Sijia and Hua, Jing and Han, Zhizhong},
  booktitle = {Proceedings of the AAAI Conference on Artificial Intelligence},
  year      = {2025}
}
